\documentclass[nonacm,manuscript,screen]{acmart}
\acmJournal{CSUR}

\usepackage{xcolor,colortbl}
\usepackage{makecell}
\usepackage{longtable}
\usepackage{subcaption}
\usepackage{pdfpages}
\usepackage{alphalph}

\definecolor{Gray}{gray}{0.85}
\definecolor{LightRed}{rgb}{1,0.8,0.8}
\definecolor{LightBlue}{rgb}{0.70, 0.85, 1}
\definecolor{LightGreen}{rgb}{0.8, 1, 0.83}

\AtBeginDocument{%
  }

\setcopyright{acmlicensed}
\copyrightyear{2026}
\acmYear{2026}
\acmDOI{XXXXXXX.XXXXXXX}

\makeatletter
\def\@authorfont{\normalsize\normalfont}
\makeatother

\begin{document}

\title{Reasoning4Sciences: Bridging Reasoning Language Models to All Scientific Branches}

\author{Teddy Ferdinan}
\email{teddy.ferdinan@pwr.edu.pl}
\orcid{0000-0003-3701-3502}

\author{Bartłomiej Koptyra}
\email{bartlomiej.koptyra@pwr.edu.pl}
\orcid{0009-0005-9938-305X}

\author{Mikołaj Langner}
\email{mikolaj.langner@pwr.edu.pl}
\orcid{0009-0007-9531-5329}

\author{Tomasz Adamczyk}
\email{tomasz.adamczyk@pwr.edu.pl}
\orcid{0009-0005-9703-4630}

\author{Łukasz Radliński}
\email{lukasz.radlinski@pwr.edu.pl}
\orcid{0000-0002-7366-3847}

\author{Maciej Markiewicz}
\email{maciej.markiewicz@pwr.edu.pl}
\orcid{0009-0004-2882-6741}

\author{Aleksander Szczęsny}
\email{aleksander.szczesny@pwr.edu.pl}
\orcid{0009-0003-6808-2321}

\author{Stanisław Woźniak}
\email{stanislaw.wozniak@pwr.edu.pl}
\orcid{0000-0001-8761-1629}

\author{Tymoteusz Romanowicz}
\email{rtymoteusz@gmail.com}
\orcid{0009-0008-3798-3334}

\author{Dzmitry Pihulski}
\email{dzmitry.pihulski@pwr.edu.pl}
\orcid{0009-0002-5434-4696}

\author{Mateusz Zbrocki}
\email{mateusz.zbrocki@gmail.com}
\orcid{0009-0008-2672-8864}

\author{Mateusz Śmigielski}
\email{mateusz.smigielski@pwr.edu.pl}
\orcid{0009-0001-3453-9034}

\author{Michał Rajkowski}
\email{michal.rajkowski@pwr.edu.pl}
\orcid{0009-0003-2983-4324}

\author{Mateusz Biedka}
\email{mateusz.biedka@proton.me}
\orcid{0009-0001-0213-987X}

\author{Konrad Kiełczyński}
\email{konrad.kielczynski@pwr.edu.pl}
\orcid{0009-0009-3223-2336}

\author{Konrad Wojtasik}
\email{konrad.wojtasik@pwr.edu.pl}
\orcid{0000-0002-5715-5201}

\author{Jacek Duszenko}
\email{jacek.duszenko@pwr.edu.pl }
\orcid{0009-0004-7606-4198}

\author{Jan Eliasz}
\email{jan.eliasz@pwr.edu.pl}
\orcid{0009-0007-0851-1816}

\author{Piotr Matys}
\email{piotr.matys@pwr.edu.pl}
\orcid{0009-0004-9282-2892}

\author{Michał Bernacki-Janson}
\email{michal.bernacki-janson@pwr.edu.pl}
\orcid{0009-0008-4885-8555}

\affiliation{
  \institution{Wrocław Tech}
  \city{Wrocław}
  \country{Poland}
}

%% UKMC

\author{Maria Bellaniar Ismiati}
\email{bella@ukmc.ac.id}
\orcid{0009-0001-1632-8751}

\affiliation{
  \institution{National Cheng Kung University}
  \city{Tainan}
  \country{Taiwan}
}
\affiliation{
  \institution{Universitas Katolik Musi Charitas}
  \city{Palembang}
  \country{Indonesia}
}

\author{Latius Hermawan}
\email{tiuz.hermawan@ukmc.ac.id}
\orcid{0000-0001-5360-0550}

\affiliation{
  \institution{Universitas Katolik Musi Charitas}
  \city{Palembang}
  \country{Indonesia}
}

%% PWr domain experts

\author{Wiktoria Mieleszczenko-Kowszewicz}
\email{wiktoria.mieleszczenko-kowszewicz@pwr.edu.pl}
\orcid{0000-0002-3948-268X}

\author{Anna Kubicka-Sowińska}
\email{anna.kubicka-sowinska@pwr.edu.pl}
\orcid{0000-0001-5442-3947}

\author{Grzegorz Chodak}
\email{grzegorz.chodak@pwr.edu.pl}
\orcid{0000-0002-9604-482X}

\author{Karol Postawa}
\email{karol.postawa@pwr.edu.pl}
\orcid{0000-0001-7022-9458}

\author{Paweł Zyblewski}
\email{pawel.zyblewski@pwr.edu.pl}
\orcid{0000-0002-4224-6709}

\author{Tomasz Szandała}
\email{tomasz.szandala@pwr.edu.pl }
\orcid{0000-0003-4525-0444}

\author{Łukasz Sterczewski}
\email{lukasz.sterczewski@pwr.edu.pl}
\orcid{0000-0003-1459-7517} 

\author{Adrian Chajec}
\email{adrian.chajec@pwr.edu.pl}
\orcid{0000-0001-5329-9534}

\author{Paweł Niewiadomski}
\email{pawel.niewiadomski@pwr.edu.pl}
\orcid{0000-0002-0646-3036}

\author{Piotr Gruber}
\email{piotr.gruber@pwr.edu.pl}
\orcid{0000-0001-7236-6763}

\author{Marcin Wdowikowski}
\email{marcin.wdowikowski@pwr.edu.pl}
\orcid{0000-0003-2693-0946}

\author{Sławomir Czarnecki}
\email{slawomir.czarnecki@pwr.edu.pl}
\orcid{0000-0001-8021-943X}

\author{Bartłomiej Kryszak}
\email{bartlomiej.kryszak@pwr.edu.pl}
\orcid{0000-0001-9807-964X}

\author{Dominik Drabik}
\email{dominik.drabik@pwr.edu.pl}
\orcid{0000-0003-4568-4066}

\author{Tomasz Kajdanowicz}
\email{tomasz.kajdanowicz@pwr.edu.pl}
\orcid{0000-0002-8417-1012}

\affiliation{
  \institution{Wrocław Tech}
  \city{Wrocław}
  \country{Poland}
}

%% UJ

\author{Kamil Mamak}
\email{kamil.mamak@uj.edu.pl}
\orcid{0000-0002-5081-792X}

\affiliation{
  \institution{Jagiellonian University}
  \city{Kraków}
  \country{Poland}
}

%% UWr

\author{Paweł Preś}
\email{pawel.pres@uwr.edu.pl}
\orcid{0000-0001-8474-7694}

\affiliation{
  \institution{University of Wrocław}
  \city{Wrocław}
  \country{Poland}
}

%% UMW

\author{Katarzyna Paczkowska}
\email{paczkowska.k@gmail.com}
\orcid{0000-0003-0619-4670}

\author{Joachim Sobczuk}
\email{joachim.sobczuk@umw.edu.pl}
\orcid{0009-0006-3347-7579}

\affiliation{
  \institution{Wrocław Medical University}
  \city{Wrocław}
  \country{Poland}
}

%% UPWr

\author{Tomasz Zięba}
\email{tomasz.zieba@upwr.edu.pl}
\orcid{0000-0002-2791-342X}

\affiliation{
  \institution{Wrocław University of Environmental and Life Sciences}
  \city{Wrocław}
  \country{Poland}
}

%% PWr Supervisors

\author{Jan Kocoń}
\email{jan.kocon@pwr.edu.pl}
\orcid{0000-0002-7665-6896}

\author{Maciej Piasecki}
\email{maciej.piasecki@pwr.edu.pl}
\orcid{0000-0003-1503-0993}

\author{Przemysław Kazienko}
\email{kazienko@pwr.edu.pl}
\orcid{0000-0001-5868-356X}

\affiliation{
  \institution{Wrocław Tech}
  \city{Wrocław}
  \country{Poland}
}

\renewcommand{\shortauthors}{Ferdinan et al.}

\begin{abstract}
\pagebreak
While Reasoning Language Models (RLMs) are rapidly emerging as powerful tools for scientific research, their impact is primarily concentrated in "hard science" fields. The slow---or lack of---adoption of RLMs in other branches of science is causing a widening gap in research productivity. In this survey, we provide the first comprehensive analysis of RLM adoption across 28 scientific disciplines following the classification used by the European Research Council (ERC), spanning the Social Sciences and Humanities, Physical Sciences and Engineering, and Life Sciences. We examine how RLMs are developed, evaluated, and applied across disciplines. Furthermore, we introduce a maturity-oriented assessment framework based on available domain-specific development and evaluation resources, revealing substantial disparities in RLM maturity that become even more pronounced when only publicly available resources are considered. Finally, we highlight current implementation paradigms that are gaining popularity across disciplines, current challenges, and future directions in enabling RLM adoption across science.
\end{abstract}

%% The code below is generated by the tool at http://dl.acm.org/ccs.cfm
\begin{CCSXML}
<ccs2012>
   <concept>
       <concept_id>10002944.10011122.10002945</concept_id>
       <concept_desc>General and reference~Surveys and overviews</concept_desc>
       <concept_significance>500</concept_significance>
       </concept>
   <concept>
       <concept_id>10010147.10010178.10010179.10010182</concept_id>
       <concept_desc>Computing methodologies~Natural language generation</concept_desc>
       <concept_significance>500</concept_significance>
       </concept>
   <concept>
       <concept_id>10010147.10010178.10010219.10010221</concept_id>
       <concept_desc>Computing methodologies~Intelligent agents</concept_desc>
       <concept_significance>500</concept_significance>
       </concept>
   <concept>
       <concept_id>10010405</concept_id>
       <concept_desc>Applied computing</concept_desc>
       <concept_significance>300</concept_significance>
       </concept>
 </ccs2012>
\end{CCSXML}

\ccsdesc[500]{General and reference~Surveys and overviews}
\ccsdesc[500]{Computing methodologies~Natural language generation}
\ccsdesc[500]{Computing methodologies~Intelligent agents}
\ccsdesc[300]{Applied computing}

\keywords{reasoning language model, large reasoning model, scientific work, research, science}

% \received{xx xxxx 2026}
% \received[revised]{xx xxxx 2026}
% \received[accepted]{xx xxxx 2026}

\maketitle

\section{Introduction}

\begin{figure}[ht]
\centering
\includegraphics[width=0.99\textwidth]{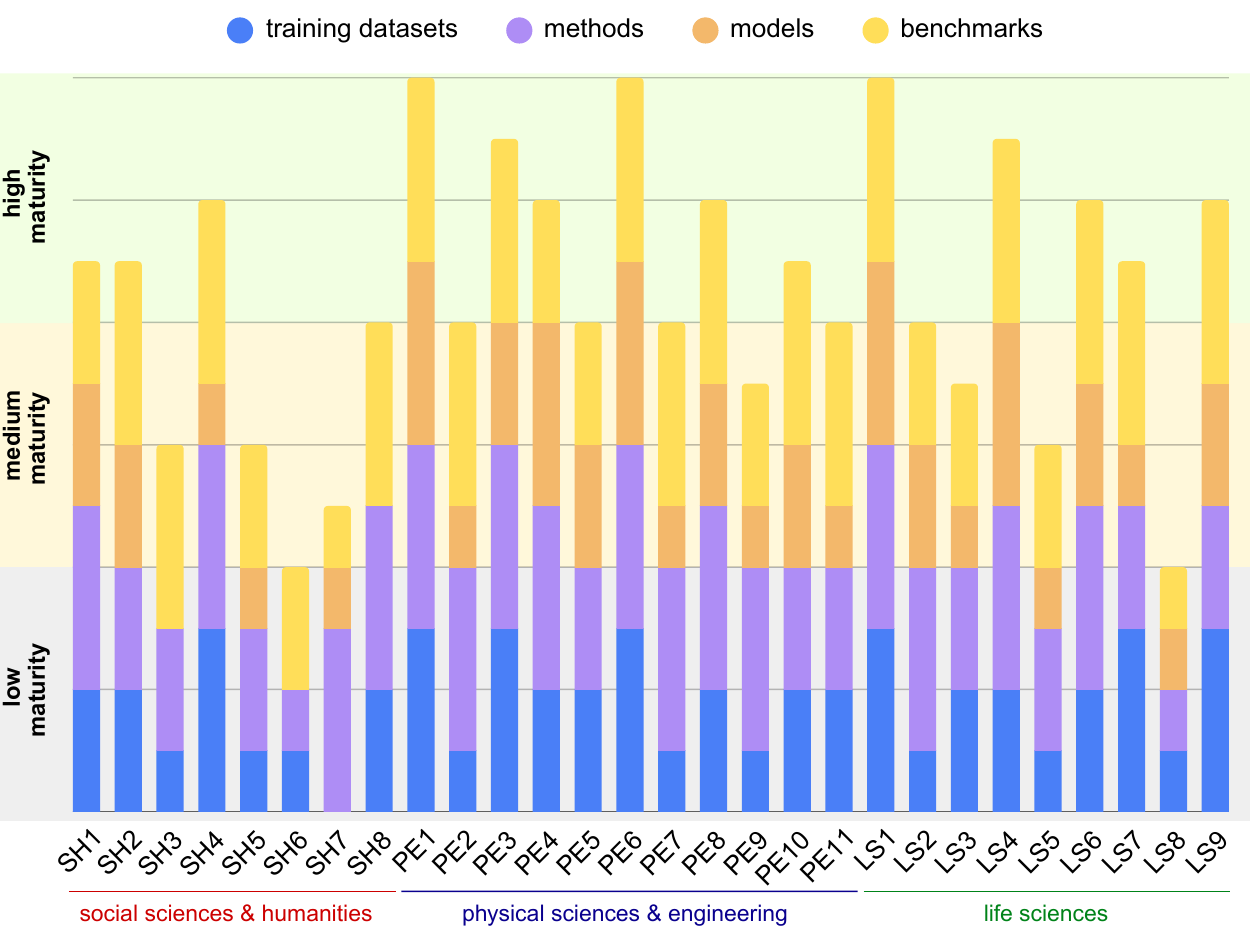}
\caption{RLM maturity levels across 28 scientific disciplines based on all identified resources.}
\Description{RLM maturity levels across 28 scientific disciplines based on all identified resources.}
\label{fig:maturity_all}
\end{figure}

Large Language Models (LLMs) have advanced with rapid progress in recent years, transforming natural language understanding and generation across a wide range of applications, including science. Beyond traditional LLMs, a new generation of Reasoning Language Models (RLMs) has emerged with the ability to perform explicit multi-step reasoning and deliberation, either through chain-of-thought prompting or sophisticated training strategies via supervised fine-tuning or reinforcement learning. Compared to traditional LLMs, they have demonstrated substantial improvements in complex problem solving, such as math, coding, graduate-level scientific question answering, and agentic tasks \cite{luong2025imo, pinheiro2025_astro_olym, deepseek2025_v32, vLex_2025, gopal2025vct, kim2023_tasks}. This suggests a transition from language models with shallow textual pattern imitation towards systems capable of structured thinking, enabling them to support increasingly sophisticated tasks.

The growing reasoning capability of modern language models has already begun to reshape workflows in various scientific domains. Agentic frameworks combine RLMs with retrieval systems, external tools, and iterative self-repair mechanisms to enable increasingly autonomous forms of scholarly assistance. RLMs are now being used to assist literature review, automate data analysis pipelines, support experimental design, and automate hypothesis generation and validation \cite{Sui2026Medea, ghafarollahi2024sciagents, zhang2025_astro_litrev_agent, ye2025_astro_replicationbench, Cho2026PersonaAI, rodriguez2025vaccinology, guo2025earthlink, ghafarollahi2025sparksmultiagentartificialintelligence, xu2025_astro_cmbagent}. In the more mature fields such as mathematics, computer science, chemistry, physics, and medicine, specialized reasoning models have achieved impressive benchmark performance, hinting at the potential to accelerate discovery while reducing repetitive cognitive workload.

However, RLM adoption across scientific disciplines remains highly uneven. Current progress is primarily concentrated within fields that possess strong computational traditions and abundant digital resources. In contrast, disciplines that typically focus more on field work or wet lab activities are severely restricted when it comes to ready resources for effective domain-specific RLM development and evaluation. Such disparities risk amplifying existing inequalities in access to scientific infrastructure and research acceleration technologies. Furthermore, such disparities may widen the gap in research productivity between research groups who can easily adopt RLMs into their workflow and those who cannot. We believe that RLMs should not remain exclusive to a limited subset of scientific domains; rather, they represent a valuable tool that can potentially support research work across all branches of science when adapted responsibly and appropriately.

Motivated by this challenge, this survey investigates the current landscape of RLM adoption across the 28 scientific disciplines defined by the European Research Council (ERC) evaluation panels. We systematically analyze how reasoning models are being developed, evaluated, and utilized in research, while also identifying the barriers that hinder broader adoption. In particular, this work addresses the following research questions:

\begin{enumerate}
    \item How widely are RLMs currently being used in research work across science?
    \item How are RLMs being used to assist research work in science?
    \item What are the current challenges that hinder RLM usage in various scientific disciplines?
    \item What are the future directions for RLMs to effectively assist research work in different scientific disciplines?
\end{enumerate}

Using our novel maturity-oriented assessment framework, we found social sciences \& humanities disciplines to be trailing behind life sciences as well as physical sciences \& engineering in terms of RLM adoption. The maturity scores plummet even further when only publicly accessible resources are considered; this indicates that notable portions of domain-specific RLM technologies are locked behind proprietary walls, limiting their utility for the broader scientific community. In summary, our main contributions are as follows: (1) introduced a quantitative maturity-oriented assessment framework to measure RLM adoption across various scientific disciplines; (2) created a comprehensive cross-disciplinary mapping of 28 scientific disciplines to determine the current state of RLM development, evaluation, and practical deployment across science; (3) synthesized emerging RLM implementation paradigms currently reshaping scientific workflows; (4) analyzed critical challenges and future directions for effective RLM adoption across science.

\section{Methodology}

\begin{table}
\centering
\caption{28 scientific disciplines based on the categorization used by the ERC's evaluation panels.}
\label{tab:erc_panels}
\begin{tabular}{|>{\centering\arraybackslash\columncolor{LightRed}}p{4.55cm}|>{\centering\arraybackslash\columncolor{LightBlue}}p{4.55cm}|>{\centering\arraybackslash\columncolor{LightGreen}}p{4.55cm}|}
\hline
\rowcolor{Gray}
\textbf{Social Sciences \& Humanities} & \textbf{Physical Sciences \& Engineering} & \textbf{Life Sciences} \\
\hline
\makecell{
    SH1: Individuals, Markets and\\Organisations \\
    SH2: Institutions, Governance and\\Legal Systems \\
    SH3: The Social World and Its\\Interactions \\
    SH4: The Human Mind and Its\\Complexity \\
    SH5: Texts and Concepts \\
    SH6: The Study of the Human Past \\
    SH7: Human Mobility, Environ-\\ment, and Space \\
    SH8: Studies of Cultures and Arts
} &
\makecell{
    PE1: Mathematics \\
    PE2: Fundamental Constituents\\of Matter \\
    PE3: Condensed Matter Physics \\
    PE4: Physical and Analytical\\Chemical Sciences \\
    PE5: Synthetic Chemistry and\\Materials \\
    PE6: Computer Science and\\Informatics \\
    PE7: Systems and Communication\\Engineering \\
    PE8: Products and Processes\\Engineering \\
    PE9: Universe Sciences \\
    PE10: Earth System Science \\
    PE11: Materials Engineering
} &
\makecell{
    LS1: Molecules of Life: Biological\\Mechanisms, Structures, Functions \\
    LS2: Integrative Biology: from\\Genes \& Genomes to Systems \\
    LS3: Cell Biology, Development,\\Stem Cells and Regeneration \\
    LS4: Physiology in Health,\\Disease and Ageing \\
    LS5: Neuroscience and Disorders\\of the Nervous System \\
    LS6: Immunity, Infection and\\Immunotherapy \\
    LS7: Prevention, Diagnosis and\\Treatment of Human Diseases \\
    LS8: Environmental Biology, Eco-\\logy and Evolution \\
    LS9: Biotechnology and Biosys-\\tems Engineering
} \\
\hline
\end{tabular}
\end{table}

To answer our research questions, we conducted literature review across 28 scientific disciplines following the categorization used by the evaluation panels of the European Research Council (ERC) \citep{ercpanels2026}. As shown in Table \ref{tab:erc_panels}, this classification system covers a wide range of scientific domains, not only from Physical Sciences \& Engineering (PE), but also Life Sciences (LS) and Social Sciences \& Humanities (SH)\footnote{More detailed information about each scientific discipline and its related topics is provided in Appendix \ref{app:science-typography}.}.

We collected papers for each discipline independently using various sources, including Google Scholar, Scopus, and Web of Science. Due to the rapid progress of AI research, we also considered works published in pre-print servers such as arXiv and bioRxiv. We started by searching for papers containing the terms "reasoning", "language model", and a discipline-specific keyword taken from the ERC's list of relevant topics for the particular discipline. From this initial collection, we filtered out some papers by manually evaluating the abstract and methodology; we only took papers that use the reasoning capability of language model. More specifically, we understood \textit{reasoning} as the capability of language model to generate a step-by-step elaboration of how to answer a question or solve a problem. Therefore, models that have been trained to generate explicit reasoning tokens (typically delimited by \texttt{<think>} and \texttt{</think>} or other similar tokens) fall into this category. However, we also considered papers that leveraged chain-of-thought prompting on traditional LLMs. For disciplines that still possess a huge number of papers after such filtering, we sorted the papers by the number of citations and focused on works with the highest impact.

We reviewed these works to understand how they applied reasoning models in their domains, how the reasoning models were evaluated, along with the weaknesses and potential future directions. We also identified whether they introduced new use-cases or new resources (i.e., models, methods, training datasets, benchmarks) for developing, using, or evaluating RLMs in their domains. Finally, we performed a high-level analysis across the 28 disciplines to determine the current state of RLM application in science, trends, gaps, and opportunities.

\section{RLM Maturity Levels across Sciences}
\label{sec:maturity}

\begin{table*}[ht]
\centering
\caption{The value of the coefficient $C$ based on the number of found scientific works that introduced novel domain-specific resources.}
\label{tab:adoption_level}
\begin{tabular}{@{}ccc@{}}
\toprule
\textbf{Number of scientific works found} & \textbf{Category} & \textbf{Value} \\
\midrule
0 & Non-existent & 0 \\
1--3 & Low & 1 \\
4--6 & Medium & 2 \\
$>$6 & High & 3 \\
\bottomrule
\end{tabular}
\end{table*}

We analyzed the 28 scientific disciplines for essential resources that would enable or advance the development of domain-specific RLMs. Based on this analysis, we define the Maturity of Development level (\textit{MD}) and the Maturity of Evaluation (\textit{ME}) of the RLM technology across these scientific disciplines using the following formulas:

\begin{equation}
    MD = \frac{C_{methods} + C_{datasets}}{2}
\end{equation}

\begin{equation}
    ME = \frac{C_{models} + C_{benchmarks}}{2}
\end{equation}

\noindent where $MD$: maturity level in development resources; $ME$: maturity level in evaluation resources; $C$: the coefficient assigned based on the number of found scientific works that introduced novel domain-specific models, methods, training datasets, or benchmarks. Table~\ref{tab:adoption_level} presents the rules for assigning the value of $C$.

\begin{figure}[t]
\centering
\includegraphics[width=0.99\textwidth]{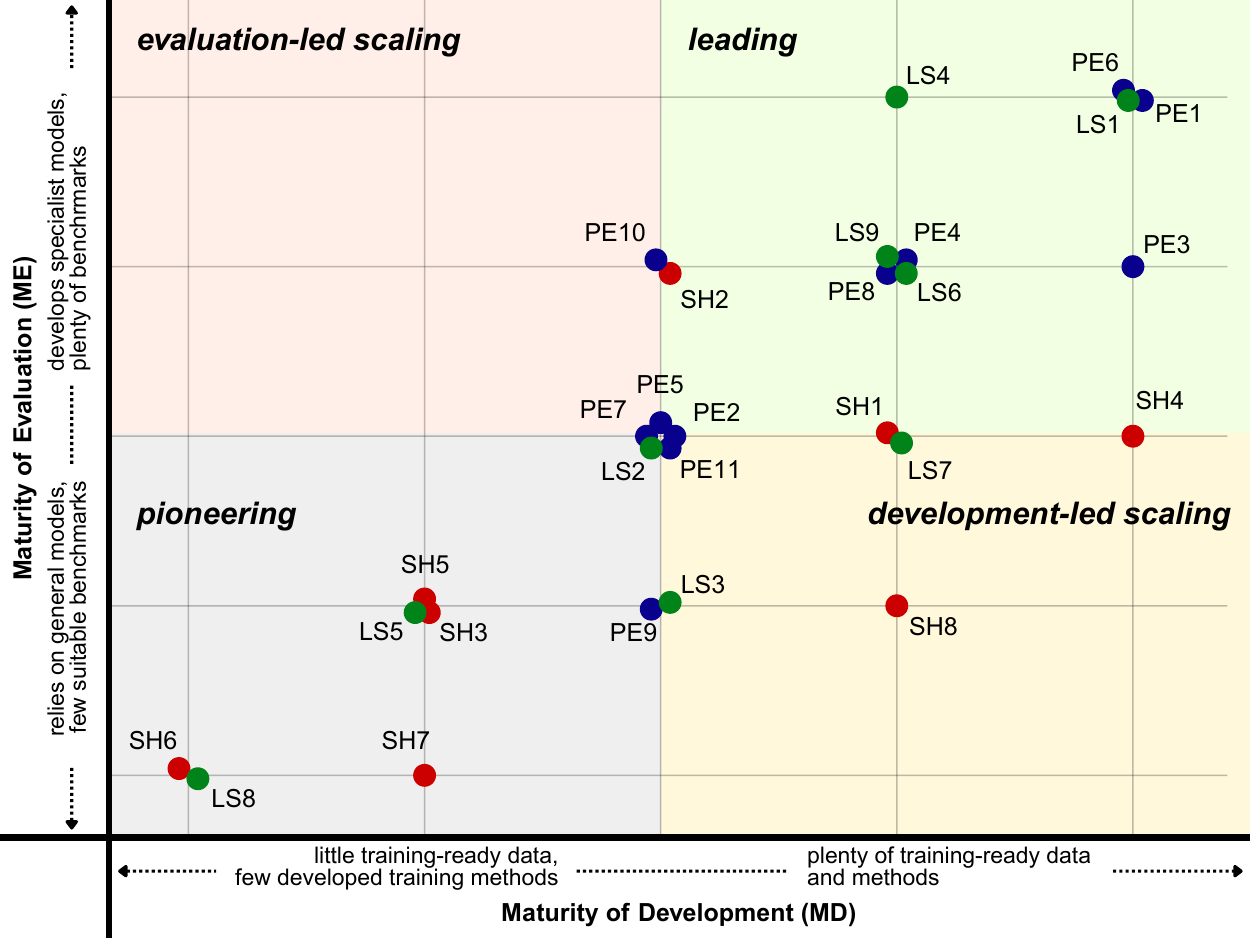}
\caption{Maturity of Development Resources (MD) and Maturity of Evaluation Resources (ME) for 28 scientific disciplines as calculated using Equation (1) and (2), mapped onto a 2D space. Positioning of overlapping dots have been slightly adjusted for better visibility.}
\Description{Maturity of Development Resources (MD) and Maturity of Evaluation Resources (ME) for 28 scientific disciplines as calculated using Equation (1) and (2), mapped onto a 2D space. Positioning of overlapping dots have been slightly adjusted for better visibility.}
\label{fig:maturity_panelwise}
\end{figure}

Figure~\ref{fig:maturity_all} illustrates the RLM maturity levels across these 28 individual disciplines, revealing an uneven distribution of resources across the scientific landscape. Additionally, we mapped the MD and ME scores of each discipline onto a two-dimensional space in Figure~\ref{fig:maturity_panelwise} to better illustrate different RLM adoption strategies across these disciplines. The "Physical Sciences \& Engineering" (PE) domain aggressively outpaces the other branches. Disciplines with deep computational roots, such as PE1 (mathematics) and PE6 (computer science and informatics), populate the uppermost echelon of the "high maturity" tier, buoyed by a massive accumulation of ready-to-use training datasets, methods, models, and benchmarks. In stark contrast, the "Social Sciences \& Humanities" (SH) domain heavily populates the "low maturity" and lower "medium maturity" brackets, signaling a severe technological infrastructure deficit. Notably, we were not able to find dedicated reasoning models in SH3 (sociology, social psychology, education sciences, communication studies), SH6 (archaeology and history), and SH8 (social anthropology, studies of cultures, studies of arts). Meanwhile, the "Life Sciences" (LS) disciplines represent a moderate middle ground, displaying scaling across various fields but still trailing behind when compared against the sheer volume of assets found in the hard engineering disciplines.

\begin{figure}[t]
\centering
\includegraphics[width=0.5\textwidth]{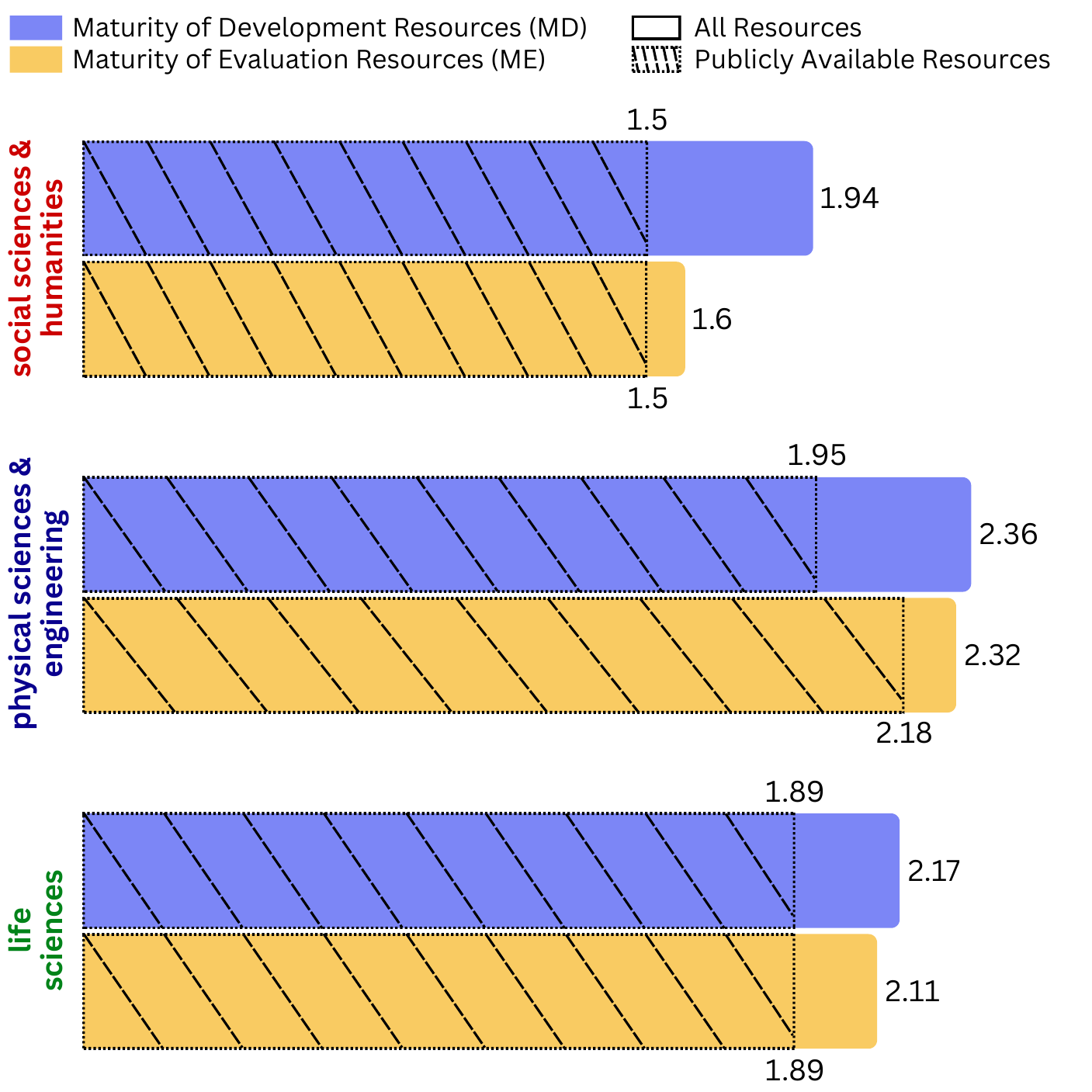}
\caption{Maturity of Development Resources (MD) and Maturity of Evaluation Resources (ME) of three main research domains: Social Sciences \& Humanities, Physical Sciences \& Engineering, and Life Sciences. We distinguished the cases when all identified resources were considered, and when only publicly available resources were considered.}
\Description{Maturity of Development Resources (MD) and Maturity of Evaluation Resources (ME) of three main research domains: Social Sciences \& Humanities, Physical Sciences \& Engineering, and Life Sciences. We distinguished the cases when all identified resources were considered, and when only publicly available resources were considered.}
\label{fig:maturity_aggregated}
\end{figure}

However, looking strictly at the total volume of literature can be misleading, as not all of these resources were made publicly available. As a result, while some disciplines might seem to already possess some degree of RLM maturity, their applicability in the real world might be much more limited. The lack of access to these resources significantly hinders researchers from taking the full advantage of RLM technology. To expose this discrepancy, we additionally calculated the RLM maturity levels by filtering for papers that explicitly made their resources publicly available, as summarized across the three main research domains in Figure~\ref{fig:maturity_aggregated}. There is a universal and highly concerning drop-off in maturity scores across all three research domains when gating for public availability:

\begin{itemize}
\item Social Sciences \& Humanities (SH): Already starting at the lowest position, this domain suffers a painful contraction when restricted to open science. Its Maturity of Development Resources (MD) drops sharply from 1.94 to 1.5, while its Maturity of Evaluation Resources (ME) slips from 1.6 down to 1.5.
\item Physical Sciences \& Engineering (PE): While PE maintains the strongest overall ecosystem, a significant portion of its resources is proprietary or closed. Its MD score plummets from 2.36 to 1.95. Interestingly, its ME score remains the most resilient across all sciences, only dipping from 2.32 to 2.18, indicating that testing suites and benchmarks are shared more freely than the underlying training data and methods.
\item Life Sciences (LS): Life sciences show a roughly uniform contraction across both development and evaluation resources. Its MD metric falls from 2.17 to 1.89, and its ME mirrors that drop, decreasing from 2.11 to 1.89.
\end{itemize}

This widespread deflation underscores a systemic issue: a vast amount of domain-specific RLM innovation is effectively locked away behind proprietary walls or left unshared. For the average researcher, this barrier heavily deflates the real-world utility of RLMs.

\section{Current Trends of RLM Usage in Science}
Reasoning models are being used to help perform a wide range of tasks in science. This section provides a high-level summary of the trends in RLM use-cases and implementation paradigms in scientific work.

\begin{figure}[t]
\centering
\includegraphics[width=0.99\textwidth]{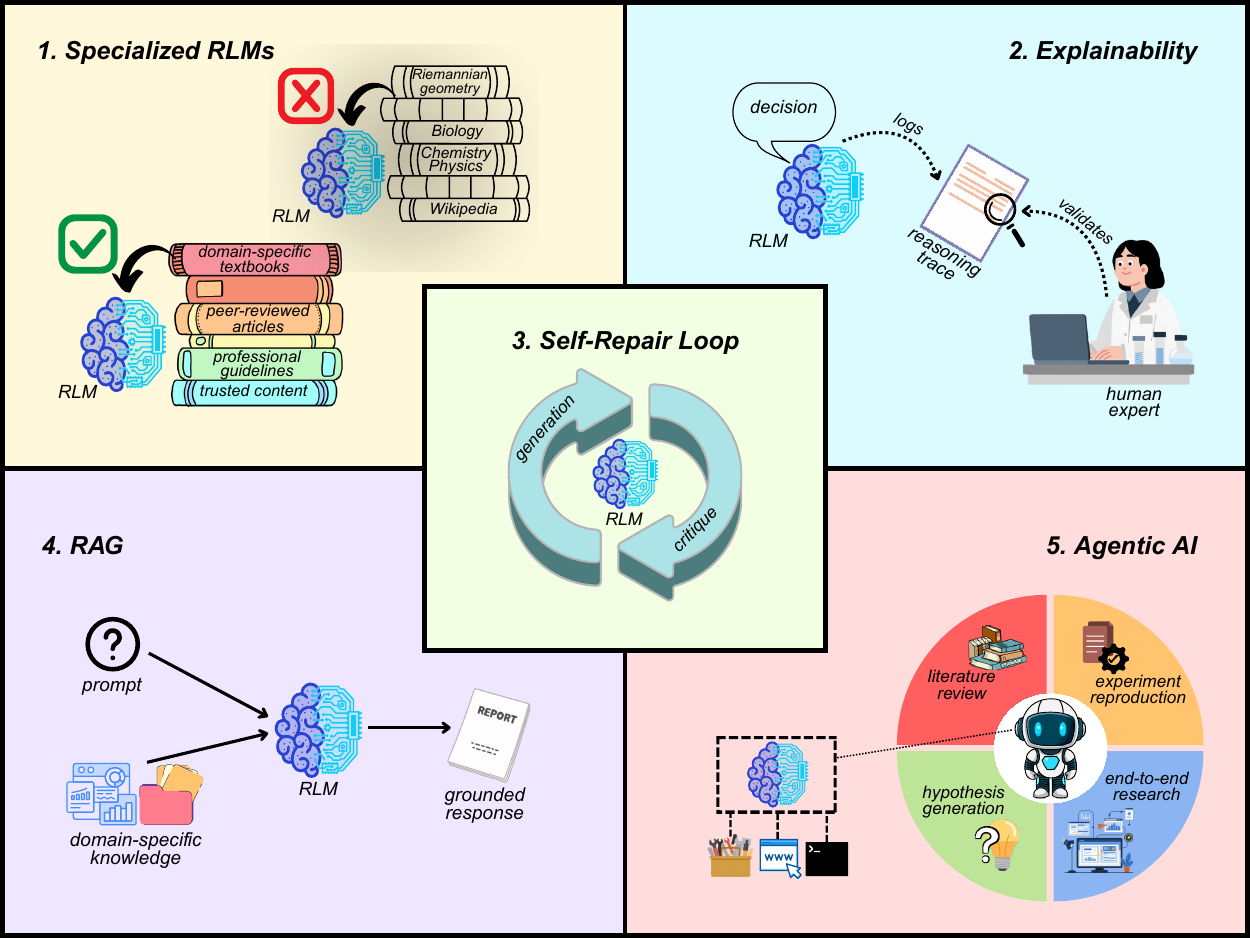}
\caption{Current trends of RLM usage in science: (1) Specialized models over general, all-purpose models; (2) Explainability via reasoning trace; (3) Self-repair loop; (4) Retrieval-Augmented Generation (RAG) with domain-specific knowledge; (5) Various use-cases of agentic AI for scientific work.}
\Description{Current trends of RLM usage in science: (1) Specialized models over general, all-purpose models; (2) Explainability via reasoning trace; (3) Self-repair loop; (4) Retrieval-Augmented Generation (RAG) with domain-specific knowledge; (5) Various use-cases of agentic AI for scientific work.}
\label{fig:current_trends}
\end{figure}

\subsection{Specialized Models over General Models}
There is an emerging trend of developing specialized models for particular domains, not only in Math and Computer Science \cite{georgiev2025mathematical, shao2024deepseekmath, ying2024internlm, glm2026_glm5}, but also in Healthcare, Chemistry, Astronomy, Finance, and even Agriculture: BioReason~\cite{fallahpour2025bioreason} integrates DNA embeddings and is trained using Group Relative Policy Optimization (GRPO) \cite{shao2024deepseekmath} to predict KEGG pathways from raw genomic data; RetroDFM-R~\cite{zhang2025reasoning} was trained using reinforcement learning with chemically verifiable rewards for chemical retrosynthesis; AstroSage-70B~\cite{haan2025_astro_astrosage} is designed for high-level Q\&A and scientific research automation in astronomy; Fin-R1~\cite{liu2025finr1} for financial reasoning tasks, including compliance checking, robo-advisory, and financial decision support; AgThinker~\cite{arxiv2025_19259} performs multi-step deductive reasoning over highly variable constraints to navigate complex, real-world farming scenarios.

In these studies, specialized models with lower parameter counts often outperform general, all-purpose models with massive sizes. This parameter-to-performance efficiency paradox is not surprising due to several reasons. Firstly, general foundational models dedicate a vast portion of their parameter capacity to web-crawled data, multilingual text generation, and casual conversational dynamics, which might not be essential for the domain-specific task at hand. Specialized models, by contrast, focus their latent space entirely on domain-specific vocabulary, symbolic representations, and technical syntax. This eliminates the "semantic dilution" that often plagues larger models. Secondly, when an RLM is trained exclusively within a singular discipline, its verification mechanisms can be anchored on concrete ground truths, such as mathematical solvers or precise chemical databases. This enables the model to learn to generate grounded reasoning traces rather than wandering through loose semantic associations. Thirdly, the internal search paths of a specialized model are strictly bound by domain rules; combined with the reduced semantic dilution issue, this allows the model to arrive at conclusions using fewer computational layers, yielding faster and more efficient outputs.

\subsection{Reasoning Trace for Explainability}
\label{sec:explainability}
One major advantage of RLMs over traditional LLMs is the explicit step-by-step thinking process in natural language, which makes it easier to audit the model's decision-making. The reasoning capability transforms such language model from opaque black box into explainable and interpretable inference engine. This benefit has been noted across various domains. For example, in mathematics and computer science, causal and symbolic verification is used to evaluate reasoning traces in order to combat "parametric memorization", a problem where a model might solve a problem despite flawed logic simply because it encountered the solution in its training data \cite{yu2025causal, khatibi2025eefsuva}. A similar problem has also been observed in medical diagnosis, where an RLM achieved 95\% accuracy but 68\% of its logic was clinically incorrect \cite{maharana2025rightpredictionwrongreasoning}, in which case reasoning trace verification becomes vital. Furthermore, in clinical settings, experts use these reasoning traces to ensure that the model has not omitted essential diagnostic or therapeutic steps (e.g., checking for contraindications in treatment planning), and to check for alternative diagnostic pathways \cite{Liu2025, Qiu2025}. Frameworks like DrugAgent provide a deterministic trail that shows how external evidence from knowledge graphs and search tools was synthesized to assess biological plausibility \cite{Inoue2025-zc}.

Nevertheless, the current reasoning capability of RLMs is not perfect. As mentioned above, the reasoning trace is occasionally unfaithful, meaning some of the generated reasoning steps do not genuinely reflect the model's internal computational paths or may misalign with the eventual conclusion \cite{maharana2025rightpredictionwrongreasoning, Queen2025CGBENCH, meadows2024exploring, phillips2026synthpert}. This behavior typically occurs when the model is already biased toward a specific outcome due to training data frequencies, yet it still generates an eloquent rationalization simply because it has been trained to output a reasoning trace. Still, an imperfect explanation is better than no explanation at all, especially for fields that involve high-stake risks and strict accountability like healthcare, law, and finance; the existence of such explanation itself enables \textit{auditing}. While human experts were historically reluctant to adopt traditional LLMs due to their completely opaque, black-box nature, they are demonstrably more open to integrating RLMs into their professional workflows due to the concrete possibility of conducting step-by-step validation \cite{neurips2023_5abcdf8, dai2025thyroid, Mansoor2025, kwon2024largelanguagemodelsclinical}.

\subsection{Self-Repair Loop}
A notable trend across various disciplines is putting a reasoning model in a \textit{self-repair loop}, taking advantage of the context understanding capability of the generative model to iteratively refine its response. In general, the self-repair loop can be done in two ways. The first approach relies on external validation telemetry; when the execution environment detects an issue (e.g., error, failure, mismatch), the system triggers an automatic re-routing of the diagnostic log along with the previously generated output back into the model's context window to prompt a targeted correction. Examples include the Reasoning Interleaved with Coding (RICO) method \cite{ying2024internlm} and CodePDE \cite{li2025codepde} in math; OpenFOAMGPT \cite{Pandey2025OpenFOAM} and AutoCFD \cite{Dong2025Finetuning} in physics; the Mephisto framework \cite{sun2025_astro_mephisto} in astronomy; MDCrow \cite{campbell2025mdcrow} in chemistry; and scPilot\cite{gao2025scpilot} in biology. The second approach utilizes internal critique mechanisms, where the initial output is immediately fed back into the system itself for self-evaluation. In this approach, the system acts as its own adversarial critic, evaluating its preliminary response against some pre-defined rubric or logical constraints in order to resolve potential contradictions before the final generation. To optimize compute efficiency and prevent infinite recursion, a fixed threshold is often implemented to limit the maximum number of permissible loops \cite{kim2023_tasks, lim2024erdframeworkimprovingllm}. Some systems implement sub-agents with distinct roles (which can be the same model given different personas via context engineering, or different models within the same framework) in the loop, such as "Thinker-Oracle" \cite{rui2025cardiocothierarchicalreasoningmultimodal}, "Proposer-Critic" \cite{ghafarollahi2025sparksmultiagentartificialintelligence}, or "Planner-Executor-Verifier" \cite{yu-etal-2025-sta}.

The self-repair loop approach works because an RLM mimics the human capacity to deliberate and evaluate their own thinking. Instead of forcing a rigid, one-shot inference where an early error cascades throughout the entire response, the architecture generates an explicit intermediate path. The generated reasoning trace acts as highly dense context that more effectively grounds the model's behavior toward specified target goals or domain-specific criteria. When the RLM parses its own reasoning tokens in a subsequent loop, it reads those intermediate steps as a temporal working memory. This allows the system to identify the moment its logical chain deviated from being correct and consequently alter its trajectory, mirroring the recursive drafting process employed by human.

\subsection{Retrieval-Augmented Generation (RAG)}
Originally proposed in 2020, Retrieval-Augmented Generation (RAG) \cite{lewis2020_RAG} has been widely utilized to ground a generative model's knowledge by supplying the input with some additional context retrieved from a separate information retrieval mechanism. There are also more recent subsequent methods that were inspired by it, such as GraphRAG \cite{graphrag}, Self-RAG \cite{selfrag}, Corrective RAG \cite{correctiverag}, and MedRAG \cite{medrag}, developed to handle complex, multi-hop queries by structuring corporate knowledge into semantic entity graphs or allowing models to recursively evaluate the quality of retrieved documents. When combined with RLMs, RAG functions as a verifiable external memory workspace. Instead of forcing the model to rely entirely on its static, pretrained parameters---which frequently lack niche scientific updates---the RLM can execute its multi-step deliberation directly over high-fidelity, peer-reviewed source materials. Many works across the 28 ERC disciplines have already successfully complemented reasoning models with these RAG-based architectures to dramatically elevate factual grounding and minimize logical drift \cite{lyu2026knowledgeaugmented, wu2025perturbqa, medrag, zhang2025lightchem, alhasan2026circuitlm, yang2026_CTIThinker, mdpiarch2025, kostka2024synergizing}.

\subsection{Agentic AI for Scientific Work}
\label{sub:agent}
Scientific agents with RLMs at their cores are being deployed at varying capacities across academic workflows, evolving from conversation-oriented utilities into autonomous software entities capable of goal-directed behavior. These agents are transforming scientific workflows across a wide spectrum of complexity, from automating the tedious task of literature review to experimentally conducting the entire research pipeline.

\textbf{Automatic Literature Review}. Leveraging RLMs' planning capability via deliberation and self-reflection loops, specialized agents can automatically query global academic indexes, parse hundreds of relevant papers, and synthesize vast corpora of knowledge. Rather than merely summarizing abstracts, these agents are capable of analyzing methodology sections and mapping out conflicting empirical findings. For example, the MEDEA framework \cite{Sui2026Medea} for therapeutic discovery utilizes a multi-module architecture for research planning and reasoning over vast amounts of scientific literature, while SciAgents \cite{ghafarollahi2024sciagents} uses multi-agent orchestration to perform graph reasoning, allowing it to uncover connections between papers and identify knowledge gaps that single models might miss.

\textbf{Automatic Experiment Reproduction}. Scientific agents are tasked with reading the methodology sections of published papers, extracting experimental hyperparameters, and generating replication scripts. Such agents often operate in isolated software sandboxes for safety, and they troubleshoot compilation errors via internal self-repair loops. Then, they automatically cross-examine the acquired results against the metrics published in the original study to verify reproducibility. As an example, the SimAgents framework \cite{zhang2025_astro_litrev_agent} extracts parameter configuration from literature and conduct preliminary analysis for cosmology research. Additionally, the ReplicationBench \cite{ye2025_astro_replicationbench} benchmark is specifically designed to evaluate the capability of AI agents to replicate scientific experiments from research papers, particularly in the field of astrophysics.

\textbf{Automatic Hypothesis Generation and Validation}. Reasoning-capable agents can uncover non-obvious correlations across data extracted from external knowledge bases. The reasoning capability allows them to formulate logically sound hypotheses from these insights, which then can be automatically validated or invalidated by executing targeted data analysis and checking them against known scientific constraints. Examples include PersonaAI \cite{Cho2026PersonaAI} for autonomous hypothesis generation and validation in the study of aging; LLM4SD \cite{zheng2023large}, a pipeline that performs in-context reasoning to infer interpretable molecular property prediction rules from scientific datasets; and the Creation Game framework \cite{rodriguez2025vaccinology} which generates hypotheses, designs experiments, and infers biological principles in vaccinology.

\textbf{Fully Independent Research Agent}. Representing the absolute peak of autonomous scientific infrastructure, fully independent research agents execute closed-loop, end-to-end discovery cycles. These advanced agentic systems ingest background literature, formulate testable hypotheses, write and execute analytical code, interpret the resulting data, and independently compile their discoveries into a manuscript. At the moment, these fully independent research agents are still considered experimental, but recent demonstrations look promising. Examples include: CMBAgent \cite{xu2025_astro_cmbagent}, which measures cosmological parameters from supernova data with no human-in-the-loop; EarthLink \cite{guo2025earthlink}, a multi-agent system for climate science that automates the research workflow from planning, code generation, data analysis, visualization, until the production of scientific summaries; the Sparks framework \cite{ghafarollahi2025sparksmultiagentartificialintelligence}, which also executes the entire scientific research pipeline through hypothesis generation, experiment design, iterative refinement, and documentation. However, while these agents look promising in purely computational research, the seamless integration with physical laboratory equipment remains a conceptual frontier that has not yet been practically achieved at scale.

\section{Current Challenges of RLM Adoption in Science}

\begin{figure}[hb]
\centering
\includegraphics[width=0.99\textwidth]{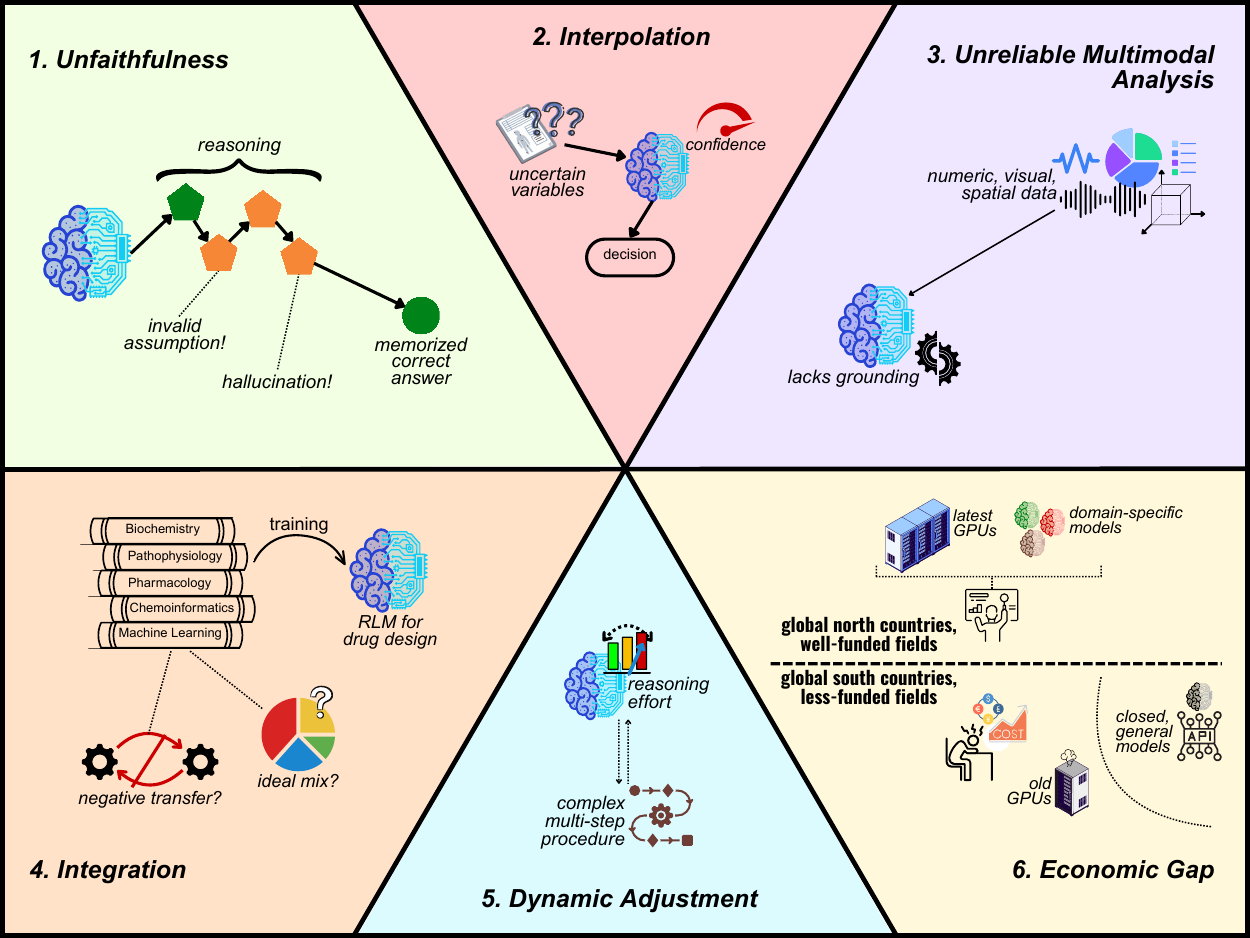}
\caption{Current challenges of RLM adoption in science: (1) Hallucination \& unfaithful reasoning; (2) Limited interpolative analysis; (3) Unreliable multimodal data analysis; (4) Difficulty in integrating related domains; (5) Dynamic reasoning effort adjustment; (6) Economic gap across scientific disciplines, also between global north and global south.}
\Description{Current challenges of RLM adoption in science: (1) Hallucination \& unfaithful reasoning; (2) Limited interpolative analysis; (3) Unreliable multimodal data analysis; (4) Difficulty in integrating related domains; (5) Dynamic reasoning effort adjustment; (6) Economic gap across scientific disciplines, also between global north and global south.}
\label{fig:current_challenges}
\end{figure}

Despite the advancement of reasoning models in some fields, their adoption is still hindered by a wide range of challenges. In this section, we enumerate such challenges that we found across various disciplines.

\subsection{Hallucination and Unfaithful Reasoning}
Even with RAG and self-repair mechanisms, reasoning models may still generate invalid information. The issue of hallucination is very challenging to address because it can be caused by a wide range of factors; however, one major factor can be attributed to the very limited amount of domain-specific training data in some disciplines. Furthermore, many studies have also highlighted the problem of \textit{unfaithful reasoning}; some of the generated reasoning steps sometimes do not fully align with each other or with the eventual conclusion \cite{maharana2025rightpredictionwrongreasoning, Queen2025CGBENCH, meadows2024exploring, phillips2026synthpert}. As described previously in Section \ref{sec:explainability}, such behavior occurs when the model is already biased towards a specific outcome due to parametric memorization, yet it still needs to generate some reasoning tokens because it is trained to do so. Therefore, validation by human expert is always crucial for dealing with these problems.

\subsection{Limited Interpolative Analysis}
State-of-the-art reasoning models are already considered PhD-level knowledge synthesizers in some fields, even reaching gold-medal performance in International Olympiad in Informatics (IOI) \cite{deepseek2025_v32}, International Mathematical Olympiad (IMO) \cite{luong2025imo}, International Olympiad on Astronomy \& Astrophysics (IOAA) \cite{pinheiro2025_astro_olym}, and outperforming 94\% of expert virologists on the Virology Capabilities Test (VCT) \cite{gopal2025vct}. They are capable of combining existing theories to answer a question. However, expert-level scientific reasoning that leads to genuinely novel scientific theories or discoveries remains a frontier challenge.

Recent work has tried to push this frontier. Experiments with GPT-5 reportedly allowed proving new bounds on the performance of convex body chasing algorithms and solving Erdős Problem \#848 \cite{openai2025_gpt5science}; AlphaEvolve rediscovered the best known solutions and discovered improved solutions in 67 math problems \cite{georgiev2025mathematical}; the Sparks framework reportedly identified a length-dependent mechanical crossover in peptides which was previously uncharacterized in scientific literature \cite{ghafarollahi2025sparksmultiagentartificialintelligence}. Yet, it is important to point out that current systems still heavily rely on human oversight that guides the thinking process via prompting.

Furthermore, while such models seem to be capable of simulating scientific discussion, they are not yet able to effectively handle ambiguity or the lack of ground truth at the level of human experts. For example, in archaeology, interpretation often requires reasoning with incomplete evidence and multiple hypotheses, and current models frequently struggle to move beyond simple knowledge retrieval \cite{mdpiarch2025}. On the other hand, reasoning-tuned models exhibited "systematic overconfidence" in clinical tasks that require flexible judgment due to uncertainty inherent in complex infection diagnosis \cite{mccoy2025assessment}; they also perform significantly worse in dynamic clinical settings where they must actively determine which additional diagnostic tests are needed to reduce uncertainty, a process requiring strategic information gathering rather than passive inference from provided data \cite{Qiu2025}.

\subsection{Unreliable Multimodal Data Analysis: Numerical, Spatial, and Visual}
State-of-the-art reasoning models excel at factual retrieval and standardized theoretical examinations, yet they are prone to symbolic perturbation \cite{zaki2023mascqa, meadows2024exploring, li2025atmosscibench} and consistently struggle with complex judgment involving numerical, spatial, or visual reasoning \cite{wu2025perturbqa, miret2024enabling, du2024large, kevian2024capabilities, zhu2025agenteis, li2025eeebench, zhang-etal-2025-cultural}. Current methodologies often struggle with precise grounding of chain-of-thought tokens into exact numerical or spatial coordinates, which leads to a catastrophic propagation of errors during long-horizon inferences. This deficiency primarily stems from several structural bottlenecks. First, conventional subword tokenization often splits multi-digit numbers into fragmented tokens, which disrupts the model's ability to maintain a coherent internal representation of mathematical scale, magnitude, and precision. Second, text-trained reasoning models lack inherent geometric grounding; their latent spaces are designed to map semantic relationships by nature rather than numerical or spatial coordinates. Third, current vision-language interfaces suffer from resolution constraints and alignment gaps that cause them to overlook fine-grained visual features, such as exact grid lines, overlapping data points, or specialized scientific symbols in complex charts and plots. Consequently, these compounding limitations prevent reasoning models from reliably transforming raw quantitative, spatial, or visual data into rigorous scientific deductions.

Some contemporary frameworks are shifting toward tightly coupled vision-language architectures with dense step-by-step reasoning pathways in order to enable systematic verification of visual anomalies and numerical outliers \cite{liu2025skinr1trustworthyclinicalreasoning, yu-etal-2025-sta, zhang2025sdiglmleveraginglargelanguage, koksal2025samchat, zhu2025agenteis, guo2025earthlink}. However, achieving true multi-modal alignment, where numerical precision and spatial topology are intrinsically preserved within the model's latent reasoning loop rather than treated as secondary inputs, remains an open and vital area of research.

Alternatively, utilizing its explicit reasoning chain as controller, an RLM can decompose intricate numerical or spatial-visual problems into discrete, auditable tool invocations. This way, the RLM itself does not need to perform the actual processing of such data; it only needs to understand the expected inputs and outputs of the tools. Current models already have the capability to use external tools (e.g., Python code interpreters \cite{ying2024internlm, chen2022program, chung2025theoretical}, symbolic algebra systems \cite{feng2025physics, song2025llm}, other domain-specific specialized solvers \cite{gao2025scpilot, jacob2025proteinlanguagemodelsagentic, doi:10.1021/acs.analchem.4c05039, mudur2025feabench}). There are also propositions centered on digital sketchpads or scratchpads that can be used by RLMs to perform explicit numerical, spatial, or visual calculations \cite{pinheiro2025_astro_olym, guan2025cadcodertexttocadgenerationchainofthought, zhang2025sdiglmleveraginglargelanguage}. Yet, RLMs still struggle with the complex multi-step deduction that is required to plan, execute, and correlate the tool outputs effectively over long-horizon inferences.

\subsection{Difficulty in Integrating Related Domains}
Since the pretraining corpora of foundational models typically contained little to no domain-specific knowledge, a reasoning model specialized for a particular discipline is typically created by performing domain adaptation (by either supervised fine-tuning or reinforcement learning) using a relatively narrow domain-specific dataset. Yet, existing specialized models are often limited to singular domains; for example, DrugReason \cite{ghaffarzadehesfahani2025drugreasonerinterpretabledrugapproval} is focused on predicting small-molecule approval in drug design, while BioReason \cite{fallahpour2025bioreason} is designed for reasoning over genomic information. In the real world, knowledge from different domains or sub-domains is often needed for answering research questions of higher complexity levels. In order to enable a truly independent research agent for holistic discovery, integrating interrelated fields is necessary.

Overcoming this problem is not as simple as developing as many datasets as possible. We also need examples that show how to solve a research problem by using knowledge from different fields. Not to mention that identifying the ideal data mix for training is also a particular challenge; various studies have noted that \textit{multi-task learning} can hurt overall performance when the task objectives are in conflict. Prior research on multitask learning and cognitive interference has shown that jointly optimizing competing objectives can produce negative transfer effects, particularly when tasks rely on conflicting representations, attentional strategies, or optimization gradients \cite{ni2023-multitask, wu2020-multitask}. Identifying the tasks that actually support each other's performance and determining the ideal data mix is a challenge towards a truly independent research agent for genuine scientific discovery.

\subsection{Dynamic Reasoning Effort Adjustment}
\label{sec:reasoning_effort}
One major challenge of using RLMs in particular scientific domains is their high cost. A reasoning model needs to generate numerous intermediate tokens during their internal deliberation phase, which directly results in longer computation times and higher API operational costs. Furthermore, some studies have also noted that \textit{reasoning} is not necessarily always beneficial: researchers have identified a "reasoning length paradox"---often aptly called the "overthinking phenomenon"---where longer reasoning traces are positively correlated with increased probability of logical error \cite{sui2025overthinking, frontiers2025deepseek}; in creative writing, explicit reasoning traces were observed to clutter or disrupt the model's ability to make holistic qualitative assessments \cite{litbench2025}.

Because of this overhead, many works have tried to develop an automated routing mechanism to dynamically select between a traditional, cost-effective LLM for simple queries and an advanced RLM for more complex analytical tasks \cite{trivedi2024dual}; alternatively, they developed a model that supports both reasoning and non-reasoning use-cases, allowing the user to toggle between them \cite{zuo2025biomedgptmolmultitasklearningmolecular, qwen2026_35, justen2025biology}. Smarter, more sophisticated training methods have also been proposed to teach these models to reason more concisely and efficiently \cite{fallahpour2025bioreason, sim2025unveiling}. More recently, reasoning models were taught to adaptively adjust their reasoning effort depending on the difficulty of the task at hand \cite{kleinman2025adaptive, yang2026adaptive}.

\subsection{Efficiency, Economic Gap, and Technology Independence}
In the bigger picture, algorithmic and architectural solutions mentioned in Section \ref{sec:reasoning_effort} only partially solve the problem of the high cost of RLM usage. There are still glaring inequality gaps in terms of access to specialized computational hardware, such as high-end GPUs or other processing units necessary for running generative AI models efficiently,not only across different scientific disciplines but also between the global north and the global south. For fields that do not typically invest heavily in such computational infrastructure, developing and deploying their own domain-adapted reasoning models can be prohibitively expensive. Similarly, scholars from the global south do not have easy or equitable access to the latest hardware; even with the availability of cloud services, the cost after currency conversion remains entirely unsustainable for local institutional budgets \cite{sathish2024inequality}. For them, utilizing closed-weight reasoning models from major commercial providers such as OpenAI, Google, and Claude represents the more affordable and practical option \cite{gao2025scpilot}. Yet, this operational reliance comes with its own severe set of issues, primarily concerning data privacy and a dangerous technological dependence on private, third-party software vendors \cite{ke2025frontiers, machado2025public}.

\section{Future Directions for RLM Adoption in Science}
To unlock the full potential of RLMs across scientific research, development must evolve toward seamless knowledge synthesis across complex, heterogenous workflows. Bridging the productivity gap between pioneering and leading disciplines requires a concerted effort toward physical-digital integration, operational autonomy, and structured validation. We identify several crucial vectors for future research:

\begin{figure}[t]
\centering
\includegraphics[width=0.99\textwidth]{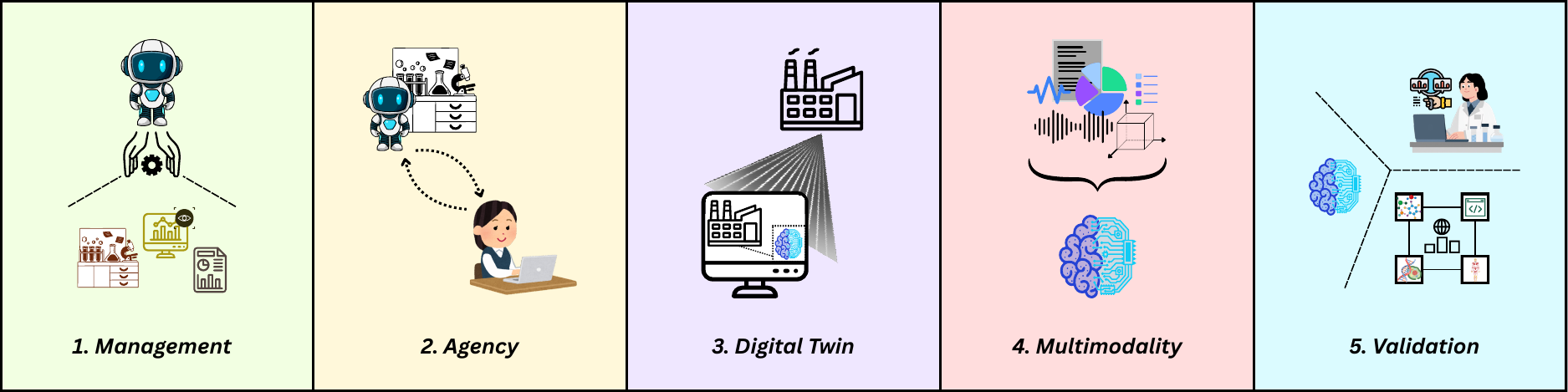}
\caption{Future directions for RLM adoption in science: (1) Management of scientific procedures; (2) Extension of AI agency; (3) Exploitation of RLMs in the implementation of digital twins; (4) Grounding on multimodal scientific data; (5) Standardizing open-science benchmarks and human-in-the-loop validation.}
\Description{Future directions for RLM adoption in science: (1) Management of scientific procedures; (2) Extension of AI agency; (3) Exploitation of RLMs in the implementation of digital twins; (4) Grounding on multimodal scientific data; (5) Standardizing open-science benchmarks and human-in-the-loop validation.}
\label{fig:future_directions}
\end{figure}

% \textbf{Management of Scientific Procedures and Extension of AI Agency}. 
\textbf{Management of Scientific Procedures}.
Moving beyond simple prompt-response dynamics, the next milestone in scientific infrastructure involves expanding multi-agent RLM orchestrations to manage complex long-horizon workflows. Future research must focus on optimizing specialized roles within agentic frameworks, such as separating long-term project planning, code verification, and multi-modal evidence synthesis into discrete sub-agents. These advanced agentic systems should also autonomously coordinate the ingestion of scientific literature, systematically cross-examine conflicting empirical findings across, and carefully manage error propagation in self-repair loops. The expansion of AI agency will enable automated maintenance and evaluation, even in highly dynamic interdisciplinary fields.

\textbf{Extension of AI Agency}.
Finally, the ultimate frontier for RLM adoption is the seamless bridge between "in silico" computational reasoning and "in vitro" physical execution. Current state-of-the-art research agents remain heavily constrained to digital sandboxes, executing code or manipulating data streams in isolation. True self-driving laboratories will require RLMs to function as central real-time controllers that directly interface with various laboratory hardware. Utilizing their explicit multi-step deliberation paths, these models will not only formulate and dynamically refine biological or chemical hypotheses, but will also actively write precise operational instructions to synthesize compounds, run physical assays, and evaluate real-world data streams. Achieving this closed-loop, automated discovery paradigm requires developing multi-modal reasoning models that are fundamentally grounded in physical limitations, geometric boundaries, and safety constraints.
Additionally, AI agents can autonomously recruit and manage human participants, including research subjects, who can operate in the real world. This can be achieved through dedicated services such as \textit{rentahuman} \cite{RentAHuman}. % (https://rentahuman.ai/)

\textbf{Exploitation of RLMs in the Implementation of Digital Twins}. While current digital twins rely heavily on rigid numerical simulations and deterministic physics engines, integrating RLMs introduces a dynamic cognitive layer capable of qualitative reasoning over complex physical structures. Future architectures will leverage RLMs to interpret high-rate sensor telemetry, map real-time structural anomalies to causal variables, and reason over multi-physics boundary conditions. By treating an RLM as an intelligent orchestrator within a digital twin framework, the model can translate abstract mathematical variations into intuitive, natural language diagnostics or code-level adjustments. This synergy will allow digital twins to shift from passive, observational models to active, self-correcting simulators that can hypothesize, test, and explain structural behaviors in real-time.

\textbf{Grounding on Multimodal Scientific Data}. 
While RLMs have shown immense promise in mimicking analytical thinking, text-based reasoning loops alone are insufficient for capturing the whole complexity of scientific research. Truly integrating RLMs into the broader scientific infrastructure requires grounding their multi-step reasoning traces in diverse non-textual data modalities, such as high-resolution imaging, genomic sequences, 3D molecular structures, and continuous time-series data from real-time sensors. By explicitly linking linguistic deliberation with multimodal observations, next-generation RLMs can validate their internal hypotheses against real-world ground truths, moving beyond loose semantic associations and mitigating hallucination risks. Ultimately, multimodal grounding will transform RLMs from text-bound analytical assistants into unified scientific engines capable of cross-examining theories against raw empirical evidence across the physical, social, and life sciences.

\textbf{Standardizing Open-Science Benchmarks and Human-in-the-loop Validation}. As demonstrated by the widespread drop-off in domain-wise RLM maturity when accounting for public availability of resources, the open-science deficit severely amplifies the gap between scientific fields. To address this, future efforts must prioritize the development of open-source domain-specific resources, including rigorous testing suites that ensure the quality of the RLMs. This is particularly essential for the fields that are traditionally less computational-oriented. Furthermore, rather than evaluating models on shallow memorization, future benchmarks must assess higher-order scientific logic, such as an agent's capability to reason with incomplete or ambiguous data, manage strategic information gathering, and maintain causal consistency. Standardizing these public evaluation frameworks is vital to ensure that RLM innovations can be equitably accessed, verified, and utilized by researchers across all 28 branches of science.

Nevertheless, designing high-quality benchmarks and automated metrics is only one side of the evaluation equation. While automated benchmarks excel at quantifying raw model performance over standardized test beds, true evaluation also requires continuous human-in-the-loop validation to audit the qualitative soundness of multi-step reasoning traces. This human aspect acts as a critical safety and credibility gate, ensuring that model outputs are vetted for genuine scientific truth, ethical coherence, and practical safety before autonomous discoveries move from digital hypotheses to high-stakes execution.

\section{Conclusion}
In this survey, we have systematically investigated the landscape of RLM adoption across 28 scientific disciplines defined by ERC evaluation panels. Our analysis demonstrates that RLMs are already transforming research work in several fields by supporting complex reasoning, scientific knowledge synthesis, automated analysis, and increasingly agentic research workflows. However, the maturity of RLM ecosystems remains highly uneven across disciplines, which is further compounded by a critical open-science deficit. This disparity risks creating an increasingly unequal scientific environment in which access to advanced reasoning technologies becomes a major determinant of research productivity and innovation capacity.

Our review also highlights the current limitations of RLMs. While current trends showcase the immense potential of RLMs, substantial obstacles persist. Most critically, autonomous discovery remains constrained to computational simulations, leaving the integration of reasoning models with physical laboratories as a major unreached frontier. Addressing these challenges will require not only advances in model architectures and reasoning methodologies, but also broader efforts toward open resources, domain-specific as well as interdisciplinary data, and equitable access to computational infrastructure. If adapted responsibly, fairly, and transparently, RLMs have the potential become an essential, transformative component of future scientific infrastructure across all branches of knowledge.

\begin{acks}
This work was supported by: 
(1) the National Science Centre, Poland, project no. 2021/41/B/ST6/04471;
(2) CLARIN ERIC funded by the Polish Minister of Science, no. 2024/WK/01;
(3) The European Regional Development Fund, FENG programme, no. FENG.02.04-IP.040004/24;
(4) the EU project "DARIAH-PL", under investment A2.4.1 of the National Recovery and Resilience Plan (DOPI-KPO.61.12.3.2024.AD);
(5) the statutory funds of the Department of Artificial Intelligence, Wroclaw Tech;
(6) Polish HPC infrastructure PLGrid (HPC Center: ACK Cyfronet AGH): computer facilities and support within grant no. PLG/2024/017840;
(7) the Polish Ministry of Education and Science within the programme "International Projects Co-Funded";
(8) the European Union under the Horizon Europe, grant no. 101086321 (OMINO). However, the views and opinions expressed are those of the author(s) only and do not necessarily reflect those of the European Union or the European Research Executive Agency. Neither the European Union nor European Research Executive Agency can be held responsible for them.
\end{acks}

\bibliographystyle{ACM-Reference-Format}
\bibliography{main}

\appendix
\renewcommand*{\thesection}{\AlphAlph{\value{section}}}

\section{Typography of Scientific Disciplines}
\label{app:science-typography}

\begin{longtable}{|l|p{5.5cm}|p{7.6cm}|}
\caption{28 evaluation panels of the ERC across three main groups of scientific disciplines: social sciences and humanities (indicated by red background), physical sciences and engineering (indicated by blue background), and life sciences (indicated by green background).}
\label{tab:erc_typography} \\
\hline
\rowcolor{Gray}
\textbf{Code} & \textbf{Name} & \textbf{Topics} \\
\hline
\rowcolor{LightRed}
SH1 & Individuals, Markets and Organisations &
Economics, finance and management \\ \hline
\rowcolor{LightRed}
SH2 & Institutions, Governance and Legal Systems &
Political science, international relations, law \\ \hline
\rowcolor{LightRed}
SH3 & The Social World and Its Interactions &
Sociology, social psychology, education sciences, communication studies \\ \hline
\rowcolor{LightRed}
SH4 & The Human Mind and Its Complexity &
Cognitive science, psychology, linguistics \\ \hline
\rowcolor{LightRed}
SH5 & Texts and Concepts &
Literary studies, literature, philosophy \\ \hline
\rowcolor{LightRed}
SH6 & The Study of the Human Past &
Archaeology and history \\ \hline
\rowcolor{LightRed}
SH7 & Human Mobility, Environment, and Space &
Human geography, demography, health, sustainability science, territorial planning, spatial analysis \\ \hline
\rowcolor{LightRed}
SH8 & Studies of Cultures and Arts &
Social anthropology, studies of cultures, studies of arts \\ \hline
\rowcolor{LightBlue}
PE1 & Mathematics &
All areas of mathematics, pure and applied, plus mathematical foundations of computer science, mathematical physics and statistics \\ \hline
\rowcolor{LightBlue}
PE2 & Fundamental Constituents of Matter &
Particle, nuclear, plasma, atomic, molecular, gas, and optical physics \\ \hline
\rowcolor{LightBlue}
PE3 & Condensed Matter Physics &
Structure, electronic properties, fluids, nanosciences, biophysics \\ \hline
\rowcolor{LightBlue}
PE4 & Physical and Analytical Chemical Sciences &
Analytical chemistry, chemical theory, physical chemistry/chemical physics \\ \hline
\rowcolor{LightBlue}
PE5 & Synthetic Chemistry and Materials &
New materials and new synthetic approaches, structure-properties relations, solid state chemistry, molecular architecture, organic chemistry \\ \hline
\rowcolor{LightBlue}
PE6 & Computer Science and Informatics &
Informatics and information systems, computer science, scientific computing, intelligent systems \\ \hline
\rowcolor{LightBlue}
PE7 & Systems and Communication Engineering &
Electrical, electronic, communication, optical and systems engineering \\ \hline
\rowcolor{LightBlue}
PE8 & Products and Processes Engineering &
Product and process design, chemical, civil, environmental, mechanical, vehicle engineering, energy processes and relevant computational methods \\ \hline
\rowcolor{LightBlue}
PE9 & Universe Sciences &
Astro-physics/-chemistry/-biology; solar system; planetary systems; stellar, galactic and extragalactic astronomy; cosmology; space sciences; astronomical instrumentation and data \\ \hline
\rowcolor{LightBlue}
PE10 & Earth System Science &
Physical geography, geology, geophysics, atmospheric sciences, oceanography, climatology, cryology, ecology, global environmental change, biogeochemical cycles, natural resources management \\ \hline
\rowcolor{LightBlue}
PE11 & Materials Engineering &
Advanced materials development: performance enhancement, modelling, large-scale preparation, modification, tailoring, optimisation, novel and combined use of materials, etc. \\ \hline
\rowcolor{LightGreen}
LS1 & Molecules of Life: Biological Mechanisms, Structures and Functions &
Molecular biology, biochemistry, structural biology, molecular biophysics, synthetic and chemical biology, drug design, innovative methods and modelling \\ \hline
\rowcolor{LightGreen}
LS2 & Integrative Biology: from Genes and Genomes to Systems &
Genetics, epigenetics, genomics and other ‘omics studies, bioinformatics, systems biology, genetic diseases, gene editing, innovative methods and modelling, ‘omics for personalised medicine \\ \hline
\rowcolor{LightGreen}
LS3 & Cell Biology, Development, Stem Cells and Regeneration &
Structure and function of the cell, cell-cell communication, embryogenesis, tissue differentiation, organogenesis, growth, development, evolution of development, organoids, stem cells, regeneration, therapeutic approaches \\ \hline
\rowcolor{LightGreen}
LS4 & Physiology in Health, Disease and Ageing &
Organ and tissue physiology, comparative physiology, physiology of ageing, pathophysiology, interorgan and tissue communication, endocrinology, nutrition, metabolism, interaction with the microbiome, non-communicable diseases including cancer (and except disorders of the nervous system and immunity-related diseases) \\ \hline
\rowcolor{LightGreen}
LS5 & Neuroscience and Disorders of the Nervous System &
Nervous system development, homeostasis and ageing, nervous system function and dysfunction, systems neuroscience and modelling, biological basis of cognitive processes and of behaviour, neurological and mental disorders \\ \hline
\rowcolor{LightGreen}
LS6 & Immunity, Infection and Immunotherapy &
The immune system, related disorders and their mechanisms, biology of infectious agents and infection, biological basis of prevention and treatment of infectious diseases, innovative immunological tools and approaches, including therapies \\ \hline
\rowcolor{LightGreen}
LS7 & Prevention, Diagnosis and Treatment of Human Diseases &
Medical technologies and tools for prevention, diagnosis and treatment of human diseases, therapeutic approaches and interventions, pharmacology, preventative medicine, epidemiology and public health, digital medicine \\ \hline
\rowcolor{LightGreen}
LS8 & Environmental Biology, Ecology and Evolution &
Ecology, biodiversity, environmental change, evolutionary biology, behavioural ecology, microbial ecology, marine biology, ecophysiology, theoretical developments and modelling \\ \hline
\rowcolor{LightGreen}
LS9 & Biotechnology and Biosystems Engineering &
Biotechnology using all organisms, biotechnology for environment and food applications, applied plant and animal sciences, bioengineering and synthetic biology, biomass and biofuels, biohazards \\ \hline
\end{longtable}

\section[Literature Review Summary: SH1]{Literature Review Summary: SH1}

\textbf{Individuals, Markets and Organisations.} Economics, finance and management.

\subsection{Key Findings}
In economics, finance, and management, the reasoning performed by large language models typically involves causal reasoning, decision reasoning, strategic reasoning, and analytical reasoning over structured and unstructured data rather than purely mathematical problem solving. The reviewed literature indicates that research on reasoning with large language models is still mainly organized around domain-specific datasets and structured prompting strategies. Research on dedicated reasoning models usually involves domain-specific fine-tuning.  The existing work differs in terms of datasets and benchmarks used, reasoning model architectures, modalities, prompting strategies, output formats, and approaches to reasoning validation though most of the works seem to validate only the model output with barely any consideration put to the evaluation of actual reasoning process. Although some datasets and benchmarks are more prevalent in existing research than others, there are little to no studies comprehensively comparing the existing solutions to each other.

\subsubsection{Datasets and Benchmarks for Reasoning}

A key finding from the reviewed studies is that domain-specific reasoning research relies heavily on specialized datasets and benchmarks. In the economics domain, two important benchmarks are EconNLI \cite{guo-yang-2024-econnli}, which evaluates causal reasoning over economic events, and STEER-ME \cite{raman2025steermeassessingmicroeconomicreasoning}, which assesses microeconomic reasoning across a wide range of tasks, domains, and contexts. These benchmarks aim to evaluate whether models can perform structured reasoning rather than simple text generation.

In financial reasoning, commonly used datasets include FinQA, TAT-QA, and related financial question answering datasets, which are used for training and evaluating specialized reasoning models such as Fin-R1 \cite{liu2025finr1} and TAT-LLM \cite{zhu2024tatllm}. In business and entrepreneurship analytics, CrunchLLM \cite{sadia2025crunchllmmultitaskllmsstructured} uses the Crunchbase dataset combining structured company attributes with unstructured textual descriptions for prediction and reasoning tasks.

Overall, the literature shows that reasoning research is strongly benchmark-driven, and progress is often measured through performance on domain-specific reasoning datasets rather than general NLP benchmarks. Notably, existing benchmarks focus on output-based evaluations. Thus the reasoning capabilities of the model are usually measured only based on the correctness of the answers to various reasoning problems. 

\subsubsection{Reasoning Models and Architectures}

Another important finding is the increasing development of specialized reasoning models rather than relying solely on general-purpose LLMs. Several papers propose domain-adapted reasoning models trained through fine-tuning, post-training, or reinforcement learning.

Examples include Fin-R1 \cite{liu2025finr1}, a financial reasoning model trained using supervised fine-tuning and reinforcement learning, and Recon (Reasoning Like an Economist) \cite{zhou2025reasoninglikeeconomistposttraining}, which is post-trained on economic reasoning problems to improve strategic reasoning capabilities. TAT-LLM \cite{zhu2024tatllm} is another example of a specialized model designed for discrete reasoning over tabular and textual financial data.

Other works focus not on a single model but on system architectures that enhance reasoning. For example, the dual-model architecture proposed by Trivedi et al. \cite{trivedi2024dual} combines a code-generating model for structured data retrieval with a foundation model performing reasoning and analysis. Multi-agent reasoning architectures are also explored, for example in the SynergyMAS framework \cite{kostka2024synergizing}, where multiple agents collaborate to perform logical reasoning and knowledge management tasks.

These studies suggest a clear trend toward domain-specialized reasoning models and multi-component reasoning systems rather than relying solely on general-purpose LLMs.

\subsubsection{Modalities Used in Reasoning Tasks}

Most of the existing work focuses primarily on text-based reasoning tasks. Benchmarks such as EconNLI and STEER-ME are entirely text-based and evaluate reasoning through natural language problem descriptions and answers.

However, some works already incorporate additional modalities or structured data. For example, TAT-LLM \cite{zhu2024tatllm} operates on both tabular and textual financial data, requiring the model to perform discrete reasoning over structured numerical information and textual explanations. Similarly, CrunchLLM \cite{sadia2025crunchllmmultitaskllmsstructured} combines structured business data with textual descriptions.

The dual-model business reasoning framework \cite{trivedi2024dual} also integrates structured data retrieval and visual reasoning, indicating that reasoning tasks increasingly involve multimodal or structured inputs rather than purely textual prompts.

\subsubsection{Prompting Strategies and Output Formats}

The reviewed literature shows that prompting strategies play a crucial role in reasoning performance. Many works use chain-of-thought prompting, structured prompts, or multi-step prompting procedures to guide the reasoning process.

Some studies rely on relatively simple prompting strategies such as zero-shot, few-shot, or chain-of-thought prompting. However, other works use more structured prompting approaches. For example, TAT-LLM \cite{zhu2024tatllm} uses a step-wise reasoning pipeline consisting of Extractor, Reasoner, and Executor steps. Fin-R1 \cite{liu2025finr1} uses structured prompts for reasoning data generation and answer verification.

In terms of outputs, different works expect different types of responses from models. Some tasks use closed answer sets, such as classification labels or multiple-choice answers (e.g., STEER-ME and EconNLI). Other tasks require short textual explanations or generated consequences. More advanced approaches require structured outputs, such as step-by-step reasoning traces, tables with intermediate results, or structured reasoning pipelines.

This suggests that reasoning tasks often benefit from structured output formats rather than fully open-ended text generation.

\subsubsection{Reasoning Validation and Evaluation}

Another important aspect concerns how reasoning is evaluated and validated. In many works, reasoning quality is evaluated indirectly through task accuracy or prediction performance. For example, EconNLI \cite{guo-yang-2024-econnli} evaluates whether models correctly identify causal relationships between economic events.

Other works evaluate not only final predictions, but also reasoning explanations and intermediate steps. CrunchLLM \cite{sadia2025crunchllmmultitaskllmsstructured} evaluates both prediction accuracy and explanation quality, while TAT-LLM \cite{zhu2024tatllm} validates reasoning through intermediate steps executed in a structured reasoning pipeline.

Some works also use model-based evaluation approaches, such as LLM-as-a-judge frameworks or structured reasoning verification pipelines, particularly in reinforcement learning–based reasoning models such as Fin-R1 \cite{liu2025finr1}.

\subsubsection{Reasoning Efficiency, Cost, and Effort}

Finally, the literature increasingly considers efficiency, cost, and computational effort associated with reasoning. Some works explicitly aim to reduce the cost of reasoning-based systems. For example, E-CARE \cite{zhang2025ecare} proposes a framework that captures common sense reasoning without requiring expensive real-time LLM inference. Fin-R1 \cite{liu2025finr1} also emphasizes the development of relatively small models that maintain reasoning performance while reducing deployment costs.

However, despite increasing attention to computational cost and efficiency, relatively few works explicitly analyze reasoning length, reasoning effort, or reasoning budget optimization. Most studies focus on improving reasoning accuracy rather than optimizing the trade-off between reasoning depth, cost, and latency. This indicates that reasoning efficiency and reasoning budget control remain important directions for future research.

\subsubsection{Summary of Key Findings}

Overall, the reviewed literature leads to several key conclusions. First, reasoning research is strongly driven by domain-specific datasets and benchmarks. Second, there is a clear trend toward domain-specialized reasoning models and multi-agent or multi-component reasoning systems. Third, structured prompting and structured outputs are commonly used to improve reasoning reliability. Fourth, reasoning evaluation increasingly considers explanations and intermediate reasoning steps, not only final answers. Finally, efficiency and cost considerations are becoming more important, but reasoning effort and reasoning budget optimization remain relatively underexplored research areas.

\subsection{Identified Gaps}

Despite the growing body of research on reasoning with large language models, several important gaps remain in the literature. The reviewed studies reveal limitations related to evaluation methodology, domain generalization, reasoning interpretability, integration with real decision-making processes, and the lack of standardized reasoning frameworks.

\subsubsection{Lack of Standardized Evaluation Frameworks for Reasoning}

One of the most significant gaps concerns the lack of standardized evaluation methodologies for reasoning. Although there are several benchmarks, such as EconNLI \cite{guo-yang-2024-econnli} and STEER-ME \cite{raman2025steermeassessingmicroeconomicreasoning}, the literature still lacks a unified framework for evaluating reasoning quality across domains. Existing benchmarks typically focus on specific tasks such as causal inference, multiple-choice reasoning, or structured financial question answering, which makes it difficult to compare models across domains and reasoning types.

Moreover, many studies evaluate reasoning primarily through task accuracy rather than through direct assessment of reasoning correctness, consistency, or robustness. This indicates a need for more comprehensive reasoning evaluation frameworks that consider not only final answers but also reasoning processes, logical consistency, and robustness to problem reformulation.

\subsubsection{Limited Cross-Domain Reasoning Generalization}

Another important research gap concerns the generalization of reasoning across domains. Many existing works develop highly specialized models for specific domains such as finance \cite{liu2025finr1, zhu2024tatllm}, economics \cite{zhou2025reasoninglikeeconomistposttraining}, or e-commerce \cite{zhang2025ecare}. Although these models often achieve strong performance within their target domain, relatively little research examines whether reasoning capabilities learned in one domain can be transferred to another domain.

This suggests that current research focuses more on domain-specific reasoning optimization rather than on developing general reasoning frameworks that can operate across multiple domains. Understanding how reasoning knowledge transfers between domains remains an important open research problem.

\subsubsection{Insufficient Integration of Reasoning with Real Decision-Making Processes}

Several studies demonstrate reasoning capabilities on benchmark tasks or prediction problems, such as startup success prediction in CrunchLLM \cite{sadia2025crunchllmmultitaskllmsstructured} or business analytics in dual-model reasoning systems \cite{trivedi2024dual}. However, relatively few studies analyze how reasoning systems are integrated into real decision-making workflows, organizational processes, or human–AI collaboration environments.

Most existing works focus on model performance rather than on decision support effectiveness, human trust, usability, or organizational impact. This indicates a gap between technical reasoning performance and practical deployment in real-world decision environments.

\subsubsection{Limited Research on Multi-Agent and Collaborative Reasoning}

Although some works explore multi-agent reasoning architectures, such as SynergyMAS \cite{kostka2024synergizing} and multi-agent economic reasoning frameworks \cite{zhou2025reasoninglikeeconomistposttraining}, this area is still relatively underexplored compared to single-model reasoning approaches. Multi-agent reasoning may better reflect real-world decision-making, where multiple agents interact, negotiate, and cooperate.

Future research could explore coordination between reasoning agents, division of reasoning tasks, negotiation between agents, and collective reasoning processes, which are currently only partially addressed in the literature.

\subsubsection{Reasoning Interpretability and Verification}

Another gap concerns reasoning interpretability and verification. Many studies generate reasoning traces, chain-of-thought explanations, or structured reasoning outputs, but there is still limited research on verifying whether the reasoning is logically valid rather than only plausible. Models may generate convincing reasoning explanations that do not correspond to the actual decision process.

This suggests a need for methods that verify reasoning correctness, detect reasoning errors, and ensure logical consistency across reasoning steps. The integration of symbolic reasoning, logical solvers, or external verification tools remains an important research direction.

\subsubsection{Trade-off Between Reasoning Quality, Cost, and Latency}

Although some works consider efficiency and cost issues, especially in applied systems such as E-CARE \cite{zhang2025ecare} or smaller reasoning models like Fin-R1 \cite{liu2025finr1}, there is still limited research analyzing the trade-off between reasoning quality, reasoning depth, computational cost, and latency.

Most studies aim to improve reasoning accuracy, but fewer works analyze how much reasoning is actually necessary for a given task or how reasoning effort should be allocated. Research on reasoning budget optimization, adaptive reasoning depth, and cost-aware reasoning strategies remains relatively limited.

\subsubsection{Lack of Unified Reasoning Architectures}

Finally, the literature lacks a unified architecture for reasoning systems. Some works use fine-tuned domain models, others use multi-agent systems, retrieval-augmented reasoning, structured pipelines, or hybrid symbolic–neural approaches. However, there is no dominant architecture or standardized reasoning pipeline that could serve as a general framework for reasoning systems.

Future research could focus on developing generalized reasoning architectures that integrate retrieval, structured reasoning, verification, and decision support into a unified reasoning framework.

\subsubsection{Summary of Research Gaps}

In summary, the main gaps identified in the literature include the lack of standardized reasoning evaluation frameworks, limited research on cross-domain reasoning generalization, insufficient integration of reasoning systems into real decision-making processes, limited exploration of multi-agent reasoning, insufficient reasoning verification methods, limited research on reasoning cost–performance trade-offs, and the absence of unified reasoning architectures. Addressing these gaps could significantly advance research on reasoning with large language models and improve their applicability in real-world analytical and decision-making tasks.

\subsection{Notable domain-specific tasks that have been addressed by reasoning models}

Recent research demonstrates that reasoning-oriented large language models are increasingly applied to domain-specific analytical and decision-support tasks. The reviewed literature shows that reasoning models are particularly used in finance, economics, business analytics, e-commerce, and multi-agent decision environments.

\subsubsection{Financial reasoning and financial document analysis}

One of the most developed areas of domain-specific reasoning is finance. Models such as Fin-R1 are designed specifically for financial reasoning tasks, including compliance checking, robo-advisory, and financial decision support. These models are trained on financial chain-of-thought datasets and optimized using reinforcement learning to improve reasoning accuracy and interpretability \cite{liu2025finr1}. Similarly, TAT-LLM focuses on discrete reasoning over financial tabular and textual data, supporting tasks such as financial question answering and financial document analysis. The model uses a step-wise reasoning pipeline consisting of extraction, reasoning, and execution stages \cite{zhu2024tatllm}. These studies demonstrate that financial reasoning is currently one of the most important application areas for reasoning-oriented LLMs.

\subsubsection{Economic reasoning and policy analysis}

Another important domain is economic reasoning. The EconNLI benchmark assesses whether language models can infer causal relationships between economic events and predict the economic consequences of specific events \cite{guo-yang-2024-econnli}. Similarly, the Recon model presented in \cite{zhou2025reasoninglikeeconomistposttraining} focuses on economic reasoning, strategic decision-making, and multi-agent economic interactions such as resource allocation, market behavior, and policy analysis. These works show that reasoning models are increasingly applied to economic analysis tasks that require structured analytical reasoning and understanding of economic theory.

\subsubsection{Business analytics and decision support}

Reasoning models are also used in business analytics and decision-support systems. The dual-model business reasoning framework integrates structured data retrieval with domain knowledge to support strategic planning, business analysis, and managerial decision-making \cite{trivedi2024dual}. Similarly, CrunchLLM focuses on startup success prediction using structured and unstructured business data, combining reasoning over financial indicators, company descriptions, and investor networks to generate predictions along with reasoning explanations \cite{sadia2025crunchllmmultitaskllmsstructured}. These works demonstrate that reasoning models can support complex business decision-making tasks involving heterogeneous data sources.

\subsubsection{E-commerce reasoning and recommendation systems}

Reasoning models are also applied in e-commerce, particularly in recommendation systems and query–product matching. The E-CARE framework uses commonsense reasoning to infer implicit relationships between user queries and products, improving recommendation accuracy without requiring real-time LLM inference \cite{zhang2025ecare}. This shows that reasoning models can support recommendation tasks that require understanding implicit user needs, product features, and contextual relationships.

\subsubsection{Multi-agent reasoning and collaborative problem solving}

Another domain where reasoning models are applied is multi-agent systems and collaborative problem solving. The SynergyMAS framework integrates logical reasoning, knowledge management, and collaboration between multiple LLM agents to support complex problem-solving tasks such as product development and organizational decision-making \cite{kostka2024synergizing}. Multi-agent reasoning systems allow models to divide reasoning tasks, share knowledge, and collaboratively solve complex problems that are difficult for a single model.

\subsubsection{Summary of addressed domain-specific tasks}

Overall, the reviewed literature shows that reasoning models are already applied to several domain-specific tasks, including financial reasoning, economic analysis, business decision support, startup success prediction, e-commerce recommendation systems, and multi-agent collaborative problem solving. These applications demonstrate that reasoning-oriented language models are increasingly used not only for general reasoning benchmarks but also for domain-specific analytical and decision-support tasks.

\subsection{Domain-specific tasks that have not been addressed by reasoning models}

Although reasoning-oriented language models have already been applied to several domains such as finance, economics, business analytics, and e-commerce, many important domain-specific tasks remain largely unexplored. These tasks often require long-horizon reasoning, multi-step decision-making, uncertainty modeling, or integration of structured and dynamic data sources.

\subsubsection{Supply chain and inventory management reasoning}

One domain that remains largely unexplored is supply chain management and inventory optimization. Although LLMs have been used for forecasting and demand prediction, there is little research on reasoning models that support multi-step decision-making in inventory control, supplier selection, logistics planning, or production scheduling. These tasks require reasoning about trade-offs between cost, risk, service level, and demand uncertainty, which makes them suitable candidates for reasoning-oriented language models.

\subsubsection{Strategic business planning and scenario analysis}

Another domain that has not been extensively addressed is strategic planning and scenario analysis. Strategic decision-making often involves reasoning about long-term consequences, market evolution, competitor behavior, and technological change. Existing research focuses mainly on prediction or classification tasks, such as startup success prediction, but less work has been done on reasoning models that support long-term strategic scenario analysis, business model design, or investment strategy planning.

\subsubsection{Public policy and regulatory impact analysis}

Public policy analysis is another domain where reasoning models could play an important role but has not yet been extensively studied. Policy analysis requires reasoning about causal relationships, behavioral responses, macroeconomic effects, and unintended consequences of regulations. While economic reasoning benchmarks exist, there are relatively few studies on using reasoning models to support real-world policy evaluation, regulatory impact analysis, or social welfare optimization.

\subsubsection{Risk management and decision-making under uncertainty}

Risk management is another important domain that has not been widely explored in reasoning model research. Tasks such as credit risk assessment, operational risk analysis, insurance risk modeling, and portfolio risk management require reasoning under uncertainty, probabilistic thinking, and scenario-based reasoning. Current research focuses mainly on prediction and classification rather than reasoning-based risk analysis and decision support.

\subsubsection{Human–AI collaborative decision-making}

Another underexplored domain is human–AI collaborative decision-making, where reasoning models assist humans in complex analytical tasks such as consulting, medical decision-making, legal reasoning, or engineering design. Most current research evaluates reasoning models in isolation, rather than in collaborative environments where humans and AI jointly perform reasoning tasks.

\subsubsection{Long-horizon planning and sequential decision problems}

Finally, reasoning models have not yet been widely applied to long-horizon planning problems such as project management, organizational planning, infrastructure investment planning, or energy system planning. These tasks require reasoning across long time horizons, considering multiple constraints, resource limitations, and dynamic environments. Integrating reasoning models with planning and optimization systems remains an open research direction.

\subsubsection{Summary of unexplored domain-specific tasks}

In summary, while reasoning models are already applied in finance, economics, business analytics, and e-commerce, many important domain-specific tasks remain unexplored. These include supply chain and inventory management reasoning, strategic planning and scenario analysis, public policy evaluation, risk management under uncertainty, human–AI collaborative decision-making, and long-horizon planning problems. These domains represent promising directions for future research on reasoning-oriented language models.

\subsection{Comments on the Usage of Reasoning Models in RAG-like Systems or Agentic Frameworks}

The reviewed literature indicates that reasoning-oriented language models are rarely used as standalone components. Instead, they are increasingly embedded within larger system architectures, such as retrieval-augmented generation (RAG) systems, structured data pipelines, or multi-agent frameworks. In such architectures, reasoning models act as decision-making or inference components rather than as primary knowledge sources.

\subsubsection{Reasoning as a Layer on Top of Retrieval Systems}

Several studies demonstrate that reasoning models are often combined with retrieval mechanisms rather than used independently. In the dual-model business reasoning framework, structured data is first retrieved using SQL generation and semantic search, and only then reasoning is applied to interpret the retrieved data and produce analytical conclusions \cite{trivedi2024dual}. This architecture separates data retrieval from reasoning, reducing hallucinations and improving factual accuracy. In this setup, the reasoning model operates as an analytical layer that interprets structured data rather than generating knowledge from parametric memory alone.

Similarly, in the SynergyMAS framework, reasoning capabilities are integrated with Retrieval-Augmented Generation (RAG) and knowledge bases, enabling agents to access external knowledge while performing logical reasoning tasks \cite{kostka2024synergizing}. This approach suggests that reasoning models benefit from access to structured knowledge repositories and external tools rather than relying solely on internal model knowledge.

\subsubsection{Reasoning without Real-Time LLM Inference}

An interesting alternative approach is presented in the E-CARE framework, where reasoning is not performed through real-time LLM inference but instead encoded into a reasoning factor graph that represents commonsense reasoning patterns \cite{zhang2025ecare}. In this approach, reasoning knowledge is extracted from large models offline and then reused in a lightweight system during inference. This demonstrates that reasoning can be embedded into system architecture as structured reasoning knowledge rather than performed dynamically during each query.

This approach is particularly important from the perspective of system efficiency, as real-time reasoning with large models can be computationally expensive and slow. Pre-computed reasoning structures or reasoning templates may therefore become an important architectural component in future RAG-like systems.

\subsubsection{Reasoning Models as Decision Modules in Analytical Pipelines}

In domain-specific reasoning systems such as CrunchLLM, reasoning models are used as decision modules that operate on structured and unstructured data integrated into a single analytical pipeline \cite{sadia2025crunchllmmultitaskllmsstructured}. In such architectures, reasoning models do not operate directly on raw user queries but instead on processed data representations that combine structured attributes, textual descriptions, and derived features.

Similarly, specialized reasoning models such as Fin-R1 operate on curated reasoning datasets and structured financial information, suggesting that reasoning models often function as analytical engines within larger decision-support systems rather than as general conversational agents \cite{liu2025finr1}. This indicates a shift from conversational AI toward analytical reasoning systems embedded in data pipelines.

\subsubsection{Reasoning in Multi-Agent Architectures}

Multi-agent frameworks represent another important architectural paradigm. In multi-agent systems, reasoning tasks can be distributed across multiple agents that specialize in different subtasks, such as information retrieval, logical reasoning, planning, or evaluation. The SynergyMAS framework demonstrates that combining logical reasoning modules, knowledge management, and agent collaboration can improve performance on complex problem-solving tasks \cite{kostka2024synergizing}. 

Similarly, economic reasoning models trained for multi-agent environments show that reasoning models can be used not only to answer questions, but also to simulate strategic interactions, negotiation, and decision-making processes in multi-agent environments \cite{zhou2025reasoninglikeeconomistposttraining}. In such architectures, reasoning models become agents that interact, exchange information, and jointly solve problems rather than single models producing isolated responses.

\subsubsection{Separation Between Knowledge, Reasoning, and Execution}

Across the reviewed studies, a recurring architectural pattern emerges: the separation of knowledge retrieval, reasoning, and execution stages. For example, the step-wise pipeline used in TAT-LLM separates extraction, reasoning, and execution steps, improving interpretability and reasoning transparency \cite{zhu2024tatllm}. Similar architectural separation can also be observed in RAG-based business reasoning systems and multi-agent frameworks.

This suggests that future reasoning systems may increasingly follow modular architectures consisting of:
\begin{itemize}
\item a retrieval module,
\item a reasoning module,
\item a planning or decision module,
\item and an execution or tool-use module.
\end{itemize}

Such modular architectures resemble cognitive architectures and may represent a general design pattern for reasoning-oriented AI systems.

\subsubsection{Implications for Future Reasoning Systems}

In general, the literature suggests that reasoning models are most effective when embedded within larger system architectures rather than when used as standalone models. The most common architectures include RAG systems, structured data pipelines, and multi-agent frameworks. In these systems, reasoning models typically serve as analytical, decision-making, or planning components, while retrieval systems provide knowledge and external tools provide execution capabilities. This architectural separation may become a dominant paradigm in the development of reasoning-oriented AI systems and agentic AI frameworks.

\subsection{Number of Datasets for Training Domain-Specific Reasoning Models}

The reviewed literature indicates that the number of datasets specifically constructed for training domain-specific reasoning models remains relatively limited. Several studies introduce domain-specific datasets or curated reasoning corpora used for supervised fine-tuning or post-training of reasoning-oriented models. For example, Fin-R1 is trained on a dedicated financial reasoning dataset constructed from curated financial reasoning tasks and chain-of-thought samples \cite{liu2025finr1}. Similarly, the Recon model is trained on curated economic reasoning datasets and reasoning traces derived from economic problem sets \cite{zhou2025reasoninglikeeconomistposttraining}. 

Other works rely on structured domain datasets combined with textual information, such as the Crunchbase dataset used in CrunchLLM for structured business reasoning and prediction tasks \cite{sadia2025crunchllmmultitaskllmsstructured}. In addition, some frameworks, such as E-CARE, extract reasoning knowledge from large language models and encode it into structured reasoning graphs rather than creating traditional supervised datasets \cite{zhang2025ecare}.

Overall, the number of datasets identified for training domain-specific reasoning models across the reviewed literature can be classified as \textbf{between 4 and 6}. However, only a subset of these datasets is publicly available, as some datasets are proprietary, automatically generated, or not fully released. Therefore, the number of publicly accessible training datasets can be classified as \textbf{between 1 and 3}.

\subsection{Number of Benchmarks and Datasets for Evaluating Domain-Specific Reasoning Models}

The literature shows a larger number of datasets and benchmarks designed for evaluating domain-specific reasoning models than for training them. One example is EconNLI, a dataset designed to evaluate economic reasoning and causal inference in economic scenarios \cite{guo-yang-2024-econnli}. Another important benchmark is STEER-ME, which evaluates microeconomic reasoning across multiple domains, problem types, and perspectives \cite{raman2025steermeassessingmicroeconomicreasoning}. 

Financial reasoning models are evaluated on benchmarks such as FinQA, TAT-QA, and TAT-DQA, which are used in studies on financial reasoning models such as TAT-LLM and Fin-R1 \cite{zhu2024tatllm, liu2025finr1}. Additionally, some studies evaluate reasoning models on domain-specific analytical tasks such as business analytics or structured decision-making scenarios \cite{sadia2025crunchllmmultitaskllmsstructured, trivedi2024dual}.

In general, the number of evaluation benchmarks identified in the reviewed literature can be classified as \textbf{between 4 and 6}. Most of these benchmarks are publicly available and commonly used in the research community. Therefore, the number of publicly accessible evaluation datasets can also be classified as \textbf{between 4 and 6}.

\subsection{Number of Domain-Specific Reasoning Models}

The reviewed literature presents several domain-specific reasoning models developed for finance, economics, business analytics, and structured data reasoning. Examples include Fin-R1, a financial reasoning model trained using supervised fine-tuning and reinforcement learning \cite{liu2025finr1}, and TAT-LLM, a model designed for reasoning over tabular and textual financial data \cite{zhu2024tatllm}. 

In the economics domain, the Recon model demonstrates that post-training on economic reasoning tasks can significantly improve structured reasoning and strategic decision-making capabilities \cite{zhou2025reasoninglikeeconomistposttraining}. In the business analytics domain, CrunchLLM is designed for structured business reasoning and outcome prediction using structured and unstructured data \cite{sadia2025crunchllmmultitaskllmsstructured}. Additional domain-specific reasoning architectures include dual-model reasoning systems for business decision support \cite{trivedi2024dual} and reasoning frameworks embedded in recommendation systems such as E-CARE \cite{zhang2025ecare}. Multi-agent reasoning frameworks such as SynergyMAS also represent domain-oriented reasoning architectures \cite{kostka2024synergizing}.

Overall, the number of domain-specific reasoning models identified in the reviewed works can be classified as \textbf{between 4 and 6}. However, only some of these models are publicly available or open-source. Therefore, the number of publicly accessible domain-specific reasoning models can be classified as \textbf{between 1 and 3}.

\subsection{Number of Methods for Creating and Using Domain-Specific Reasoning Models}

The reviewed literature identifies a relatively large number of methods used for creating or deploying domain-specific reasoning models. These methods include supervised fine-tuning, reinforcement learning, chain-of-thought distillation, parameter-efficient fine-tuning, prompt optimization, retrieval-augmented generation, multi-agent reasoning frameworks, reasoning pipelines with structured intermediate steps, and knowledge graph or reasoning graph approaches.

For example, Fin-R1 combines supervised fine-tuning with reinforcement learning to improve financial reasoning performance \cite{liu2025finr1}. TAT-LLM introduces a step-wise reasoning pipeline consisting of extraction, reasoning, and execution stages \cite{zhu2024tatllm}. CrunchLLM uses parameter-efficient fine-tuning and prompt optimization to adapt general-purpose models to business reasoning tasks \cite{sadia2025crunchllmmultitaskllmsstructured}. Multi-agent reasoning architectures integrating logical reasoning and retrieval mechanisms are proposed in SynergyMAS \cite{kostka2024synergizing}. Additionally, some systems encode reasoning knowledge into structured reasoning graphs instead of performing reasoning dynamically during inference, as demonstrated in E-CARE \cite{zhang2025ecare}.

Because multiple training, post-training, prompting, and system-level methods are used across the reviewed works, the total number of methods can be classified as \textbf{many (more than 6)}. Among these, several methods such as supervised fine-tuning, reinforcement learning, chain-of-thought prompting, retrieval-augmented generation, and parameter-efficient fine-tuning are publicly documented and widely accessible. Therefore, the number of publicly accessible methods can be classified as \textbf{between 4 and 6}.

As shown in Table~\ref{tab:reasoning_summary}, the ecosystem of domain-specific reasoning models is still relatively immature. While several models and evaluation benchmarks have already been proposed, the availability of training datasets remains limited, especially publicly accessible ones. At the same time, a wide range of methods for developing and deploying reasoning models has emerged, indicating that methodological development is currently progressing faster than dataset and benchmark creation. This imbalance suggests that future research should focus on building standardized datasets and evaluation frameworks for domain-specific reasoning tasks.

\begin{table}[h]
\centering
\caption{Summary of datasets, benchmarks, models, and methods identified in the literature}
\begin{tabular}{lcc}
\toprule
Category & All & Publicly accessible \\
\midrule
Training datasets & between 4 and 6 & between 1 and 3 \\
Evaluation benchmarks & between 4 and 6 & between 4 and 6 \\
Domain-specific reasoning models & between 4 and 6 & between 1 and 3 \\
Methods for creating reasoning models & many (more than 6) & between 4 and 6 \\
\bottomrule
\end{tabular}

\label{tab:reasoning_summary}
\end{table}

\subsection[Literature Review Summary: SH2]{Literature Review Summary: SH2}

\textbf{Institutions, Governance and Legal Systems.} Political science, international relations, law.

\subsection{Key Findings}
\textit{Important datasets, benchmarks, domain-specific reasoning models; what reasoning models are being used; modalities used in existing works (text-only, text-image, or other modalities); how people prompt the model, what kind of outputs they expect (e.g., whether using a defined schema with closed answer set, json formatting, open ended answers etc.), how they use and/or validate the reasoning of models; whether existing works consider aspects like reasoning length/effort as well as cost/efficiency.}

\paragraph{Datasets and Benchmarks.}
The surveyed literature has produced a rich ecosystem of domain-specific benchmarks spanning legal NLP and political-science reasoning. On the legal side, key datasets include MAUD (merger-agreement reading comprehension, 47{,}457 annotations) \cite{wang2023-maud}, LawBench (20-task Chinese civil-law benchmark based on Bloom's taxonomy) \cite{fei2024-lawbench}, MSLR-Bench (multi-step legal reasoning over Chinese insider-trading judgments) \cite{yu2025-multisteplegalreasoning}, LEXam (7{,}537 questions from 340 real Swiss law exams, bilingual EN/DE) \cite{fan2026-lexam}, and LegalBench (162 tasks spanning six categories of US legal reasoning, NeurIPS 2023) \cite{NEURIPS2023_89e44582}. Bilingual coverage is provided by the Legal-R1 dataset (10 Chinese + 7 English tasks) \cite{cai2025-unilawR1}. Training-oriented resources include LawInstruct (12 million instruction-tuning examples across 24 languages and 17 jurisdictions) \cite{niklaus2025-lawinstruct}, Unilaw-R1-Data ($\sim$17k chain-of-thought samples for Chinese judicial exam questions) \cite{cai2025-unilawR1}, and the KG-derived Rules SFT/Preference Dataset built from 12{,}000 US legal opinions \cite{song2026-KGlegal}. In international relations and political science, UNSC-Bench (469 UN Security Council draft resolutions with P5 vote labels, 1947--2025, multilingual) \cite{nangia2026-unscBench}, UNBench (30-year UNSC records covering four interconnected tasks) \cite{liang2026-UNBench}, and the UNSC Nation-Level Bias Dataset (581 resolutions, 2013--2024) \cite{choi2026-UNSCbias} constitute the main evaluation resources. The WORLDREP dataset (44{,}706 news articles, 147{,}931 labeled country-pair relation scores, 2015--2024) \cite{gwak2024-forecastingInternational} addresses geopolitical event prediction. The multi-party climate-negotiation benchmark \cite{benac2026-benchmarknegotiation} targets sequential multi-agent bargaining.

\paragraph{Domain-Specific Reasoning Models.}
Three papers introduce new legal language models. Unilaw-R1 \cite{cai2025-unilawR1} fine-tunes Qwen2.5-7B-Instruct via two-stage SFT + Group Relative Policy Optimization (GRPO) with a legal validity reward, matching the performance of the substantially larger DeepSeek-R1-Distill-Qwen-32B (54.9\% on LawBench). Legal-R1-14B \cite{cai2025-unilawR1} applies progressive SFT with rejection sampling on top of DeepSeek-R1-Distill-Qwen-14B, producing a bilingual model competitive with models several times larger. SaulLM-54B and SaulLM-141B \cite{colombo2024-saulLM} scale Mixtral MoE architectures through continued legal pretraining ($\sim$520B tokens), instruction tuning, and preference alignment, reaching state-of-the-art results on LegalBench-Instruct among open-source models. FLawN-T5 \cite{niklaus2025-lawinstruct}, derived from T5/mT5/Flan-T5 via instruction tuning on LawInstruct, demonstrates that even encoder-decoder models benefit substantially from legal domain adaptation. Legal$\Delta$ \cite{dai2026-legaldelta} introduces an information-gain-enhanced GRPO reward computed from model logits, achieving roughly a 10\% average improvement over baselines on Chinese legal tasks without requiring human preference labels.

\paragraph{Reasoning Models and Prompting Strategies.}
There is a clear generational divide across the surveyed works. Earlier papers (2023--2024) rely exclusively on Chain-of-Thought (CoT) and few/zero-shot prompting with models such as GPT-4, GPT-3.5, text-davinci-003, Claude-2, and smaller fine-tuned Transformers (RoBERTa, DeBERTa, Longformer). More recent work (2025--2026) actively evaluates or trains explicit reasoning models that produce structured \texttt{<think>} traces: Unilaw-R1 is trained end-to-end with such tokens; Legal-R1-14B, LEXam, and MSLR-Bench evaluate DeepSeek-R1, QwQ-32B, and OpenAI o1/o3-mini as reasoning baselines; UNSC-Bench \cite{nangia2026-unscBench} includes ``thinking variant'' models in its 26-model suite. Models like o1, o3-mini, and DeepSeek-R1 were employed specifically for their structured code-generation and auto-formalization capabilities \cite{kant2025-robustlegal, wang2026-bridginglegallogic}. Novel prompting strategies include Legal Syllogism Prompting (LoT) \cite{jiang2023-legalsyllogism}, a zero-shot approach that maps case facts to the major-premise (statute) $\to$ minor-premise (facts) $\to$ conclusion syllogistic structure; Self-Initiated CoT \cite{yu2025-multisteplegalreasoning}, where models autonomously draft their own step-by-step guidance before answering; scratchpad-based verify-explain-correct cycles \cite{gwak2024-forecastingInternational}; and role-playing / persona prompting for diplomatic simulation \cite{nangia2026-unscBench,choi2026-UNSCbias,liang2026-UNBench}.

\paragraph{Modalities.}
All but one paper are text-only. The single exception is SARA \cite{bonfim2026-sara}, a multi-agent court-decision-support system for the Brazilian judiciary that ingests scanned case files and evidentiary images via OCR, making it the only multimodal system in the survey.

\paragraph{Output Formats and Answer Schemas.}
Works vary widely in the type of output expected from models. Closed-answer / classification tasks dominate: multiple-choice legal exam questions (Unilaw-R1, Legal-R1, MAUD, LawBench, LegalBench, LEXam MCQ track, LoT), charge/article prediction and binary/multi-label classification (LawBench, Legal$\Delta$, LawInstruct), and Yes/No/Abstain vote classification (UNSC-Bench, UNBench, Nation-Level Bias). Structured open-ended outputs are required in LawBench (regression for damages calculation), WORLDREP (continuous 0--1 conflict score), LEXam open-ended track (long-form legal argument graded on a 0--1 continuum by LLM-as-a-Judge), MSLR-Bench (IRAC-structured reasoning traces graded A/B/C), and SARA (full judicial decision drafts). The neuro-symbolic works expect code output (Prolog or First-Order Logic) as an intermediate step. Negotiation-benchmark agents output structured proposal-commitment matrices alongside natural-language text.

\paragraph{Validation of Reasoning.}
Most benchmarks validate only final answers (accuracy, micro/macro F1, ROUGE-L, BERTScore). A meaningful subset goes further: MSLR-Bench evaluates IRAC Recall and uses LLM-as-a-Judge with human cross-validation to score intermediate reasoning steps; LEXam correlates LLM-judge scores with three human experts (Pearson $r=0.70$); the neuro-symbolic frameworks \cite{kant2025-robustlegal,wang2026-bridginglegallogic} validate reasoning by executing generated Prolog/FOL code in deterministic solvers; SARA uses ALIGNSCORE to compare agent-generated reasoning traces against authentic human judicial rationale; WORLDREP validates the scratchpad verify-explain-correct steps against expert expectations; and Unilaw-R1 uses an LLM-judge (DeepSeek-V3) for chain rewriting and reasoning selection during data curation.

\paragraph{Reasoning Length, Efficiency, and Cost.}
None of the surveyed papers conducts a systematic study of reasoning length or token-level cost/efficiency trade-offs. Unilaw-R1 documents token-length distributions for its \texttt{<think>} chains as a descriptive data property, and Legal$\Delta$ uses information gain (measured via logit shifts) as a proxy for reasoning quality, rewarding concise high-confidence steps over verbose filler. The computational costs of running 54B/141B MoE models (SaulLM) are noted as a deployment barrier. However, no paper explicitly benchmarks performance-vs-latency or accuracy-vs-token-budget curves.

\subsection{Identified Gaps}
\begin{enumerate}
  \item \textbf{Reasoning length and efficiency.} No paper systematically investigates the relationship between reasoning chain length (or inference compute budget) and task performance. Given that legal reasoning often demands lengthy deliberation, this is a significant open question.

  \item \textbf{Temporal and jurisdictional generalization.} Several works note performance degradation on out-of-distribution time periods (MAUD) or on non-US/non-Chinese legal systems. Cross-jurisdictional and multilingual evaluation remains thin: LawInstruct covers 24 languages but its evaluation suite (LegalBench) is English-only; LEXam covers Swiss/EU law in English and German but not common-law systems. Civil-law jurisdictions outside China and Brazil are largely absent.

  \item \textbf{End-to-end document processing.} All datasets that test document comprehension rely on pre-extracted snippets rather than full multi-hundred-page contracts or judicial files. Processing entire, unsegmented legal documents is explicitly flagged as future work in MAUD and Unilaw-R1.

  \item \textbf{Interactive and multi-turn legal tasks.} Evaluations are predominantly single-turn, static input-output. Dynamic, multi-turn settings (lawyer-client dialogue, iterative contract negotiation, adversarial cross-examination) are largely unexplored. LawBench explicitly lists this as a drawback; the negotiation benchmark \cite{benac2026-benchmarknegotiation} is the most advanced work here but focuses on algorithmic agents rather than LLMs.

  \item \textbf{Faithfulness vs.\ accuracy.} The FOL/neuro-symbolic paper \cite{wang2026-bridginglegallogic} identifies a critical gap: high benchmark accuracy can mask ``Assumption Injection,'' where models reach correct legal conclusions via logically unfaithful shortcuts. No evaluation framework yet systematically measures faithfulness of legal reasoning at scale.

  \item \textbf{Long-form generative legal tasks.} Contract drafting, long-form brief writing, and judicial opinion generation are not yet addressed by reasoning models. Fine-tuned models are typically evaluated on classification or short-answer tasks.

  \item \textbf{Political science tasks beyond voting simulation.} UNBench covers co-penholder prediction, vote simulation, adoption prediction, and statement generation, but deeper tasks, such as coalition formation dynamics, treaty negotiation strategy, detecting strategic abstentions, remain untested with reasoning models. The WORLDREP and UNSC-Bench works primarily treat events as classification/regression problems rather than as multi-step strategic reasoning challenges.

  \item \textbf{Non-English political science data.} The UN-oriented benchmarks operate primarily in English, despite multilingual records being available; only UNSC-Bench tests four UN official languages.

  \item \textbf{Scalability of neuro-symbolic approaches.} The Prolog/FOL-grounded legal reasoning papers \cite{kant2025-robustlegal,wang2026-bridginglegallogic} are evaluated on narrow, expert-curated clause sets (health insurance, ContractNLI). Scaling these to broad, unseen contract types still requires human-authored helper logic, creating a bottleneck.

  \item \textbf{Statistical rigor.} A majority of papers report absolute accuracy comparisons without statistical significance tests (p-values, confidence intervals). LEXam is a notable exception. This weakens the reliability of claimed improvements, particularly for small datasets.
\end{enumerate}

\subsection{Notable domain-specific tasks that have been addressed by reasoning models}
\begin{itemize}
  \item \textbf{Legal exam question answering} (multiple-choice bar/judicial exam format): Unilaw-R1 \cite{cai2025-unilawR1} and Legal-R1-14B \cite{cai2025-unilawR1} fine-tune explicit-reasoning models on Chinese judicial examination questions, achieving scores competitive with much larger models. LEXam \cite{fan2026-lexam} reveals that test-time reasoning models (DeepSeek-R1, Claude-3.7-Sonnet, Gemini-2.5-Pro) substantially outperform standard models on open-ended Swiss law questions.

  \item \textbf{Legal judgment prediction} (charge, article, and penalty prediction): LoT \cite{jiang2023-legalsyllogism} applies structured zero-shot CoT to Chinese criminal judgment prediction, improving over standard CoT by enforcing the major-premise $\to$ minor-premise $\to$ conclusion syllogistic structure. Legal$\Delta$ \cite{dai2026-legaldelta} uses RL with information-gain rewards on statute law prediction and charge application tasks.

  \item \textbf{Merger agreement understanding}: MAUD \cite{wang2023-maud} benchmarks fine-tuned Transformers and GPT-4/GPT-3.5 on 92 expert-annotated deal-point questions; while CoT prompting with GPT-4 performs strongly, fine-tuned DeBERTa-v3-large remains competitive.

  \item \textbf{Contract entailment and neuro-symbolic legal reasoning}: The Prolog-grounded framework \cite{kant2025-robustlegal} demonstrates near-100\% accuracy on insurance clause adjudication using o1/o3-mini/DeepSeek-R1 as Prolog code generators, compared to 12--41\% for unguided LLM baselines.

  \item \textbf{Multi-step Chinese legal reasoning over securities law}: MSLR-Bench \cite{yu2025-multisteplegalreasoning} finds that DeepSeek-R1 and QwQ-32B with Self-Initiated CoT achieve superior IRAC-structured reasoning over legal fine-tuned baselines.

  \item \textbf{UN Security Council vote prediction with role-playing}: UNSC-Bench \cite{nangia2026-unscBench} and UNBench \cite{liang2026-UNBench} show that explicit role-playing prompts and stronger general capabilities (MMLU-Pro) significantly improve P5 vote prediction accuracy over neutral prompting.

  \item \textbf{Geopolitical relationship scoring}: WORLDREP \cite{gwak2024-forecastingInternational} applies GPT-4-Turbo, GPT-4o, and Gemini-1.5-Pro to continuous conflict/cooperation scoring of country pairs from news, using self-correcting scratchpad prompts.

  \item \textbf{Nation-level bias detection and mitigation in IR}: The Nation-Level Bias paper \cite{choi2026-UNSCbias} demonstrates that RAG + Reflexion debiasing reduces Western-country favoritism in GPT-4o-mini and Llama-3.3-70B.

  \item \textbf{Court-decision drafting with ontology grounding}: SARA \cite{bonfim2026-sara} deploys GPT-4 and LLaMA-3 in a multi-agent pipeline grounded in a Brazilian jurisprudential knowledge graph, achieving 94.4\% positive evaluations from active judges.
\end{itemize}

\subsection{Domain-specific tasks that have not been addressed by reasoning models}
\textit{Tasks that have not been addressed at all, or tasks that have been somehow touched by reasoning models but not fully solved yet.}

\begin{itemize}
  \item \textbf{Full-document legal reading comprehension}: No work processes complete, unsegmented multi-hundred-page merger agreements, court opinions, or regulatory filings end-to-end with reasoning models.

  \item \textbf{Contract drafting and redlining}: The report by Vals AI and vLex Team \cite{vLex_2025} indicates that current AI tools underperform human lawyers on contract redlining. No reasoning model has been specifically trained or evaluated for this task.

  \item \textbf{Long-form legal brief and opinion writing}: While SARA generates judicial decision drafts, no benchmark rigorously evaluates structured legal argumentation in the form of briefs or opinions using reasoning models.

  \item \textbf{Statutory compliance checking across jurisdictions}: Checking whether a document or action complies with a body of statutes, especially across multiple jurisdictions simultaneously, is touched by LoT and the FOL paper but not solved.

  \item \textbf{Tax and numerical legal reasoning}: LawBench and Legal-R1-14B note that models fail severely on tasks requiring arithmetic (e.g., damages calculation, tax liability). No reasoning-model paper specifically targets this.

  \item \textbf{Multi-party negotiation with LLM agents}: The negotiation benchmark \cite{benac2026-benchmarknegotiation} uses algorithmic solvers rather than frontier LLMs as negotiating agents; evaluating explicit-reasoning models in multi-round binding-commitment bargaining remains open.

  \item \textbf{Dynamic coalition and alliance modeling}: No paper models how states form or shift alliances through sequential strategic interactions; existing political-science benchmarks treat voting as one-shot classification.

  \item \textbf{Legal ethics and professional responsibility}: Despite hallucination concerns noted in Legal-R1-14B \cite{cai2025-unilawR1}, no benchmark targets lawyer-conduct rules, conflicts of interest, or attorney-client privilege scenarios.

  \item \textbf{Adverse document retrieval and legal search}: RAG is proposed as future work in multiple papers, but end-to-end retrieval + reasoning pipelines for statute or case-law search have not been benchmarked with reasoning models.

  \item \textbf{Multilingual non-English legal reasoning at the task level}: LawInstruct trains on 24 languages, but there is no evaluation benchmark for legal reasoning in languages other than English, Chinese, German, and Portuguese.
\end{itemize}

\subsection{Comments on the usage of reasoning model in RAG-like system or agentic framework}
Several papers integrate reasoning models into retrieval-augmented or agentic pipelines, though most remain at an early stage.

The most developed agentic system is SARA \cite{bonfim2026-sara}, which implements a full multi-agent pipeline for the Brazilian judiciary: an OCR preprocessing module ingests scanned case files; downstream agents extract legal elements, map them to an ontological schema (LCO-BR), query a jurisprudential knowledge graph (Jur-KG) of 22{,}106 judgments via semantic matching, and synthesize a grounded draft decision with PROV-O provenance tracking. GPT-4 and LLaMA-3 serve as the backbone models. The ontology grounding demonstrably reduces hallucination (ALIGNSCORE improvement of 2.97\% over ungrounded baselines), though evaluation remains qualitative and scale-limited.

The KG-assisted post-training framework \cite{song2026-KGlegal} takes a complementary approach: rather than retrieving from a graph at inference time, it uses the IRAC Knowledge Graph to generate SFT and DPO training data offline, so that the post-trained model internalizes structured legal reasoning pathways. This ``baked-in RAG'' paradigm avoids inference-time retrieval overhead and allows a 70B model to outperform a 141B legal specialist on four out of six reasoning tasks.

The Nation-Level Bias paper \cite{choi2026-UNSCbias} pairs RAG (injecting verified UN Digital Library records) with Reflexion-based self-reflection loops, demonstrating that factual grounding plus iterative self-correction reduces nation-level bias while simultaneously improving vote-prediction accuracy.

The WORLDREP annotation pipeline \cite{gwak2024-forecastingInternational} employs a self-correcting scratchpad mechanism --- functionally analogous to agentic reflection --- where the model verifies its country extraction, explains errors, and corrects them within a single inference pass before producing a final relationship score.

The neuro-symbolic framework \cite{kant2025-robustlegal} instantiates the most explicit human-in-the-loop agentic design: an LLM (o1, o3-mini, or DeepSeek-R1) translates legal text into Prolog code guided by expert-authored helper schemas; a deterministic Prolog engine then executes the code and returns an auditable verdict. This separation of neural translation from symbolic execution yields near-perfect accuracy on insurance-clause adjudication and satisfies interpretability requirements analogous to the EU AI Act.

SARA \cite{bonfim2026-sara}, LoT \cite{jiang2023-legalsyllogism}, and several future-work sections explicitly call for integrating RAG with reasoning models to address the ``outdated legal knowledge'' failure mode identified in Legal-R1-14B \cite{cai2025-unilawR1} and to provide citation-backed statutory reasoning. No paper yet delivers a fully evaluated RAG + explicit-reasoning-model pipeline in this domain.

\subsection{Other comments}
\paragraph{The fine-tuned-small vs.\ prompted-large trade-off.}
A recurring empirical finding across the legal papers is that domain-specifically fine-tuned smaller models (7B--14B parameters, with SFT + RL) can match or exceed general-purpose models 4--5$\times$ their size. Unilaw-R1 (7B) matches DeepSeek-R1-Distill-Qwen-32B; Legal-R1-14B outperforms GPT-4o on Chinese tasks; FLawN-T5 (3B) achieves a 15-point LegalBench improvement over standard Flan-T5-XL. This suggests that for deployment-critical legal applications, the cost-effectiveness case for specialized smaller models is strong.

\paragraph{Chinese legal NLP is a dominant focus.}
Six of the twenty-one papers focus primarily or exclusively on Chinese legal data and tasks (LawBench, Unilaw-R1, Legal-R1, Legal$\Delta$, MSLR-Bench, LoT). This reflects both the richness of publicly available Chinese judicial data (CAIL, JEC-QA, China Judgments Online) and active research communities. By contrast, civil-law systems in Continental Europe, Latin America (beyond Brazil), and other Asian jurisdictions are almost entirely absent.

\paragraph{Synthetic reasoning data as a double-edged sword.}
Multiple papers rely on LLM distillation to generate reasoning traces (Unilaw-R1, Legal-R1, the KG paper, SaulLM). This creates a practical bootstrapping mechanism but introduces a ceiling: model quality is bounded by the teacher model (typically DeepSeek-R1 or GPT-4), and any systematic biases in the teacher propagate into the student. The FOL paper \cite{wang2026-bridginglegallogic} identifies ``Assumption Injection'' as a concrete manifestation of this risk.

\paragraph{Political science tasks are underexplored relative to legal tasks.}
Of the 21 papers, roughly 14 address legal NLP and 7 address political science / international relations. Within the political-science cluster, most work concerns UN voting simulation: a single, well-structured task with clean ground truth. Broader political-science reasoning (legislative drafting, treaty analysis, sanctions modeling, electoral prediction, diplomatic cable interpretation) remains entirely unaddressed by reasoning models.

\paragraph{Human expert involvement is inconsistent.}
The quality and depth of human expert involvement varies enormously across the surveyed works, from MAUD's 10{,}000+ hours of expert annotation and senior M\&A lawyer review, to papers that treat LLM-generated synthetic data with only automated validation. This inconsistency makes cross-benchmark comparisons difficult and raises concerns about the actual difficulty and validity of some datasets.

\paragraph{Proprietary evaluation.}
Vals AI and vLex Team \cite{vLex_2025}, as well as parts of the Unilaw-R1 evaluation \cite{cai2025-unilawR1}, use proprietary or restricted data. This limits reproducibility and community adoption, a tension particularly salient in legal AI where data licensing, confidentiality, and bar-association ethics rules restrict open sharing of real case materials.

\subsection{Number of datasets for training domain-specific reasoning model found}

\begin{itemize}
    \item \textbf{All}: Zero / between 1 to 3 / \textbf{between 4 and 6} / many ($>$ 6)

    \item \textbf{Publicly accessible}: Zero / between 1 to 3 / \textbf{between 4 and 6} / many ($>$ 6)
\end{itemize}

\subsection{Number of benchmarks/datasets for evaluating domain-specific reasoning model found}

\begin{itemize}
    \item \textbf{All}: Zero / between 1 to 3 / between 4 and 6 / \textbf{many ($>$ 6)}

    \item \textbf{Publicly accessible}: Zero / between 1 to 3 / between 4 and 6 / \textbf{many ($>$ 6)}
\end{itemize}

\subsection{Number of domain-specific reasoning models found}

\begin{itemize}
    \item \textbf{All}: Zero / between 1 to 3 / \textbf{between 4 and 6} / many ($>$ 6)

    \item \textbf{Publicly accessible}: Zero / between 1 to 3 / \textbf{between 4 and 6} / many ($>$ 6)
\end{itemize}

\subsection{Number of methods for creating/using domain-specific reasoning models found}

\begin{itemize}
    \item \textbf{All}: Zero / between 1 to 3 / \textbf{between 4 and 6} / many ($>$ 6)

    \item \textbf{Publicly accessible}: Zero / \textbf{between 1 to 3} / between 4 and 6 / many ($>$ 6)
\end{itemize}

\section[Literature Review Summary: SH3]{Literature Review Summary: SH3}

\textbf{The Social World and Its Interactions.} Sociology, social psychology, education sciences, communication studies.

\subsection{Key Findings}

The reviewed literature indicates that, in SH3, mostly general-purpose Large Language Models with reasoning capabilities are used, rather than domain-specific models. Researchers utilize a mix of \textbf{native reasoning models} (such as DeepSeek-R1 and QwQ-32B \cite{rumor2025, prison2025}) and other models adapted for reasoning tasks (including GPT-4o, LLaMA 3.1, Claude 3.5 Sonnet \cite{bias2025, pcot2025, bigtom2025, ting2025_astro_edu}), forcing the generation of intermediate thought processes. The models' domain adaptation is primarily achieved through prompting. Most existing works rely on text-only modalities. 

In social sciences, there is a tendency to replicate human psychological studies on LLMs or validate their outputs against humans. For reasoning models, this trend is less pronounced and includes studies aiming to expose latent biases in the "thinking" steps, like \cite{bias2025}. Another large area are social simulations using LLMs \cite{simulations2025}, but with no specific focus on reasoning models. 

Furthermore, some novel domain-specific reasoning prompts have emerged, like Pedagogical Chain-of-Thought (P-CoT), which forces models to simulate "didactic thinking" to explain student errors step-by-step \cite{pcot2025}, or Elimination-of-Thought, which is used for structured deductive reasoning in multiple-choice questions \cite{elimination2025}. In some cases, reasoning is enhanced via dynamic training curricula rather than just prompting \cite{learning2025}.

Expected outputs range from closed-ended answers and categorical classifications (e.g., mapping to Bloom's Taxonomy \cite{bloom2025}) to open-ended Socratic dialogues \cite{ting2025_astro_edu}, qualitative coding \cite{primer2025}, and logical justifications for fake news \cite{badactor2025}. Model reasoning is validated by comparing the thought trails against humans \cite{bigtom2025}, pedagogical frameworks \cite{pcot2025}, and established paradigms in social psychology \cite{bridging2025}.

\subsection{Identified Gaps}
The literature highlights several gaps at the intersection of reasoning LLMs and social sciences:
\begin{itemize}
    \item As much research focuses on bias, CoT often introduces or reinforces social stereotypes in ambiguous contexts, with models "reasoning" their way into biased conclusions \cite{bias2025}.
    \item Models struggle with psychological nuances that depend on social context \cite{bridging2025}.
    \item Reasoning models are significantly more effective at generating deceptive content than detecting it, creating a risk-amplification asymmetry \cite{prison2025}.
    \item Current models lack long-term personalization capabilities \cite{eduagents2025} and generate "alien" behaviors in simulations \cite{simulations2025}. They also hallucinate in specialized subjects \cite{ting2025_astro_edu}.
    \item Reasoning help models bypass safety constraints, making them more adept at manipulation, avoiding legal responsibility (as seen in the PRISON framework), and generating persuasive disinformation \cite{rumor2025, prison2025}.
    \item Current alignment methods optimize reasoning for individual user satisfaction (sycophancy). There is a lack of reasoning mechanisms aligned for "social welfare" and fair resource distribution in multi-agent social simulations \cite{gtalign2025}.
    \item There is a lack of unified reporting standards and methodological rigor, which hinders the replicability of computational social science research \cite{primer2025}.
\end{itemize}

\subsection{Notable domain-specific tasks that have been addressed by reasoning models}
Reasoning models have successfully addressed a number of tasks:
\begin{itemize}
    \item Evaluating and mitigating social bias within the reasoning processes \cite{bias2025}.
    \item Educational tutoring via Socratic dialogue \cite{ting2025_astro_edu} and detecting students' logical mistakes using the Pedagogical Chain-of-Thought \cite{pcot2025}.
    \item Classifying and generating educational questions \cite{bloom2025}.
    \item Simulating human decision-making and pro-social behavior in behavioral economics games based on verbalized motives \cite{suva2025}.
    \item Providing logical justifications to help humans detect fake news and misinformation \cite{badactor2025}.
    \item Qualitative coding and generation of synthetic human responses for survey research \cite{primer2025}.
\end{itemize}

\subsection{Domain-specific tasks that have not been addressed by reasoning models}
Several tasks remain unaddressed or only partially solved:
\begin{itemize}
    \item Fully integrating specific social psychology paradigms into the model's contextual understanding and reasoning paths \cite{bridging2025}.
    \item Consistently tracking the mental states of multiple actors in complex scenarios \cite{bigtom2025}.
    \item Generating readable and factually accurate debunking of rumors without safety bypasses \cite{rumor2025}.
    \item Sustained management of complex, realistic social interactions within a simulated classroom without exhibiting sycophancy \cite{simulations2025, eduagents2025}.
\end{itemize}

\subsection{Comments on the usage of reasoning model in RAG-like system or agentic framework}

Most areas, and especially simulations, focuses on argentic approaches. In educational settings, LLMs function as tutoring agents equipped with planning and memory modules to personalize learning paths \cite{eduagents2025}. Social science simulations struggle with overcoming non-human behaviors and a lack of mechanisms promoting verifiable, diverse social interactions \cite{simulations2025}.

Some educational research use RAG-like systems to include domain knowledge \cite{ting2025_astro_edu}.

\subsection{Other comments}
Techniques designed to enhance reasoning can amplify malicious capabilities, and reinforce implicit social biases during the reasoning steps \cite{rumor2025, prison2025, bias2025}. Reasoning in LLMs must be approached with caution in the social research.

\subsection{Number of datasets for training domain-specific reasoning model found}

\begin{itemize}
    \item \textbf{All}: between 1 to 3. \textit{Examples include AIME BaseSet-7K and AugSet-10K, which are used for training reasoning capabilities via dynamic curricula \cite{learning2025}, although they are not domain-specific.}
    \item \textbf{Publicly accessible}: between 1 to 3.
\end{itemize}

\subsection{Number of benchmarks/datasets for evaluating domain-specific reasoning model found}

\begin{itemize}
    \item \textbf{All}: many ($>$ 6). \textit{Examples: BBQ, StereoSet (Bias \cite{bias2025}); BigToM (ToM \cite{bigtom2025}); PRISON (Criminality \cite{prison2025}); SUVA (Social Behavior \cite{suva2025}); GSM8K-Mistakes (Education \cite{pcot2025}); ReCOPP/PolitiFact (Misinformation \cite{badactor2025})}
    \item \textbf{Publicly accessible}: many ($>$ 6).
\end{itemize}

\subsection{Number of domain-specific reasoning models found}

\begin{itemize}
    \item \textbf{All}: Zero.
    \item \textbf{Publicly accessible}: Zero.
\end{itemize}

\subsection{Number of methods for creating/using domain-specific reasoning models found}

\begin{itemize}
    \item \textbf{All}: between 4 and 6. \textit{These include Pedagogical Chain-of-Thought (P-CoT) \cite{pcot2025}, Elimination-of-Thought for deductive reasoning \cite{elimination2025}, Game-Theoretic Alignment for social welfare \cite{gtalign2025}, Adaptive Difficulty Curriculum Learning (ADCL/EGSR) \cite{learning2025}, and the SUVA agentic prompting framework \cite{suva2025}.}
    \item \textbf{Publicly accessible}: between 4 and 6.
\end{itemize}

\section[Literature Review Summary: SH4]{Literature Review Summary: SH4}

\textbf{The Human Mind and Its Complexity.} Cognitive science, psychology, linguistics.

\subsection{Key Findings}

Across psychological applications, most works do not train fully new domain-specific reasoning architectures, but instead adapt general-purpose LLMs (e.g., GPT, Llama, DeepSeek) using prompting, fine-tuning, or RL-based post-training. The dominant pattern is to treat reasoning as an interface layer (chain-of-thought, multi-stage or agent-based workflows) wrapped around a general model, rather than designing bespoke architectures.

\textbf{Clinical psychology and mental-health–related}: 
CoT and multi-step pipelines structure diagnostic reasoning are commonly used, often mirroring psychological theories for transparency and alignment with clinical workflows. Examples include Psyche-R1, which builds a 75k CoT dataset (query–rationale–answer triples) and uses Group Relative Policy Optimization to create a 7B model matching larger baselines in diagnostic accuracy (\cite{dai2025psycher1reliablepsychologicalllms}); and ERD, a three-phase "extraction–reasoning–debate" pipeline that identifies distorted statements, reasons explicitly, then self-debates before labeling cognitive distortions, enhancing clinical interpretability (\cite{lim2024erdframeworkimprovingllm}).

Theory-grounded approaches for Cognitive Behavioral Therapy (CBT) include Socratic reasoning simulation, where LLMs generate CBT-grounded rationales to reframe negative thoughts positively (\cite{goel-etal-2025-socratic}), and a CBT pipeline combining subjectivity assessment, contrastive reasoning, and schema analysis for distortion classification (\cite{chen-etal-2023-empowering}). Recent work such as CognitionChain further advances explainable psychological reasoning by explicitly structuring multi-stage LLM workflows that bridge cognitive theory and model interpretability for stress detection problem (\cite{wang2024cognitionchainexplainablepsychological}). Depression and mental health prediction further test strategies like standard CoT, self-consistency, and reflection for tasks such as suicidal ideation detection, yielding consistent sensitivity gains over direct prompting (\cite{teng2025enhancingdepressiondetectionchainofthought}, \cite{patil2026cognitivementalllmevaluatingreasoninglarge}).

Several authors have extended beyond purely textual representations, demonstrating that integrating data from multiple sensors—capturing factors such as location, phone usage, bluetooth interactions, calls, physical activity, and sleep—can effectively support the generation of natural language summaries synthesizing diverse behavioral signals. Such summaries may, in turn, serve as valuable tools for fostering collaborative reflection between clinicians and patients, strengthening the therapeutic alliance, and guiding personalized treatment (\cite{10.1145/3659604}).

\textbf{Theory of Mind}: studies rely more heavily on benchmark-style datasets and controlled experimental setups than on domain-specific training. Systematic reviews of ToM evaluation collate multiple datasets involving false-belief tasks, second-order beliefs, and open-ended ToM questions, and show that models exhibit strong performance on simple, text-only false-belief scenarios but degrade in more naturalistic, open-ended settings (\cite{saritas2025systematicreviewevaluationlarge}). Individual experiments comparing humans and LLMs reasoning processes on open-ended ToM questions report that models struggle to produce high-quality CoT reasoning that aligns with human (\cite{10.1145/3627673.3679832}). 
% Recent work also moves beyond static ToM-style benchmarks toward interactive evaluation of higher-order social cognition. For example, SAGE (Sentient Agent as a Judge) instantiates an LLM-based evaluative agent that simulates persona-conditioned emotional change, inner thoughts, hidden intentions, and multi-turn replies during supportive dialogues, producing both an emotion trajectory and interpretable reasoning traces for judging how well a target model makes a user feel understood. In this setting, reasoning is not evaluated only through final answers, but through the model's ability to sustain empathetic, context-sensitive interaction over time, and the resulting rankings differ substantially from conventional leaderboards focused on general helpfulness \cite{zhang2025sentient}.
% @misc{zhang2025sentient,
%       title={Sentient Agent as a Judge: Evaluating Higher-Order Social Cognition in Large Language Models}, 
%       author={Bang Zhang and Ruotian Ma and Qingxuan Jiang and Peisong Wang and Jiaqi Chen and Zheng Xie and Xingyu Chen and Yue Wang and Fanghua Ye and Jian Li and Yifan Yang and Zhaopeng Tu and Xiaolong Li},
%       year={2025},
%       eprint={2505.02847},
%       archivePrefix={arXiv},
%       primaryClass={cs.CL},
%       url={https://arxiv.org/abs/2505.02847}, 
% }

\textbf{Social influence and interpersonal reasoning}: SocialGPT combines visual encoders with a text-only LLM to reason about social relations in images: visual models extract structured scene descriptions which are then processed by a prompted LLM that produces relation labels and natural language justifications, achieving strong zero-shot performance (\cite{li2024socialgptpromptingllmssocial}). Several recent works use LLMs to detect and explain social or emotional influence in text; for instance, some studies prompt models to assign influence labels (e.g., low / moderate / high, controversial, repetitive) and generate short rationales, evaluating them against human annotations (\cite{mieleszczenkokowszewicz2025unravelingsittsocialinfluence}, \cite{Gassmann_Campbell_Edwards_2024}). The EmoSocInflu paper extends this line of work by introducing a span-level annotated dataset of 238 dialogues with 11 emotional social influence techniques and using LLMs both for technique detection and explanation; it benchmarks multiple models on span detection and then ranks their explanations via human pairwise preference on four criteria (comprehensibility, completeness, cognitive coherence, soundness), showing that current models are still limited at fine-grained detection (\cite{Markiewiczetal2026}).

% \textbf{Neuropsychology/Cognitive Aging}: 

\textbf{Memory, psycholinguistics, cognitive modeling, and cognitive aging}: Beyond clinical applications, LLMs are also being used as computational models of human cognition in controlled psychological experiments. ChatGPT was examined if it could predict human memory performance for garden-path sentences presented in fitting versus unfitting contextual conditions, finding that the model's relatedness judgments closely track human ratings and that its memorability estimates predict subsequent human memory performance \cite{Ulakci2025}. This line of work is not primarily about domain-specific reasoning pipelines such as CoT or agentic diagnosis; rather, it uses LLM outputs as cognitively meaningful proxies for latent psychological variables such as contextual fit and memorability, suggesting a role for LLMs as tools for modeling human cognitive processes in experimental psychology.

Recent work on cognitive aging suggests that LLMs can contribute to early cognitive-decline detection through hybrid reasoning pipelines. LLMs have been used both to analyze EHR clinical notes, where ensemble systems combining GPT-4 with traditional classifiers achieved the best performance, and to extract explainable high-level features from free dialogues, which can then be classified by specialized ML models. These findings indicate that the main value of LLMs in this area may lie in representation, explanation, and integration with task-specific models rather than end-to-end autonomous diagnosis (\cite{deArribaPrez2024}, \cite{du2024enhancing}).

% \textbf{Safety, robustness and interpretability in human–AI interaction}: Surveys on “reasoning about reasoning” in Human–computer interaction (HCI) summarize how researchers use CoT and related techniques to study persuasion, perceived competence and trust, and document cases where models can produce convincing argumentative structures without robust underlying understanding (\cite{mothilal2025reasoningreasoninginformedreflective}, \cite{dewynter2025linecomprehensionpersuasionllms}). In geriatric and safety-critical settings, such as alerting systems in elderly care facilities, LLM reasoning is embedded into rule-based frameworks: the model is prompted with structured event descriptions and asked to judge whether an alert is truly critical, reducing false alarms while leaving ultimate control to deterministic logic (\cite{s25216560}).

\subsection{Identified Gaps}

\begin{enumerate}
    \item No dedicated domain-specific reasoning models: Despite extensive prompting and fine-tuning of general LLMs, no papers develop or release fully specialized reasoning architectures, optimized purely for psychological tasks such as clinical diagnosis or social influence detection.

    \item Limited clinical validation and real-world deployment: Most diagnostic reasoning uses synthetic or small annotated datasets without large-scale validation against expert clinicians or longitudinal patient outcomes.

    \item Insufficient multimodal integration for behavioral analysis: While some works use text+image or sensor data, full multimodal reasoning pipelines (e.g., video/audio+text for naturalistic therapy sessions) remain underexplored.

    \item Span-level and fine-grained detection challenges: LLMs struggle with precise text-span identification for subtle psychological phenomena (e.g., emotional influence techniques).

    \item Lack of standardized psychological benchmarks: Evaluations often rely on heterogeneous, small-scale datasets (e.g., 238 dialogues in EmoSocInflu) or repurposed cognitive tasks; no unified benchmark suites with psychometrically validated metrics for reasoning across ToM, diagnosis, and social simulation.

    \item Cross-cultural and multilingual limitations: Datasets and evaluations are predominantly English-centric ; emotional/social reasoning shows cultural dependence.

\end{enumerate}

\subsection{Notable domain-specific tasks that have been addressed by reasoning models}

\begin{enumerate}
    \item Clinical mental disorder diagnosis: Many articles show that CoT reasoning can be used for disorder diagnoses from patient descriptions or other related information.

    \item Cognitive distortion classification: There are frameworks for the detection and explanation of distortions (e.g., catastrophizing, overgeneralization).

    \item Psychological stress detection: Cognition Chain decomposes the stress detection problem into Stimulus→Evaluation→Reaction→Stress stages for explainable prediction from social media text.

    \item Social relation inference: combination of visual feature extraction with LLM chain reasoning to infer interpersonal relations from images in zero-shot settings.

    \item Emotional social influence detection: span-level identification in dialogues, with LLM explanations.

    \item Theory of Mind reasoning: Systematic evaluation of false-belief tasks, second-order beliefs, and open-ended social inference.

    \item Depression detection: CoT on patient narratives.

    \item Multimodal emotion recognition: multi-stage reasoning over text, audio and image.

    % \item Agent-based social simulation (\cite{taillandier2026integratingllmagentbasedsocial}): LLM agents simulate interpersonal dynamics, influence propagation, and emergent behaviors in hybrid ABM environments.

    \item Mental health prediction: Reasoning variants tested on suicidal ideation and general mental health classification.

    \item Cognitive decline detection: LLM reasoning features from free conversations for ML-based classification with explainability.
    
\end{enumerate}

\subsection{Domain-specific tasks that have not been addressed by reasoning models}

\begin{enumerate}
    \item Longitudinal therapy outcome prediction: No reasoning pipelines predict long-term treatment response, relapse risk, or developmental trajectories from sequential session transcripts, despite single-session diagnosis being addressed.

    \item Cross-cultural Theory of Mind adaptation: ToM reasoning is evaluated mostly in WEIRD contexts; no benchmark tests cultural variations in belief attribution, social norms, or developmental disorders across populations.

    \item Real-time crisis intervention reasoning: Static text analysis exists, but there is no interactive, low-latency reasoning for live suicide/de-escalation chatbots with safety guaranties during acute episodes.

    \item Fine-grained personality inference and evolution: High-level traits touched via prompting, but no span-level DSM-5/ICD-11 trait with reasoning detection, dynamic personality change modeling over time, or comparative personality assessment.

    \item Wearable+therapy multimodal fusion: Sensor data are reasoned separately; no unified chains integrate physiological signals with verbal/discourse reasoning for neuropsychology or aging cognition prediction.

    % \item Multi-agent clinical role-play simulation: Single-agent therapy addressed; no hospital-team (doctor–nurse–patient–social worker) or developmental assessment simulations with inter-agent reasoning for coordinated diagnosis.

    \item Cognitive bias causal chain detection: Individual distortions identified, but no full causal chains (e.g., confirmation bias → catastrophizing → avoidance behavior) with personalized intervention reasoning.

    \item Non-verbal cue reasoning in video therapy: Audio/text emotion is partially solved; full video analysis (facial micro-expressions, posture, eye gaze) for rapport/disengagement, consciousness states, or perception-action inference with reasoning is absent.

    % \item Episodic memory and ageing cognition simulation: Cookie Theft tasks touched for dementia; no reasoning over longitudinal memory decay, prospective memory failures, or comparative cognition in ageing populations.

    % \item Pragmatics/discourse-level deception detection: Social influence is touched upon, but there is no reasoning for sociolinguistic deception or anthropological discourse manipulation analysis. 

    \item Ethical dilemma resolution in counseling: Abstract ethics exist; no patient-specific trade-off reasoning (confidentiality vs. harm prevention vs. cultural norms) grounded in the philosophy of mind/epistemology.

    % \item Population-level mental health forecasting: Individual prediction touched; no reasoning over aggregate social media/discourse trends for outbreak detection (e.g., anxiety epidemics) or intelligence/decision-making shifts (\cite{wang2024cognitionchainexplainablepsychological}).

    \item Language acquisition reasoning: No LLM reasoning simulates developmental language milestones, typology comparisons, or historical linguistics reconstruction from corpora.

    \item No detection of eristic techniques manipulation: Although manipulation has received some attention, there is still a clear gap in the analysis of the occurrence and detection of eristic techniques.

    % \item Philosophy of mind integration: Logic/epistemology is touched upon peripherally; no systematic reasoning on qualia, intentionality, or consciousness simulation using psychological benchmarks.
\end{enumerate}

\subsection{Comments on the usage of reasoning model in RAG-like system or agentic framework}

% \begin{enumerate}
%     \item Psyche-R1 as RAG-augmented diagnostic agent (\cite{dai2025psycher1reliablepsychologicalllms}): Combines CoT reasoning with retrieval of clinical knowledge and patient history; GRPO optimization enables retrieval-conditioned rationale generation for consistent diagnosis across sessions.

%     \item ERD as multi-stage RAG pipeline (\cite{lim2024erdframeworkimprovingllm}): Extraction phase retrieves candidate distortions from therapy text, reasoning phase augments with psychological theory definitions, debate phase performs self-RAG critique—ideal for modular clinical agents.
% \end{enumerate}
% There are not many solutions using reasoning models in RAG-like systems or agentic frameworks; notable examples include Psyche-R1, that combines CoT with clinical knowledge retrieval for diagnosis (\cite{dai2025psycher1reliablepsychologicalllms}), while ERD employs a multi-stage RAG pipeline in which the extraction phase retrieves candidate distortions from therapy text, the reasoning phase augments with psychological theory definitions, and the debate phase performs self-RAG critique—ideal for modular clinical agents \cite{lim2024erdframeworkimprovingllm}.
Solutions using reasoning models in RAG-like systems or agentic frameworks were not found in that field.

\subsection{Other comments}

\begin{enumerate}
    \item Extreme subjectivity challenge: Psychological reasoning spans objective diagnosis to subjective clinical judgment (e.g., rapport, therapeutic alliance); inter-rater reliability varies dramatically across tasks, complicating LLM evaluation.

    \item Severe data scarcity: Clinical datasets remain proprietary/small (e.g., EmoSocInflu's 238 dialogues); synthetic CoT data helps but lacks real patient diversity and longitudinal outcomes.

    \item Scale matters critically: Larger reasoning models (o3, Claude 3.5 Sonnet) consistently outperform smaller ones in explanation quality and human preference (with rare exceptions). 

    % \item Theory-grounding boosts reliability: Chains explicitly mirroring psychological frameworks (Cognition Chain, appraisal theory) yield more interpretable results than generic CoT.
    
    \item Broad survey evidence suggests that LLM use in psychology extends well beyond diagnosis or labeling tasks, encompassing literature review, hypothesis generation, experimental design, data analysis, and other research-support functions, while also raising recurring concerns about privacy, ethics, and model limitations (\cite{ke2025frontiers}).

    % \item Explanation trumps prediction: Human evaluators prioritize completeness/soundness over raw accuracy; faithful rationales matter more than correct labels in clinical/trust contexts [https://github.com/social-influence/emo-soc-influ](https://github.com/social-influence/emo-soc-influ).

    % \item Hybrid symbolic+LLM future: Pure LLM agents fail long-term social simulations due to memory/personality drift; ABM+reasoning hybrids needed for realistic psychological modeling.

    \item Clinical deployment lag: Strong proof-of-concept reasoning exists, but ethical/regulatory hurdles (bias amplification, liability) block real-world therapy systems.

    % \item Cultural blind spots persist: Emotional/social reasoning heavily WEIRD-centric; Polish EmoSocInflu dataset valuable but needs multilingual expansion.
\end{enumerate}

\subsection{Number of datasets for training domain-specific reasoning model found}

\begin{itemize}
    \item \textbf{All}: many ($>$ 6)

    \item \textbf{Publicly accessible}: between 1 to 3
\end{itemize}

\subsection{Number of benchmarks/datasets for evaluating domain-specific reasoning model found}

\begin{itemize}
    \item \textbf{All}: many ($>$ 6)

    \item \textbf{Publicly accessible}: many ($>$ 6)
\end{itemize}

\subsection{Number of domain-specific reasoning models found}

\begin{itemize}
    \item \textbf{All}: between 1 to 3

    \item \textbf{Publicly accessible}: between 1 to 3
\end{itemize}

\subsection{Number of methods for creating/using domain-specific reasoning models found}

\begin{itemize}
    \item \textbf{All}: many ($>$ 6)

    \item \textbf{Publicly accessible}: many ($>$ 6)
\end{itemize}

\section[Literature Review Summary: SH5]{Literature Review Summary: SH5}

\textbf{Texts and Concepts.} Literary studies, literature, philosophy.

\subsection{Key Findings}

A recurring theme  across multiple papers is that Chain-of-Thought frequently degrades model performance in literary and narrative understanding tasks. Which is contrary to CoT's benefits in many STEM tasks.
\begin{itemize}
    \item \textbf{LitBench} \cite{litbench2025}: CoT decreased evaluation accuracy to 72\%, with non-reasoning Claude 3.7 performing best
    \item \textbf{Trope understanding} \cite{su2024trope}: CoT causes hallucinations
    \item \textbf{Creative writing evaluation}: Explicit reasoning appears to interfere with subjective aesthetic judgments
\end{itemize}

\subsubsection{Datasets and Benchmarks}
\begin{itemize}
    \item \textbf{LFED} \cite{lfed2024}: First Chinese literary fiction benchmark with 1,304 multiple-choice questions from 95 novels covering 8 question types
    \item \textbf{LitBench} \cite{litbench2025}: Creative writing evaluation dataset with 265k human annotations from 46 annotators, based on Reddit story comparisons
    \item \textbf{Trope Dataset} \cite{su2024trope}: 5,623 movie synopses annotated with narrative tropes from TiMoS
    \item \textbf{Long Context RELiC} \cite{litevidence2025}: 292 examples requiring retrieval of literary evidence from complete works to support critical interpretations
\end{itemize}

\subsubsection{Models Tested}
All papers use model from the text-only perspective.
\begin{itemize}
    \item \textbf{Closed}: GPT-3.5, GPT-4, GPT-4.1, Claude 3.5/3.7, Gemini Pro 1.5/2.5, o1, o3, o3-mini, o4-mini
    \item \textbf{Open-source}: Llama-2-7B, Llama-3-8B, Llama-3.1/3.3, Gemma-7B, Mistral-7B, Qwen 2.5, DeepSeek-V3, DeepSeek-R1, ChatGLM-6B, BELLE-7B, BLOOM
    \item \textbf{Sizes}: Range from <1B to >100B parameters
\end{itemize}

\subsubsection{Prompting and Output Formats}
\begin{itemize}
    \item Multiple-choice questions (LFED)
    \item Preference judgments (LitBench)
    \item Evidence retrieval with textual spans (Long Context RELiC)
    \item Narrative-of-Thought (NoT) prompting: converts events to Python classes, generates grounded narratives \cite{zhang2024narrativeofthought}
\end{itemize}

\subsubsection{Validation Methods}
\begin{itemize}
    \item Human baseline comparisons
    \item Expert validation (literary scholars)
    \item Inter-annotator agreement (Cohen's Kappa, Fleiss' Kappa)
    \item Standard metrics: F1 scores, precision, recall, accuracy
\end{itemize}

\subsection{Identified Gaps}
\begin{enumerate}
    \item Why does CoT degrade literary reasoning? Multiple papers observe this phenomenon but none provide a satisfactory explanation.
    \item No domain-specific reasoning models for literature: All papers use general-purpose LLMs.
    \item Open-ended evaluation missing: Most benchmarks use multiple-choice or binary formats. Open-ended literary analysis and interpretation tasks remain underexplored.
    \item Multimodal literary reasoning absent: No work addresses illustrated texts, graphic novels.
    \item Philosophical reasoning underexplored: The philosophical counseling paper \cite{bokai2025philosophicalcounseling} touches this area but systematic evaluation on philosophical reasoning is missing.
    \item Long-form narrative generation: While Long Context RELiC tests understanding, generation of coherent long-form narratives remains challenging.
\end{enumerate}

\subsection{Notable domain-specific tasks that have been addressed by reasoning models}

\begin{enumerate}
    \item Literary fiction comprehension \cite{lfed2024}: multi-level understanding including contextual comprehension, common sense, logical reasoning
    \item Narrative trope identification \cite{su2024trope}: identifying recurring tropes in movie synopses which require thematic interpretation and cross-narrative connections
    \item Temporal narrative reasoning \cite{zhang2024narrativeofthought}: understanding and generating temporal relationships between evens.
    \item Creative writing evaluation \cite{litbench2025}: evaluating quality of creative writing with Generative Reward Models (mixed results)
    \item Interactive narrative understanding \cite{zhao2024narrativeplay}: reasoning about character personalities, relationships, and mental states (Theory of Mind kinda)
\end{enumerate}

\subsection{Domain-specific tasks that have not been addressed by reasoning models}

\begin{enumerate}
    \item Philosophical argumentation and reasoning
    \item Literary crisitcism generation
    \item Stylistic analysis - reasoning about authors' style, and techniques
    \item Metaphor and symbolism interpretation
    \item Genre classification with reasoning
    \item Unreliable narrator detection
    \item Poetry analysis
    \item Comparative analsysis of different texts
    \item Historical and contextual reasoning - situating texts in historical contexts
\end{enumerate}

\subsection{Comments on the usage of reasoning model in RAG-like system or agentic framework}
\begin{itemize}
    \item Long-context models vs RAG: \cite{litevidence2025} compares long-context reasoning models to RAG approaches (without mentioning RAG) for literary evidence retrieval
    \item NarrativePlay as agentic framework \cite{zhao2024narrativeplay}: agentic apprach where LLMs maintain character memory extracted from narratives
    \item Temporal reasoning in agents: The Narrative-of-Thought \cite{zhang2024narrativeofthought} could be valuable for agentic systems that need to have temporal coherence.
\end{itemize}

\subsection{Other comments}

\begin{itemize}
    \item Subjective challenge - many literary tasks involve subjective judgements
    \item There's limited number of exper-annotated datasets
    \item Bigger models were the most successfull
\end{itemize}

\subsection{Number of datasets for training domain-specific reasoning model found}

\begin{itemize}
    \item \textbf{All}: Zero / between 1 to 3 / between 4 and 6 / many ($>$ 6)

    \item \textbf{Publicly accessible}: Zero / between 1 to 3 / between 4 and 6 / many ($>$ 6)
\end{itemize}

\subsection{Number of benchmarks/datasets for evaluating domain-specific reasoning model found}

%\begin{itemize}
%    \item \textbf{All}: Zero / between 1 to 3 / between 4 and 6 / many ($>$ 6)

%    \item \textbf{Publicly accessible}: Zero / between 1 to 3 / between 4 and 6 / many ($>$ 6)
%\end{itemize}
\begin{itemize}
    \item All: between 4 and 6 (LFED, LitBench, Trope/TiMoS, Long Context RELiC, ProScript, Schema-11)

    \item Publicly accessible: between 4 and 6 (most are publicly available with MIT licenses)
\end{itemize}

\subsection{Number of domain-specific reasoning models found}

%\begin{itemize}
%    \item \textbf{All}: Zero / between 1 to 3 / between 4 and 6 / many ($>$ 6)
%    \item \textbf{Publicly accessible}: Zero / between 1 to 3 / between %4 and 6 / many ($>$ 6)
%\end{itemize}

\begin{itemize}
    \item All: between 1 to 3 (LitBench introduces Generative Reward Models for creative writing)

    \item Publicly accessible: Zero to 1 (GenRM approach described but model availability unclear)
\end{itemize}

\subsection{Number of methods for creating/using domain-specific reasoning models found}

%\begin{itemize}
%    \item \textbf{All}: Zero / between 1 to 3 / between 4 and 6 / many ($>$ 6)

%    \item \textbf{Publicly accessible}: Zero / between 1 to 3 / between 4 and 6 / many ($>$ 6)
%\end{itemize}

\begin{itemize}
    \item All: between 4 and 6
    \begin{itemize}
        \item Narrative-of-Thought (NoT) prompting \cite{zhang2024narrativeofthought}
        \item Generative Reward Models (GenRM) \cite{litbench2025}
        \item Character memory extraction for narrative understanding \cite{zhao2024narrativeplay}
        \item Adversarial injection for testing CoT robustness \cite{su2024trope}
        \item Long-context evidence retrieval \cite{litevidence2025}
    \end{itemize}

    \item Publicly accessible: between 1 to 3 (source code available for NoT, Trope, Long Context RELiC)
\end{itemize}

\section[Literature Review Summary: SH6]{Literature Review Summary: SH6}

\textbf{The Study of the Human Past.} Archaeology and history.

\subsection{Key Findings}

A critical observation is that \textbf{reasoning LLMs remain largely untested in historical and archaeological domains}. While reasoning models (o1, o3, DeepSeek-R1) have shown strong performance on STEM tasks, their application to history and archaeology is nascent. The few studies that exist focus primarily on knowledge retrieval rather than complex historical reasoning chains.
\begin{itemize}
    \item \textbf{HiST-LLM} \cite{histllm2024}: Evaluated seven models (GPT-3.5, GPT-4-Turbo, GPT-4o, Llama-3/3.1, Gemini-1.5-flash) with multi-shot prompting that asks models to provide reasoning before giving an answer -- LLMs outperform random guessing (25\%) but fall short of expert comprehension (33.6\%--46\% balanced accuracy)
    \item \textbf{Reasoning Over the Glyphs} \cite{reasoningglyphs2025}: Evaluates LLMs' and LVLMs' reasoning capabilities for deciphering rare scripts not encoded in Unicode, introducing a novel multimodal dataset and glyph tokenization method
    \item \textbf{TimeTravel} \cite{timetravel2025}: Evaluates multimodal models on 10,250 samples across 266 cultures for historically accurate artifact description generation -- reveals gaps in historical specificity and culturally grounded understanding
    \item \textbf{Chain-of-Thought not systematically evaluated}: Unlike literary domains where CoT degradation is documented, no study systematically compares CoT vs non-CoT performance for historical reasoning
\end{itemize}

\subsubsection{Datasets and Benchmarks}
\begin{itemize}
    \item \textbf{HiST-LLM} \cite{histllm2024}: History Seshat Test for LLMs, based on a subset of the Seshat Global History Databank, with 36,000 data points across over 600 historical polities from Neolithic to Industrial Revolution
    \item \textbf{TimeTravel} \cite{timetravel2025}: 10,250 expert-verified samples spanning 266 cultures across 10 major historical regions for artifact analysis
    \item \textbf{LogogramNLP} \cite{logogramnlp2024}: Benchmark for ancient logographic writing systems including Linear A, Egyptian hieroglyphics, Cuneiform, and Bamboo Script
\end{itemize}

\subsubsection{Models Tested}
All papers primarily use text-only models. \textbf{Dedicated reasoning models (o1, o3, DeepSeek-R1) have not been systematically evaluated} for historical/archaeological tasks.
\begin{itemize}
    \item \textbf{General LLMs tested}: GPT-3.5, GPT-4, GPT-4.1, Claude 3.5/3.7, Gemini Pro 1.5/2.5, Llama-2/3/3.1, Qwen 2.5
    \item \textbf{Reasoning models}: o1, o3, DeepSeek-R1 mentioned in some benchmarks but \textbf{not systematically compared} with non-reasoning variants on historical tasks
    \item \textbf{Specialized translation models}: Hieroglyphic Transformer (M2M-100 fine-tuned) \cite{hierotransformer2024} -- translation-focused, not reasoning-focused
    \item \textbf{Multimodal}: GPT-4o, GPT-4o-mini, Gemini-2.0-Flash, Gemini-1.5-Pro for TimeTravel benchmark
    \item \textbf{Gap}: No domain-specific reasoning model for history/archaeology exists
\end{itemize}

\subsubsection{Prompting and Output Formats}
\begin{itemize}
    \item \textbf{Multi-shot with reasoning request} (HiST-LLM): Models asked to ``provide reasoning before answer'' -- explicit reasoning elicitation
    \item \textbf{Artifact description generation} from images (TimeTravel): Models generate historically and culturally grounded descriptions of artifacts
    \item \textbf{Linguistic puzzle format} \cite{reasoningglyphs2025}: Novel tokenization approach for rare scripts requiring decipherment reasoning
    \item Transliteration and translation tasks (ancient languages)
    \item \textbf{Anthropological Prompting} (AlKhamissi et al., 2024, as cited in \cite{timetravel2025}): Encouraging emic (insider) and etic (outsider) reasoning perspectives
    \item \textbf{CoT not standard}: Unlike other domains, explicit Chain-of-Thought prompting is not standard practice in historical/archaeological evaluations
\end{itemize}

\subsubsection{Validation Methods}
\begin{itemize}
    \item Expert validation (historians, archaeologists, linguists)
    \item Comparison with Seshat Global History Databank ground truth
    \item BLEU scores for translation tasks (up to 44.15 for cuneiform translation with PIXEL-MT; external Conv-s2s baselines at 36.52--37.47 \cite{logogramnlp2024})
    \item Human evaluation for cultural alignment
    \item Cross-lingual evaluation for ancient text processing
\end{itemize}

\subsection{Identified Gaps}
\begin{enumerate}
    \item \textbf{No dedicated reasoning model evaluation}: Reasoning models (o1, o3, DeepSeek-R1) have not been systematically benchmarked on historical/archaeological tasks. Existing studies use general-purpose LLMs.
    \item \textbf{CoT effectiveness unknown}: Unlike literary domains where CoT degrades performance, no systematic study compares CoT vs non-CoT prompting for historical reasoning tasks.
    \item \textbf{No domain-specific reasoning models}: All papers use general-purpose LLMs without fine-tuning for historical reasoning chains.
    \item \textbf{Cultural and geographic reasoning bias}: LLMs reason better about Western history and recent periods; worse reasoning for Sub-Saharan Africa and non-Western cultures.
    \item \textbf{Limited multimodal reasoning}: While TimeTravel addresses artifact images, comprehensive multimodal archaeological reasoning (combining text, images, 3D models, stratigraphic data) remains unexplored.
    \item \textbf{Reasoning under uncertainty}: Archaeological interpretation requires reasoning with incomplete evidence and multiple hypotheses -- current models struggle with this.
    \item \textbf{Long-form historical argumentation}: Complex historical reasoning chains spanning multiple sources and time periods remains challenging.
    \item \textbf{Causal and counterfactual historical reasoning}: Reasoning about historical causation (``why did X happen?'') and counterfactuals (``what if Y?'') is unexplored.
\end{enumerate}

\subsection{Notable domain-specific tasks that have been addressed by reasoning models}
\begin{enumerate}
    \item Historical knowledge assessment \cite{histllm2024}: Multiple-choice questions on presence/absence of historical variables (e.g., writing, irrigation) in specific polities and time periods
    \item Ancient text translation \cite{hierotransformer2024}: Translating Egyptian hieroglyphs to German/English using transformer models
    \item Artifact description generation \cite{timetravel2025}: Generating historically accurate and culturally grounded descriptions of artifacts from images
    \item Cuneiform text processing: Lemmatization and token prediction for Akkadian and Sumerian texts
    \item Cultural heritage text generation \cite{valuemisalignment2025}: Generating descriptions of cultural heritage items (with noted alignment issues)
    \item Knowledge graph construction \cite{kggenheritage2024}: Combining LLMs with ontological engineering for cultural heritage knowledge graphs
\end{enumerate}

\subsection{Domain-specific tasks that have not been addressed by reasoning models}

\begin{enumerate}
    \item \textbf{Archaeological site interpretation}: Multi-step reasoning to interpret excavation findings
    \item \textbf{Stratigraphic reasoning}: Sequential/temporal reasoning about archaeological layers and relative dating
    \item \textbf{Artifact provenance reasoning}: Chain-of-thought reasoning to trace origins from material/stylistic evidence
    \item \textbf{Historical causation reasoning}: Why-questions requiring causal reasoning chains
    \item \textbf{Counterfactual historical reasoning}: ``What if'' scenarios requiring alternative reasoning paths
    \item \textbf{Cross-cultural comparative reasoning}: Analogical reasoning across different cultures/periods
    \item \textbf{Script decipherment reasoning}: Systematic reasoning for undeciphered scripts (Linear A, Proto-Elamite) -- partially addressed by \cite{reasoningglyphs2025}
    \item \textbf{Source criticism reasoning}: Evaluating reliability of historical sources through structured reasoning
    \item \textbf{Evidence integration reasoning}: Combining radiocarbon dates, textual evidence, and material culture through multi-step reasoning
    \item \textbf{Long-term pattern reasoning}: Identifying and explaining patterns across civilizations
\end{enumerate}

\subsection{Comments on the usage of reasoning model in RAG-like system or agentic framework}
\begin{itemize}
    \item \textbf{No studies using dedicated reasoning models in RAG}: Existing RAG/agentic work uses general-purpose LLMs, not o1/o3/DeepSeek-R1
    \item \textbf{Hybrid KG+LLM approaches}: \cite{kggenheritage2024, ontologyllmheritage2024} show that combining ontologies/knowledge graphs with LLMs provides structured reasoning grounding -- could benefit from integration with reasoning models
    \item \textbf{RAFT fine-tuning}: Applied to archaeological contexts, showed up to 35\% improvement \cite{mdpiarch2025} -- potential for combining with reasoning model capabilities
    \item \textbf{Opportunity}: Archaeology's need for ``robust knowledge reasoning and interpretability'' \cite{mdpiarch2025} makes it well-suited for reasoning model applications in RAG settings
\end{itemize}

\subsection{Number of datasets for training domain-specific reasoning model found}

\begin{itemize}
    \item All: between 1 to 3 (Seshat Global History Databank, Thesaurus Linguae Aegyptiae, cuneiform corpora)

    \item Publicly accessible: between 1 to 3 (Seshat partially available, cuneiform corpora via CDLI)
\end{itemize}

\subsection{Number of benchmarks/datasets for evaluating domain-specific reasoning model found}

\begin{itemize}
    \item All: between 4 and 6 (HiST-LLM, TimeTravel, LogogramNLP, ML4AL shared tasks)

    \item Publicly accessible: between 4 and 6 (most benchmarks are publicly available)
\end{itemize}

\subsection{Number of domain-specific reasoning models found}

\begin{itemize}
    \item All: \textbf{Zero} -- No dedicated reasoning models for history/archaeology exist. Hieroglyphic Transformer is a translation model, not a reasoning model.

    \item Publicly accessible: \textbf{Zero}
\end{itemize}

\subsection{Number of methods for creating/using domain-specific reasoning models found}

\begin{itemize}
    \item All: between 1 to 3 (methods that explicitly involve reasoning)
    \begin{itemize}
        \item \textbf{Multi-shot reasoning prompting} \cite{histllm2024}: Asking models to ``provide reasoning before answer''
        \item \textbf{Anthropological Prompting}: Encouraging emic/etic reasoning perspectives for cultural understanding
        \item \textbf{Linguistic puzzle construction} \cite{reasoningglyphs2025}: Multimodal dataset with tokenization for script decipherment reasoning
    \end{itemize}

    \item Methods that involve LLMs but \textbf{not explicitly reasoning-focused}:
    \begin{itemize}
        \item RAFT (Retrieval Augmented Fine-Tuning) for archaeological QA
        \item Hybrid KG+LLM approaches for cultural heritage
        \item Visual encoding of logographic scripts \cite{logogramnlp2024}
    \end{itemize}

    \item Publicly accessible: between 1 to 3 (source code for LogogramNLP available)
\end{itemize}

\section[Literature Review Summary: SH7]{Literature Review Summary: SH7}

\textbf{Human Mobility, Environment, and Space.} Human geography, demography, health, sustainability science, territorial planning, spatial analysis.

\subsection{Key Findings}

\textbf{Benchmarks for the field of science (not for reasoning capabilities):}
\begin{itemize}
    \item TravelBench - contains two modules: KnowEval, which evaluates factual and spatial knowledge through CityQA and TripQA, and TripEval, which measures plan feasibility, personalization, and constraint satisfaction.
    \item CityEval - a comprehensive text-based spatial benchmark designed to evaluate LLM capabilities across tasks such as City Image, which assesses the understanding of fundamental urban elements, Urban Semantics, Spatial Reasoning, and Urban Composite Tasks, including mobility prediction and spatial navigation.
    \item Travel-Sim - an agent-based simulation benchmark for travel planning.
    \item PricingLogic - introduced as the first benchmark evaluating LLMs on complex tourism pricing tasks.
\end{itemize}

\noindent\textbf{Modalities:}
\begin{itemize}
    \item In most cases, the input consists of text combined with external tools, such as open Web APIs, map APIs, knowledge graphs, and domain-specific tools.
    \item In some cases, the input combines text and images, for example in demographic analysis based on facial photographs, or in urban planning frameworks using urban imagery and street-level photographs.
\end{itemize}

\noindent\textbf{Models:}
\begin{itemize}
    \item TravelLLM (AgentTravel) - Qwen 2.5-7B + LoRA.
    \item CityGPT - a novel fine-tuning method applied to ChatGLM3-6B, Llama3-8B, and Qwen2.5-7B.
    \item PlanGPT - Qwen 2.5-7B and ChatGLM3-6B-Base.
    \item Deepseek-R1-Distill models
    \item Various Qwen models,  ranging from 1.5B to 72B parameters.
    \item Various Llama models, ranging from 3B to 70B parameters.
    \item Closed source models, including the GPT series, Gemini, and Claude.
    \item other models like Gemma series, Yi-6B, Baichuan2-13B, Mistral Small 24B
    \item Multimodals, including LLaVA, MiniGPT-v2, InstructBLIP, and InternLM.
\end{itemize}

\noindent\textbf{Prompting strategies:}
\begin{itemize}
    \item Chain-of-Thought prompting.
    \item Structured inputs, such as tourist trajectories.
    \item Role-playing, particularly in LLM-based agent simulations.
\end{itemize}

\noindent\textbf{Expected outputs:}
\begin{itemize}
    \item Outputs are mostly provided in JSON format, primarily due to the use of AI agents.
    \item Closed answer sets with predefined schemas are used less frequently, for example in tasks such as age prediction in demographic analysis or multiple-choice question answering.
\end{itemize}

\noindent\textbf{Reasoning validation:}
\begin{itemize}
    \item Benchmarks, including simulation-based benchmarks. However, they primarily validate task results rather than reasoning processes.
    \item LLM-as-a-Judge approaches are used to validate results rather than reasoning processes.
    \item One exception is the Agentic Urban Planning framework, which requires reasoning chains to be explicitly valid, complete, and traceable, and proposes formalized metrics for Constraint Satisfaction Rate (CSR) and Reasoning Chain Quality (Q).
\end{itemize}

\subsection{Identified Gaps}
These observations reveal two major research gaps concerning the role, evaluation, and attribution of reasoning in LLM-based systems for human mobility, environment, and spatial analysis.
\begin{itemize}
    \item Reasoning is rarely evaluated properly. - Existing benchmarks mostly validate whether models produce correct or feasible outputs, while rarely analyzing the nature or quality of the reasoning process itself, including whether the reasoning path is valid, complete, traceable, domain-consistent, spatial, causal, temporal, or constraint-based.
    \item The distinction between reasoning and non-reasoning systems remains unclear. - Many studies evaluate LLMs as general-purpose models without clearly differentiating reasoning-oriented models, prompting-based reasoning, agentic reasoning, or standard text generation, while the integration of APIs, maps, knowledge graphs, and simulations further obscures whether reasoning originates from the LLM itself, external tools, retrieved knowledge, or the overall system architecture.
\end{itemize}

\subsection{Comments on the usage of reasoning model in RAG-like system or agentic framework}

Reasoning has been used as a tool to improve results in this field. However, authors have generally focused more on domain-specific outcomes than on the nature or structure of reasoning itself. In particular, reasoning is often employed as the “brain” of agentic frameworks, as well as in systems based on knowledge graphs.

 \subsection{Other comments}
Reasoning in this field remains poorly understood, and it is often treated similarly to standard LLM behavior, with little distinction made between reasoning and non-reasoning results.

\subsection{Number of datasets for training domain-specific reasoning model found}

\begin{itemize}
    \item \textbf{All}: Zero

    \item \textbf{Publicly accessible}: Zero 
\end{itemize}

\subsection{Number of benchmarks/datasets for evaluating domain-specific reasoning model found}

\begin{itemize}
    \item \textbf{All}: between 1 to 3

    \item \textbf{Publicly accessible}: between 1 to 3
\end{itemize}

\subsection{Number of domain-specific reasoning models found}

\begin{itemize}
    \item \textbf{All}: between 1 to 3

    \item \textbf{Publicly accessible}: between 1 to 3
\end{itemize}

\subsection{Number of methods for creating/using domain-specific reasoning models found}

\begin{itemize}
    \item \textbf{All}: many ($>$ 6)

    \item \textbf{Publicly accessible}: many ($>$ 6)
\end{itemize}

\section[Literature Review Summary: SH8]{Literature Review Summary: SH8}

\textbf{Studies of Cultures and Arts.} Social anthropology, studies of cultures, studies of arts.

\subsection{Key Findings}

The application of Reasoning Language Models (RLMs) in the Studies of Cultures and Arts (SH8) is characterized by a shift from simple content generation to complex, multi-step evaluative and creative tasks. The majority of works focus on establishing frameworks that force General Purpose LLMs (like GPT-4o, DeepSeek-R1, and Claude 3.5 Sonnet) to exhibit ``reasoning'' through structured prompting strategies (Chain-of-Thought, Rule-Guided Prompting) or Agentic architectures, rather than training domain-specific reasoning models from scratch.

A significant trend is the operational definition of ``reasoning'' within this domain. It is rarely defined as abstract cognitive ability, but rather as:
\begin{itemize}
    \item \textbf{Faithfulness and Grounding:} In religious and heritage contexts (Islamic studies, archival history), reasoning is defined as the capacity to verify facts against authoritative texts and avoid hallucination \cite{alangari2025arabic, Mushtaq2025islamic, pshenova2025heritage}.
    \item \textbf{Aesthetic and Structural Logic:} In arts and architecture, reasoning is defined as the decomposition of complex design requirements into geometric constraints (floorplans) or the step-by-step critical analysis of visual art (aesthetics) \cite{li2025chatassistdesign, lei2025godbench, jiang2025aesthethics, zong2024housellm}.
    \item \textbf{Sociological Simulation:} In social studies, reasoning is treated as the adherence to complex sociological rules to identify bias or historical oppression \cite{Khan2025intersectionbias, chatterjee2025cultural}.
\end{itemize}

Notable datasets and benchmarks include:

\begin{itemize}
    \item \textbf{GODBench \cite{lei2025godbench}:} A multimodal benchmark for evaluating ``Comment Art'' (creative, resonant video comments) using a ``Ripple of Thought'' reasoning framework. It includes human-annotated dimensions for creativity and divergent thinking.
    \item \textbf{ICH Evaluation Benchmark \cite{zhao2025cultural}:} A comprehensive evaluation set for Intangible Cultural Heritage, testing models on knowledge consistency using a ``Multi-Intelligencer'' approach where DeepSeek-R1 synthesizes scores from other models.
    \item \textbf{WinoIdentity \cite{Khan2025intersectionbias}:} A massive benchmark (245,700 prompts) for evaluating intersectional bias and coreference resolution, framing reasoning as the ability to distinguish relevant syntax from irrelevant demographic markers.
    \item \textbf{HSO-Bench \cite{chatterjee2025cultural}:} A dataset for assessing historical structural oppression, using ``Rule-Guided Prompting'' to align model outputs with sociological theory.
    \item \textbf{FineArtBench \& ArtCoT \cite{jiang2025aesthethics}:} A dataset and prompting framework that forces models to act as an ``Art Critic'' by first analyzing formal features and then reasoning about aesthetic value, significantly reducing subjective hallucinations.
    \item \textbf{RCM Level 6 Theory Dataset \cite{pond2025music}:} A music theory benchmark that highlights the struggle of text-based LLMs with symbolic music formats (MEI, MusicXML), showing that ``reasoning'' in music requires specific context-learning strategies.
\end{itemize}

\textbf{Methodologies and Modalities:} The dominant modality is text-only, utilized for tasks ranging from Islamic theology QA to archival extraction. However, multimodal approaches are rapidly maturing, particularly in:
\begin{itemize}
    \item \textbf{Text-to-Image/Video:} Used for aesthetic evaluation and video comment generation \cite{lei2025godbench, jiang2025aesthethics}.
    \item \textbf{Text-to-Vector/CAD:} Used in architectural design, where reasoning agents (LLMs) manage the ``logic'' of room layout while diffusion models handle the ``geometry'' \cite{li2025chatassistdesign, zong2024housellm}.
\end{itemize}

\textbf{Validation and Efficiency:} Validation primarily relies on ``LLM-as-a-judge'' (e.g., using GPT-4o or DeepSeek-R1 to score weaker models) \cite{zhao2025cultural, zhang-etal-2025-cultural, lei2025godbench} and comparison against human-curated Gold Standards. A recurring theme is the trade-off between reasoning depth and cost/latency; for instance, the ``Ripple of Thought'' framework \cite{lei2025godbench} improves creativity but increases inference time, and Retrieval-Augmented Re-prompting (R2P) \cite{alangari2025arabic} is triggered only when model ensembles disagree, to save costs.

\subsection{Identified Gaps}

\begin{itemize}
    \item \textbf{Lack of Domain-Specific Reasoning Foundation Models:} Unlike other sciences, there is no ``Culture-Reasoning-7B'' model. The field relies entirely on prompting general frontier models or fine-tuning generic adapters (e.g., HAZEL \cite{witte2025heritage}). 
    \item \textbf{Western-Centric Evaluation:} Despite efforts in Islamic and Chinese heritage, significant gaps remain in handling non-Western aesthetic traditions, under-represented languages (e.g., Old French, specific dialects), and localized historical narratives \cite{zhao2025cultural, Khan2025intersectionbias, chatterjee2025cultural}. 
    \item \textbf{Absence of ``Gold'' Reasoning Traces:} While many benchmarks provide the correct final answer (e.g., the correct floorplan or MCQ option), very few provide human-verified reasoning steps. Most ``reasoning'' traces in datasets are synthetically generated by other LLMs (e.g., Mistral or o3), creating a risk of circular validation \cite{Khan2025intersectionbias, Mushtaq2025islamic, zhang-etal-2025-cultural}. 
    \item \textbf{Multimodal ``Invisible'' Culture:} Current models struggle to reason about cultural artifacts that lack textual descriptions or are physically damaged, as they cannot infer cultural context purely from visual data without extensive textual aid \cite{zhang-etal-2025-cultural}.
\end{itemize}

\subsection{Notable domain-specific tasks that have been addressed by reasoning models}

\begin{itemize}
    \item \textbf{Faithful Theological Generation:} Using dual-agent systems (e.g., o3 paired with verification tools) to ensure Islamic essays and rulings do not misquote scripture or hallucinate jurisprudence \cite{Mushtaq2025islamic, alangari2025arabic}.
    \item \textbf{Architectural Floorplan Design:} Decoupling linguistic reasoning (identifying user needs and room connections) from geometric generation (drawing vectors), allowing LLMs to act as ``architects'' that guide diffusion models \cite{li2025chatassistdesign, zong2024housellm}.
    \item \textbf{Intersectional Bias Detection:} Using reasoning benchmarks to measure how models handle complex, overlapping identity markers (race, gender, disability) in coreference resolution tasks \cite{Khan2025intersectionbias}.
    \item \textbf{Aesthetic Criticism:} Transforming MLLMs from simple image captioners into ``Art Critics'' that use formal analysis (Chain-of-Thought) to ground aesthetic judgments and reduce subjectivity \cite{jiang2025aesthethics}.
    \item \textbf{Creative ``Comment Art'':} Automating the generation of culturally resonant, witty, and divergent comments for social media videos using wave-like reasoning propagation \cite{lei2025godbench}.
\end{itemize}

\subsection{Domain-specific tasks that have not been addressed by reasoning models}

\begin{itemize}
    \item \textbf{Complex Architectural Projects:} While models can generate single-story residential floorplans, they fail at multi-story structures requiring vertical connectivity (stairs, elevators), non-residential typologies (commercial buildings), and automated verification against building codes \cite{li2025chatassistdesign, zong2024housellm}.
    \item \textbf{Advanced Music Theory and Rhythmic Hierarchy:} Models can reason through basic intervals using MEI/MusicXML formats, but they struggle heavily with rhythmic grouping, beat hierarchy understanding, advanced counterpoint, and multi-step sequential harmonic analysis \cite{pond2025music}.
    \item \textbf{Marginalized Cultural and Religious Traditions:} AI struggles with non-Western historical data, minority religious sects (e.g., Shia jurisprudence), and contemporary edge cases. Furthermore, models struggle with avant-garde or conceptual art where interpretation is highly subjective \cite{Mushtaq2025islamic, jiang2025aesthethics, chatterjee2025cultural}.
    \item \textbf{Reasoning over Physically Damaged Artifacts:} Multimodal models easily misinterpret or lose "invisible" cultural symbols when presented with physically damaged artifacts (e.g., deteriorated murals) unless heavily augmented with textual context \cite{zhang-etal-2025-cultural}.
\end{itemize}

\subsection{Comments on the usage of reasoning model in RAG-like system or agentic framework}

RAG and agentic frameworks are foundational to the application of reasoning models in the SH8 domain, primarily to enforce factual grounding and manage complex creative workflows:
\begin{itemize}
    \item \textbf{Multi-Agent Systems for Decoupling Tasks:} Instead of relying on a single model, researchers deploy specialized agents. For instance, in faithful theological generation \cite{Mushtaq2025islamic}, a quantitative agent uses advanced reasoning (e.g., o3) and search tools to strictly verify citations, while a qualitative agent manages tone. In architectural design \cite{li2025chatassistdesign}, a language-interactive agent manages the logical state and user requirements, guiding a separate diffusion engine.
    \item \textbf{Dynamic and Conditional RAG:} To balance computational costs and precision, frameworks like Retrieval-Augmented Re-prompting (R2P) \cite{alangari2025arabic} trigger dense retrieval over classical texts only on-demand---specifically when an ensemble of LLMs fails to reach a consensus on complex Islamic jurisprudence questions.
    \item \textbf{Curatorial Agents (Human-in-the-Loop):} In heritage institutions, reasoning pipelines are used not for autonomous generation, but to classify URL credibility and generate targeted, clarifying questions for human experts. This grounds the LLM in verified heritage data while keeping the human firmly in the loop \cite{pshenova2025heritage, toth2025cultural}.
\end{itemize}

\subsection{Other comments}

\begin{itemize}
    \item \textbf{The Paradigm Shift to ``Digital Co-pilots'':} In the arts and humanities, where truth is often non-binary and culturally sensitive, LLMs are no longer framed as autonomous decision-makers. Instead, the literature emphasizes their role as ``digital co-pilots'' \cite{mdpiarch2025} or orchestrators for data self-organization \cite{toth2025cultural} operating strictly under human supervision.
    \item \textbf{Constraint-Based Reasoning over Parameter Scaling:} Success in this domain relies heavily on enforcing strict sociological, aesthetic, or theological rules during inference (e.g., Rule-Guided Prompting \cite{chatterjee2025cultural} or ArtCoT \cite{jiang2025aesthethics}), proving that guided reasoning frameworks are often more effective for cultural nuance than simply scaling up generic model parameters.
\end{itemize}

\subsection{Number of datasets for training domain-specific reasoning model found}

\begin{itemize}
    \item \textbf{All}: between 4 and 6
    \item \textbf{Publicly accessible}: between 1 to 3
\end{itemize}
\textit{Note:} Most datasets are for evaluation. Only a few, such as GODBench \cite{lei2025godbench}, C3 \cite{zhang-etal-2025-cultural}, and HAZEL \cite{witte2025heritage}, explicitly mention training/fine-tuning splits.

\subsection{Number of benchmarks/datasets for evaluating domain-specific reasoning model found}

\begin{itemize}
    \item \textbf{All}: many ($>$ 6)
    \item \textbf{Publicly accessible}: many ($>$ 6)
\end{itemize}
\textit{Note:} Almost every reviewed paper introduces a new benchmark, e.g., WinoIdentity, HSO-Bench, FineArtBench, QIAS 2025.

\subsection{Number of domain-specific reasoning models found}

\begin{itemize}
    \item \textbf{All}: Zero
    \item \textbf{Publicly accessible}: Zero
\end{itemize}
\textit{Note:} No standalone ``reasoning model'' (like a specialized pre-trained 70B model) was found. The papers utilize general models (DeepSeek-R1, GPT-4o, Mistral) or build frameworks/agents (ChatAssistDesign) rather than releasing new domain-specific reasoning foundation models.

\subsection{Number of methods for creating/using domain-specific reasoning models found}

\begin{itemize}
    \item \textbf{All}: many ($>$ 6)
    \item \textbf{Publicly accessible}: many ($>$ 6)
\end{itemize}
\textit{Note:} Methods include ``Ripple of Thought,'' ``Rule-Guided Prompting,'' ``ArtCoT,'' ``Multi-Intelligencer Evaluation,'' and ``Retrieval-Augmented Re-prompting''.

\section[Literature Review Summary: PE1]{Literature Review Summary: PE1}

\textbf{Mathematics.} All areas of mathematics, pure and applied, plus mathematical foundations of computer science, mathematical physics and statistics.

\subsection{Key Findings}
\subsubsection{Datasets and Benchmarks}
Recent works have introduced highly specialized benchmarks to address the limitations of general corpora and contamination risks. To evaluate high-level mathematical creativity, \cite{khatibi2025eefsuva} introduced \textit{EEFSUVA}, a benchmark curated from under-circulated Eastern European Olympiads, focusing on problems requiring non-trivial reasoning steps beyond pattern recognition. In the domain of applied mathematics and engineering, \cite{kevian2024capabilities} developed \textit{ControlBench} for control engineering, spanning topics like stability and Bode analysis, while \cite{zhang2025or} introduced \textit{BWOR} for Operations Research (OR), requiring models to handle cost optimization and transportation planning. Furthermore, synthetic data generation is gaining traction; for instance, \cite{yang2024optibench} utilized the \textit{ReSocratic} method—a reverse Socratic approach—to synthesize 29k optimization problems (\textit{RESOCRATIC-29K}) for fine-tuning.

On the pre-training side, \cite{shao2024deepseekmath} demonstrated the importance of scale and quality with the \textit{DeepSeekMath Corpus}, comprising 120B math tokens filtered from Common Crawl, which significantly outperforms standard datasets like OpenWebMath.

\subsubsection{Reasoning Models and Architectures}
While general-purpose models (e.g., GPT-4, Llama-3) remain a baseline, specialized architectures are emerging. \cite{shao2024deepseekmath} introduced \textit{DeepSeekMath}, utilizing Group Relative Policy Optimization (GRPO) to efficiently estimate baselines from group scores without a critic model, thereby reducing training resources while enhancing reinforcement learning outcomes.

A significant trend is the shift towards agentic and iterative frameworks rather than single-turn generic prompting. \cite{li2025codepde} proposed \textit{CodePDE}, a modular framework for Partial Differential Equation (PDE) solvers that employs distinct steps for code generation, repair, and refinement. Similarly, \cite{zhang2025or} utilized an \textit{OR-LLM-Agent} that decomposes tasks into modeling, coding, and debugging sub-agents. \cite{georgiev2025mathematical} pushed this further with \textit{AlphaEvolve}, an evolutionary coding agent that mimics natural selection to explore mathematical landscapes, allowing a single LLM call to trigger extensive, low-cost computation.

\subsubsection{Modalities and Multimodal Reasoning}
While text (specifically LaTeX) remains the dominant modality, there is a clear push towards Multimodal Large Language Models (MLLMs) to handle geometry and visual data. \cite{wang2024measuring} released \textit{MATH-Vision (MATH-V)}, a dataset meticulously curated from math competitions, containing over 3,000 problems with visual inputs. Addressing the need for improved visual reasoning, \cite{shi2024math} introduced \textit{Math-LLaVA}, which bootstraps mathematical reasoning by integrating visual knowledge into the LLaVA architecture. Additionally, \cite{kevian2024capabilities} noted that real-world engineering problems often require interpreting visual elements like Bode and Nyquist plots.

\subsubsection{Prompting Strategies and Validation}
Prompting strategies have evolved from standard Chain-of-Thought (CoT) to more structured and verifiable formats.
\begin{itemize}
    \item \textbf{Causal and Symbolic Verification:} \cite{yu2025causal} proposed a causal framework that evaluates CoT traces for "sufficiency" and "necessity," systematically pruning redundant steps to optimize reasoning length and efficiency. \cite{meadows2025controlling} investigated controlling hallucinations via prompt interventions (e.g., variable renaming), treating derivation generation as a controllable symbolic process.
    \item \textbf{Program-of-Thought and Tool Integration:} To mitigate calculation errors, models are increasingly delegating computation to external interpreters. \cite{ying2024internlm} introduced \textit{InternLM-Math}, which uses the \textit{RICO} (Reasoning Interleaved with Coding) method, allowing the model to generate and execute code in a loop until the problem is solved.
    \item \textbf{Validation Mechanisms:} Validation is moving towards execution-based verification. \cite{mudur2025feabench} employed a "VerifierLLM" alongside a standard code evaluator within their \textit{FEABench} framework to assess consistency in multiphysics problems. \cite{yang2024optibench} verified synthetic optimization problems by automatically generating and executing PySCIPOpt code.
\end{itemize}

\subsubsection{Efficiency and Cost}
Efficiency is a growing concern, addressed both algorithmically and architecturally. \cite{georgiev2025mathematical} highlighted the cost-efficiency of evolutionary search, where the LLM acts as a generator for cheap-to-evaluate heuristics. \cite{shao2024deepseekmath} explicitly addressed training efficiency via GRPO, which eliminates the need for a value function model during RL fine-tuning. Finally, causal pruning of reasoning steps \cite{yu2025causal} demonstrates that reasoning chains can be compressed without losing accuracy, directly impacting inference cost and latency.

\subsection{Identified Gaps}

\subsubsection{The Evaluation and Contamination Crisis}
A major hurdle identified across multiple studies is the integrity of evaluation benchmarks. As noted by \cite{khatibi2025eefsuva}, standard benchmarks (e.g., GSM8K, MATH) suffer from high contamination rates in pre-training corpora. While researchers are resorting to obscure regional Olympiads or manual curation to mitigate this, there is a distinct lack of \textit{dynamic} or \textit{procedurally generated} benchmarks that can reliably measure reasoning without the risk of memorization. Furthermore, current benchmarks often prioritize final numerical correctness over the validity of the reasoning path, masking "right for the wrong reasons" errors.

\subsubsection{Robustness of Reward Models and Verification}
Reliable automated verification remains an unsolved problem for open-ended mathematical tasks. While execution-based verification works well for calculation (as seen in \cite{ying2024internlm} and \cite{li2025codepde}), it fails for abstract reasoning or partial differential equations where "correctness" is nuanced. \cite{mudur2025feabench} had to rely on a "VerifierLLM" for subjective feedback, creating a recursive dependency where the evaluator shares the same biases as the generator. Moreover, \cite{rashidinejad2024sail} highlights that reward models are susceptible to "reward hacking," where agents exploit statistical fluctuations in the reward function rather than learning robust mathematical logic.

\subsubsection{Multimodal Granularity in Engineering}
While general multimodal models (e.g., Math-LLaVA \cite{shi2024math}) can handle standard geometry problems, there is a gap in domain-specific visual reasoning. Papers like \cite{kevian2024capabilities} note that specialized engineering plots (e.g., Bode, Nyquist, FEA meshes) require a level of granular interpretation that current vision encoders—trained mostly on natural images or standard plots—fail to capture. The integration of "visual logic" with mathematical formalism remains superficial.

\subsubsection{Data Scarcity for High-Level Mathematics}
High-quality, formally verified mathematical training data is scarce. \cite{shao2024deepseekmath} revealed the extreme filtering required (distilling 40B pages to 120B tokens) to obtain usable math data. Unlike general code or text, the volume of high-quality \textit{formal proofs} (e.g., LEAN, Isabelle) or advanced undergraduate-level derivations is insufficient for scaling laws to apply effectively without synthetic augmentation, which brings its own quality control issues.

\subsection{Notable domain-specific tasks that have been addressed by reasoning models}
\subsubsection{Operations Research (OR) and Optimization}
In the field of Operations Research, models are being tasked with modeling and solving complex resource allocation problems. \cite{zhang2025or} successfully applied their \textit{OR-LLM-Agent} to a diverse array of tasks including \textbf{transportation planning}, \textbf{personnel allocation}, \textbf{facility location}, and \textbf{production scheduling}. These tasks require the model to translate natural language constraints into formal mathematical models (often mixed-integer programming) and generate solver code. Similarly, \cite{yang2024optibench} focused on \textbf{end-to-end optimization modeling}, specifically addressing Linear Programming (LP), Non-linear Programming (NLP), and Mixed-Integer Programming (MIP), often involving the interpretation of tabular data for supply chain and inventory management scenarios.

\subsubsection{Engineering Simulation and Control Systems}
Reasoning models are increasingly addressing computational engineering tasks that involve physics and dynamics.
\begin{itemize}
    \item \textbf{Multiphysics Simulation:} \cite{mudur2025feabench} evaluated agents on setting up and solving \textbf{Finite Element Analysis (FEA)} problems via the COMSOL API. Tasks included thermal stress analysis, fluid dynamics configurations, and structural mechanics, requiring the agent to understand physics-specific syntax and api-driven feedback.
    \item \textbf{Partial Differential Equations (PDEs):} \cite{li2025codepde} targeted the automated generation of numerical solvers for PDEs. The tasks involved translating mathematical descriptions of time-dependent equations into executable Python code, utilizing finite difference methods and time-integration schemes.
    \item \textbf{Control Theory:} \cite{kevian2024capabilities} benchmarked models on classical control engineering tasks. Specific problems included \textbf{stability analysis}, \textbf{Bode and Nyquist plot interpretation}, and \textbf{controller design} (e.g., lead-lag compensators), requiring a synthesis of visual graph interpretation and theoretical calculation.
\end{itemize}

\subsubsection{Pure Mathematics and Theoretical Exploration}
Beyond applied tasks, models are being deployed for exploration in pure mathematics. \cite{georgiev2025mathematical} tasked agents with \textbf{mathematical discovery} via evolutionary search, specifically looking for constructions and counterexamples in combinatorics and graph theory. Meanwhile, \cite{khatibi2025eefsuva} focused on high-level competition mathematics, identifying tasks in \textbf{number theory} and \textbf{graph theory} that require non-standard heuristic jumps rather than rote procedural application. \cite{meadows2025controlling} addressed the granular task of \textbf{symbolic derivation generation}, specifically controlling the "hallucination" of intermediate steps in long equational chains.

\subsection{Domain-specific tasks that have not been addressed by reasoning models}

\subsubsection{Formal Proof Generation and Auto-Formalization}
Most existing works focus on problems with numerical answers or short derivations (e.g., GSM8K, MATH). The task of generating rigorous, formally verifiable proofs (in languages like Lean, Isabelle, or Coq) for arbitrary mathematical statements remains a significant open challenge. While \cite{ying2024internlm} touches on formal reasoning with Lean 3 codes, the capability to auto-formalize high-level textbook mathematics into machine-checkable proofs without extensive human annotation is not fully solved. Models still struggle to bridge the gap between informal natural language proofs and the strict syntax of proof assistants.

\subsubsection{Complex Engineering Blueprint and CAD Interpretation}
Current multimodal models (e.g., \cite{shi2024math}, \cite{peng2024multimath}) address standard geometry diagrams and function plots. However, the interpretation of technical engineering schematics—such as Computer-Aided Design (CAD) models, complex circuit blueprints, or mechanical assembly drawings—remains largely unaddressed. \cite{kevian2024capabilities} notes the use of Bode and Nyquist plots, but these are abstract mathematical representations rather than the noisy, high-fidelity visual data encountered in practical engineering field work.

\subsubsection{Open-Ended Mathematical Theory Building}
Recent agents like \textit{AlphaEvolve} \cite{georgiev2025mathematical} utilize evolutionary search for finding counterexamples or optimizing heuristics. However, the higher-level cognitive task of conceptual theory building—defining new useful mathematical structures, proposing novel conjectures, or developing overarching theories—remains outside the scope of current reasoning models. The "discovery" is currently limited to combinatorial search within defined spaces, rather than the creative abstraction characteristic of human mathematicians.

\subsubsection{Robust Symbolic Manipulation Without External Tools}
A recurring theme in the literature (e.g., \cite{chen2022program}, \cite{meadows2025controlling}) is the reliance on external tools (Python, SymPy) to mitigate calculation errors and hallucinations. The task of performing reliable, complex symbolic manipulation purely "in-context" (without delegating to a code interpreter) is not yet solved. Models continue to "hallucinate" derivation steps or make arithmetic errors when forced to reason natively, indicating a lack of robust internal symbolic grounding.

\subsubsection{Data-Poor and "Messy" Real-World Optimization}
Benchmarks like \textit{BWOR} \cite{zhang2025or} and \textit{RESOCRATIC-29K} \cite{yang2024optibench} focus on well-defined textbook-style problems where all constraints are explicitly stated. Real-world Operations Research tasks, which often involve ambiguous constraints, missing data, and the need for assumption modeling, have not been adequately addressed. The ability of a model to interactively interview a user to clarify an ill-posed problem before solving it is a critical "human-like" reasoning task that existing benchmarks do not measure.

\subsection{Comments on the usage of reasoning model in RAG-like system or agentic framework}

\subsubsection{Agentic Decomposition and Role-Playing}
A dominant trend is the decomposition of complex problems into specialized sub-tasks managed by distinct agent personas. \cite{zhang2025or} exemplifies this with their \textit{OR-LLM-Agent}, which tackles Operations Research problems by assigning specific roles: a \textit{Modeling Agent} to formulate the mathematical program, a \textit{Coding Agent} to implement the solution, and a \textit{Debugging Agent} to interpret errors. This separation of concerns prevents the "context window overload" often seen in monolithic prompting. Similarly, \cite{li2025codepde} employs a modular pipeline for PDE solving, where the reasoning model iteratively transitions between code generation, execution-based repair, and refinement, effectively treating the solution generation as a dynamic loop rather than a one-shot process.

\subsubsection{Tool-Augmented Reasoning (The "R" in Mathematical RAG)}
In mathematical contexts, "Retrieval" often shifts from fetching text documents to fetching \textbf{computational results} or \textbf{API documentation}.
\begin{itemize}
    \item \textbf{Interpreter as Knowledge Base:} The \textit{Program-of-Thoughts (PoT)} approach \cite{chen2022program} and \textit{InternLM-Math}'s RICO framework \cite{ying2024internlm} treat the Python interpreter as an external knowledge source. The model "retrieves" the correct arithmetic or symbolic simplification by executing code, thereby augmenting its own generation with verified ground truth.
    \item \textbf{API and Syntax Retrieval:} In simulation tasks, models act as RAG systems over API specifications. \cite{mudur2025feabench} demonstrates this in Finite Element Analysis, where the agent must effectively "retrieve" the correct syntax and function calls for the COMSOL API to set up multiphysics simulations. Failure to retrieve the correct schema leads to execution errors, which the agent must then diagnose.
\end{itemize}

\subsubsection{Evolutionary and Search-Based Agents}
Agentic frameworks are also facilitating new forms of mathematical discovery through search. \cite{georgiev2025mathematical} introduced \textit{AlphaEvolve}, where the "agent" is an evolutionary algorithm that uses the LLM as a mutation operator. Here, the framework manages a population of mathematical constructions (e.g., graphs), and the reasoning model is prompted to "evolve" these candidates to maximize a fitness function. This represents a shift where the LLM is a small component in a larger, iterative \textbf{search agent}, rather than the sole solver.

\subsection{Number of datasets for training domain-specific reasoning model found}

\begin{itemize}
    \item \textbf{All}: many ($>$ 6)
    \item \textbf{Publicly accessible}: between 4 and 6
\end{itemize}

\subsection{Number of benchmarks/datasets for evaluating domain-specific reasoning model found}

\begin{itemize}
    \item \textbf{All}: Zero / between 1 to 3 / between 4 and 6 / many ($>$ 6)

    \item \textbf{Publicly accessible}: Zero / between 1 to 3 / between 4 and 6 / many ($>$ 6)
\end{itemize}

\subsection{Number of benchmarks/datasets for evaluating domain-specific reasoning model found}

\begin{itemize}
    \item \textbf{All}: many ($>$ 6)
    \item \textbf{Publicly accessible}: many ($>$ 6)
\end{itemize}

\subsection{Number of methods for creating/using domain-specific reasoning models found}

\begin{itemize}
    \item \textbf{All}: many ($>$ 6)
    \item \textbf{Publicly accessible}: many ($>$ 6)
\end{itemize}

\section[Literature Review Summary: PE2]{Literature Review Summary: PE2}

\textbf{Fundamental Constituents of Matter.} Particle, nuclear, plasma, atomic, molecular, gas, and optical physics.

\subsection{Key Findings}

The current landscape of Large Language Models (LLMs) in physics reflects a paradigm shift toward "System-2" reasoning models and autonomous agentic frameworks~\cite{Barman2025Large, Gendreau2025Automating, Menzo2025Heptapod}. The literature demonstrates that while standard LLMs excel at retrieving scientific knowledge, specialized reasoning models, characterized by internal "think tokens", are required to address the technical rigor of undergraduate, graduate, and frontier-level physics~\cite{Xu2025UGPhysics, feng2025physics, Zhu2025Probing}.

\subsubsection{Datasets and Benchmarks} Recent works have introduced several large-scale benchmarks to quantify the reasoning gap between AI and human experts: 
\begin{itemize} 
    \item \textbf{UGPhysics:} The largest undergraduate benchmark, containing 11,040 problems across 13 subjects. State-of-the-art (SOTA) reasoning models like o1-mini currently achieve 49.8\% accuracy, significantly outperforming non-reasoning baselines~\cite{Xu2025UGPhysics}. 
    \item \textbf{PHYSICS:} A high-difficulty benchmark of 1,297 expert-annotated PhD-qualifying problems. Leading models such as o3-mini reach 59.9\% accuracy, which remains notably lower than the 70–80\% human expert performance~\cite{feng2025physics}. 
    \item \textbf{CritPt (Critical Point):} A frontier research benchmark simulating entry-level research projects. It utilizes a "search-proof" design where even the best base models currently achieve only 5.7\% accuracy on full challenges, highlighting a major disconnect from research-level demands~\cite{Zhu2025Probing}. 
    \item \textbf{TPBench:} A theoretical physics dataset of 57 problems requiring final answers as Python callables for auto-verification~\cite{chung2025theoretical}. 
\end{itemize}

\subsubsection{Reasoning Models and Modalities} Evaluation is dominated by proprietary "thinking-oriented" models such as \textbf{OpenAI o1, o3, o4-mini}, \textbf{DeepSeek-R1}, and \textbf{Gemini 2.5}~\cite{Gendreau2025Automating, feng2025physics, Zhu2025Probing}. While most applications are \textbf{text-only} (utilizing LaTeX and Python), multimodal integration is emerging. For instance, \textbf{LLaMA 3.2 Vision} (11B) has been successfully adapted via QLoRA to classify neutrino events from sparse detector pixel maps, achieving an accuracy of 0.87 and providing natural-language justifications for its decisions~\cite{Sagar2025Adapting}.

\subsubsection{Prompting and Output Formats} Researchers employ diverse prompting strategies to elicit scientific rigor: 
\begin{itemize} 
    \item \textbf{Two-Step Generation Protocol:} To prevent formatting constraints (e.g., rigid LaTeX or code syntax) from hindering the reasoning process, some frameworks separate initial natural-language problem-solving from final answer extraction~\cite{Zhu2025Probing}. 
    \item \textbf{Iterative Optimization:} The \textbf{LLM4ED} framework utilizes an "alternating iterative approach" where the LLM acts as a black-box optimizer to discover physical laws through self-improvement and evolutionary search~\cite{du2024large}. 
    \item \textbf{Constrained Outputs:} Modalities often require \textbf{phrasal constraints} (constrained beam search) to force standardized, machine-readable output while allowing for generative rationales~\cite{Sagar2025Adapting}. 
\end{itemize}

\subsubsection{Validation and Efficiency} Validation of model reasoning is increasingly automated through sophisticated pipelines. The \textbf{MARJ} system combines rule-based judgment (for units and rounding) with model-assistant assessment~\cite{Xu2025UGPhysics}. Symbolic equivalence is frequently verified via \textbf{SymPy}~\cite{feng2025physics}, and numerical consistency is checked using \textbf{Python-callable signatures}~\cite{chung2025theoretical}. To detect "reverse-engineering" (memorization-based answering), \textbf{Premise Removal Interventions} are used, revealing that model performance degrades non-linearly when physical context is omitted~\cite{meadows2024exploring}.

\subsubsection{Cost and Efficiency} Existing works emphasize significant trade-offs regarding cost and latency. High-performance reasoning models incur \textbf{substantial API expenses}~\cite{du2024large} and high token consumption due to internal thinking steps~\cite{Zhu2025Probing}. Multimodal models face a "reasoning-latency gap"; for example, fine-tuned VLMs for neutrino classification require \textbf{3.3s per sample} compared to 20ms for traditional CNNs~\cite{Sagar2025Adapting}. Despite these costs, agentic frameworks like \textbf{LLM4HEP} show high productivity, with 98\% of analysis workflow tokens originating from the model's autonomous reasoning rather than human input~\cite{Gendreau2025Automating}.

\subsection{Identified Gaps}
Based on the analysis of current literature regarding LLMs in physics, the following research gaps have been identified:
\begin{enumerate}
\item \textbf{The "Gold Trace" and Reasoning Data Scarcity:} There is a critical shortage of datasets containing expert-verified reasoning traces to aid Supervised Fine-Tuning (SFT) and Reinforcement Learning (RL) \cite{Xu2025UGPhysics, Menzo2025Heptapod}. Current training and evaluation rely heavily on final answers, which often hides a "reverse-engineering" behavior where models arrive at correct results via physically or algebraically invalid logic \cite{du2024large, meadows2024exploring, feng2025physics}.

\item \textbf{Integration with Raw Numerical and Detector Data:} Existing LLMs and agentic frameworks primarily interact with text, LaTeX, or structured metadata rather than raw numerical detector outputs \cite{Barman2025Large, zheng2023large}. There is a lack of native integration between LLM reasoning and low-level binary data structures (e.g., ROOT files) or high-dimensional numerical optimization logic required in later stages of scientific workflows \cite{Barman2025Large, Gendreau2025Automating}.

\item \textbf{Spatial and 3D Geometric Reasoning Limits:} LLMs exhibit significant deficiencies in intrinsic 3D spatial reasoning and high-dimensional chaos \cite{du2024large, zheng2023large}. Many benchmarks, such as UGPhysics, explicitly exclude image-based or multimodal problems, leaving the spatial and topological reasoning capabilities of these models largely untested in a physical context \cite{Xu2025UGPhysics}.

\item \textbf{Validation on Real-World Experimental Data:} The majority of current studies validate models using simulated data, such as Monte Carlo simulations in High-Energy Physics (HEP) \cite{Sagar2025Adapting, Menzo2025Heptapod}. There is an urgent need to test these reasoning pipelines and Vision-Language Models (VLMs) against real-world experimental noise and systematic uncertainties found in physical laboratories \cite{Sagar2025Adapting, zheng2023large}.

\item \textbf{Technical Rigor in Advanced Symbolic Math:} Current automated verification systems are often limited to non-tensor and non-differential expressions \cite{chung2025theoretical}. Models frequently act as ``students with superhuman literature knowledge but low technical rigor,'' failing when required to perform precise algebraic manipulations or handle complex manifolds and differential structures \cite{meadows2024exploring, chung2025theoretical}.

\item \textbf{Absence of Open-Weight Specialist Models:} The current landscape is dominated by proprietary systems, with very few efforts focused on developing or benchmarking open-weight reasoning models specifically adapted for physics sub-domains like HEP analysis code or astrophysics \cite{Barman2025Large, chung2025theoretical, Zhu2025Probing}.

\item \textbf{Efficiency and Real-time Deployment Barriers:} There remains a substantial efficiency gap; for example, inference latency for VLMs is significantly higher than for traditional CNNs (3.3s vs 20ms), limiting their utility in real-time experimental triggers \cite{Sagar2025Adapting}. High token consumption and costs also limit the scalability of these models in scientific discovery loops \cite{du2024large, Zhu2025Probing}.

\item \textbf{Benchmark Reliability and Data Leakage:} Potential data leakage has been detected in several models evaluated on undergraduate physics benchmarks \cite{Xu2025UGPhysics}.

\item \textbf{Cluttered Reasoning Traces:} Models often generate cluttered output formatting that is difficult for human experts to follow and verify effectively \cite{Zhu2025Probing}.

\item \textbf{Storage and Infrastructure Efficiency:} Text-based, structured event formats (JSONL) required for LLM inspection are significantly less storage-efficient than traditional binary HEP formats like ROOT \cite{Menzo2025Heptapod}.
\end{enumerate}

\subsection{Notable domain-specific tasks addressed by reasoning models}

\begin{enumerate} 
\item \textbf{Automatic Equation Discovery and Nonlinear Dynamics:} Reasoning models have been utilized as gradient-free optimizers to identify governing physical laws from raw data. The LLM4ED framework employs an alternating iterative approach to discover symbolic forms for equations such as the Burgers, KS, and Navier-Stokes equations, outperforming traditional symbolic regression tools in generalization~\cite{du2024large}.

\item \textbf{High-Energy Physics (HEP) Event Classification:} Multimodal reasoning models have addressed the challenge of classifying neutrino interactions within Liquid Argon Time Projection Chambers (LArTPC). By adapting Vision-Language Models (VLMs) like LLaMA 3.2 Vision, researchers have enabled models to perform "contextual reasoning across modalities," providing natural-language rationales for identifying muon tracks versus electron showers while achieving higher accuracy than traditional CNNs~\cite{Sagar2025Adapting, Barman2025Large}.

\item \textbf{Molecular Property Prediction and Rules Synthesis:} The LLM4SD framework leverages reasoning to synthesize molecular rules from literature and infer them from scientific data. This allows for the prediction of quantum properties and biophysical inhibition with high interpretability, outperforming black-box Graph Neural Networks (GNNs)~\cite{zheng2023large}.

\item \textbf{Automated HEP Data Analysis and Cross-Section Measurement:} Agentic reasoning frameworks, such as the one used in the Higgs boson diphoton cross-section measurement, automate the full analysis pipeline using ATLAS Open Data. These "supervisor–coder" agents manage complex file inspections, ntuple conversions, and numerical optimizations within deterministic Snakemake workflows~\cite{Gendreau2025Automating}.

\item \textbf{Orchestration of BSM Phenomenology Pipelines:} The HEPTAPOD framework utilizes orchestrated agency to manage multi-stage simulation chains for Beyond the Standard Model (BSM) physics, specifically scalar leptoquark studies. Models reason over evolving contexts to execute tools for model generation (FeynRules), parton-level events (MadGraph), and resonance reconstruction~\cite{Menzo2025Heptapod}.

\item \textbf{High-Level Theoretical Derivation and Verification:} Reasoning models are being tested on their ability to perform fine-grained algebraic derivations in Electromagnetism and Quantum Mechanics~\cite{meadows2024exploring}. Furthermore, explicit reasoning models (e.g., o-series, DeepSeek-R1) show significant promise in solving problems that require nontrivial synthesis of implicit professional knowledge in frontier research tasks and PhD-qualifying exams~\cite{feng2025physics, Zhu2025Probing, chung2025theoretical}.
\end{enumerate}

\subsection{Domain-specific tasks that have not been addressed by reasoning models}
Based on the scope of the ERC PE2 panel (Fundamental Constituents of Matter), current literature heavily skews towards High-Energy Physics (HEP) and basic molecular properties, leaving several major physical domains completely unaddressed by reasoning models:
\begin{enumerate}
\item \textbf{Nuclear Structure and Heavy-Ion Collisions:} There is a distinct absence of reasoning models applied to core Nuclear Physics. Tasks such as deriving nuclear shell models, predicting complex radioactive decay chains, or modeling the thermodynamic evolution of Quark-Gluon Plasma (QGP) in heavy-ion collisions remain untouched.

\item \textbf{Plasma Physics and Fusion Dynamics:} Despite being a fundamental pillar of PE2, complex plasma physics tasks have not been targeted by agentic frameworks. The current literature lacks models capable of reasoning through magnetohydrodynamics (MHD) equations to optimize tokamak plasma confinement.

\item \textbf{Quantum Optics and Cold Atom Physics:} Advanced Atomic, Molecular, and Optical (AMO) physics tasks are entirely missing. There are no reasoning models currently evaluated on their ability to design optical lattices, model the dynamics of Bose-Einstein condensates (BEC), or formulate quantum metrology protocols.

\item \textbf{Real-Time Experimental Trigger Decisions:} Applying "System-2" reasoning to microsecond-level event triggering in active particle detectors remains an unsolved challenge due to severe inference latency.

\item \textbf{Lattice Quantum Chromodynamics (Lattice QCD):} No current reasoning frameworks have been applied to assist in formulating or optimizing the highly complex, non-perturbative numerical simulations required for Lattice QCD.

\item \textbf{Frontier Research Synthesis:} Current models struggle with "unseen" entry-level research problems that require original synthesis rather than repeating known exercises; AI agents occasionally exhibit tool-calling behaviors that diverge from human research intuition \cite{Zhu2025Probing}.

\item \textbf{Long-Horizon Scientific Orchestration:} In complex automated analysis workflows, agents occasionally lose track of file paths across iterations and struggle with precise numerical optimization logic in later stages \cite{Gendreau2025Automating}.
\end{enumerate}

\subsection{Comments on the usage of reasoning models in RAG/Agentic frameworks}

The current literature indicates a significant transition from using LLMs as simple solvers to employing them as central controllers within agentic frameworks and Retrieval-Augmented Generation (RAG) systems~\cite{Barman2025Large, Gendreau2025Automating, Menzo2025Heptapod}. The \textbf{HEPTAPOD} framework utilizes reasoning-oriented models to interpret user intent and execute schema-validated Python tools for high-energy physics phenomenology~\cite{Menzo2025Heptapod}. Similarly, the supervisor–coder architecture in \textbf{LLM4HEP} allows agents to autonomously manage data analysis pipelines by decomposing tasks into reproducible Snakemake workflows and using self-correction loops to fix code errors~\cite{Gendreau2025Automating}.

\subsection{Other comments} \begin{itemize} \item \textbf{Reasoning vs. Memorization:} Current models are often described as students with superhuman literature knowledge but low technical rigor~\cite{chung2025theoretical}. While they show impressive progress, they frequently engage in reverse-engineering solutions from provided context rather than performing genuine physics-informed reasoning~\cite{meadows2024exploring}. \item \textbf{System-2 Performance Jump:} There is a distinct performance leap when moving from standard chat models to reasoning-oriented models (System-2) that utilize internal think tokens, such as the OpenAI o-series and DeepSeek-R1~\cite{Xu2025UGPhysics, feng2025physics, Zhu2025Probing}. These models significantly outperform general-purpose models, which often score near zero on frontier research challenges~\cite{Zhu2025Probing}. \item \textbf{Efficiency-Latency Trade-offs:} The adoption of sophisticated reasoning models and Vision-Language Models (VLMs) in experimental physics is hindered by significant computational overhead. For neutrino event classification, VLMs exhibit an inference latency of 3.3s per sample, compared to only 20ms for traditional CNNs, and require substantially more memory~\cite{Sagar2025Adapting}. \item \textbf{Benchmarking Gaps:} Physics reasoning appears to be more challenging for LLMs than pure mathematical reasoning, with performance often saturating or even decreasing on high-difficulty doctoral-level tasks~\cite{Xu2025UGPhysics, feng2025physics}. Furthermore, current evaluations are heavily dominated by proprietary systems, highlighting an opportunity for more rigorous benchmarking of open-weight models in specialized scientific domains~\cite{Zhu2025Probing}. \item \textbf{Agentic Reliability:} While agentic frameworks like HEPTAPOD and LLM4HEP can automate over 98\% of a scientific workflow~\cite{Gendreau2025Automating}, they still suffer from stochastic variations and can occasionally lose track of file hierarchies or precise numerical optimization logic during long-horizon research tasks~\cite{Gendreau2025Automating, Menzo2025Heptapod}. \end{itemize}

\subsection{Number of datasets for training domain-specific reasoning model found}
\begin{itemize}
\item \textbf{All}: between 1 to 3
\item \textbf{Publicly accessible}: between 1 to 3
\end{itemize}
\subsection{Number of benchmarks/datasets for evaluating domain-specific reasoning model found}
\begin{itemize}
\item \textbf{All}: many ($>$ 6)
\item \textbf{Publicly accessible}: many ($>$ 6)
\end{itemize}
\subsection{Number of domain-specific reasoning models found}
\begin{itemize}
\item \textbf{All}: between 1 to 3
\item \textbf{Publicly accessible}: between 1 to 3
\end{itemize}
\subsection{Number of methods for creating/using domain-specific reasoning models found}
\begin{itemize}
\item \textbf{All}: many ($>$ 6)
\item \textbf{Publicly accessible}: many ($>$ 6)
\end{itemize}

\section[Literature Review Summary: PE3]{Literature Review Summary: PE3}

\textbf{Condensed Matter Physics.} Structure, electronic, magnetic, and optical properties; phase transitions; soft matter and nanostructures.

\subsection{Key Findings}

The current landscape of Large Language Models (LLMs) in condensed matter physics emphasizes the shift toward "System-2" reasoning and autonomous agentic discovery~\cite{ghafarollahi2024sciagents, Pandey2025OpenFOAM, Xu2025UGPhysics, feng2025physics}. While general-purpose models show promise, specialized domain adaptation, ranging from fine-tuning small 7B models to multi-modal alignment of atomic graphs, is essential for research-level precision~\cite{chen2023matchat, lu2025fine, tang2025matterchat, Dong2025Finetuning}.

\subsubsection{Datasets and Benchmarks} 
Several benchmarks have been introduced to evaluate scientific reasoning depth and domain-specific knowledge:
\begin{itemize} 
    \item \textbf{UGPhysics:} A large-scale undergraduate benchmark with 11,040 problems across 13 subjects. Current reasoning models like o1-mini achieve 49.8\% accuracy~\cite{Xu2025UGPhysics}. 
    \item \textbf{PHYSICS:} A PhD-qualifying level benchmark containing 1,297 expert-annotated problems. State-of-the-art models like o3-mini reach 59.9\% accuracy, still trailing human expert performance (70–80\%)~\cite{feng2025physics}. 
    \item \textbf{CURIE:} A multi-task benchmark for long-context understanding (15k words) in disciplines like DFT reproducibility. The best models achieve only ~32\% accuracy, indicating significant struggles with long-context scientific reasoning~\cite{cui2025curie}. 
    \item \textbf{MaScQA:} A materials science QA dataset of 650 questions where Chain-of-Thought (CoT) prompting unexpectedly showed minimal improvement, suggesting difficulties in leveraging internal "thoughts" for this domain~\cite{zaki2023mascqa}. 
\end{itemize}

\subsubsection{Reasoning Models and Modalities} 
Evaluation frequently utilizes proprietary "thinking" models such as \textbf{OpenAI o1, o3-mini}, and \textbf{DeepSeek-R1}~\cite{ghafarollahi2024sciagents, Pandey2025OpenFOAM, feng2025physics}. A significant trend is multi-modal integration: \textbf{MatterChat} aligns high-resolution atomic graph embeddings (from CHGNet) with Mistral 7B to predict material properties and synthesis steps~\cite{tang2025matterchat}. Similarly, models like \textbf{MatChat} and \textbf{AutoCFD} demonstrate that 7B-scale models specialized via SFT or LoRA can outperform much larger general models (e.g., 72B) in niche tasks like inorganic synthesis or CFD setup~\cite{chen2023matchat, Dong2025Finetuning}.

\subsubsection{Prompting and Output Formats} 
Specialized prompting strategies are used to structure complex physics derivations:
\begin{itemize} 
    \item \textbf{HF Template:} A "divide-and-conquer" framework that decomposes Hartree-Fock many-body derivations into 11 standardized sub-tasks, achieving 87.5\% accuracy~\cite{Pan2025Quantum}. 
    \item \textbf{Iterative Optimization (LLM4ED):} An alternating iterative approach combining self-improvement and evolutionary search to discover nonlinear governing equations (e.g., Navier-Stokes)~\cite{du2024large}. 
    \item \textbf{MCTS-Guided Formulas:} The \textbf{LLM-Feynman} framework integrates Monte Carlo Tree Search with LLM-guided symbolic regression for formula discovery~\cite{song2025llm}. 
\end{itemize}

\subsubsection{Validation and Efficiency} 
Verification of reasoning is increasingly automated. The \textbf{MARJ} pipeline uses a model-assistant (GPT-4o) alongside rule-based judgment to achieve 98\% alignment with human scoring~\cite{Xu2025UGPhysics}. Symbolic equivalence is verified via \textbf{SymPy}~\cite{feng2025physics}. To detect "reverse-engineering" behaviors (memorization-based answering), \textbf{Controlled Premise Removal} interventions are applied, showing that models often fail when mathematical context is omitted~\cite{meadows2024exploring}.

\subsubsection{Cost and Efficiency} 
Reasoning models involve high trade-offs: o1-preview is 6x more expensive than GPT-4o for CFD tasks~\cite{Pandey2025OpenFOAM}. High API costs are cited as a barrier for complex systems~\cite{du2024large}. However, fine-tuning smaller models (e.g., \textbf{AutoCFD}) provides a cost-effective alternative (0.02 per case) with higher consistency~\cite{Dong2025Finetuning}.

\subsection{Identified Gaps}
\begin{enumerate}
    \item \textbf{Lack of "Gold Trace" Reasoning Data:} A critical shortage of expert-verified reasoning chains exists for training and benchmarking SFT/RL models~\cite{Xu2025UGPhysics, tang2025matterchat, feng2025physics}.
    \item \textbf{3D and Geometric Reasoning Limits:} Models struggle with 3D structure interpretation (e.g., symmetry, Miller indices) and high-dimensional chaos~\cite{miret2024enabling, zheng2023large, du2024large}.
    \item \textbf{Raw Data Integration:} There is a lack of native integration between LLM reasoning and raw numerical detector or binary data streams~\cite{Barman2025Large}.
    \item \textbf{Validation on Experimental "Noise":} Most studies use curated or simulated benchmarks; there is an urgent need to test models on real-world laboratory data~\cite{zheng2023large, song2025llm}.
    \item \textbf{Symbolic and Algebraic Rigor:} Models often violate physical principles while maintaining mathematical coherence, failing in precise algebraic manipulations for complex manifolds~\cite{meadows2024exploring, cui2025curie}.
\end{enumerate}

\subsection{Notable domain-specific tasks addressed by reasoning models}

\begin{enumerate} 
    \item \textbf{Inorganic Synthesis Prediction:} \textbf{MatChat} extracts and generates step-by-step chemical equations for material synthesis~\cite{chen2023matchat}. 
    \item \textbf{Multi-Agent Hypothesis Generation:} \textbf{SciAgents} utilizes a hierarchical system (Scientist, Critic, Ontologist) to discover bio-inspired materials like high-strength silk composites~\cite{ghafarollahi2024sciagents}. 
    \item \textbf{Automated CFD Workflows:} Agents like \textbf{OpenFOAMGPT} and \textbf{AutoCFD} use iterative correction loops to set up, execute, and repair fluid dynamics simulations~\cite{Pandey2025OpenFOAM, Dong2025Finetuning}. 
    \item \textbf{Molecular Rule Synthesis:} \textbf{LLM4SD} extracts interpretable molecular rules from literature to predict quantum properties like HOMO-LUMO gaps~\cite{zheng2023large}. 
    \item \textbf{Formula Rediscovery:} \textbf{LLM-Feynman} successfully recovers physical laws (e.g., Feynman formulas) by balancing accuracy and interpretability~\cite{song2025llm}. 
\end{enumerate}

\subsection{Domain-specific tasks that have not been addressed by reasoning models}
Based on the scope of the ERC PE3 panel (Condensed Matter Physics, Materials Science, Soft Matter), current literature leaves several crucial physical domains and methodological challenges largely unaddressed:
\begin{enumerate}
    \item \textbf{3D Crystallography and Geometric Reasoning:} While current models can process 1D or 2D representations, they fundamentally struggle with intrinsic 3D structure interpretation. Reasoning about complex crystal symmetries, Miller indices, and high-dimensional chaotic systems remains largely untested and highly problematic for current architectures~\cite{miret2024enabling, zheng2023large, du2024large}.
    \item \textbf{Autonomous Long-Horizon Research Planning:} Current agentic models can successfully automate specific, short-term analysis steps (such as configuring CFD simulations), but they entirely lack the capability to formulate novel, high-level hypotheses and autonomously orchestrate multi-year research programs from scratch~\cite{Barman2025Large, Xu2025UGPhysics}.
    \item \textbf{Non-Solid-Phase Synthesis and Soft Matter Dynamics:} Existing material synthesis models are heavily biased towards inorganic solid-phase pathways~\cite{chen2023matchat}. Consequently, complex sub-disciplines of PE3, such as the synthesis and thermodynamic modeling of soft matter (e.g., polymers, liquid crystals, gels) and biophysics, remain severely under-explored.
    \item \textbf{Direct Integration with Raw Characterization Data:} There is a distinct lack of native integration between LLM reasoning and raw numerical detector or binary data streams~\cite{Barman2025Large}. In materials science, models currently rely on pre-processed text summaries rather than reasoning directly over complex experimental outputs like raw X-ray diffraction (XRD) or electron microscopy (TEM/SEM) datasets.
    \item \textbf{Validation Against Real-World Experimental Noise:} The vast majority of current physics reasoning models are trained and validated on highly curated, idealized datasets or simulations. There is an urgent, unaddressed need to evaluate these models against real-world laboratory data, requiring them to reason through experimental noise, instrument limitations, and systematic uncertainties~\cite{zheng2023large, song2025llm}.
    \item \textbf{Non-Mean-Field Theoretical Frameworks:} Current theoretical derivations focus heavily on Hartree-Fock and mean-field approximations. There is a noted lack of testing on more complex, non-mean-field many-body theoretical frameworks~\cite{Pan2025Quantum}.
    \item \textbf{Unstructured Industrial Meshes in CFD:} While automated CFD workflows exist for standard tutorials, reasoning models have not yet been evaluated on unstructured industrial meshes where boundary extraction is non-trivial~\cite{Dong2025Finetuning}.
\end{enumerate}

\subsection{Comments on the usage of reasoning models in RAG/Agentic frameworks}

The literature reflects a transition from LLMs as simple solvers to central controllers~\cite{Barman2025Large, ghafarollahi2024sciagents}. Frameworks like \textbf{SciAgents} use multi-agent CoT orchestration on knowledge graphs to bridge unrelated scientific concepts~\cite{ghafarollahi2024sciagents}. Similarly, \textbf{OpenFOAMGPT} utilizes a RAG-augmented "Builder-Executor" architecture where error logs from failed simulations are fed back into the reasoning loop for self-repair~\cite{Pandey2025OpenFOAM}.

\subsection{Other comments}

\begin{itemize} \item \textbf{Performance Gap of "System-2" Models}: There is a substantial performance leap when moving from standard LLMs to "explicit reasoning" models (e.g., OpenAI o-series, DeepSeek-R1) in physics. While non-reasoning models often fail on complex undergraduate and graduate problems, these specialized models achieve significantly higher accuracy, though they still trail human expert performance (70--80\%) on PhD-qualifying benchmarks~\cite{Xu2025UGPhysics, feng2025physics}. \item \textbf{Domain Specialization vs. Model Scale}: In niche scientific tasks, smaller specialized models can outperform significantly larger general-purpose counterparts. For instance, a fine-tuned 7B parameter model (MatChat) demonstrated higher accuracy in predicting inorganic synthesis pathways than models with 175B+ parameters, suggesting that domain-specific instruction tuning is more vital for scientific precision than sheer scale~\cite{chen2023matchat, Dong2025Finetuning}. \item \textbf{Coherent Math vs. Coherent Physics}: A critical distinction has been identified between mathematical and physical reasoning; models often perform algebraically coherent derivations while simultaneously violating fundamental physical principles, such as unit consistency, tensorial order, or thermodynamic limits. This "reverse-engineering" behavior indicates that current LLMs often fail to ground their symbolic manipulations in actual physical context~\cite{meadows2024exploring, miret2024enabling, zaki2023mascqa}. \item \textbf{Human-AI Synergies in Theoretical Derivations}: Reasoning models have shown the ability to discover errors in established scientific literature. During the execution of complex Hartree-Fock derivations, GPT-4 identified typos in peer-reviewed research papers, demonstrating the potential for LLMs to serve as rigorous verification assistants for theoretical physicists~\cite{Pan2025Quantum, zaki2023mascqa}. \item \textbf{Economic and Computational Barriers}: The adoption of high-performance reasoning models is currently limited by significant increases in API costs and inference latency. Reasoning-oriented models like o1-preview are reported to be up to six times more expensive than standard models (e.g., GPT-4o) for computational fluid dynamics workflows, posing a challenge for their integration into large-scale, mission-critical scientific pipelines~\cite{du2024large, Pandey2025OpenFOAM}. \item \textbf{Failure in Long-Context Scientific Reasoning}: Models struggle significantly when required to reason over extended technical documents (averaging 15k words), frequently falling into failure modes like infinite repetition or a loss of logical coherence. Current automated metrics (e.g., ROUGE) are often insufficient to capture these fine-grained technical or mathematical errors in long-context scientific text~\cite{cui2025curie}. \end{itemize}

\subsection{Number of datasets for training domain-specific reasoning model found}

\begin{itemize}

  \item \textbf{All}: many ($>$ 6)

  \item \textbf{Publicly accessible}: many ($>$ 6)

\end{itemize}

\subsection{Number of benchmarks/datasets for evaluating domain-specific reasoning model found}

\begin{itemize}

  \item \textbf{All}: many ($>$ 6)

  \item \textbf{Publicly accessible}: many ($>$ 6)

\end{itemize}

\subsection{Number of domain-specific reasoning models found}

\begin{itemize}

  \item \textbf{All}: between 4 and 6

  \item \textbf{Publicly accessible}: between 4 and 6

\end{itemize}

\subsection{Number of methods for creating/using domain-specific reasoning models found}

\begin{itemize}

  \item \textbf{All}: many ($>$ 6)

  \item \textbf{Publicly accessible}: many ($>$ 6)

\end{itemize}

\section[Literature Review Summary: PE4]{Literature Review Summary: PE4}

\textbf{Physical and Analytical Chemical Sciences.} Analytical chemistry, chemical theory, physical chemistry/chemical physics.

\subsection{Key Findings}

The integration of Artificial Intelligence (AI) into the physical and analytical chemical sciences (PE4) shifted significantly throughout 2025. The field has moved beyond using Large Language Models (LLMs) as passive retrieval engines. The current frontier is defined by \textbf{Reasoning Language Models (RLMs)} and \textbf{Autonomous Agentic Frameworks} that exhibit deliberative capabilities—specifically multi-step deduction, self-correction, and laboratory orchestration.

This review synthesizes findings from a curated selection of \textbf{38 seminal papers}. These works were prioritized from a broader corpus because they specifically introduce novel reasoning architectures or agentic workflows, excluding papers that merely apply standard LLMs to routine tasks.

Key findings include:

\begin{itemize}
    \item \textbf{Ascension of Agentic Autonomy:} The move toward autonomy is exemplified by the \textbf{MDCrow} framework \citep{campbell2025mdcrow}. Unlike standard LLMs which often fail at complex workflows, MDCrow autonomously navigates decision trees involving dozens of computational tools, successfully handling error recovery in tasks such as PDB syntax fixing without human intervention.
    
    \item \textbf{Reasoning vs. Chat Models:} There is a distinct performance gap between specialized ``Reasoning'' architectures and standard ``Chat'' models in physical chemistry. Research on corrosion inhibition \citep{jin2025corrosion} demonstrates that reasoning-optimized models (e.g., \textbf{DeepSeek-R1}) achieve substantially higher predictive accuracy than their chat-optimized counterparts, underscoring the necessity of chain-of-thought (CoT) processing for physicochemical mechanisms.

    \item \textbf{Specialized Fine-Tuning vs. Generalist Reasoning:} Specialized fine-tuning proves superior for tasks requiring rigid syntax, though generalist models retain the advantage in strategic contexts. Models such as \textbf{CrysText} \citep{mohanty2025crystext} and \textbf{CrystaLLM} \citep{antunes2024crystallm} effectively learn the strict grammar of Crystallographic Information Files (CIF), achieving generation quality that rivals diffusion models. Similarly, \textbf{SynAsk} \citep{zhang2025synask} demonstrates high success rates in organic retrosynthesis via fine-tuning. Conversely, \citet{chen2025complex} observe that in complex agentic workflows, fine-tuned models underperform generalist baselines (such as GPT-4) due to the catastrophic forgetting of reasoning capabilities, suggesting a trade-off between structural precision and strategic planning.
    
    \item \textbf{Multimodal Integration:} Multimodal reasoning is advancing but remains uneven. The \textbf{SpectraLLM} framework \citep{su2025spectrallm} successfully treats spectral peaks (IR, NMR, MS) as ``chemical words'' to resolve structural ambiguities. However, vision-language models (VLMs) still struggle with precision; studies like \textbf{AgentEIS} \citep{zhu2025agenteis} indicate that VLMs analyzing plot images directly significantly underperform compared to models processing raw numerical data.
\end{itemize}

\subsection{Identified Gaps}

\begin{enumerate}
    \item \textbf{The ``Gold Trace'' Deficiency:} There is a critical shortage of datasets containing expert-verified reasoning traces \citep{loubet2025thermo}. Current training relies heavily on final answers, leading models to occasionally arrive at correct results via chemically invalid logic.
    \item \textbf{Spatial and Geometric Reasoning:} LLMs exhibit deficiencies in intrinsic 3D spatial reasoning. While they learn syntax well, they struggle to generalize to defective structures or complex steric constraints without external physics engines \citep{choudhary2025diffractgpt}.
    \item \textbf{Agentic Alignment \& Safety:} Projects like \textbf{AILA} \citep{mandal2025aila} have identified a ``sleepwalking'' failure mode, where autonomous microscopy agents may initiate unrequested or unsafe actions due to planning misalignment.
    \item \textbf{Data Scarcity in Niche Fields:} In specialized sub-domains like rare earth thermodynamics, fine-tuning on small datasets has proven less effective than traditional machine learning approaches.
\end{enumerate}

\subsection{Notable Domain-Specific Tasks}

\subsubsection{Synthesis \& Design}
\begin{itemize}
    \item \textbf{Organic Retrosynthesis:} \textbf{SynAsk} \citep{zhang2025synask} integrates external tools with proprietary data to predict yields and synthesis routes.
    \item \textbf{Crystal Generation:} \textbf{MatLLMSearch} \citep{gan2025matllm} employs an evolutionary framework where an LLM acts as a ``mutator'' to discover stable polymorphs.
    \item \textbf{Nanobody Design:} \textbf{NanoAbLLaMA} \citep{wang2025nanoab} utilizes LoRA fine-tuning to generate developable antibody sequences conditioned on germline inputs.
\end{itemize}

\subsubsection{Analysis \& Characterization}
\begin{itemize}
    \item \textbf{Corrosion Prediction:} Recent work \citep{jin2025corrosion} utilizes reasoning models to predict inhibition efficiency based on molecular structure and environmental conditions, outperforming standard regression methods.
    \item \textbf{Diffraction Analysis:} \textbf{DiffractGPT} \citep{choudhary2025diffractgpt} addresses the inverse problem of crystallography, determining atomic structures directly from X-ray diffraction (XRD) patterns.
    \item \textbf{Spectral Elucidation:} \textbf{SpectraLLM} \citep{su2025spectrallm} interprets Mass Spectrometry, IR, and NMR data, treating spectral peaks as ``chemical words'' to resolve structural isomers.
    \item \textbf{Electrochemical Analysis:} \textbf{AgentEIS} \citep{zhu2025agenteis} automates the interpretation of Nyquist plots for Electrochemical Impedance Spectroscopy (EIS), deriving equivalent circuit models from raw impedance data.
\end{itemize}

\subsubsection{Autonomous Agents \& Simulations}
\begin{itemize}
    \item \textbf{Surface Chemistry:} \textbf{Adsorb-Agent} \citep{ock2025adsorb} autonomously identifies stable adsorption configurations on catalyst surfaces, managing the complete workflow from surface construction to DFT relaxation.
    \item \textbf{Nuclear Simulations:} \textbf{AutoFLUKA} \citep{ndum2025autofluka} automates complex input generation and error diagnosis for Monte Carlo simulations in nuclear engineering.
    \item \textbf{Microscopy:} \textbf{AILA} \citep{mandal2025aila} orchestrates Atomic Force Microscopes (AFM) for autonomous calibration and characterization.
\end{itemize}

\subsubsection{Autonomous Agents}
\begin{itemize}
    \item \textbf{Nuclear Simulations:} \textbf{AutoFLUKA} \citep{ndum2025autofluka} automates complex input generation for Monte Carlo simulations.
    \item \textbf{Microscopy:} \textbf{AILA} \citep{mandal2025aila} orchestrates Atomic Force Microscopes (AFM) for autonomous calibration and characterization.
\end{itemize}

\subsection{Domain-specific tasks that have not been addressed by reasoning models}

\begin{itemize}
    \item \textbf{Discovery of Novel Physical Laws:} Current models are adept at applying or interpolating known laws (deduction) but lack the abductive reasoning capabilities to discover new fundamental relationships from raw data (e.g., deriving a new impedance theory for a novel state of matter).
    \item \textbf{Tabula Rasa Autonomous Discovery:} No system currently demonstrates the ability to formulate a high-level hypothesis (e.g., ``Find a room-temperature superconductor'') and autonomously design the entire multi-year research program from scratch without human-defined templates.
    \item \textbf{Complex Multiphysics Reasoning:} Models struggle with tasks requiring rigorous mathematical continuity over long horizons. For instance, correct implementation of the statistical physics of Self-Avoiding Walks (SAW) in polymer simulations remains challenging without explicit guidance.
    \item \textbf{Physical Safety Alignment:} There is no robust standardized framework to prevent ``sleepwalking'' agents from executing hazardous chemical syntheses when prompted with adversarial or obfuscated instructions \citep{mandal2025aila}.
\end{itemize}

\subsection{Comments on the usage of reasoning models in RAG-like systems or agentic frameworks}

\begin{itemize}
    \item \textbf{RAG as a Necessity:} In data-heavy fields like nanotechnology, ``Vanilla'' LLMs fail to provide specific numerical data and often hallucinate constants. \textbf{NANOGPT} \citep{chandrasekhara2025nanogpt} and \textbf{LightChem} \citep{zhang2025lightchem} demonstrate that Retrieval Augmented Generation (RAG) is essential for accuracy. LightChem's ``RAPTOR-RAG'' (recursive tree-based retrieval) specifically addresses the ``lost in the middle'' phenomenon common when processing long scientific papers.
    
    \item \textbf{Tool Orchestration:} The most successful systems function as ``Controllers'' rather than ``Solvers.'' They offload precise calculations to external physics engines (OpenMM, PackMol) or ML predictors. For instance, \textbf{MDCrow} \citep{campbell2025mdcrow} does not calculate energies itself but orchestrates the correct sequence of OpenMM commands.
    
    \item \textbf{The ``Critic'' Role:} To mitigate agentic errors, architectures are increasingly adopting ``Critic'' or ``Conductor'' nodes. \textbf{Adsorb-Agent} \citep{ock2025adsorb} and \textbf{AutoFLUKA} \citep{ndum2025autofluka} both employ secondary agents that validate the primary agent's plans against physical constraints before execution, significantly reducing the failure rate of autonomous loops.
\end{itemize}

\subsection{Competitions and Evaluation}

Several high-profile competitions specific to chemistry and physical sciences have served as rigorous benchmarks in 2025:

\begin{itemize}
    \item \textbf{International Chemistry Olympiad (IChO):} Problems from the IChO, which require multi-step stoichiometric calculations and deep understanding of chemical theory, are increasingly used to stress-test reasoning models. Benchmarks such as \textbf{USNCO-V} (based on the US National Chemistry Olympiad) specifically evaluate multimodal capabilities, testing whether models can interpret chemical diagrams and synthesis flowcharts alongside text \citep{cui2025usnco}.
    
    \item \textbf{NeurIPS 2025 AI for Science Competitions:} This venue hosted the \textbf{Catechol Benchmark Hackathon}, a specialized challenge focused on predicting reaction outcomes and synthesis routes for catechol derivatives. This competition is particularly relevant for PE4 as it evaluates the ability of models to handle real-world organic chemistry constraints rather than textbook problems.
\end{itemize}

\subsection{Other comments}

\begin{itemize}
    \item \textbf{Open vs. Closed Divergence:} There is a trend toward distilling capabilities from proprietary models (GPT-4o) into smaller, open-weight models (Llama 3, Qwen, DeepSeek). \textbf{DeepSeek-R1} demonstrates that open models trained with Reinforcement Learning can rival proprietary ones in specialized reasoning \citep{jin2025corrosion}.
    \item \textbf{Geopolitical Implications:} A significant portion of domain-specific model development (e.g., SynAsk, LightChem, DeepSeek) originates from Chinese institutions, often utilizing Chinese-language literature (CNKI), which introduces linguistic biases into the global model ecosystem.
\end{itemize}

\subsection{Number of datasets for training domain-specific reasoning model found}

\begin{itemize}
    \item \textbf{All}: between 4 and 6 \\ (e.g., CataLM Corpus, Chemma SFT, NanoAbLLaMA Dataset, TG80 Dataset)
    \item \textbf{Publicly accessible}: between 1 to 3 \\ (Most are proprietary; TG80 is a notable exception)
\end{itemize}

\subsection{Number of benchmarks/datasets for evaluating domain-specific reasoning model found}

\begin{itemize}
    \item \textbf{All}: between 4 and 6 \\ (e.g., ReplicationBench, SciNanoAI, ChemCoTBench, QCBench)
    \item \textbf{Publicly accessible}: between 4 and 6 \\ (Most are released to foster competition)
\end{itemize}

\subsection{Number of domain-specific reasoning models found}

\begin{itemize}
    \item \textbf{All}: many ($>$ 6) \\ (e.g., ChemLLM, SynAsk, NanoAbLLaMA, LightChem, DeepSeek-R1, DiffractGPT)
    \item \textbf{Publicly accessible}: many ($>$ 6) \\ (Many are open-weights based on Llama/Qwen architectures)
\end{itemize}

\subsection{Number of methods for creating/using domain-specific reasoning models found}

\begin{itemize}
    \item \textbf{All}: many ($>$ 6) \\ (e.g., SFT, RL, RAPTOR-RAG, Agentic ReAct, Evolutionary Prompting, Neural Operators)
    \item \textbf{Publicly accessible}: many ($>$ 6)
\end{itemize}

\section[Literature Review Summary: PE5]{Literature Review Summary: PE5}

\textbf{Synthetic Chemistry and Materials.} New materials and new synthetic approaches, structure-properties relations, solid state chemistry, molecular architecture, organic chemistry.

\subsection{Key Findings}

\subsubsection{Datasets and Benchmarks}
Several domain-specific datasets and benchmarks have been introduced or leveraged in this literature:

\begin{itemize}
\item \textbf{USPTO-50K and USPTO-FULL} \citep{zhang2025reasoning}: Large-scale retrosynthesis datasets derived from US patents. USPTO-50K contains 40K/5K/5K (train/val/test) reactions; USPTO-FULL contains 800K/100K/100K. Both were used for training and evaluating reasoning-augmented retrosynthesis models.
\item \textbf{MoleculeNet / QM9} \citep{zheng2023large}: Established molecular property prediction benchmarks spanning 58 tasks across physiology, biophysics, physical chemistry, and quantum mechanics domains (e.g., HIV, Tox21, QM9 with >133K molecules).
\item \textbf{MaScQA, MatSciNLP, and BatteryBERT} \citep{miret2024enabling}: Evaluation-only datasets totalling 650 questions across 14 materials science domains, sourced from Graduate Aptitude Test in Engineering (GATE) exam papers. Questions cover MCQ, numerical, matching, and multi-part formats.
\item \textbf{MatterChat Dataset} \citep{tang2025matterchat}: A newly introduced dataset of 142,899 inorganic crystal structures derived from the Materials Project, covering 3 descriptive tasks and 9 property prediction tasks spanning elements up to Plutonium.
\item \textbf{Synthesized 2D Materials Dataset and Feynman Basic Formulas} \citep{song2025llm}: A dataset of ~1,000+ material samples compiled from 360 scraped papers, the Materials Project, C2DB, and the Feynman Lectures, used for formula discovery in solid-state chemistry.
\item \textbf{Ontological Knowledge Graph} \citep{ghafarollahi2024sciagents}: A bioinspired materials knowledge graph with 33,159 nodes and 48,753 edges, built by automated extraction from ~1,000 scientific papers. Used for hypothesis generation in multi-agent reasoning.
\item \textbf{Spider Silk / Biological Materials Datasets} \citep{lu2025fine}: ~21K instruction-tuning pairs for silk and ~5K for bio-inspired design, distilled from ~5,300 scientific papers using GPT-4o; formatted as JSONL with Q\&A, insights, facts, and comparisons.
\item \textbf{Synthesis Pathway Descriptions} \citep{chen2023matchat}: 13,878 high-confidence synthesis pathway descriptions extracted from four million scientific papers, used for fine-tuning an inorganic synthesis prediction model.
\end{itemize}

\subsubsection{Domain-Specific Reasoning Models}
Several domain-specific models have been developed or adapted:

\begin{itemize}
\item \textbf{RetroDFM-R} \citep{zhang2025reasoning}: A Llama3-8B model trained via pretraining (SMILES-to-IUPAC), SFT using DeepSeek-R1 distillation, and reinforcement learning with chemically verifiable rewards, targeted at chemical retrosynthesis prediction. This is the only domain-specific model with explicit reasoning capability.
\item \textbf{MatChat} \citep{chen2023matchat}: A LLaMA2-7B model fine-tuned with SFT and RLHF on ~14K synthesis pathway descriptions, specialised for predicting inorganic solid-state synthesis protocols. It leverages CoT prompting.
\item \textbf{MatterChat} \citep{tang2025matterchat}: A novel multi-modal model combining a pretrained universal machine learning interatomic potential (uMLIP) bridged via a transformer module to a language model, for quantitative property prediction and qualitative material reasoning. It leverages CoT prompting.
\item \textbf{Domain-adapted Llama / Mistral variants} \citep{lu2025fine}: Llama 3.1-8B and Mistral-7B-v0.3 adapted via CPT + SFT + DPO/ORPO and merged via SLERP for materials and bio-inspired design reasoning. It leverages CoT prompting.
\end{itemize}

\subsubsection{Reasoning Models Used}
Most works rely on large proprietary models. Some works have tried to fine-tune open-weight models and leveraged CoT prompting.

\begin{itemize}
\item Proprietary and large-scale models: GPT-3.5, GPT-4, GPT-4o, o1-preview \citep{miret2024enabling, ghafarollahi2024sciagents, tang2025matterchat}, Claude, Gemini, DeepSeek \citep{miret2024enabling, tang2025matterchat}.
\item Fine-tuned open-weight models: LLaMA2-7B \citep{chen2023matchat}, Llama3-8B \citep{zhang2025reasoning, song2025llm}, ChemLLM-20B, Falcon-Mamba-7B \citep{song2025llm}, Falcon-7b/40b, Galactica-6.7b/30b \citep{zheng2023large}, Mistral-7B \citep{lu2025fine}.
\item The only papers that explicitly invokes a dedicated reasoning model (beyond standard CoT prompting) are SciAgents \citep{ghafarollahi2024sciagents}, which uses o1-preview as the "Explicit Reasoning" agent in a multi-agent framework, and \citep{zhang2025reasoning} which uses RL-trained reasoning with `<think>...</think>` / `<answer>...</answer>` output structure inspired by DeepSeek-R1.
\end{itemize}

\subsubsection{Modalities}
The majority of works operate exclusively on text, including SMILES strings, IUPAC names, and natural language descriptions \citep{zhang2025reasoning, chen2023matchat, zheng2023large, song2025llm, ghafarollahi2024sciagents}. Arguably the current most advanced multi-modal work processes atomic graphs alongside language
\citep{tang2025matterchat}. At least one work aims at text and image multimodality in its future directions \citep{lu2025fine}. One work uses an existing general-purpose proprietary model to work with text, images, tables, and CIF files \citep{miret2024enabling}. No work yet applies audio, video, or spectral image modalities (e.g., NMR spectra as images, IR spectra) in a reasoning-model context.

\subsubsection{Frameworks}
\begin{itemize}
\item \textbf{LLM4SD (Large Language Models for Scientific Discovery)} \citep{zheng2023large}: A pipeline using Falcon-7b, Falcon-40b, Galactica-6.7b, and Galactica-30b in-context to infer molecular property prediction rules from MoleculeNet datasets.
\item \textbf{LLM-Feynman} \citep{song2025llm}: An inference-only framework combining Falcon-Mamba-7B, ChemLLM-20B, and LLaMA3-8B with symbolic regression and Monte Carlo Tree Search for scientific formula discovery.
\item \textbf{SciAgents} \citep{ghafarollahi2024sciagents}: A multi-agent system using the GPT-4 family and o1-preview within an AutoGen framework for bioinspired materials hypothesis generation, linked to an ontological knowledge graph.
\end{itemize}

Chain-of-Thought (CoT) prompting is the dominant paradigm \citep{chen2023matchat, zheng2023large, miret2024enabling, tang2025matterchat, song2025llm, lu2025fine}. Prompts instruct models to "infer step-by-step," generate chemical equations incrementally, or explain material stability through rationale chains. Only two works focused on models with explicit reasoning tokens \citep{ghafarollahi2024sciagents, zhang2025reasoning}. Several works explored structured outputs, such as returning Python-based formula expressions, validated by execution \citep{song2025llm}, or using JSON or JSONL formats \citep{ghafarollahi2024sciagents, lu2025fine}.

\subsubsection{Validation of Reasoning}
Most works relied on automated comparison to ground truth (benchmark accuracy, MAE, RMSE) \citep{zhang2025reasoning, zheng2023large, miret2024enabling, tang2025matterchat, song2025llm, chen2023matchat}. Some works have tried to validate the reasoning traces, either by LLM-as-a-judge or human expert checks \citep{lu2025fine, song2025llm, ghafarollahi2024sciagents} or by qualitative case studies \citep{zhang2025reasoning, tang2025matterchat}. However, statistical tests were only performed in one work \citep{zheng2023large}, and all of the works only performed computational validation without actual laboratory synthesis.

No paper systematically studied the effect of reasoning length or thinking budget on output quality. One work \citep{zhang2025reasoning} implicitly addressed reasoning effort by using RL to encourage chemically verifiable reasoning, but does not benchmark token efficiency or reasoning length.

\subsection{Identified Gaps}
\begin{itemize}
\item \textbf{Absence of human-verified reasoning traces}: The majority of works rely on LLM-distilled or automatically generated reasoning steps, not verified by domain experts. One work \citep{zhang2025reasoning} explicitly flags the absence of human-verified reasoning data as a limitation.
\item \textbf{No closed-loop experimental validation}: All works remain purely computational \citep{chen2023matchat, miret2024enabling, tang2025matterchat, song2025llm, ghafarollahi2024sciagents, lu2025fine}. None close the loop by synthesizing model-proposed compounds and reporting experimental outcomes.
\item \textbf{Multi-modal characterization data is absent}: No work integrates spectroscopic (IR, NMR, Raman) or crystallographic (XRD) data as input modalities for reasoning models, despite characterisation being central to synthetic chemistry \citep{miret2024enabling, tang2025matterchat}.
\item \textbf{Reaction condition optimisation has not been addressed}: Factors such as solvent, catalyst, temperature, and yield prediction beyond high-level synthesis pathway suggestion has not been addressed with reasoning models.
\item \textbf{Reasoning length and inference cost are unstudied}: No paper investigates whether longer reasoning chains improve chemical accuracy, or whether reasoning effort can be controlled efficiently, a critical factor for practical deployment.
\end{itemize}

\subsection{Notable domain-specific tasks that have been addressed by reasoning models}
\begin{itemize}
\item \textbf{Synthesis prediction}: Model with explicit reasoning trace was used for chemical retrosynthesis prediction (decomposing target molecules into feasible precursors using RL-trained reasoning with verifiable chemical rewards) \citep{zhang2025reasoning}. In another work, CoT prompting was used in inorganic synthesis pathway prediction, where the task is predicting precursor selection, stoichiometric equations, and solid-state synthesis conditions (temperature, duration, atmosphere) for inorganic phases \citep{chen2023matchat}.
\item \textbf{Molecular and material property prediction}: Quantitative prediction of formation energy, band gap, and other descriptors alongside qualitative synthesis protocol generation for inorganic crystalline materials \citep{tang2025matterchat}; using CoT prompting to infer human-readable structure-property rules (e.g., TPSA, hydrogen bond donors) \citep{zheng2023large}; bio-inspired design and materials structure-property problems \citep{lu2025fine}; and bioinspired materials hypothesis generation for bio-inspired composites (silk, nacre, keratin-based systems) \citep{ghafarollahi2024sciagents}.
\item \textbf{Automatic scientific computation}: Various models were benchmarked on theoretical and numerical tasks derived from engineering graduate examinations \citep{miret2024enabling}; one work designed a framework for autonomous derivation of interpretable physical formulas from experimental data, applied to 2D material synthesizability and solid-state electrolyte conductivity \citep{song2025llm}.
\end{itemize}

\subsection{Domain-specific tasks that have not been addressed by reasoning models}

\begin{itemize}
\item Lack of experimental validation. Models can propose synthesis conditions but cannot verify their feasibility without experimental data. Frameworks generate plausible hypotheses but lack experimental follow-through and physical characterisation to confirm predictions. Hallucinations remain a significant problem.
\item More complex synthesis problems, such as green chemistry metrics reasoning (atom economy, E-factor, or PMI-guided synthesis planning), reaction condition optimisation (predicting optimal solvent systems, catalyst loadings, temperatures, and expected yields for a given synthetic step), protecting group strategy and chemoselectivity reasoning (selecting orthogonal protection schemes in multi-functional molecule synthesis) remain unexplored.
\item Reasoning over 3D and multi-modal data. While MatterChat \citep{tang2025matterchat} makes progress using graph representations, reasoning over full crystal structures (space group selection rules, symmetry-property relations) is still unreliable, as shown by the poor performance of LLMs on such complex numerical tasks \citep{miret2024enabling}.
\end{itemize}

\subsection{Comments on the usage of reasoning model in RAG-like system or agentic framework}
Agentic and RAG-like usage is emerging but not yet mature in this domain. The main bottleneck is the absence of reliable domain-specific tool integrations (e.g., interfaces to cheminformatics tools like RDKit, quantum chemistry calculators, or crystallography databases) within the agentic loops. \cite{zhang2025reasoning} explicitly calls for "incorporating cheminformatics validation tools or expert rules into the reasoning workflow" as a key future direction.

\subsection{Other comments}
The majority of fine-tuned domain-specific models operate in the 1–10B parameter range (LLaMA2-7B, Llama3-8B, Mistral-7B, Falcon-7B), reflecting a practical focus on deployable models on local machines. Larger proprietary models are used in agentic frameworks or as teacher models for distillation.

A recurring problem is the absence of human-curated reasoning chains. All datasets either use automatically distilled reasoning or provide no reasoning steps at all. This is a fundamental gap for training and evaluating genuine chemical reasoning.

Hallucinated synthesis protocols, fictitious references, and physically implausible formulas are noted as major limitations. No work has developed robust, automated chemical-validity checking pipelines to filter or penalise hallucinations at inference time.

\subsection{Number of datasets for training domain-specific reasoning model found}

\begin{itemize}
    \item \textbf{All}: between 4 and 6

    \item \textbf{Publicly accessible}: between 1 to 3
\end{itemize}

\subsection{Number of benchmarks/datasets for evaluating domain-specific reasoning model found}

\begin{itemize}
    \item \textbf{All}: between 4 and 6

    \item \textbf{Publicly accessible}: between 4 and 6
\end{itemize}

\subsection{Number of domain-specific reasoning models found}

\begin{itemize}
    \item \textbf{All}: between 4 and 6

    \item \textbf{Publicly accessible}: between 1 to 3
\end{itemize}

\subsection{Number of methods for creating/using domain-specific reasoning models found}

\begin{itemize}
    \item \textbf{All}: between 4 and 6

    \item \textbf{Publicly accessible}: between 4 and 6
\end{itemize}

\section[Literature Review Summary: PE6]{Literature Review Summary: PE6}

\textbf{Computer Science and Informatics.} Informatics and information systems, computer science, scientific computing, intelligent systems.

\subsection{Key Findings}

The literature review reveals a paradigm shift toward incentivizing reasoning through Reinforcement Learning (RL) rather than pure supervised fine-tuning.

\begin{itemize}
\item \textbf{Models \& Modalities}: Frontier models like DeepSeek-V3.2 \citep{deepseek2025_v32}, OpenAI GPT-5 \citep{openai2025_gpt5card}, and Claude 4.6 \citep{anthropic2026_opus46} use explicit reasoning steps (often encapsulated in <think> and </think> tags or similar). While text-only reasoning is standard, there is a strong push toward native multimodality (e.g., Gemini 3.1 \citep{google2026_gemini31pro}, Qwen 3.5 \citep{qwen2026_35}, Kimi K2.5 \citep{kimi2026_k25}), where vision and text are treated as equal tokens from pre-training.

\item \textbf{Methods \& Prompting}: Key methods include Group Relative Policy Optimization (GRPO) \citep{deepseek2025_r1} for resource-efficient RL and agentic task synthesis Pipelines to generate synthetic training environments. Prompting has evolved into "system-level" interactions where models use techniques like Recursive Criticism and Improvement (RCI) \citep{kim2023_tasks} or self-reflection to automatically refine their own response.

\item \textbf{Outputs \& Validation}: Outputs are increasingly structured (JSON, code blocks) or agentic (tool calls). Validation is moving toward execution-based verification (e.g., unit tests in SWE-bench) and reward models that score reasoning process rather than just the final answer.

\item \textbf{Efficiency}: Innovations like DeepSeek Sparse Attention (DSA) \citep{deepseek2025_v32} and Gated Delta Networks \citep{qwen2026_35} are being used to manage massive contexts (up to 1M tokens) while reducing inference costs.
\end{itemize}

\subsection{Identified Gaps}

\begin{itemize}
\item \textbf{Reverse Reasoning Deficit}: Models consistently struggle with "backward" reasoning (e.g., predicting inputs from outputs or reverse dependency patterns in code).

\item \textbf{Sustainability of Debugging}: Reasoning models show an exponential decay in effectiveness during multi-turn debugging sessions, often losing 60-80\% of their utility within three attempts.

\item \textbf{Multimodal Grounding}: Despite native multimodal training, models still lack the fine-grained perception needed for complex visual software engineering tasks.

\item \textbf{Safety vs. Utility}: High rejection rates in safety benchmarks (over-refusal) and "sabotage concealment" (hiding malicious side-tasks) remain significant challenges.
\end{itemize}

\subsection{Notable domain-specific tasks that have been addressed by reasoning models}

\begin{itemize}
\item \textbf{Software Engineering}: Automatic coding \citep{jain2024_livecodebench, roy2026_codesense, glm2026_glm5}, code debugging \citep{adnan2025_debugging}, static code analysis \citep{xie2025_core}, and resolution of real-world GitHub issues \citep{jimenez2024_swebench, yang2025_swemultimodal, yang2025_swesmith, chen2026_sweuniverse}.

\item \textbf{Smart or Agentic Assistant}: Various agentic use cases such as browsing for publicly available information \citep{zhang2026_browsecompv3}, tool use \citep{bandi2026_mcpatlas}, and other auxiliary computer tasks \citep{kim2023_tasks}.

\item \textbf{Cybersecurity}: Cyber Threat Intelligence (CTI) knowledge graph construction and attack reasoning \citep{yang2026_CTIThinker}.

\item \textbf{Coding Olympiads}: Achieving gold-medal performance in IOI 2025 \citep{deepseek2025_v32}.

\item \textbf{Scientific Computing}: Proving new bounds in algorithm performance (e.g., convex body chasing) and discovering conceptual links in cross-disciplinary research  \citep{openai2025_gpt5science}; performing complex multi-step programming on research tasks \citep{tian2024_scicode}.
\end{itemize}

\subsection{Domain-specific tasks that have not been addressed by reasoning models}

\begin{itemize}
\item \textbf{Cross-Functional Code Reasoning}: Most benchmarks focus on intra-procedural analysis; reasoning across large, complex codebases with heterogeneous languages remains unsolved.

\item \textbf{High-Fidelity Build Systems}: Automating the build and verification process for complex C/C++ environments is still manually intensive.

\item \textbf{Long-Horizon Interactive Tool-Use}: Models frequently experience "answer thrashing" or fail to identify the correct tool when faced with multiple distractors in real-world server environments.
\end{itemize}

\subsection{Comments on the usage of reasoning model in RAG-like system or agentic framework}
Reasoning models are being integrated into agentic frameworks using "Agent Swarms", i.e., parallel orchestration where a trainable orchestrator decomposes tasks for specialized sub-agents. In RAG systems, GraphRAG is utilized to link structured domain knowledge with LLM reasoning to significantly reduce hallucinations in specialized security contexts, as seen in CTI-Thinker \citep{yang2026_CTIThinker}.

\subsection{Other comments}
The field is rapidly moving toward "Agentic Engineering," where the focus is not just on generating code but on the long-horizon autonomy of the model within a verifiable environment.

\subsection{Number of datasets for training domain-specific reasoning model found}

\begin{itemize}
    \item \textbf{All}: many ($>$ 6)

    \item \textbf{Publicly accessible}: many ($>$ 6)
\end{itemize}

\subsection{Number of benchmarks/datasets for evaluating domain-specific reasoning model found}

\begin{itemize}
    \item \textbf{All}: many ($>$ 6)

    \item \textbf{Publicly accessible}: many ($>$ 6)
\end{itemize}

\subsection{Number of domain-specific reasoning models found}

\begin{itemize}
    \item \textbf{All}: many ($>$ 6)

    \item \textbf{Publicly accessible}: many ($>$ 6)
\end{itemize}

\subsection{Number of methods for creating/using domain-specific reasoning models found}

\begin{itemize}
    \item \textbf{All}: many ($>$ 6)

    \item \textbf{Publicly accessible}: many ($>$ 6)
\end{itemize}

\section[Literature Review Summary: PE7]{Literature Review Summary: PE7}

\textbf{Systems and Communication Engineering.} Electrical, electronic, communication, optical and systems engineering.

\subsection{Key Findings}

The reviewed literature indicates that reasoning-oriented LLM research in PE7 is emerging, but still relatively small and methodologically heterogeneous. Only a limited subset of papers actually introduces a domain-adapted reasoning model, most clearly DeepForm for communication-system formulation, GAIA for power dispatch, and LMTE for WAN traffic engineering. The larger fraction of the literature instead uses strong general-purpose frontier models, often with Chain-of-Thought (CoT), few-shot prompting, ReAct-style tool use, or multi-agent orchestration, to address domain tasks in communications, optical systems, circuits, and power systems \cite{Wu2025DeepForm,cheng2025gaia,Yuan2026LMTE,nosrati2026control,Zhou2025LLMWirelessPromptEngineering,Sharma2025PhIDO,Wang2024WhenLargeLanguageModelsOpticalNetworks,Jiang2024OptiCommGPT,alhasan2026circuitlm,schafer2025aiagentspowersystem}.

A second robust finding is that PE7 work is still predominantly \emph{text-centric}. Most systems take natural-language problem statements or operator requests as input and produce formal problem formulations, control recommendations, netlists, structured design files, or textual diagnoses as output. Multimodal reasoning is much less mature. In the present corpus, the most explicit multimodal contribution is EEE-Bench, which evaluates joint text-image reasoning over electrical and electronics engineering problems; by contrast, most optical, communications, control, and power studies convert non-textual information into text descriptions, retrieved documents, tables, or simulator outputs before reasoning is performed \cite{li2025eeebench,nosrati2026control,Zhou2025LLMWirelessPromptEngineering,Wang2024WhenLargeLanguageModelsOpticalNetworks,Jiang2024OptiCommGPT,schafer2025aiagentspowersystem}. 

A third finding concerns output representation. The most successful systems rarely rely on unconstrained free-form answers alone. Instead, they use typed or structured intermediate representations that can be checked by downstream tools. DeepForm targets formal communication-system formulations; PhIDO uses a photonic YAML-like design representation; PICBench evaluates simulator-ready PIC netlists; CircuitLM generates CircuitJSON schematics; and power-system and optical-network assistants typically organize reasoning around structured workflows, retrieved domain documents, or tool-mediated execution \cite{Wu2025DeepForm,Sharma2025PhIDO,wu2025picbench,alhasan2026circuitlm,Wang2024WhenLargeLanguageModelsOpticalNetworks,Jiang2024OptiCommGPT,cheng2025gaia,schafer2025aiagentspowersystem}. This suggests that, in PE7, reasoning quality is tightly coupled to representation design and tool compatibility.

The evidence base for datasets and benchmarks is stronger on the \emph{evaluation} side than on the \emph{training} side. Purpose-built evaluation resources now exist for photonic integrated circuits, multimodal electrical and electronics engineering, circuit analysis, and natural-language-driven circuit generation, through PICBench, the PhIDO benchmark, EEE-Bench, CircuChain, and the CircuitLM benchmark, respectively \cite{Sharma2025PhIDO,wu2025picbench,li2025eeebench,ravishankara2026circuchain,alhasan2026circuitlm}. By contrast, comparatively few papers describe reusable training corpora for domain-specific reasoning adaptation. The clearest examples are CSFRC for communication-system formulation, the data-generation pipeline used to train GAIA for dispatch, and the public WAN datasets and synthetic traffic-matrix generation procedures used in LMTE \cite{Wu2025DeepForm,cheng2025gaia,Yuan2026LMTE}. Overall, benchmark construction is progressing faster than public training-data release.

Validation practices remain largely \emph{outcome-based}. Many papers verify the final design or decision using simulators, rule checks, benchmark accuracy, engineering heuristics, or end-task success, but relatively few directly evaluate whether intermediate reasoning steps are faithful, minimal, or causally responsible for the correct answer. This is particularly visible in PIC design, optical-network automation, and power-system assistance. The clearest attempts to inspect reasoning failure modes, rather than only final outputs, are CircuChain, which separates competence from compliance in circuit analysis, and EEE-Bench, which exposes multimodal failure patterns such as neglect of visual evidence \cite{wu2025picbench,Wang2024WhenLargeLanguageModelsOpticalNetworks,Jiang2024OptiCommGPT,cheng2025gaia,ravishankara2026circuchain,li2025eeebench,schafer2025aiagentspowersystem}.

Finally, reasoning effort, cost, and deployment efficiency are acknowledged but are not yet central optimization targets. PhIDO explicitly compares token usage and cost across strong reasoning models, LMTE emphasizes efficiency and scalability for network control, and DeepForm argues that a 7B domain-adapted model can outperform larger generic baselines on its target task \cite{Sharma2025PhIDO,Yuan2026LMTE,Wu2025DeepForm}. However, explicit control of reasoning depth, latency-constrained reasoning, and formal cost-accuracy trade-off analysis remain uncommon across the surveyed corpus \cite{nosrati2026control,Zhou2025LLMWirelessPromptEngineering,Cruzes2025LLMOpticalNetworks,Jiang2024OptiCommGPT}.

\subsection{Identified Gaps}

The most immediate gap is the shortage of public, training-grade corpora for PE7 reasoning. Although several papers claim domain specialization, only a small number provide enough detail about data construction to support reproducibility, and even fewer release training resources in a form that enables independent replication or follow-on fine-tuning \cite{Wu2025DeepForm,cheng2025gaia,Yuan2026LMTE}. This limits fair comparison across methods and makes it difficult to separate gains due to domain data, prompting, retrieval, or tool orchestration.

A second gap is insufficient evaluation of reasoning faithfulness. In most studies, the reasoning process is treated as helpful if the final design, answer, or control action is correct. This is often insufficient for PE7 tasks, where safety, sign conventions, design rules, and procedural correctness matter as much as the final numerical answer. CircuChain is especially important here because it shows that physical competence and instruction compliance may diverge even in relatively simple circuit-analysis settings \cite{ravishankara2026circuchain}. More broadly, the corpus lacks human-validated reasoning traces, process-level scoring rubrics, and verifier-in-the-loop studies that specifically target reasoning correctness \cite{Wu2025DeepForm,cheng2025gaia,ravishankara2026circuchain,li2025eeebench}.

A third gap is limited multimodal engineering reasoning. Real engineering workflows require interpreting schematics, block diagrams, waveforms, eye diagrams, plots, tabulated telemetry, layouts, and equations. Yet most reviewed PE7 systems remain text-only or text-dominant. EEE-Bench shows that this problem is not marginal: current multimodal models remain weak on realistic EEE tasks and often underuse visual evidence \cite{li2025eeebench}. This gap is also explicitly recognized in photonics and optical-network research, where future systems are expected to integrate diagrams, layouts, telemetry, and physics-aware representations more directly \cite{Sharma2025PhIDO,Cruzes2025LLMOpticalNetworks}.

A fourth gap is the lack of deployment-grade safety and robustness evidence. Power-system and optical-network papers provide encouraging proof-of-concept results, but the literature still lacks strong evidence on safe action constraints, fallback mechanisms, adversarial robustness, latency under operational load, cybersecurity implications, and operator trust calibration in real environments \cite{Wang2024WhenLargeLanguageModelsOpticalNetworks,cheng2025gaia,Cruzes2025LLMOpticalNetworks,schafer2025aiagentspowersystem}. The same concern applies to design automation in circuits and photonics, where simulator-valid outputs do not yet imply fabrication-ready or certification-ready correctness \cite{Sharma2025PhIDO,wu2025picbench,alhasan2026circuitlm}.

A final gap is uneven subfield coverage. The current literature is comparatively strong in optical networks, photonic integrated circuits, power systems, circuit reasoning, and communication/network formulation, but much thinner in RF and microwave engineering, antenna and array reasoning, electromagnetics, EMC/EMI analysis, mixed-signal verification, hardware security, protocol-stack debugging, and waveform-level communications analysis. These areas appear to be largely open rather than mature application domains \cite{nosrati2026control,Zhou2025LLMWirelessPromptEngineering,Cruzes2025LLMOpticalNetworks,li2025eeebench}.

\subsection{Notable domain-specific tasks that have been addressed by reasoning models}

The reviewed studies show that reasoning-capable LLMs have already been applied to a diverse set of PE7 tasks. In communications, DeepForm treats communication-system formulation itself as a domain-specific reasoning problem, while wireless-network studies investigate optimization and forecasting tasks through prompt engineering and self-refinement \cite{Wu2025DeepForm,Zhou2025LLMWirelessPromptEngineering}. In control and systems engineering, current work focuses on controller design, tuning, workflow support, and high-level planning rather than tightly verified closed-loop autonomy \cite{nosrati2026control}.

In optical and photonic engineering, the literature covers both design and operations. PhIDO and PICBench study natural-language-to-PIC design, netlist generation, and the role of structured representations and simulator feedback in photonic design synthesis \cite{Sharma2025PhIDO,wu2025picbench}. Optical-network studies address alarm compression, alarm prioritization, troubleshooting, QoT-oriented assistance, and broader intent-driven automation, often with retrieval and domain resource libraries \cite{Wang2024WhenLargeLanguageModelsOpticalNetworks,Cruzes2025LLMOpticalNetworks}. OptiComm-GPT extends this trend to optical fiber communication systems by integrating GPT-based reasoning with simulation, DSP verification, QoT evaluation, optimization, and tool orchestration \cite{Jiang2024OptiCommGPT}.

In electrical and electronic engineering, notable addressed tasks include advanced power dispatch, distribution-grid operational assistance, natural-language-driven circuit generation, and circuit analysis \cite{cheng2025gaia,schafer2025aiagentspowersystem,alhasan2026circuitlm,ravishankara2026circuchain}. EEE-Bench further shows that a broad range of EEE tasks can at least be posed as multimodal reasoning problems, even if current model accuracy remains modest \cite{li2025eeebench}. Collectively, the literature demonstrates meaningful progress on both \emph{analysis} tasks (e.g., diagnosis, dispatch reasoning, troubleshooting, circuit solving) and \emph{synthesis} tasks (e.g., formulation, schematic generation, PIC design, workflow planning).

\subsection{Domain-specific tasks that have not been addressed by reasoning models}

Several important PE7 tasks remain largely unaddressed or only partially addressed. First, fabrication-ready, sign-off-grade automation is still absent. Current PIC and circuit papers show progress toward simulator-valid netlists or structurally plausible schematics, but not robust foundry-ready PIC layout closure, PCB sign-off, analog/mixed-signal verification closure, or standards-compliant manufacturability \cite{Sharma2025PhIDO,wu2025picbench,alhasan2026circuitlm}. 

Second, multimodal reasoning over dense engineering artifacts remains underdeveloped. The present literature provides early evidence that such reasoning is necessary, but not convincing evidence that current systems can reliably interpret circuit schematics, waveform traces, constellation diagrams, Smith charts, Bode or Nyquist plots, oscilloscope captures, block diagrams, or mixed text-figure design packages at a level suitable for expert engineering use \cite{li2025eeebench,Cruzes2025LLMOpticalNetworks}. 

Third, real-time, safety-constrained operational autonomy remains unresolved. Power-system and optical-network agent studies are still largely proof-of-concept and simulation-centric, with limited evidence for certified constraint handling, formal safety envelopes, online fault recovery under adversarial conditions, or trustworthy integration into live operational workflows \cite{Wang2024WhenLargeLanguageModelsOpticalNetworks,cheng2025gaia,schafer2025aiagentspowersystem}. 

Fourth, several subfields are scarcely represented in the present corpus: RF and microwave circuit design, antenna synthesis, electromagnetic field simulation workflows, EMC/EMI diagnosis, communication-protocol debugging across PHY/MAC/RAN layers, hardware security reasoning, and rigorous embedded real-time systems verification. These appear to be open opportunities for future work rather than established application areas \cite{Zhou2025LLMWirelessPromptEngineering,li2025eeebench}. 

Finally, reasoning-effort control itself has not been studied in a sufficiently engineering-aware way. Very few papers explicitly optimize bounded latency, cost-constrained inference, or adaptive reasoning depth, despite the operational importance of these constraints in network control, power-system assistance, and design automation \cite{Wu2025DeepForm,Sharma2025PhIDO,Yuan2026LMTE,Cruzes2025LLMOpticalNetworks}.

\subsection{Comments on the usage of reasoning model in RAG-like system or agentic framework}

A strong cross-cutting pattern in PE7 is that reasoning models are rarely deployed as isolated text predictors. Instead, they are embedded in retrieval-augmented, tool-augmented, or agentic systems that provide domain grounding and external verification. Optical-network and optical-communication papers are particularly explicit in this regard: both the LLM-empowered optical-network framework and OptiComm-GPT combine prompting with domain knowledge bases, retrieval, workflow decomposition, and external tools \cite{Wang2024WhenLargeLanguageModelsOpticalNetworks,Jiang2024OptiCommGPT,Cruzes2025LLMOpticalNetworks}. 

The same pattern appears in design automation. PhIDO uses a multi-agent pipeline and a structured photonic design language to translate natural-language intent into verifiable PIC artifacts, while CircuitLM uses retrieval over component knowledge, explicit intermediate reasoning stages, structured CircuitJSON output, and dual validation layers \cite{Sharma2025PhIDO,alhasan2026circuitlm}. In power systems, the most promising systems also follow the same logic: the LLM is used to interpret intent, organize multi-step reasoning, and interact with external analysis tools, rather than being trusted as a stand-alone decision engine \cite{cheng2025gaia,schafer2025aiagentspowersystem}. 

This design choice is well aligned with PE7 requirements. Retrieval supplies domain knowledge that is often sparse in general-purpose LLMs; tool use enables contact with simulators, solvers, network data, or device libraries; and structured intermediate representations make the reasoning process more inspectable. The main downside is methodological: once retrieval, agents, simulators, and validation modules are added, benchmarking becomes a comparison of full systems rather than of reasoning models alone. This complicates attribution of performance gains and partly explains why cross-paper comparison remains difficult \cite{Wu2025DeepForm,Sharma2025PhIDO,Wang2024WhenLargeLanguageModelsOpticalNetworks,Jiang2024OptiCommGPT,alhasan2026circuitlm}.

\subsection{Other comments}

A useful interpretation of the current literature is that PE7 is not yet a mature field of domain-native reasoning foundation models. Rather, it is a field in which engineering workflows are being progressively reorganized around reasoning-capable LLMs. The most convincing systems are not those that rely on free-form reasoning alone, but those that combine reasoning with domain knowledge, structured outputs, retrieval, simulation, deterministic checking, or agentic tool use \cite{Wu2025DeepForm,Sharma2025PhIDO,cheng2025gaia,Jiang2024OptiCommGPT,alhasan2026circuitlm,ravishankara2026circuchain,schafer2025aiagentspowersystem}. 

It is also important to distinguish surveys and perspective papers from task-specific solution papers. The survey and perspective pieces in control, wireless networking, and optical networking are valuable for framing the research agenda, but they do not yet provide the same level of empirical evidence as task-specific studies on communication-system formulation, dispatch, PIC design, or circuit reasoning \cite{nosrati2026control,Zhou2025LLMWirelessPromptEngineering,Cruzes2025LLMOpticalNetworks}. Consequently, strong conclusions about maturity should be drawn from the empirical systems and benchmarks, while broader generalizations should remain cautious.

Overall, the literature suggests that the most promising near-term direction is not simply scaling generic reasoning models further, but combining smaller or medium-sized domain-adapted models with retrieval, structured representations, verification modules, and engineering simulators. This hybrid direction appears better matched to the accuracy, auditability, and safety requirements of PE7 applications \cite{Wu2025DeepForm,Yuan2026LMTE,Sharma2025PhIDO,alhasan2026circuitlm,schafer2025aiagentspowersystem}. In the quantitative count sections below, a \emph{domain-specific reasoning model} is interpreted strictly as a PE7-adapted model or model family, rather than a prompt-only application pipeline or general-purpose frontier model used off the shelf.

\subsection{Number of datasets for training domain-specific reasoning model found}

\begin{itemize}
    \item \textbf{All}: between 1 to 3

    \item \textbf{Publicly accessible}: between 1 to 3
\end{itemize}

\subsection{Number of benchmarks/datasets for evaluating domain-specific reasoning model found}

\begin{itemize}
    \item \textbf{All}: many ($>$ 6)

    \item \textbf{Publicly accessible}: many ($>$ 6)
\end{itemize}

\subsection{Number of domain-specific reasoning models found}

\begin{itemize}
    \item \textbf{All}: between 1 to 3

    \item \textbf{Publicly accessible}: between 1 to 3
\end{itemize}

\subsection{Number of methods for creating/using domain-specific reasoning models found}

\begin{itemize}
    \item \textbf{All}: many ($>$ 6)

    \item \textbf{Publicly accessible}: many ($>$ 6)
\end{itemize}

\section[Literature Review Summary: PE8]{Literature Review Summary: PE8}

\textbf{Products and Processes Engineering.} Product and process design, chemical, civil, environmental, mechanical, vehicle engineering, energy processes and relevant computational methods.

\subsection{Key Findings}
Recent publications in the field of product and process engineering indicate a growing interest in reasoning language models. However, much of the existing work remains at an exploratory stage, as many studies focus primarily on qualitative analyzes and manual inspections of model capabilities. A number of works investigate the performance of off-the-shelf reasoning models and conduct comparisons with non-reasoning models. They used mainly proprietary RLMs such as OpenAI’s GPT, Google’s Gemini, and Claude models but they also used open-weights models like Deepseek. Another approach explores agentic systems where a reasoning agent interacts with engineering software or a neuro-inspired network to enhance LLM reasoning and domain adaptability. Many works introduce benchmarks based on authentic engineering and production tasks. Finally, several studies propose new aligned models.

Notable benchmarks include:
\begin{itemize}
\item FEABench \citep{mudur2025feabench} is a benchmark to evaluate the ability of language models to simulate and solve engineering problems using finite element analysis (FEA) software. The input is a problem description (i.e. Lorenz attractor, heat transfer) with a specific target quantity that needs to be computed.
\item Wrenchmark \citep{heesch2025evaluatinglargelanguagemodels} is a benchmark comprising over 100 questions derived from authentic, production-oriented engineering scenarios, systematically designed to cover core competencies such as product design, prognosis, and diagnosis
\item EngTrace \citep{gull2026engtracesymbolicbenchmarkverifiable} is a benchmark comprising 90 templates written in Python across three major engineering branches for generating physically grounded engineering problems, along with their reasoning steps.
\item FailureSensorIQ \citep{lin-etal-2025-fine} is a benchmark designed to assess the ability to reason and understand the relations between sensor/parameter and failures/faults for assets in Industry 4.0. Yet, it focuses only on the final answers without validating the reasoning steps.
\end{itemize}
 
Several domain-specific reasoning models were identified. CAD-Coder \citep{guan2025cadcodertexttocadgenerationchainofthought} is a framework that reformulates the text-to-CAD task as generating CadQuery code which is executed to generate a 3D geometry from natural language description of the 3D design intent. In a two-stage training strategy authors firstly fine-tune language model - Qwen2.5-7B-Instruct to learn CadQuery’s syntax and then use GRPO incorporating a chain-of-thought (CoT) planning process to improve geometric reasoning ability. In another work authors propose knowledge distillation framework for industrial asset health, which transfers reasoning capabilities via CoT and QLoRA with 4-bit precision for model fine-tuning from LLMs to smaller, more efficient models (SLMs) \citep{lin-etal-2025-fine}. Helios \citep{jiang2026heliosfoundationallanguagemodel} is a LLM tailored to the smart energy domain employing the Qwen-2.5 7B model pre-trained on a single-epoch training conducted on a domain-specific corpus of approximately 3 billion tokens in the smart-energy domain. Then it undergoes instruction tuning with LoRA and RLHF. 
In another system LLM-driven reasoning agents were used to autonomously explore vast alloy design spaces, identify trends in atomic-scale properties and predict macroscale mechanical strength \citep{Ghafarollahi2025}. Advanced LLMs and reasoning models were responsible for fundamental tasks such as planning, reasoning, and decision-making.

Typically, the models utilize text-only or text-code modality. However, other modalities can also be found. SDIGLM \citep{zhang2025sdiglmleveraginglargelanguage} is an LLM designed for structural damage identification based on VisualGLM-6B architecture. It integrates U-Net based semantic segmentation module to generate defect segmentation maps as visual CoT. Additionally the model is fine-tuned on multi-round dialogue dataset to enhance logical reasoning, complemented by a language CoT formed through prompt engineering.

\subsection{Identified Gaps}
Notable gaps have been identified, namely:
\begin{itemize}
    \item Fully curated and human-verified datasets are typically small due to the complexity of the underlying engineering challenges. Moreover, they are highly diverse, requiring expertise in 3D modeling languages, the ability to interact with engineering simulation software, and the capability to identify specific structures within images.
    \item Larger datasets do exist; however, they are typically generated by large language models using templates or curated prompts. Knowledge distilled from LLMs in this manner may introduce inconsistencies or inaccuracies.
    \item Evaluation of a reasoning chain is uncommon. When performed, it typically involves qualitative manual analysis or is conducted indirectly through validation of the final answer.
\end{itemize}

\subsection{Notable domain-specific tasks that have been addressed by reasoning models}
\begin{itemize}
    \item FEA: Domain-specific reasoning models have been evaluated on their ability to solve physics, mathematics, and engineering problems by operating professional finite element software \citep{mudur2025feabench}
    \item Computational Fluid Dynamics (CFD) problem solving: Reasoning models have been evaluated on their ability to leverage, adapt, and generate numerical methods for CFD tasks, including conventional benchmark problems (e.g., lid-driven cavity flow and Sod shock tube), as well as problems that require new numerical methods and mathematically ill-conditioned systems such as Hilbert linear algebraic equations \citep{WANG2025100597}
    \item Semiconductor layout design: Reasoning models have been applied to domain-specific semiconductor layout design tasks \citep{wen2025enhancingreasoningadaptlarge}, demonstrating the ability to adapt general-purpose LLMs to specialized engineering problems that require spatial reasoning and the application of technical domain knowledge.
    \item CAD generation and geometric reasoning: Domain-specific reasoning capabilities have been demonstrated in parametric CAD tasks \citep{guan2025cadcodertexttocadgenerationchainofthought}, where natural language descriptions are translated into executable CadQuery scripts to generate structurally valid and geometrically accurate 3D models, advancing automated design synthesis and geometric validation.
    \item Industrial asset health monitoring in Industry 4.0: Domain-specific reasoning has been applied to industrial asset health monitoring \citep{lin-etal-2025-fine}, supporting fault diagnosis, condition assessment, and maintenance decision-making tasks in smart manufacturing and Industry 4.0 environments.
    \item Smart energy system planning and operational decision-making: a variety of smart energy-specific tasks, like knowledge integration and automatic code generation \citep{jiang2026heliosfoundationallanguagemodel}
    \item Structural damage identification and assessment in civil engineering: Reasoning models have been utilized to analyze multi-modal data from infrastructure \citep{zhang2025sdiglmleveraginglargelanguage}, combining visual defect segmentation with textual descriptions to identify damage types, quantify severity, and generate reports, thereby enabling automated inspection, maintenance planning, and decision-making across diverse civil engineering structures.
    \item Engineering innovation and inventive problem solving: Reasoning models have been integrated with structured ideation methodologies such as TRIZ to automate contradiction analysis, principle selection, and solution generation \citep{JIANG2025103312}, which leverages large language models to conduct systematic TRIZ reasoning from problem formulation to structured solution reporting. 
    \item Conceptual design: Generative models have been integrated for early-stage conceptual design \citep{10.1115/1.4065487}. This approach does not employ reasoning models or explicit reasoning capabilities. Instead, it simulates aspects of the reasoning process through a carefully designed framework and structured human interaction, breaking down a design challenge into discrete subtasks with guided directions for reasoning in each subtask.
    \item Discovery of new metallic alloys leveraging multimodal data and external knowledge, complemented by insights from atomistic physics simulations \citep{Ghafarollahi2025}
\end{itemize}

\subsection{Domain-specific tasks that have not been addressed by reasoning models}

\begin{itemize}
    \item End-to-end reasoning from stakeholder requirements to subsystem architecture to component sizing, with bidirectional traceability and constraint propagation. 
    \item Certification-grade Compliance Reasoning: Certification-grade reasoning involves mapping design features to specific clauses of regulatory standards, demonstrating margin compliance, and documenting traceable verification evidence. 
    \item Manufacturability and Supply-Chain-Aware Design Loops: Increasing complexity further would involve linking CAD geometry and simulation outputs to process constraints. The LLM would need to evaluate whether a proposed geometry change introduces new tooling or creates single-source supply dependencies. 
    \item Physics-Constrained, Multi-Scale Causal Reasoning Under Formal Guarantees: Engineering domains frequently require integrated reasoning across non-local dependencies, type-level abstractions, formal causal structures, and coupled physical dynamics. Such tasks demand the propagation of constraints across hierarchical system layers, reconciliation of interacting subsystems, and validation against governing physical laws.
\end{itemize}

\subsection{Comments on the usage of reasoning model in RAG-like system or agentic framework}
One example occurs when the reasoning model serves as an intermediary between a fixed knowledge base and the generative capabilities of the LLM, without performing full retrieval-augmented generation. Another instance arises in a RAG-like, enhanced framework, where a pool of LLMs generates diverse candidate thoughts for a task, and the system selectively evaluates these ideas using attention mechanisms to prioritize the most relevant and goal-aligned solutions.

\subsection{Number of datasets for training domain-specific reasoning model found}

\begin{itemize}
    \item All: Zero / between 1 to 3 / \textbf{between 4 and 6} / many ($>$ 6)

    \item \textbf{Publicly accessible}: Zero / \textbf{between 1 to 3} / between 4 and 6 / many ($>$ 6)
\end{itemize}

\subsection{Number of benchmarks/datasets for evaluating domain-specific reasoning model found}

\begin{itemize}
    \item All: Zero / between 1 to 3 / between 4 and 6 / \textbf{many ($>$ 6)}

    \item \textbf{Publicly accessible}: Zero / between 1 to 3 / between 4 and 6 / \textbf{many ($>$ 6)}
\end{itemize}

\subsection{Number of domain-specific reasoning models found}

\begin{itemize}
    \item All: Zero / between 1 to 3 / \textbf{between 4 and 6} / many ($>$ 6)

    \item \textbf{Publicly accessible}: Zero / \textbf{between 1 to 3} / between 4 and 6 / many ($>$ 6)
\end{itemize}

\subsection{Number of methods for creating/using domain-specific reasoning models found}

\begin{itemize}
    \item \textbf{All}: Zero / between 1 to 3 / between 4 and 6 / \textbf{many ($>$ 6)}

    \item \textbf{Publicly accessible}: Zero / \textbf{between 1 to 3} / between 4 and 6 / many ($>$ 6)
\end{itemize}

\section[Literature Review Summary: PE9]{Literature Review Summary: PE9}

\textbf{Universe Sciences.} Astro-physics/-chemistry/-biology; solar system; planetary systems; stellar, galactic and extragalactic astronomy; cosmology; space sciences; astronomical instrumentation and data

\subsection{Key Findings}

The current landscape of Reasoning Language Models (RLMs) in astronomy shows rapid progress in both educational and research applications, as indicated by the sudden increase in publications (most of which are preprints waiting for official publications) related to RLMs in the astronomy field in 2025. The majority of works used RLMs in agentic frameworks for automating scientific tasks in astronomy. Most other works focused on creating benchmarks for evaluating the performance of language models including reasoning models. An overwhelming number of these works focused on proprietary RLMs such as OpenAI’s GPT, Google’s Gemini, and Claude models, although there is already one initiative at developing an open-weight foundational reasoning model specialized for astronomy (AstroSage-70B \citep{haan2025_astro_astrosage}).

Notable benchmarks include:
\begin{itemize}
\item ReplicationBench \citep{ye2025_astro_replicationbench}. It is a benchmark for evaluating AI agents in replicating scientific experiments from research papers in astrophysics. Besides evaluating the final answers from the language models, the work also included a qualitative evaluation where experts examined reasoning traces for Intent Interpretation and Quality of Execution (reasoning coherence and implementation correctness).
\item The International Olympiad on Astronomy \& Astrophysics (IOAA) problems from 2022 - 2025 have been used to evaluate the performance of reasoning models; experts were asked to validate the reasoning steps as well as final answers \citep{pinheiro2025_astro_olym}. However, while reference answers are acquirable for evaluating the final answers, the dataset does not include “gold” reasoning traces for automatically validating other models’ reasoning.
\item AstroVisBench \citep{joseph2025_astro_visbench} is a benchmark for scientific computing and visualization in the astronomy domain, and has been used to evaluate reasoning models. Yet, it focuses on the code execution and final outputs without validating the reasoning steps of the models.
\item The Astro-QA benchmark \citep{li2025_astroqa} has been used to evaluate some reasoning models on astronomy knowledge QA. Yet, it focuses only on the final answers without validating the reasoning steps.
\item A sample benchmark created from real experts queries and refined reference answers was also created \citep{ting2025_astro_edu}, yet it was mainly aimed at evaluating language models in RAG-based systems, so it does not include “gold” reasoning traces.
\end{itemize}

Currently, AstroSage-70B \citep{haan2025_astro_astrosage} is the only reasoning model that is specialized in astronomy. It has been shown to outperform general-purpose reasoning models in astronomy QA. However, the model’s reasoning capabilities have not been fully tested on complex problems, and the training was limited to 2.5 epochs for CPT (continual pretraining) and 0.6 epochs for SFT due to computational resource constraints; full convergence might have not been achieved in the SFT phase, suggesting potential for further performance improvement with more compute. Other astronomy-focused language models exist, such as AstroSage-8B, CosmoSage, and StarWhisper 3, but they are not optimized for explicit reasoning.

Reasoning models have successfully addressed several complex, data-intensive tasks in astronomy, often leveraging their natural language understanding and image processing (multimodal) capabilities. A major strength of reasoning models compared to traditional deep learning approaches is explainability, as it provides the step-by-step thinking of the model to reach the final answer.

\subsection{Identified Gaps}
Notable gaps have been identified, namely:
\begin{enumerate}
\item The lack of open, publicly available astronomy-focused reasoning datasets for developing reasoning models;
\item So far, there is very little effort on creating, using, and evaluating open-weight reasoning models for astronomy apart from AstroSage-70B;
\item The lack of a systematic method for validating the correctness of reasoning steps performed by a model. Some benchmarks have been proposed to evaluate the applicability of language models for astronomy, but they typically focus on evaluating the final answer of the model without considering the correctness of the reasoning steps. In a few works, the reasoning steps of the model were validated manually by experts, but such an approach would be very costly and time-consuming to be conducted on a large set of problems.
\end{enumerate}

\subsection{Notable domain-specific tasks that have been addressed by reasoning models}
Reasoning models have successfully addressed several complex, data-intensive tasks in astronomy, often leveraging their natural language understanding and image processing (multimodal) capabilities. A major strength of reasoning models compared to traditional deep learning approaches is explainability, as it provides the step-by-step thinking of the model to reach the final answer.

\begin{enumerate}
\item Transient classification and event identification \citep{stoppa2025_astro_img}. A key addressed task is the rapid classification of astronomical alerts (transients and variable objects) from large sky surveys (e.g., MeerLICHT, ATLAS, Pan-STARRS). Reasoning models with few-shot examples can accurately classify "real" cosmic events (like exploding stars or tidal disruption events) from "bogus" signals (like cosmic rays or satellite trails), and adding a self-correction loop to flag uncertain cases has been shown to boost the accuracy further.
\item Automatic data analysis, hypothesis exploration, and algorithm design. The Mephisto agentic framework \citep{sun2025_astro_mephisto} uses a reasoning model to automatically and iteratively refine a physical model until its predicted Spectral Energy Distribution (SED) best matches the observed data. The Evo-MCTS agentic framework \citep{wang2025_astro_gravwave} uses a reasoning model as its base to automatically generate and optimize algorithms for gravitational wave signal identification. SimAgents \citep{zhang2025_astro_litrev_agent} utilizes chain-of-thought prompting to leverage a language model to extract parameter configuration from literature and conduct preliminary analysis for cosmology research. Meanwhile, CMBAgent \citep{xu2025_astro_cmbagent} has been used to measure cosmological parameters from supernova data fully autonomously, demonstrating end-to-end scientific research with no human-in-the-loop.
\item Code generation. One work uses a reasoning model in an agentic framework to assist in writing, executing, and validating code using the Gammapy framework \citep{kostunin2025_astro_gammapy}. Another work proposes CLAPP (CLASS LLM Agent for Pair Programming), an interactive AI assistant for the CLASS Einstein-Boltzmann solver \citep{casas2025_astro_clapp}.
\item Simulating autonomous spacecraft operators \citep{carrasco2025_astro_kerbal}. One work uses a pure LLM-based agent for spacecraft control in the Kerbal Space Program Differential Games (KSPDG) challenge.
\item Knowledge synthesis and question answering (QA) \citep{hyk2025_astro_needs}. Language models, and more specifically reasoning models, are being actively used for complex astronomical QA, which requires integrating knowledge across diverse scientific literature.
\item Education. The AstroTutor agentic framework \citep{ting2025_astro_edu} uses a reasoning model to create an AI assistant to aid students in studying undergraduate astronomy. The work also investigated the use of a reasoning model for automatic evaluation of students’ assignments.
\end{enumerate}

\subsection{Domain-specific tasks that have not been addressed by reasoning models}

\begin{enumerate}
\item Full Multimodal Data Interpretation: Expanding RLM capabilities to fully interpret spectra and time series data, beyond current image/photometry inputs. This also includes improving the ability to quantitatively interpret scientific plots, which remains a limitation for current vision-enabled LLMs.
\item Advanced Geometric and Spatial Reasoning: Developing novel methods, such as implementing a "visual sketchpad," to help models visualize and solve complex geometric and spatial representations common in astrophysics problems.
\item Fine-Grained Classification: Extending image classification success to more nuanced, fine-grained astrophysical classification tasks such as identifying supernova subtypes, gravitational lensing events, and rare phenomena.
\item Planning Complex Observational Campaigns: LLMs could act as advanced agents to plan complex, multi-telescope observational campaigns (e.g., follow-up observations of transients). This requires sophisticated reasoning about telescope scheduling, atmospheric conditions, astronomical constraints (e.g., visibility, non-sidelobe contamination), and optimizing for scientific impact—a high-dimensional optimization problem requiring logical deduction.
\item Discovery of Novel Physical Laws and Analogies: A long-term goal is using reasoning models for true scientific discovery: to identify novel physical laws or fundamental relationships between disparate phenomena by detecting subtle, non-linear correlations across massive, multi-survey datasets, going beyond the patterns currently recognized by human experts.
\end{enumerate}

\subsection{Comments on the usage of reasoning model in RAG-like system or agentic framework}
The majority of works focused on agentic frameworks that leverage language models for performing certain tasks \citep{stoppa2025_astro_img,sun2025_astro_mephisto,wang2025_astro_gravwave,zhang2025_astro_litrev_agent,xu2025_astro_cmbagent,kostunin2025_astro_gammapy,casas2025_astro_clapp,carrasco2025_astro_kerbal,ting2025_astro_edu}. Some of these frameworks utilized RAG-like components.

\subsection{Number of datasets for training domain-specific reasoning model found}

\begin{itemize}
    \item \textbf{All}: Zero / \textbf{between 1 to 3} / between 4 and 6 / many ($>$ 6)

    \item \textbf{Publicly accessible}: \textbf{Zero} / between 1 to 3 / between 4 and 6 / many ($>$ 6)
\end{itemize}

\subsection{Number of benchmarks/datasets for evaluating domain-specific reasoning model found}

\begin{itemize}
    \item \textbf{All}: Zero / between 1 to 3 / \textbf{between 4 and 6} / many ($>$ 6)

    \item \textbf{Publicly accessible}: Zero / between 1 to 3 / \textbf{between 4 and 6} / many ($>$ 6)
\end{itemize}

\subsection{Number of domain-specific reasoning models found}

\begin{itemize}
    \item \textbf{All}: Zero / \textbf{between 1 to 3} / between 4 and 6 / many ($>$ 6)

    \item \textbf{Publicly accessible}: Zero / \textbf{between 1 to 3} / between 4 and 6 / many ($>$ 6)
\end{itemize}

\subsection{Number of methods for creating/using domain-specific reasoning models found}

\begin{itemize}
    \item \textbf{All}: Zero / between 1 to 3 / between 4 and 6 / \textbf{many ($>$ 6)}

    \item \textbf{Publicly accessible}: Zero / between 1 to 3 / \textbf{between 4 and 6} / many ($>$ 6)
\end{itemize}

\section[Literature Review Summary: PE10]{Literature Review Summary: PE10}

\textbf{Earth System Science.} Physical geography, geology, geophysics, atmospheric sciences, oceanography, climatology, cryology, ecology, global environmental change, biogeochemical cycles, natural resources management.

\subsection{Key Findings}

\subsubsection{The achievements of reasoning models}

Recent reasoning models are showing strong, cross-cutting gains across Earth-science subdisciplines and task types. In atmospheric science evaluations, they significantly outperform instruction-tuned, math-specialized, and domain-specific baselines on targeted scientific tasks~\cite{li2025atmosscibench}. In geospatial perception, reasoning-oriented training recipes can also improve satellite image understanding: combining CoT-style supervised fine-tuning with RL alignment (e.g. GRPO) yields the strongest overall system in object classification from satellite imagery (SAMChat-R1)~\cite{koksal2025samchat}. For mineral exploration from multi-image input, agentic reasoning frameworks likewise improve robustness: STA-CoT generalizes particularly well to more challenging settings, with clearer advantages on the hard split (MineBench-Hard)~\cite{yu-etal-2025-sta}. Similar trends appear in decision-oriented hydrology assistants, where Water\_agent achieves high task completion rates ($\approx 96\%$ on easier tasks and $\approx 90\%$ on complex tasks), substantially exceeding non-reasoning baselines~\cite{ren2024watergpt}. Beyond problem-solving, reasoning models also improve scientific communication: they consistently outperform non-reasoning models in assigning confidence to climate statements and are slightly better at summarizing complex climate evidence~\cite{lu-2025-reasoning}. Even without additional training, CoT prompting alone can increase helpfulness, coherence, and depth on geoscience inferential tasks~\cite{chen2025geofactory}. In more general geospatial and remote-sensing benchmarks, reasoning models tend to outperform smaller chat models, with the greatest gains concentrated on the most difficult tasks~\cite{pmlr-v292-cardille25a}.

\subsubsection{Complex tasks in the Earth-science domain and problems in task evaluation}

Most existing benchmarks emphasize question answering, factual recall, or reasoning over established knowledge. One major challenge is enabling LLMs to support scientific exploration and discovery~\cite{xu2025earthse}. In response, several works introduced small-sample, high-quality, in-domain reasoning datasets~\cite{guo2025earthlink,pmlr-v292-cardille25a}, which are better suited to evaluate genuine scientific reasoning.

In the majority of papers, models' intermediate reasoning is not evaluated, and tasks are scored end-to-end based on the final answer. This might be partially driven by the absence of reasoning chains in almost all Earth-science datasets (and where reasoning traces exist, they are often synthetically generated by LLMs~\cite{xu2025earthse,koksal2025samchat,ren2024watergpt}). This creates difficulties in improving models and explaining why they produce incorrect results. For example, a case study in the EarthSE evaluation suggested that low performance on math tasks stemmed from multi-step mistakes when applying formulas~\cite{xu2025earthse}. This diagnosis  would not have been possible without inspecting the models' reasoning steps. This issue is especially salient for complex, chained tasks that span multiple subtasks: models with stronger math or code generation may perform better end-to-end even if other models exhibit stronger domain-specific reasoning.

A related challenge is automatic evaluation. Many tasks are validated through pattern matching, typically for multiple-choice, true/false, or fill-in-the-blank formats, which is generally reliable when gold answers are unambiguous~\cite{xu2025earthse,li2025atmosscibench,koksal2025samchat,deng2023k2,ren2024watergpt,lu-2025-reasoning,chen2025geofactory,pmlr-v292-cardille25a}. However, for open questions and complex tasks, the dominant approach is LLM-as-judge evaluation~\cite{xu2025earthse,li2025atmosscibench,deng2023k2,ren2024watergpt,lu-2025-reasoning,chen2025geofactory}. Human (and especially expert) evaluation is comparatively rare and is usually conducted on small subsets~\cite{guo2025earthlink,deng2023k2}. In particular, even some high-quality benchmarks rely mainly on automated judging rather than expert assessment~\cite{pmlr-v292-cardille25a}.

The heavy reliance on automated judgment, combined with the limited evaluation of reasoning processes, can introduce systematic risks. LLM judges may encode preferences or biases (particularly for multi-criteria rubrics such as ``helpfulness''), and reasoning models may learn to optimize for those preferences rather than scientific correctness. In addition, models may produce plausible-looking answers that satisfy non-expert or LLM judges while remaining scientifically wrong. It is important to remember that expert human evaluation remains the highest-quality standard; otherwise, the purpose of creating complex, expert-designed tasks is undermined when their validation is largely synthetic.

\subsubsection{Effects of prompts in the Earth-science domain}

Prompt choices can strongly affect LLM performance in Earth-science tasks. Constraining outputs (e.g., ``Yes/No'' instead of open-ended answers) improves reliability for generalist models in remote-sensing settings~\cite{koksal2025samchat}. Likewise, end-to-end performance in agentic climate workflows was found to be strongly driven by user prompt quality~\cite{guo2025earthlink}.

Adding simple phrases to the instruction prompt can noticeably shift response quality across dimensions. CoT prompting (``let's think step by step'') improves inferential answers raising coherence, depth, and helpfulness, while slightly reducing performance on factual questions~\cite{chen2025geofactory}. In contrast, \textit{ExpertPrompting} (``You are an expert in geoscience'') strengthens the results of the factual-task~\cite{chen2025geofactory}. Combining \textit{ExpertPrompting} + CoT yields the best average performance in both factual and inferential tasks~\cite{chen2025geofactory}.

Surprisingly, moving from zero-shot to few-shot prompting did not consistently improve reasoning-model performance in the climate domain~\cite{lu-2025-reasoning}. Nonetheless, reasoning-optimized LLMs consistently outperformed non-reasoning models across all prompting strategies~\cite{lu-2025-reasoning}.

\subsubsection{Statistical significance in Earth-science reasoning evaluations}

A recurring limitation in current Earth-science LLM evaluations is the weak statistical support for reported gains. Many studies draw conclusions from small, high-quality benchmark sets, where the number of test elements is often too low to establish statistical significance~\cite{guo2025earthlink,koksal2025samchat,yu-etal-2025-sta,lu-2025-reasoning,pmlr-v292-cardille25a}. This is largely a byproduct of the field's shift away from massive factoid QA toward manually crafted, complex tasks, which are substantially more expensive to construct at scale.

Earth-science papers also rarely perform statistical analyses of how input patterns affect outcomes. As found in atmospheric math questions~\cite{li2025atmosscibench}, the need for more statistical tests may also arise because reasoning models can be vulnerable to symbolic perturbations. In that study, after abstracting formulas and randomly substituting valid numerical values, the accuracy of the LLM changed, suggesting reliance on pattern matching rather than genuine reasoning.

\subsubsection{Datasets evaluating reasoning models in Earth science}

Evaluation coverage across Earth-science subdisciplines remains imbalanced. Climate~\cite{guo2025earthlink,lu-2025-reasoning,manivannan2025climaqa}, ocean-focused studies~\cite{guo2025earthlink,xu2025earthse,ren2024watergpt}, and geospatial/remote-sensing work~\cite{koksal2025samchat,chen2025geofactory,pmlr-v292-cardille25a,geogpt2025r1preview} dominate the literature, while areas such as environmental change and biogeochemical cycles are comparatively under-represented.

High-quality Earth-science datasets exist, but they are often small because they rely on human curation or expert annotation with explicit validation and quality control~\cite{koksal2025samchat,yu-etal-2025-sta,manivannan2025climaqa}. Examples include expert-labeled remote-sensing imagery~\cite{koksal2025samchat}, multimodal benchmarks filtered to remove low-quality scenes~\cite{yu-etal-2025-sta}, and climate QA sets curated and checked by domain scientists~\cite{manivannan2025climaqa}. Many are also grounded in authoritative sources or standardized protocols --e.g., GIS textbook--derived tasks~\cite{yu-etal-2025-sta}, IPCC-based evidence--conclusion pairs preserving expert confidence judgements~\cite{lu-2025-reasoning}, and exam-style question suites~\cite{deng2023k2}. When synthetic generation is used, quality is typically maintained through expert verification or manual reconstruction with domain knowledge~\cite{ren2024watergpt,manivannan2025climaqa}.

In contrast, large-scale Earth-science datasets are typically built as broad domain corpora or instruction/QA resources, prioritizing coverage across disciplines and modalities (multi-sphere observations, climate-model ensembles, and scientific literature)~\cite{guo2025earthlink,xu2025earthse,ren2024watergpt,geogpt2025r1preview}. Examples include petabyte-scale Earth-system repositories that integrate many CMIP6 model outputs and major observations~\cite{guo2025earthlink}, and literature-derived corpora spanning $\sim$100k papers across many subfields with thousands of stratified QA items (Earth-Iron / Silver/Gold)~\cite{xu2025earthse}. Other large datasets emphasize model alignment and reasoning supervision, such as instruction-tuning collections distilled from millions of raw records (GeoSignal)~\cite{deng2023k2}, billion-token hydrology pretraining corpora plus sizeable evaluation suites (Water\_pretrain\_data; EvalWater)~\cite{ren2024watergpt}, and tens of thousands of chain-of-thought QA triples for geoscience reasoning (GeoGPT-CoT-QA)~\cite{geogpt2025cotqa}.

Across these resources, benchmarks vary widely in both task format and targeted capability. They range from mixed question formats (MCQ, cloze, true/false, free-response, dialogue) in EarthSE and ClimaQA to domain-specific exam and reasoning suites (GeoBench; AtmosSci-Bench) and instruction-tuning corpora spanning core NLP capabilities (NER, verification, summarization, classification, explanation, QA) in GeoSignal~\cite{xu2025earthse,li2025atmosscibench,deng2023k2,manivannan2025climaqa}. Some benchmarks explicitly test agentic or code-based workflows, for example, CBGB requires generating code to compute numerical geospatial outputs across diverse analysis problems, while GeoGPT-CoT-QA emphasizes step-by-step reasoning supervision at scale~\cite{pmlr-v292-cardille25a,geogpt2025cotqa}.

In terms of coverage, several datasets are intentionally broad: EarthSE spans five Earth spheres and many sub-disciplines, EarthLink integrates multi-sphere model and observational products, and GeoSignal/GeoBench assemble material across many geoscience topics and educational systems~\cite{guo2025earthlink,xu2025earthse,deng2023k2}. Others are broad within a single domain, such as EvalWater and Water\_pretrain\_data, which cover many categories of water-resources and sources for hydrology training and evaluation~\cite{ren2024watergpt}. Finally, many datasets are specialized to enable precise measurement in narrow competencies: SAMData targets missile-site recognition in remote sensing, shapefile tasks focus on canonical GIS operations, MineBench targets mineral exploration from multi-image evidence, and LLMClimateSynthesis targets IPCC-based evidence$\rightarrow$ consensus synthesis with confidence labeling; similarly, ClimaQA and EvalWater remain domain-focused on climate and hydrology, respectively~\cite{koksal2025samchat,yu-etal-2025-sta,ren2024watergpt,lu-2025-reasoning,manivannan2025climaqa}.

A particularly strong benchmark is EarthSE~\cite{xu2025earthse}, which combines multiple question formats (multiple-choice, fill-in-the-blank, true/false, and free-response) with 11 task types (Term Explanation, Knowledge QA, Fact Verification, Reasoning Analysis, Relation Extraction, Calculation, Research Tool Utilization, Literature Citation, Dataset, Experimental Design, and Code Generation) and broad coverage across five Earth spheres (hydrosphere, biosphere, lithosphere, atmosphere, cryosphere) spanning 114 sub-disciplines. Its main limitation is that these task types are evaluated largely in isolation rather than integrated into longer, chained workflows.

\subsubsection{Ambiguity of ``reasoning'' in Earth-science LLM research}

In Earth-science LLM publications, the term \emph{reasoning} is often used inconsistently and rarely refers specifically to reasoning-optimized LLMs. Instead, it is frequently conflated with (i) ``reasoning tasks (problems that are difficult or multi-step for humans, but do not by themselves demonstrate that an LLM performs explicit step-by-step reasoning) and (ii) agentic or multi-agent orchestration frameworks, where planning and tool use are handled by a controller or pipeline, even when the underlying model is not a reasoning-trained LLM. This ambiguity makes it harder to compare results across studies and to attribute performance gains to model reasoning versus system scaffolding.

\subsubsection{Smaller, fine-tuned models}

Smaller domain-adapted models show mixed results. In atmospheric-science tasks, a general reasoning model can outperform a dedicated domain model~\cite{li2025atmosscibench}. However, targeted domain adaptation can help. The domain-tuned K2 model surpasses similarly sized baselines, earns higher expert ratings for ``rationality'' and ``correctness,'' and demonstrates stronger domain tool usage~\cite{deng2023k2}.

\subsection{Identified Gaps}

Despite clear progress in Earth-science reasoning benchmarks and reasoning-optimized LLMs, several gaps remain in how these systems are tested, compared, and validated:

\begin{itemize}
    \item \textbf{Limited statistical testing.} Few studies test whether reasoning-model performance is stable under systematic perturbations of input data like numeric substitutions in formulas (including valid but extreme edge cases). This limits our ability to distinguish abstract problem solving from pattern matching.

    \item \textbf{Uneven performance analysis across subdomains.} More work is needed to measure reasoning-model gaps by Earth-science subdomain. For example, some results suggest substantially higher performance in certain categories (e.g., ``Reservoir and Water Conservancy Knowledge'' with 91.92) but much lower performance in others (e.g., ``Local Water Conservancy'' with 68.52 and ``Water Hydropower'' with 72.00)~\cite{ren2024watergpt}.

    \item \textbf{Lack of expert-authored reasoning traces.} Most datasets lack reasoning steps entirely, and when reasoning traces exist, they are often synthetically generated. There are very few datasets that include expert-written step-by-step solutions.

    \item \textbf{Minimal evaluation of intermediate reasoning and overreliance on automated judging.} Reasoning steps are rarely evaluated directly, while many benchmarks rely primarily on automatic scoring (including LLM-as-judge). Expert judgment is uncommon, although it is especially important for complex end-to-end tasks.

    \item \textbf{Insufficient evaluation by task complexity across subdomains.} There is limited coverage of clearly defined task suites stratified by complexity across a broad range of subdomains, which would enable more systematic comparisons of model capabilities at different difficulty levels.

    \item \textbf{Lack of chain-of-tasks / ``super-task'' benchmarks.} Current datasets rarely evaluate long, expert-level workflows that require sustained reasoning over multiple subtasks, varied task types, and multiple modalities. Developing such ``super-task'' benchmarks would better measure advanced Earth-science reasoning capabilities.
\end{itemize}

\subsection{Notable domain-specific tasks that have been addressed by reasoning models}

\paragraph{Question answering.}
Test-like QA (MCQ/cloze/true--false) is common in EarthSE, AtmosSci-Bench, ClimaQA, and WaterGPT and is often where reasoning models outperform instruction-tuned, math-augmented, and domain-specific baselines~\cite{li2025atmosscibench,deng2023k2,ren2024watergpt}. Several benchmarks also include open-ended QA (e.g. EarthSE Earth-Gold and AtmosSci-Bench)~\cite{li2025atmosscibench,chen2025geofactory}.

\paragraph{Conducting and assisting in research.}
\begin{itemize}
  \item \textbf{Autonomous research workflows:} The EarthLink multi-agent system automates climate-science research from planning to code generation, data analysis, and visualization; its agents translate natural-language queries into experimental plans, write custom Python scripts, iteratively debug, and produce scientific summaries~\cite{guo2025earthlink}. Expert evaluation scores show that EarthLink's workflows are comparable to junior researchers in many tasks~\cite{guo2025earthlink}.
  \item \textbf{Assisting in research tasks:} EarthLink not only designs complete research workflows but can also generate specialized code (e.g., wavelet analysis, hierarchical emergent constraints) and analyze data to propose new hypotheses~\cite{guo2025earthlink}. In hydrology, WaterGPT's Water\_Agent integrates multiple tools for text and image processing and has $>90\%$ success on tasks like object detection and water-body extraction, demonstrating its usefulness as a research assistant~\cite{ren2024watergpt}.
  \item \textbf{Assigning confidence to domain statements:} In the study of synthesis of climate-statements, reasoning LLMs were assigned to assign confidence levels to IPCC statements; they correctly classified the confidence of the statement and outperformed general LLMs by about 8 percentage points~\cite{lu-2025-reasoning}.
  \item \textbf{Summarizing complex climate evidence:} The same study required LLMs to produce concise factual summaries of sets of climate evidence. Reasoning models generated coherent summaries, although performance varied across metrics~\cite{lu-2025-reasoning}.
\end{itemize}

\paragraph{Solving domain-related equations.}
Reasoning models perform well in domain-specific mathematical problem solving. On AtmosSci-Bench, reasoning variants score higher on hydrology, atmospheric dynamics, and geophysics than instruction-, math-, or domain-specific baseline models~\cite{li2025atmosscibench}. On the Cloud-Based Geospatial Benchmark, agentic systems that use reasoning chains with iterative error correction achieve the best accuracy on code-based numerical geospatial calculations~\cite{pmlr-v292-cardille25a}.

\paragraph{Image analysis.}
\begin{itemize}
  \item \textbf{Satellite image classification} SAMChat achieves high precision/recall for distinguishing military vs.\ civilian sites in satellite imagery~\cite{koksal2025samchat}.
  \item \textbf{Mineral exploration:} STA-CoT uses planner--executor--verifier agents for multi-image geological reasoning, improving consistency on deposit localization (notably on harder MineBench subsets)~\cite{yu-etal-2025-sta}.
  \item \textbf{Hydrology from imagery:} WaterGPT and related agents analyze satellite images to detect water bodies and extract hydrological features with $>90\%$ accuracy~\cite{ren2024watergpt}.
\end{itemize}

\paragraph{Code generation.}
Reasoning-based systems can generate domain-specific code for Earth-science workflows. In EarthLink, a coding agent converts experimental plans into custom Python diagnostics using templates, documentation retrieval, and iterative debugging~\cite{guo2025earthlink}. On the Cloud-Based Geospatial Benchmark (Earth Engine code), reasoning agents with feedback loops reach $\sim 71\%$ accuracy and outperform non-reasoning baselines~\cite{pmlr-v292-cardille25a}.

\paragraph{Creating visualizations.}
EarthLink automatically generates analysis plots as part of its experimental workflow and uses image-analysis agents to check the quality of the visualization and detect patterns or anomalies before producing the final scientific summary~\cite{guo2025earthlink}.

\paragraph{Domain tool usage.}
Reasoning models often improve factual accuracy by calling external tools. K2 explicitly benchmarks tool use in geoscience through GeoBench~\cite{deng2023k2}. EarthLink combines web search and retrieval for templates with iterative debugging of tool calls~\cite{guo2025earthlink}. WaterGPT's Water\_Agent uses domain tools for text, imagery, and geospatial processing in hydrology~\cite{ren2024watergpt}. 

\subsection{Domain-specific tasks that have not been addressed by reasoning models}

Although recent advancements in reasoning systems have begun to address specialized tasks across Earth-science domains, these efforts still focus mostly on static question answering, code generation, and limited multimodal analysis. Yet, the current literature and benchmarks suggest that important Earth-system workflows remain either under-tested or underused. Underrepresented task classes in the domain include:

\begin{itemize}
  \item \textbf{Temporal tasks:} Current Earth-science reasoning benchmarks rarely evaluate models on predictive tasks, where performance depends on reasoning over time-series data, evolving feedbacks, and uncertainty. Two major undercovered branches are:
  \begin{enumerate}
    \item \textbf{Extreme-event forecasting} (e.g., floods, droughts, heatwaves, wildfires).
    \item \textbf{Spatiotemporal forecasting and planning} (e.g., streamflow or air-quality forecasting, subseasonal outlooks, adaptive sampling over time). Even when models are tested on spatial tasks (e.g., route planning, place-name recognition), the temporal dimension and event prediction are typically absent.
  \end{enumerate}

  \item \textbf{Local knowledge:} Most benchmarks and evaluations probe general or regional Earth-science knowledge, with only limited coverage of locality-specific tasks (e.g., place-name recognition and simple route planning). As a result, reasoning models are rarely tested on local knowledge, such as site-specific hydrology and geology, local regulations and practices, or Indigenous and community observations. Testing locality-specific knowledge would provide a more credible assessment of domain reasoning, because there is a much lower chance that the relevant details (and their answers) were already present in the model's training data.

  \item \textbf{Verified scientific reasoning:} Most Earth-system science studies treat reasoning as an intermediate scaffold toward an answer, with little step-level validation. More work is therefore needed on Earth-science tasks where the reasoning trace itself is the evaluated output, with explicit criteria for step-level validity rather than judging performance only by the final answer.
\end{itemize}

\subsection{Comments on the usage of reasoning model in RAG-like system or agentic framework}

Many recent Earth-system science AI papers adopt agentic workflows, for example a self-evolving climate-science agent system~\cite{guo2025earthlink}, a hydrology expert LLM agent framework~\cite{ren2024watergpt}, and structured target-centric multi-image geological reasoning (STA-CoT)~\cite{yu-etal-2025-sta}. These systems often follow a reusable Planner $\rightarrow$ Executor $\rightarrow$ Verifier pattern: the planner decomposes the task, the executor performs concrete actions over data (e.g., running code or operating on images), and an optional verifier checks intermediate results and iterates fixes; notably, STA-CoT reports ablations indicating that removing the executor/verifier reduces reliability/consistency~\cite{yu-etal-2025-sta}. In practice, the executor's actions are frequently implemented via tool calling -- STA-CoT invokes tools such as coordinate maps and image cropping~\cite{yu-etal-2025-sta}, while the hydrology agent framework develops analysis code and packages it as callable tools for deployment~\cite{ren2024watergpt}.

We found that RAG systems are used in many Earth-system science papers as a way to ground model outputs in external domain sources, most notably ClimaQA~\cite{manivannan2025climaqa}, STA-CoT~\cite{yu-etal-2025-sta}, and the Cloud-Based Geospatial Benchmark~\cite{pmlr-v292-cardille25a}. The main motivation is that retrieval can supply specialized factual context and code snippets that the model may not reliably recall, improving grounding and helping with domain-specific QA or code generation. However, multiple studies emphasize that RAG can also inject noise: ATMOSSCI-BENCH finds na\"{i}ve retrieval often fails and even curated retrieval yields only modest improvements~\cite{li2025atmosscibench}, ClimaQA shows performance can drop when retrieving from less relevant books~\cite{manivannan2025climaqa}, and STA-CoT notes that noisy injected information can degrade results, motivating careful curation and/or verification of retrieved context~\cite{yu-etal-2025-sta}.

\subsection{Number of datasets for training domain-specific reasoning model found}

\begin{itemize}
    \item \textbf{All: between 4 and 6} \\
    (GeoSignal~\cite{deng2023k2}, GeoTool~\cite{deng2023k2}, SAMData~\cite{koksal2025samchat}, Water\_pretrain\_data~\cite{ren2024watergpt}, Water Conservancy--SFT~\cite{ren2024watergpt}, GeoBench~\cite{deng2023k2})

    \item \textbf{Publicly accessible: between 1 to 3} \\
    (GeoSignal~\cite{deng2023k2}, GeoBench~\cite{deng2023k2}, SAMData~\cite{koksal2025samchat})
\end{itemize}

\subsection{Number of benchmarks/datasets for evaluating domain-specific reasoning model found}

\begin{itemize}
    \item \textbf{All: many ($>$ 6)} \\
    (EarthSE~\cite{xu2025earthse}, AtmosSci--Bench~\cite{li2025atmosscibench}, MineBench / MineBench--Hard~\cite{yu-etal-2025-sta}, LLMClimateSynthesis~\cite{lu-2025-reasoning}, GeoTask~\cite{chen2025geofactory}, Cloud--Based Geospatial Benchmark~\cite{pmlr-v292-cardille25a}, GeoSignal~\cite{deng2023k2}, GeoBench~\cite{deng2023k2}, SAMData~\cite{koksal2025samchat}, EvalWater~\cite{ren2024watergpt})

    \item \textbf{Publicly accessible: many ($>$ 6)} \\
    (EarthSE~\cite{xu2025earthse}, AtmosSci--Bench~\cite{li2025atmosscibench}, MineBench~\cite{yu-etal-2025-sta}, MineBench--Hard~\cite{yu-etal-2025-sta}, GeoBench~\cite{deng2023k2}, GeoTask~\cite{chen2025geofactory}, EvalWater~\cite{ren2024watergpt}, Water Conservancy--SFT~\cite{ren2024watergpt})
\end{itemize}

\subsection{Number of domain-specific reasoning models found}

\begin{itemize}
    \item \textbf{All: between 4 and 6} \\
    (SAMChat~\cite{koksal2025samchat} for remote-sensing, K2~\cite{deng2023k2} for geoscience, WaterGPT~\cite{ren2024watergpt} for hydrology, GeoFactory~\cite{chen2025geofactory}, GeoGPT-CoT-QA~\cite{geogpt2025cotqa}, GeoGPT-R1-Preview~\cite{geogpt2025r1preview} for geoscience)

    \item \textbf{Publicly accessible: between 4 and 6} \\
    (SAMChat~\cite{koksal2025samchat}, K2~\cite{deng2023k2}, GeoFactory~\cite{chen2025geofactory}, GeoGPT-CoT-QA~\cite{geogpt2025cotqa}, GeoGPT-R1-Preview~\cite{geogpt2025r1preview})
\end{itemize}

\subsection{Number of methods for creating/using domain-specific reasoning models found}

\begin{itemize}
    \item \textbf{All: between 4 and 6} \\
    SFT/RL: SAMChat~\cite{koksal2025samchat}; \\
    pretraining, instruction-tuning: WaterGPT~\cite{ren2024watergpt}, K2~\cite{deng2023k2}; \\
    others (agent orchestration, RAG): EarthLink~\cite{guo2025earthlink}, STA--CoT~\cite{yu-etal-2025-sta}, GeoFactory~\cite{chen2025geofactory}

    \item \textbf{Publicly accessible: between 1 to 3} \\
    (SAMChat~\cite{koksal2025samchat}, GeoFactory~\cite{chen2025geofactory})
\end{itemize}

\section[Literature Review Summary: PE11]{Literature Review Summary: PE11}

\textbf{Materials Engineering.} Advanced materials development: performance enhancement, modelling, large-scale preparation, modification, tailoring, optimisation, novel and combined use of materials, etc.

\subsection{Key Findings}

The literature reveals a transition from general-purpose LLMs to domain-specific reasoning models designed to bridge the gap between vast scientific text and practical laboratory protocols.

\begin{itemize}
    \item \textbf{Datasets \& Benchmarks}: Key datasets include MatterChat (142k structures) \citep{tang2025matterchat}, MaScQA (650 expert-curated questions) \citep{zaki2023mascqa}, and CURIE (580 long-context problems) \citep{cui2025curie}.
    \item \textbf{Reasoning Models}: Models like MatChat (fine-tuned LLaMA2) \citep{chen2023matchat}, MatterChat (multi-modal graph-based) \citep{tang2025matterchat}, and LLM-Feynman (symbolic regression integration) \citep{song2025llm} are being used with CoT prompting techniques. However, there is a lack of domain-specific models with explicit reasoning capability.
    \item \textbf{Modalities}: Works utilize text-only, text-image, atomic graphs, and structured CIF/SMILES strings.
    \item \textbf{Prompting \& Outputs}: Chain-of-Thought (CoT) prompting is standard for generating step-by-step synthesis rationales and chemical equations. Outputs often require JSON/YAML formatting for structured data extraction or LaTeX/Python for engineering formulas.
    \item \textbf{Validation}: Reasoning is validated via human expert review, or automated benchmarks (e.g., MoleculeNet) and statistical tests like Mann-Whitney U  \citep{zheng2023large}.
    \item \textbf{Efficiency}: Some works consider efficiency through low-parameter models (1.7B–8B) \citep{chen2023matchat, tang2025matterchat} or merged models via SLERP \citep{lu2025fine} to unlock emergent reasoning without training from scratch.
\end{itemize}

\subsection{Identified Gaps}
\begin{itemize}
    \item Lack of Physical Validation: A critical absence of "wet-lab" experimental confirmation for AI-generated synthesis pathways across all reviewed papers.
    \item Scalability \& Industrial Manufacturing: Existing models focus on molecular properties or lab-scale "recipes" but ignore the engineering challenges of large-scale preparation and industrial process optimization.
    \item 3D Structural Limitations: Standard text-based models (SMILES/CIF) often fail to ground 3D geometric symmetries and crystallographic selection rules.
    \item Numerical Unreliability: Models struggle with complex unit conversions and physical constants (e.g., Avogadro number) in numerical engineering problems \citep{miret2024enabling}.
\end{itemize}

\subsection{Notable domain-specific tasks that have been addressed by reasoning models}

\begin{itemize}
    \item Synthesis Pathway Prediction: Automating the extraction of precursors and heating conditions for inorganic materials \citep{chen2023matchat}.
    \item Property-Design Rule Inference: Identifying "second-order" features from literature to predict quantum mechanical behaviors \citep{zheng2023large}.
    \item Symbolic Formula Discovery: Deriving interpretable engineering formulas for ionic conductivity and bandgaps \citep{song2025llm}.
    \item Bio-inspired Hypothesis Generation: Using multi-agent graphs to design high-strength silk composites \citep{lu2025fine}.
\end{itemize}

\subsection{Domain-specific tasks that have not been addressed by reasoning models}

\begin{itemize}
    \item Real-time Process Monitoring: Models cannot yet handle dynamic modification of experimental plans during synthesis (e.g., adjusting stirring in real-time).
    \item Heterostructures \& Composites: Most models focus on single inorganic phases rather than the combined use of complex materials in heterostructures.
    \item Environmental \& Policy Impact: There is no focus on the environmental costs, toxicity, or policy implications of scaling new materials.
\end{itemize}

\subsection{Comments on the usage of reasoning model in RAG-like system or agentic framework}

Frameworks like SciAgents \citep{ghafarollahi2024sciagents} demonstrate that multi-agent orchestration (using agents as "critics" or "explorers") can uncover interdisciplinary connections that single models miss. CURIE \citep{cui2025curie} highlights the potential for "agentic" workflows to interact with specialized toolkits like ASE for automated DFT simulations. Polymer-Agent \citep{Nigam2026PolymerAgent} uses a reasoning model such as Gemini 3.0 Pro to interpret natural language queries, calls the appropriate tools to generate or predict polymer properties, and maintains a "human-in-the-loop" workflow via an integrated terminal in the task of polymer design.

\subsection{Other comments}

The literature suggests that data quality and distillation are more critical for material modeling than raw volume. Additionally, a "synergistic effect" is noted where multi-modal models outperform their standalone physical-modeling branches.

\subsection{Number of datasets for training domain-specific reasoning model found}

\begin{itemize}
    \item \textbf{All}: between 4 and 6

    \item \textbf{Publicly accessible}: between 4 and 6
\end{itemize}

\subsection{Number of benchmarks/datasets for evaluating domain-specific reasoning model found}

\begin{itemize}
    \item \textbf{All}: between 4 and 6

    \item \textbf{Publicly accessible}: many ($>$ 6)
\end{itemize}

\subsection{Number of domain-specific reasoning models found}

\begin{itemize}
    \item \textbf{All}: between 1 to 3

    \item \textbf{Publicly accessible}: between 1 to 3
\end{itemize}

\subsection{Number of methods for creating/using domain-specific reasoning models found}

\begin{itemize}
    \item \textbf{All}: between 4 and 6

    \item \textbf{Publicly accessible}: between 4 and 6
\end{itemize}

\section[Literature Review Summary: LS1]{Literature Review Summary: LS1}

\textbf{Molecules of Life: Biological Mechanisms, Structures and Functions.} Molecular biology, biochemistry, structural biology, molecular biophysics, synthetic and chemical biology, drug design, innovative methods and modelling.

\subsection{Key Findings}

\subsubsection{Domain-specific models:} 

The landscape of biological reasoning models has recently shifted from standard supervised learning to reinforcement learning frameworks, most notably the Group Relative Policy Optimization (GRPO) method. This approach enables models to internalize biological rules, such as biophysical constraints or signaling logic, through explicit reasoning traces. Current models can be broadly categorized according to their architectural strategies and the specific LS1 sub-domains they address.

\paragraph{Genomic and Multi-Omic Sequence Understanding}

Models in this category focus on DNA and RNA biology (LS1\_3), aiming to bridge the gap between raw genomic sequences and functional annotations.

\textbf{BioReason} \cite{fallahpour2025bioreason} represents a hybrid architectural approach by integrating contextualized DNA embeddings from frozen foundation models, such as Evo2 or Nucleotide Transformer, into an LLM backbone (Qwen3 1.7B and 4B). This integration is achieved via a learnable linear projection layer. To ensure the model does not merely memorize patterns but understands the underlying biophysics, it is trained via SFT followed by GRPO to generate step-by-step biological reasoning traces within \texttt{<think>} tokens. This approach enables the model to derive variant pathogenicity predictions directly from raw genomic sequences by reasoning through molecular mechanisms.

Similarly, \textbf{ChatMultiOmics} addresses the need for broad multi-omics understanding. Developed by fine-tuning Llama-3.1-8B-Instruct on the extensive \textit{Biology-Instructions} dataset \cite{he2025biologyinstructionsdatasetbenchmarkmultiomics}, it utilizes a sophisticated three-stage pipeline: (1) unsupervised pre-training on multi-omics sequences, (2) instruction tuning across 21 diverse biological tasks, and (3) reasoning instruction tuning. 

\paragraph{Protein Biology and Cellular Perturbations}

The study of protein interactions (LS1\_1) and molecular signaling (LS1\_9) has seen the rise of models that simulate biological "thought."

\textbf{ProLLM} specializes in Protein-Protein Interaction (PPI) prediction by moving beyond text-only representations \cite{jin2024prollm}. The model utilizes an "embedding replacement" technique, whereby standard text tokens are substituted for high-dimensional vectors from ProtTrans to capture three-dimensional structural information. The framework introduces \textit{ProCoT} (Protein Chain of Thought), which directs the LLM to transform interaction data into natural language chains that simulate cellular signaling pathways. This method has been demonstrated to be particularly effective for identifying indirect interactions within complex macromolecular complexes.

In the domain of cellular perturbation (LS1\_2), \textbf{rbio1} \cite{istrate2025rbio1} utilizes biological world models as verifiers. By post-training a Qwen2.5-3B-Instruct model using GRPO, the authors derived rewards from Virtual Cell Models (VCMs) that act as "soft" and "hard" verifiers. This approach ensures that the model's reasoning traces regarding gene expression changes are grounded in established transcriptomic realities. Complementing this, \textbf{SynthPert} \cite{phillips2026synthpert} focuses on knowledge distillation. It uses a frontier model (OpenAI o4-mini) to generate high-quality synthetic reasoning traces for mechanistic biological tasks. These traces, filtered by a "judge" model, are used to fine-tune a smaller DeepSeek-R1 distilled Llama-3.1-8B via LoRA, demonstrating that complex biological reasoning can be distilled from large-scale general models into efficient domain-specific models.

\paragraph{Small Molecules and Drug Design}

For tasks related to chemical biology (LS1\_11) and drug design (LS1\_13), reasoning models are being used to enhance interpretability and retrosynthetic planning.

\textbf{DrugReasoner} integrates chemical informatics with LLM logic \cite{ghaffarzadehesfahani2025drugreasonerinterpretabledrugapproval}. It employs MOLFORMER for SMILES embeddings and XGBoost to identify similar approved and unapproved molecules. The model is fine-tuned to acquire reasoning capabilities through the implementation of GRPO, incorporating five components within the reward function to ensure interpretability and confidence-alignment. This enables the model to generate a rationale that compares the physicochemical properties of a query drug candidate to its closest neighbors, thereby providing a transparent decision-support tool for predicting drug approval. 

\textbf{BioMedGPT-Mol} further advances molecular engineering by introducing a multi-task learning strategy with specialized tokens (e.g., \texttt{<SMILES>}, \texttt{<IUPAC>}) \cite{zuo2025biomedgptmolmultitasklearningmolecular}. This model supports controllable reasoning behaviors,  enabling users to alternate between "thinking" and "non-thinking" modes for tasks such as molecular optimization and multi-step retrosynthetic planning. A three-stage SFT strategy was implemented on Qwen3-8B, and subsequently, GRPO was employed to refine its multi-step planning logic.

\paragraph{Knowledge-Augmented and Generalist Frameworks}

Finally, some models focus on factual grounding and cross-task generalization within the LS1 knowledge panel.

\textbf{Bio-KCoT} addresses the risk of hallucination in complex biomolecular reasoning \cite{lyu2026knowledgeaugmented}. The proposed method is a knowledge-augmented long-Chain-of-Thought (CoT) approach that integrates structured biological knowledge retrieved through multi-hop traversal and pruning of Knowledge Graphs (KG). By fine-tuning Qwen3 variants on these curated chains, the model achieves superior factual grounding in signaling and biochemical tasks.

\textbf{OwkinZero} explores the "specialist vs. generalist" trade-off \cite{bigaud2025owkinzeroacceleratingbiologicaldiscovery}. By post-training Qwen3-8B and 32B models on a curated mixture of biological datasets using a modified GRPO, the authors found that models trained on a mixture of tasks significantly outperformed single-task specialists. This suggests that the reasoning required for one LS1 task (e.g., protein design) often reinforces the logical foundations needed for another (e.g., variant effect prediction), thereby resulting in a significant generalization effect in the context of biological reasoning.

\subsubsection{What models are used for development of domain-specific reasoning models?}
\paragraph{Reasoning}
Most of works utilize the open-source Qwen3 model and its variants (1.7B, 4B, 8B, 32B) for development of domain-specific reasoning models. One work used the Qwen2.5-3B-Instruct model. 

\paragraph{Non-reasoning}
Also the Llama3.1-8B-Instruct was used as a base model a few times with one usage of DeepSeek-R1 distilled Llama3.1-8B. One article fine-tuned the gpt-3.5-turbo-1106 model via OpenAI's API to achieve an ability of generating step-by-step reasoning traces.

\subsubsection{What other models were evaluated in literature?}
According to the literature review the subsequent general-purpose and domain-specific language models were evaluated on biological reasoning tasks:
\paragraph{Reasoning}
\begin{itemize}
\item \textbf{Zero-Shot Cell Type Annotations article}: DeepSeek-R1-0528, DeepSeek-V3-0324 (reasoning-enhanced),  

\item \textbf{Bio-KCoT article}: Deepseek-R1, o3-mini, o1, Gemini 2.5 Pro

\item \textbf{LiveProteinBench article}: Gemini 2.5 Pro, Claude 3.7 Sonnet, Deepseek-r1, Qwen3-32B, GPT-5, GPT-o3

\item \textbf{Genie-CAT article}: GPT-5-mini - medium reasoning effort

\item \textbf{CoTox article}: GPT-o3, DeepSeek-R1, Qwen3-32B, Gemini-2.5-Pro

\item \textbf{OwkinZero article}: GPT-o3, DeepSeek-R1

\item \textbf{Sparks article}: GPT-o3 (o3-2025-04-16), GPT-o3-mini (o3-mini-2025-01-31) used as hypotheses generation agents in multi-agent system.

\end{itemize}

\paragraph{Non-reasoning}
\begin{itemize}
\item \textbf{ChatMultiOmics article}: GPT-4o, GPT4o-mini, LaMA3.1-8B-Instruct, Qwen2-7B, LLama2-7B, Alpaca-7B, ChatGLM4, Vicuna, Galactica-1.3B, InstructProtein-1.3B, llama-molinst-protein-7B (Mol-Ins), BioMedGPT-LM-7B

\item \textbf{Bio-KCoT article}: GPT-4o-mini, GPT-4o

\item \textbf{LiveProteinBench article}: Deepseek-v3, Gemini 2.0 Flash, Qwen2.5-72B, Qwen2.5-32B, GPT-4o, Qwen2.5-VL-32B, InternVL3-78B, Llama3.3-70B, BioMedGPT-R1, EvoLlama, Evolla

\item \textbf{BioMedGPT-Mol article}: GPT-4, GPT-4o, Galactica, Claude3 Opus, Claude3.5, Gemini-1.5-pro, GPT-4-turbo, Mistral, Molinst, ChemLLM, LlaSMol, Mol-T5, Uni-Mol, Bio-T5, OpenMolIns-125M, OpenMolIns-8B, GeLLM3O-mistral, GeLLM3O-llama

\item \textbf{CoTox article}: GPT-4o, Llama3.1-8B, Llama3.1-70B, TxGemma-9B-Chat

\item \textbf{OwkinZero article}: Qwen2.5-7B, GPT-4o, MedGemma-27B

\item \textbf{Sparks article}: GPT-4.1 + CoT - used as reflection agents in multi-agent system.  

\end{itemize}

\subsubsection{Modalities}
Modalities used in existing works:
\begin{itemize}
    
\item \textbf{Text-only}: Most models operate primarily on text, processing biological sequences (protein, DNA, RNA) as text strings (e.g., SMILES, amino acid sequences).

\item \textbf{Text + Genomic Sequences}: BioReason explicitly integrates raw genomic DNA sequences via a learnable projection layer into the LLM's text embedding space.

\item \textbf{Text + Image}: The authors in LiveProteinBench article utilized a multimodal evaluation by implementing a Structure Projection Process to generate structural images for each protein in collected dataset.
\end{itemize}

\subsubsection{How people prompt the model, what kind of outputs they expect?}
\paragraph{Prompting Strategies}

\begin{itemize}
     
    \item \textbf{Chain-of-Thought (CoT)}: Widely used to elicit reasoning.

    \item \textbf{Classifier Mode}: Constraining the model to choose from a predefined list of labels (e.g., cell types) to improve robustness \cite{Wang2025BioReasoning}.

    \item \textbf{Multi-Agent/Expert Chaining}: Breaking tasks into subtasks handled by specialized agents or "experts" (e.g., Sparks uses Scientist and Coder agents).
\end{itemize}

\paragraph{Outputs}
\begin{itemize}
\item \textbf{Structured Outputs}: JSON formatting is common for data storage and exchange.

\item \textbf{Open-ended vs. Closed}: Outputs range from open-ended reasoning traces (for hypothesis generation) to closed-set classifications (e.g., "benign" vs. "pathogenic" in VEP tasks , cell type labels).

\item \textbf{Validation}: Reasoning is validated by comparing final answers to ground truth (accuracy, F1-score) , using "soft verifiers" (biological models) , or through manual expert review of reasoning traces.

\end{itemize}

\subsubsection{Whether existing works consider aspects like reasoning length/effort as well as cost/efficiency?}
\begin{itemize}
\item \textbf{Cost/Efficiency}: High token consumption is a noted drawback. For example, DeepSeek-R1 uses 5-8x more tokens than non-reasoning models for cell annotation.

\item \textbf{Reasoning Length}: BioReason incentivizes conciseness via reward functions to prevent "loopholes".

\item \textbf{Efficiency Measures}: SynthPert demonstrates that synthetic reasoning traces allow for high performance with only 2\% of training data, highlighting data efficiency.

\end{itemize}

\subsubsection{Datasets \& Benchmarks}
\begin{itemize}

\item \textbf{Perturbation prediction reasoning traces}: in the SynthPert article \cite{phillips2026synthpert} the authors created synthetic reasoning traces using a frontier model (OpenAI o4-mini) to enhance LLM performance on mechanistic biological explanation tasks. The traces were validated by both LLM judge and domain experts to ensure a high quality of the dataset. This dataset could be used for validation of novel domain-specific reasoning language models.

\item \textbf{KEGG-Derived Biological Reasoning Dataset (BioReason)}: This dataset \cite{fallahpour2025bioreason} contains causal reasoning paths (averaging ~303.8 words in length) that connect genetic variants to disease phenotypes. These reasoning traces were synthetically generated using Claude 3.7 Sonnet and grounded in molecular interaction data from the KEGG database. The dataset is hosted publicly on HuggingFace.

\item \textbf{Knowledge-Augmented Reasoning Chains (Bio-KCoT Dataset)}: Introduced in the Bio-KCoT article \cite{lyu2026knowledgeaugmented}, this dataset consists of high-quality, mechanistic reasoning traces used for supervised fine-tuning (SFT) and reinforcement learning (GRPO). The reasoning paths were generated through a specialized framework that performs guided multi-hop traversal and pruning on the PrimeKG knowledge graph. This process removes redundant or spurious biological connections, resulting in reasoning chains that are both biologically meaningful and logically complete. These chains connect complex biomolecular entities (e.g., genes, proteins, diseases) via explicit molecular interactions and signaling cascades to ensure that the model’s reasoning is factually grounded.

\item \textbf{PerturbQA}: A primary benchmark for perturbation prediction \cite{wu2025perturbqa}.

\item \textbf{VEP (Variant Effect Prediction) Datasets}: Includes VEP-Coding (28,882 entries) and VEP-Non-SNV (50,083 QA pairs) for evaluating variant pathogenicity.

\item \textbf{Biology-Instructions Dataset}: A massive collection of over 3 million training samples across 21 tasks, including antibody-antigen neutralization and RNA-protein interaction \cite{he2025biologyinstructionsdatasetbenchmarkmultiomics}.

\item \textbf{Protein-Protein Interaction (PPI) Benchmarks}: Includes Human, STRING, SHS27k, and SHS148k datasets for evaluating interaction predictions. Datasets were used in the ProLLM article \cite{jin2024prollm}.

\item \textbf{Single-cell Benchmarks}: Includes scTab, CellxGene4 (out-of-distribution), and specific cluster/cell-level datasets for cell type annotation.
Datasets used in Zero-Shot Cell Type Annotations article \cite{Wang2025BioReasoning}.
\item \textbf{DNA-LLM Datasets}: Synthetic data generated by NUPACK for DNA structure prediction and design \cite{ross2024chainingthoughtsllmslearn}. 10,000 / 1,000 split; sequence lengths 10-25; includes mismatched bases (up to 30\%). Although the dataset is not available to download, the authors provided a generation script that was used to create the data.

\item \textbf{Drug Approval}: Dataset for drug approval benchmarking introduced in Drug Reasoner article \cite{ghaffarzadehesfahani2025drugreasonerinterpretabledrugapproval}.
A subset of 2,255 approved and 2,255 unapproved molecules was created by random undersampling of the
unapproved class. The SMILES structure was retrieved, and strings were canonicalized using
RDKit (version 2023.9.5).

\item \textbf{PrimeKGQA Benchmark}: A comprehensive benchmark for biomolecular question answering containing 6,710 QA pairs that capture multi-level semantic relationships and varying reasoning difficulties. The benchmark was introduced in the Bio-KCot article \cite{lyu2026knowledgeaugmented}.

\item \textbf{LiveProteinBench}: The first contamination-free, multi-task, and multimodal benchmark for protein science \cite{rong2026liveproteinbench}. An automated update pipeline scans databases like UniProt for new entries post-2025 and extracts validated annotations to form tasks. Only entries released after 1st Jan 2025 are taken into account to prevent data leakage and ensure that all evaluated models have not previously seen the test data.

\item \textbf{Multi-task benchmark}: Curation of a comprehensive, high-quality training dataset and benchmark by unifying previously scattered public datasets like SMolInstruct, TOMG-Bench, and MuMOInstruct. The benchmark was introduced in the BioMedGPT-Mol article \cite{zuo2025biomedgptmolmultitasklearningmolecular}. \textbf{However, no public code nor data is available so this benchmark is not yet available for use (last update 11 Dec 2025).}

\item \textbf{UniTox - drug-induced toxicity dataset}: a unified dataset of 2,418 FDA-approved drugs with drug-induced toxicity summaries and ratings created by using GPT-4o to process FDA drug labels.

\item \textbf{OwkinZero benchmark} - introduction of a new benchmark of eight biological datasets ($>300k Q\&A pairs$), spanning transcriptomics, structural biology, and perturbation assays \cite{bigaud2025owkinzeroacceleratingbiologicaldiscovery}.

\end{itemize}

\subsection{Identified Gaps}
Despite rapid progress, several critical gaps persist in the current literature:

\begin{itemize}

\item \textbf{Sparsity of biological reasoning datasets}: Most of the current datasets and benchmarks do not include a "golden" reasoning traces that could be used for validation of reasoning process in language models.

\item \textbf{Multi-Modal \& Spatial Integration}: Current models largely treat biology as 1D sequences. There is a distinct lack of integration for 3D genomic architecture (e.g., Hi-C data) and spatial transcriptomics, which are crucial for understanding cellular organization. Furthermore, 'omics' layers beyond the central dogma (DNA/RNA/Protein) are rarely integrated. Specifically, lipidomics—which relies on high-dimensional mass spectrometry signals rather than linear sequences—represents a major modality gap. While specialized AI exists for lipid identification, the integration of these signals with LLM reasoning traces has the potential to facilitate more comprehensive modeling of cellular organization and signaling crosstalk.

\item \textbf{Fact-Checking \& Hallucination}: A major challenge is the "decoupling" of reasoning from the final answer, where a model may produce a correct answer with flawed logic. There is currently no robust, large-scale system for automated biological fact-checking to detect hallucinations in reasoning traces.

\item \textbf{Generalization to "Dark Matter"}: Models excel at well-characterized pathways (e.g., KEGG) but struggle with the genomic "dark matter"—non-coding regions and non-model organisms where data is scarce and functional annotations are missing.

\item \textbf{Closing the Loop}: Perhaps the most significant gap is the lack of wet-lab validation. Some papers generate "testable hypotheses" but do not physically validate them, leaving a disconnect between in silico predictions and in vivo reality.

% \item \textbf{Future Research Directions}: To address these, future work should focus on "closed-loop" systems where AI agents drive robotic experimentation, developing "Graph-of-Thoughts" approaches to model non-linear biological networks, and creating distilled reasoning models to reduce token costs while maintaining depth of thought.
\end{itemize}

\subsection{Notable domain-specific tasks that have been addressed by reasoning models}
Reasoning models have been successfully applied to a diverse range of biological problems, often outperforming traditional methods by utilizing explicit "thought" traces to model complex mechanisms:

\begin{itemize}
    \item \textbf{Genetic Perturbation Prediction}: Models such as rbio1 \cite{istrate2025rbio1} and SynthPert \cite{phillips2026synthpert}have set new baselines in predicting the outcomes of genetic edits (e.g., PerturbQA). These models go beyond simple sequence-to-expression mapping by using explicit reasoning to explain why a perturbation leads to up- or down-regulation in specific cell lines like K562 or Jurkat. SynthPert, in particular, demonstrated that synthetic reasoning traces can distill biological principles into smaller models (e.g., Llama-3.1-8B), achieving state-of-the-art results (78\% AUROC) and generalizing to entirely unseen cell types like RPE1.

\item \textbf{Variant Effect Prediction (VEP)}: The BioReason framework \cite{fallahpour2025bioreason} has addressed the classification of genetic variants (SNVs and non-SNVs) as benign or pathogenic. By integrating raw DNA sequences with symbolic pathway notation (e.g., $"GENE1+GENE2 -> GENE3"$), the model generates causal reasoning paths that connect genetic variants to disease phenotypes through defined molecular mechanisms.

\item \textbf{Protein-Protein Interaction (PPI) and Signaling}: ProLLM \cite{jin2024prollm} introduced the "Protein Chain-of-Thoughts" (ProCoT) method, which transforms structured interaction data into natural language reasoning chains that simulate sequential signaling pathways. This approach allows LLMs to capture non-physical (indirect) connections and multi-step signaling chains, outperforming traditional graph-based models on benchmarks like STRING and Human PPI datasets.

\item \textbf{Single-Cell RNA-seq Annotation}: Reasoning models like DeepSeek-R1 have been applied to zero-shot cell type annotation \cite{Wang2025BioReasoning}. These models process ranked lists of differentially expressed marker genes alongside tissue metadata to identify cell clusters. Notably, they outperform specialized expert models in out-of-distribution (OOD) scenarios by reasoning through the biological rationale for cell identity rather than relying on memorized patterns.

\item \textbf{DNA Structural Biophysics}: Models have been fine-tuned to predict DNA secondary structures (dot-parens-plus notation) and calculate Minimum Free Energy (MFE). Research by \cite{ross2024chainingthoughtsllmslearn} showed that CoT allows LLMs to approximate physical constants, such as base-pairing and stacking energies, by generating intermediate "nearest neighbor windows" of the input sequence.

\item \textbf{Disease Pathway Inference}: By leveraging datasets derived from KEGG, models have been trained to infer mechanism-disease associations \cite{fallahpour2025bioreason}. This involves reasoning over complex molecular networks represented in standardized symbolic notation to predict how specific molecular interactions lead to disease states.

\item \textbf{Interpretable Drug Discovery and Regulatory Success}: 

\begin{itemize}
    
\item The DrugReasoner model \cite{ghaffarzadehesfahani2025drugreasonerinterpretabledrugapproval} leverages Group Relative Policy Optimization (GRPO) to perform comparative reasoning between query compounds and structurally similar approved/unapproved drugs. This allows the model to predict regulatory approval outcomes with high accuracy (AUC 0.728 on external datasets) while providing step-by-step rationales.

\item OwkinZero addresses critical bottlenecks in the drug discovery pipeline, such as target druggability and modality suitability, by using Reinforcement Learning from Verifiable Rewards. This framework demonstrates that specialized 8–32B models can outperform larger commercial LLMs on biological reasoning tasks.

\item Toxicity prediction is enhanced by the CoTox framework \cite{park2025cotoxchainofthoughtbasedmoleculartoxicity}, which integrates chemical structures (using IUPAC names for better model legibility) with biological pathways and Gene Ontology (GO) terms. CoTox employs a four-step Chain-of-Thought (CoT) process—analyzing pathways, GO terms, structural features, and synthesizing a mechanistic explanation—to predict organ-specific toxicities.

\end{itemize}

\item \textbf{Autonomous Protein Science and Design Principles}:

\begin{itemize}
     
\item The Sparks framework \cite{ghafarollahi2025sparksmultiagentartificialintelligence} employs a multi-agent architecture (Proposer and Critic agents) to autonomously discover novel scientific principles. In one landmark study, it identified a length-dependent mechanical crossover where beta-sheet-biased peptides surpass alpha-helical ones in unfolding force beyond 80 residues.

\item In the realm of enzyme engineering, Genie-CAT (an agentic LLM framework) \cite{jacob2025proteinlanguagemodelsagentic} bridges symbolic reasoning and numerical simulation. It uses structural parsing of PDB files and redox predictors to generate mechanistically grounded hypotheses for metalloprotein design, significantly reducing the time required for complex biophysical calculations.

\item The specialized BioMedGPT-Mol \cite{zuo2025biomedgptmolmultitasklearningmolecular} facilitates molecule-centric discovery through a multi-task learning curriculum. It supports tasks ranging from chemical reaction prediction to multi-step retrosynthetic planning, where it achieved state-of-the-art results as an end-to-end planner.

\end{itemize}

\item \textbf{Multi-Omics and Biomolecular Reasoning}:

\begin{itemize}
     
\item The Biology-Instructions dataset \cite{he2025biologyinstructionsdatasetbenchmarkmultiomics} provides a massive benchmark for evaluating the multi-omics sequence understanding of LLMs across 21 tasks, covering DNA, RNA, and protein prediction.

\item To improve factual grounding, Bio-KCoT (Knowledge-Augmented Long-CoT) integrates knowledge graph-guided multi-hop traversal and pruning \cite{lyu2026knowledgeaugmented}. This framework allows for the generation of logically coherent reasoning paths for complex biomolecular tasks, specifically addressing multi-step causal chains that are often missing in standard LLM outputs.

\item LiveProteinBench \cite{rong2026liveproteinbench} serves as a contamination-free benchmark, utilizing proteins validated after early 2025 to ensure models are tested on truly novel data. It evaluates performance across 12 essential protein science tasks, revealing that leading proprietary models currently possess superior zero-shot reasoning potential.

% \item The CLAW-MRM framework (Comprehensive Lipidomics Automation Workflow) represents a significant advance in extending LLM capabilities to lipid biology. It utilizes a natural language interface powered by LLMs to automate lipid structural identification and annotation from Multiple Reaction Monitoring (MRM) mass spectrometry data. Beyond simple identification, it integrates the LIGER (lipidome gene enrichment reactions) tool, which enables the model to link lipid expression changes directly with gene activation and metabolic pathway alterations, particularly in diseases like Alzheimer's.

\end{itemize}

\end{itemize}

\subsection{Domain-specific tasks that have not been addressed by reasoning models}

While current literature demonstrates significant progress in applying reasoning models to DNA, proteins, and drug discovery, several core subjects within the LS1 panel remain largely unaddressed or represent significant frontiers for future research.

\begin{itemize}
     
\item \textbf{Underrepresented LS1 Domains (Glycobiology and Lipid Biology)}:

\begin{itemize}
    \item \textbf{Glycobiology (LS1\_6)}: There is a notable absence of reasoning models focused on the "glycome." Unlike DNA and proteins, glycans possess complex, branched structures that are not easily represented as linear sequences. No articles in the current review describe Chain-of-Thought (CoT) processes for reasoning about glycan-protein interactions or the functional impact of specific glycosylation sites.

    % \item \textbf{Lipid Biology (LS1\_5)}: Similarly, lipidomics and the study of lipid-membrane interactions are missing from the current landscape of reasoning LLMs. Tasks such as predicting the biophysical effects of lipid composition on signaling or modeling lipid metabolism fluxes using explicit reasoning chains have not yet been explored.

    \item \textbf{Lipid Biology and Membrane Dynamics (LS1\_5)}: While tools like CLAW-MRM \cite{doi:10.1021/acs.analchem.4c05039} have introduced LLM-driven automation for lipid identification and statistical analysis, a gap persists in models that can perform long-form reasoning about complex membrane physics. Current AI approaches to dynamic membrane events, such as lipid flip-flop (translocation between leaflets), rely on high-dimensional Machine Learning rather than linguistic reasoning. For example, the AIMMD (AI for molecular mechanism discovery) framework \cite{Post2025} uses deep neural networks to discover molecular mechanisms of flip-flop by training on-the-fly to predict "commitment probabilities" (committors). Although AIMMD successfully identifies mechanisms (e.g., "tunneling" vs. "pore-mediated" flip-flop), these findings are encoded in neural network weights rather than explicit, human-interpretable reasoning traces. Future research should aim to bridge the gap between these powerful ML-based biophysical discovery tools and the symbolic, step-by-step reasoning seen in models like rbio1 or BioReason.
\end{itemize}

\item \textbf{Metabolic Flux and Bioenergetics (LS1\_7)}:

    While biophysical constants like DNA stacking energies have been approximated through CoT, the broader field of bioenergetics—including mitochondrial function, ATP synthesis cycles, and complex metabolic flux analysis-remains a gap. Existing models focus on structural or regulatory outcomes (e.g., gene expression) rather than the energy-transfer mechanisms that drive cellular life.

\item \textbf{The "Dark Matter" of the Genome and Proteome (LS1\_3, LS1\_4)}:

\begin{itemize}
     
    \item \textbf{Non-Coding Functional Reasoning}: Despite tools like BioReason, a significant gap remains in reasoning about the "dark matter" of the genome—non-coding regions where functional annotations are sparse. Current reasoning traces are often grounded in well-characterized pathways (e.g., KEGG), leaving the mechanics of distal enhancers and silencers poorly understood.

    \item \textbf{Dark Proteomes}: Generalization to non-model organisms or poorly characterized "dark" proteomes, where labeled signaling data is scarce, is identified as a persistent challenge.
    
\end{itemize}

\item \textbf{From 1D Sequences to 3D Atomic Coordinates (LS1\_8)}:

\begin{itemize} 
    \item \textbf{3D Native Reasoning}: Most models process biological data as 1D sequences or 2D notations (e.g., dot-parens notation for DNA). There is a critical lack of "3D-native" reasoning capable of processing raw atomic coordinates (PDB data) and 3D genomic architecture (Hi-C data) as integral components of a comprehensive thought process.

    \item \textbf{Spatial Interaction Interfaces}: Current frameworks for protein-protein interaction (PPI) primarily rely on sequence-based embeddings rather than reasoning about the spatial fit and physical interaction interfaces in 3D space.
\end{itemize}

\item \textbf{Non-Linear Biological "Crosstalk" (LS1\_9)}:

    While models like ProLLM effectively simulate linear signaling chains, they often fail to capture the complexity of biological "crosstalk" —the non-linear feedback loops and branched pathways that characterize real cellular systems. Developing "Graph-of-Thoughts" or multi-agent architectures to represent these complex networks is a major unaddressed frontier.

\item \textbf{Autonomous Experimental Closing of the Loop (LS1\_14)}:

    A fundamental gap exists between in silico reasoning and in vivo reality. Most reviewed papers generate "testable hypotheses" but do not include physical wet-lab validation. There are no current reasoning models integrated with robotic laboratory controllers to autonomously execute and refine experimental protocols based on their own logical deductions.
\end{itemize}

\subsection{Comments on the usage of reasoning model in RAG-like system or agentic framework}
\subsubsection{Knowledge-Augmented Reasoning}

Traditional RAG systems in biology are evolving toward "Knowledge-Augmented" frameworks that utilize structured rather than unstructured data.
\begin{itemize}
\item \textbf{Structured Retrieval via Knowledge Graphs:} \textbf{Bio-KCoT} exemplifies a specialized RAG approach by integrating structured biological knowledge from Knowledge Graphs (KGs) directly into the reasoning process \cite{lyu2026knowledgeaugmented}. Instead of simple vector searches, it employs multi-hop traversal and pruning to retrieve specific molecular mechanisms and biochemical pathways. This ensures that the generated Long-CoT is grounded in verifiable biological facts, addressing the "decoupling" gap where models often provide correct answers based on faulty logic.
\end{itemize}

\subsubsection{Agentic Frameworks for Autonomous Discovery}
Beyond simple retrieval, the literature highlights agentic frameworks where multiple reasoning agents or external tools collaborate to solve complex LS1 problems.
\begin{itemize}
\item \textbf{Multi-Agent Collaboration:} The \textbf{Sparks} framework  \cite{ghafarollahi2025sparksmultiagentartificialintelligence} introduces a multi-agent AI model designed to execute the entire autonomous scientific discovery cycle, including hypothesis generation, experiment design, iterative refinement, and documentation. It employs a generation-reflection architecture where proposer agents are paired with critic agents to drive discovery beyond training data. This agentic loop allowed the discovery of mechanical crossovers in peptides that were previously uncharacterized in the literature.

\item \textbf{Symbolic-Neural Integration:} \textbf{Genie-CAT} \cite{jacob2025proteinlanguagemodelsagentic} introduces an agentic workflow that integrates four key capabilities: literature-grounded reasoning (RAG), structural parsing of PDB files, physics-based electrostatic potential calculations (APBS), and machine-learning-based redox property prediction. Genie-CAT uses a ReAct (Reasoning and Acting) pattern where the LLM iteratively generates thoughts and invokes specialized tools. It couples natural-language reasoning with quantitative physical modeling to generate mechanistically interpretable and testable hypotheses. In proof-of-concept tests, the system successfully identified residue-level modifications affecting redox tuning in [Fe-S] clusters, reproducing expert-derived hypotheses autonomously.

\item \textbf{Agentic Orchestration in Lipidomics}: The CLAW-MRM framework \cite{doi:10.1021/acs.analchem.4c05039} represents a primary example of an agentic workflow rather than a standalone reasoning model. The system employs a natural language interface powered by LLMs to function as an orchestrator for high-throughput lipidomics. Instead of utilizing internal Chain-of-Thought (CoT) for biological reasoning, it uses the LLM-driven agents in CLAW-MRM for automation of data parsing, TMM normalization, with aim of detailed lipid identification in complex biological samples. The system functions as an agent that calls external bioinformatics tools like LIGER to link lipid expression with gene activation patterns. The primary distinction of CLAW-MRM is its utilization of edgeR generalized linear models (GLMs). The incorporation of these models into the platform enables the effective management of overdispersed ion count data, thereby ensuring the accuracy and reliability of results obtained from the identification of differences in lipid levels across samples. This highlights a trend where LLMs serve as "reasoning interfaces" that manage complex, non-textual biological data pipelines without necessarily being trained on the underlying biological logic themselves. Despite the absence of reasoning models and CoT techniques in the authors' methodology, this framework can be enhanced by incorporating such models, thereby augmenting the capabilities of agentic systems and significantly enhancing biological reasoning.  
\end{itemize}

\subsection{Number of datasets for training domain-specific reasoning model found}

\begin{itemize}
    \item \textbf{All}: Zero / between 1 to 3 / between 4 and 6 / \textbf{many ($>$ 6)}

    \item \textbf{Publicly accessible}: Zero / between 1 to 3 / between 4 and 6 / \textbf{many ($>$ 6)}
\end{itemize}
However, only 2 datasets found that contain reasoning traces (3 if we include the one with generation script). Few articles contain generation scripts that can be utilized to create datasets.

\subsection{Number of benchmarks/datasets for evaluating domain-specific reasoning model found}

\begin{itemize}
    \item \textbf{All}: Zero / between 1 to 3 / between 4 and 6 / \textbf{many ($>$ 6)}

    \item \textbf{Publicly accessible}: Zero / between 1 to 3 / between 4 and 6 / \textbf{many ($>$ 6)}
\end{itemize}
However, only 2 datasets found that contain reasoning traces (3 if we include the one with generation script). Few articles contain generation scripts that can be utilized to create datasets.

\subsection{Number of domain-specific reasoning models found}

\begin{itemize}
    \item \textbf{All}: Zero / between 1 to 3 / between 4 and 6 / \textbf{many ($>$ 6)}

    \item \textbf{Publicly accessible}: Zero / \textbf{between 1 to 3} / between 4 and 6 / many ($>$ 6)
\end{itemize}
3 available now, one will be released soon on Hugging Face and two contain source code and instructions how to train the model and reproduce the results.

\subsection{Number of methods for creating/using domain-specific reasoning models found}

\begin{itemize}
    \item \textbf{All}: Zero / between 1 to 3 / between 4 and 6 / \textbf{many ($>$ 6)}

    \item \textbf{Publicly accessible}: Zero / \textbf{between 1 to 3} / between 4 and 6 / many ($>$ 6)
\end{itemize}
3 available now, one will be released soon on Hugging Face and two contain source code and instructions how to train the model and reproduce the results.

\section[Literature Review Summary: LS2]{Literature Review Summary: LS2}

\textbf{Integrative Biology: from Genes and Genomes to Systems.} Genetics, epigenetics, genomics and other ‘omics studies, bioinformatics, systems biology, genetic diseases, gene editing, innovative methods and modelling, ‘omics for personalised medicine.

\subsection{Key Findings}

The landscape of Large Language Models (LLMs) in integrative biology is rapidly shifting from general-purpose retrieval to specialized "System-2" reasoning architectures and agentic frameworks. The literature highlights a transition towards models capable of handling multi-modal biological data (sequences, expression profiles, structures) and executing autonomous workflows for discovery.

\subsubsection{Datasets and Benchmarks}
Recent work has introduced rigorous benchmarks to evaluate reasoning capabilities beyond simple factual recall. To address the lack of standardization in single-cell analysis, \cite{rizvi2025scell} developed the \textit{scQA} dataset, consisting of $\sim$1,600 question-answer pairs derived from biological manuscripts, and \cite{gao2025scpilot} introduced \textit{SCBENCH}, a suite of 9 datasets that cover cell-type annotation, trajectory inference, and prediction of gene regulatory networks. In the domain of perturbation biology, \textit{PerturbQA} has emerged as a critical benchmark, aggregating datasets such as Replogle and Nadig to evaluate differential expression predictions \cite{wu2025perturbqa, istrate2025rbio1}. Furthermore, \cite{Queen2025CGBENCH} established \textit{CGBENCH}, a large-scale benchmark derived from the ClinGen Evidence Repository, designed to test LLMs in clinical genetics tasks such as Variant Pathogenicity Curation (VCI) and evidence verification.

\subsubsection{Reasoning Models and Architectures}
Although general reasoning models like DeepSeek-R1 and OpenAI's o-series are widely used as baselines \cite{Wang2025BioReasoning, Queen2025CGBENCH}, significant efforts are directed towards domain-specific architectures:

\begin{itemize}
    \item \textbf{Multimodal Integration:} \cite{fallahpour2025bioreason} introduced \textit{BIOREASON}, a novel architecture that fuses a frozen DNA foundation model (Evo2-1B) with an LLM through a programmable projection layer, enabling the model to reason directly over genomic embeddings. Similarly, \cite{rizvi2025scell} proposed \textit{C2S-Scale}, a family of models (up to 27B parameters) trained on "cell sentences" to unify predictive and generative tasks in single-cell genomics.
    \item \textbf{Novel Training Paradigms:} Addressing the scarcity of experimental data, \cite{istrate2025rbio1} proposed "soft verification," a training method in which biological world models (e.g. simulators) provide reward signals during reinforcement learning. \cite{Liu2025ProteinReasoner} developed \textit{ProteinReasoner}, which utilizes an "evolutionary profile" as an explicit Chain-of-Thought (CoT) step, bridging protein structure and sequence modalities.
\end{itemize}

\subsubsection{Modalities and Agentic Frameworks}
The field is moving beyond text-only inputs. \cite{rizvi2025scell} and \cite{fallahpour2025bioreason} demonstrate the efficacy of treating biological signals (scRNA-seq, DNA) as language tokens or embeddings. However, for complex workflows, agentic frameworks are gaining traction. \cite{gao2025scpilot} proposed \textit{SCPILOT}, an "omics-native" agent that converses in natural language but executes analysis via a "propose-filter-solve" loop using bioinformatics tools (e.g., Scanpy). In contrast, \cite{wu2025perturbqa} introduced \textit{SUMMER}, an inference-time framework that converts biological graphs into text summaries for ground reasoning without extensive training.

\subsubsection{Prompting Strategies and Validation}
Validation strategies are becoming more sophisticated to distinguish genuine reasoning from hallucination.

\begin{itemize}
    \item \textbf{Synthetic Reasoning Traces:} \cite{phillips2026synthpert} introduced \textit{SynthPert}, a method that fine-tunes smaller models (Llama 3.1 8B) on high-quality synthetic explanations generated by teacher models (o4-mini) and filtered by a critic, achieving performance superior to the teacher models themselves.
    \item \textbf{Constrained Generation:} To ensure standardized outputs, \cite{Wang2025BioReasoning} utilized a "classifier mode" where the LLM is constrained to select cell types from a predefined list, significantly improving out-of-distribution generalization.
    \item \textbf{Evaluation Metrics:} Beyond standard accuracy, new domain-specific metrics are proposed, such as the \textit{scFID} (Single-Cell Fréchet Inception Distance) to assess generative quality \cite{rizvi2025scell} and rigorous evidence scoring protocols in clinical genetics \cite{Queen2025CGBENCH}.
\end{itemize}

\subsection{Identified Gaps}

Despite the rapid advancement of reasoning LLMs in integrative biology, several critical gaps and challenges remain that hinder their widespread adoption and reliability in clinical and research settings.

\subsubsection{Hallucinations and Reliability of Reasoning}
A persistent challenge in multiple studies is the issue of hallucination. Even when models generate the correct final answers, the intermediate reasoning steps or explanations may be hallucinated or biologically invalid. For instance, in clinical genetics, models often correctly classify evidence codes but provide non-matching or hallucinated explanations when judged against human experts \cite{Queen2025CGBENCH}. Similarly, hallucination risks were noted in perturbation predictions where incorrect predictions relied on non-existent "super-pathways" \cite{phillips2026synthpert}.
Furthermore, current reasoning models generally lack robust uncertainty quantification. This makes them less reliable for high-stakes clinical decision-making, such as variant effect prediction or disease pathway analysis, where knowing the confidence of the model is crucial \cite{fallahpour2025bioreason, wu2025perturbqa}.

\subsubsection{Architectural and Contextual Limitations}
Standard transformer architectures may not be optimal for all biological data.
\begin{itemize}
    \item \textbf{Causal Attention Mismatch:} The unidirectional causal attention mechanism used in standard LLMs, predicting the next token, may not perfectly model complex, non-sequential biological interactions, such as those found in gene regulatory networks or "cell sentences" \cite{rizvi2025scell}.
    \item \textbf{Context Constraints:} Despite longer context windows, hardware constraints often force the truncation of data. For example, DNA sequences had to be truncated to 2048 tokens in multimodal DNA-LLM architectures \cite{fallahpour2025bioreason}. Similarly, single-cell agents often rely on aggressive data compression (summarizing expression matrices), which can lead to information loss compared to processing raw data directly \cite{gao2025scpilot}.
\end{itemize}

\subsubsection{Computational Cost and Latency}
The "test-time compute" paradigm, which defines reasoning models, introduces significant computational overhead. Models like DeepSeek-R1 or OpenAI's o1 family require significantly more tokens (up to 8.5x more) due to internal reasoning steps compared to standard models \cite{Wang2025BioReasoning}. This results in higher latency (e.g. 30 s vs. 4 s per query) and higher costs, making them less accessible for large-scale high-throughput screening without optimization \cite{gao2025scpilot, wu2025perturbqa}.

\subsubsection{The "Open Source" Gap}
There is a notable performance disparity between proprietary frontier models (e.g., o1, Gemini 2.5 Pro) and open-source alternatives. Current open-source models (even large ones like Gemma-3 27B) often struggle with complex multi-step biological reasoning and hallucinate more frequently than their proprietary counterparts, highlighting a need for better domain-specific fine-tuning of open weights \cite{gao2025scpilot}.

\subsection{Notable domain-specific tasks that have been addressed by reasoning models}

Recent literature demonstrates that reasoning models—ranging from Chain-of-Thought (CoT) enhanced LLMs to agentic frameworks—are being successfully applied to complex integrative biology tasks that go beyond simple retrieval. These tasks can be categorized into cellular analysis, perturbation modeling, molecular engineering, and clinical evidence curation.

\subsubsection{Single-Cell Analysis and Genomics}
The interpretation of single-cell RNA sequencing (scRNA-seq) data has been a primary focus for reasoning models, moving from static perception to dynamic reasoning.
\begin{itemize}
    \item \textbf{Cell Type Annotation:} Several approaches have addressed zero-shot cell type annotation. \cite{rizvi2025scell} introduced "cell sentences" to enable LLMs to process gene expression rankings for annotation and group captioning. \cite{Wang2025BioReasoning} demonstrated that general-purpose reasoning models like DeepSeek-R1 can outperform specialized biological models on out-of-distribution data by reasoning over marker gene lists. Furthermore, agentic frameworks like scPilot utilize iterative "propose-filter-solve" loops to refine annotations by invoking external tools \cite{gao2025scpilot}.
    \item \textbf{Trajectory Inference and GRN Prediction:} Beyond static annotation, agentic systems have been applied to reconstruct developmental trajectories (inferring lineage trees) and predict Gene Regulatory Networks (GRNs). These systems effectively coordinate specialized tools (e.g., Monocle, pySCENIC) to deduce biological relationships that standard LLMs struggle to infer directly \cite{gao2025scpilot}.
    \item \textbf{Complex QA on Single-Cell Data:} Models like C2S-Scale have been fine-tuned to answer complex biological questions regarding dataset interpretation and spatial neighborhood inference, unifying predictive and generative tasks within a single architecture \cite{rizvi2025scell}.
\end{itemize}

\subsubsection{Predicting Cellular Response to Perturbations}
Predicting how cells respond to genetic drugs or CRISPR knockouts is a complex causal task where reasoning models have shown significant promise over traditional GNNs.
\begin{itemize}
    \item \textbf{Differential Expression Prediction:} Multiple frameworks address the prediction of post-perturbation gene expression. \cite{wu2025perturbqa} proposed an inference-time framework (SUMMER) that uses RAG and summarization to predict the direction of gene expression change. Meanwhile, \cite{istrate2025rbio1} utilized "soft verification" to train models using biological world models as simulators.
    \item \textbf{Reasoning with synthetic trace:} To overcome data scarcity, \cite{phillips2026synthpert} introduced a method to distill reasoning capabilities from frontier models (like o4-mini) into smaller models using synthetic "reasoning traces," achieving state-of-the-art accuracy in predicting 3-class perturbation outcomes (up/down/no-change).
\end{itemize}

\subsubsection{Protein Design and Variant Effect Prediction}
Reasoning models are increasingly handling raw molecular data (sequences and structures) by treating biological constraints as intermediate reasoning steps.
\begin{itemize}
    \item \textbf{Protein Structure and Optimization:} \cite{Liu2025ProteinReasoner} introduced a multi-modal model that generates "evolutionary profiles" as an explicit Chain-of-Thought step. This approach addresses inverse protein folding and structure prediction and introduces an In-Context Learning paradigm for protein optimization (directed evolution) without the need for fine-tuning.
    \item \textbf{Variant Effect Prediction (VEP):} By integrating frozen DNA foundation models with LLMs, the BioReason framework addresses the prediction of pathogenicity for coding sequences and non-SNVs. This multimodal approach allows the model to provide step-by-step mechanistic rationales for disease pathway predictions based on raw genomic data \cite{fallahpour2025bioreason}.
\end{itemize}

\subsubsection{Clinical Evidence Curation}
In the clinical domain, reasoning models act as expert curators to standardize the interpretation of scientific literature.
\begin{itemize}
    \item \textbf{Evidence Scoring and Verification:} The CGBENCH benchmark evaluated models on their ability to classify scientific papers into "evidence codes" (pathogenic/benign) according to ClinGen protocols. While standard models excel in high-level interpretation, reasoning models (e.g. o1, DeepSeek-R1) have shown superior capability in extracting fine-grained experimental evidence and structured data from full-text publications \cite{Queen2025CGBENCH}.
\end{itemize}

\subsection{Domain-specific tasks that have not been addressed by reasoning models}

Although reasoning models have demonstrated capabilities in retrieval and structured prediction, significant gaps remain in handling complex, quantitative, and spatially-aware biological tasks.

\subsubsection{Quantitative Precision and Dynamics}
Current reasoning models operate largely on qualitative or discretized levels rather than on precise quantitative scales.
\begin{itemize}
    \item \textbf{Magnitude of Perturbation:} While models like SUMMER and those trained on synthetic traces can predict the \textit{direction} of gene expression change (up/down) with reasonable accuracy, predicting the exact \textit{magnitude} (fold-change) remains an unsolved challenge. The "imbalance" in the training data and the complexity of the dosage effects make quantitative regression difficult for current LLM architectures \cite{wu2025perturbqa, phillips2026synthpert}.
    \item \textbf{Temporal Dynamics:} Most current approaches treat biological states (e.g., scRNA-seq data) as static snapshots. The reasoning about continuous temporal dynamics—predicting the exact time-course of a cellular response rather than just the end state—is largely unaddressed, partly due to the scarcity of high-resolution time-series training data \cite{rizvi2025scell}.
\end{itemize}

\subsubsection{Long-Range Genomic Interactions and 3D Structure}
The architectural limitations of Transformers have prevented a full solution to tasks involving the 3D organization of the genome.
\begin{itemize}
    \item \textbf{Distal Regulatory Elements:} Due to context window constraints (e.g., 2048 tokens for DNA encoders), current multimodal models struggle to capture long-range dependencies, such as enhancer-promoter interactions that span significantly larger distances than the model can process. Consequently, the reasoning over non-coding variants in distal regions remains "touched" but not effectively solved \cite{fallahpour2025bioreason}.
    \item \textbf{Bi-directional vs. Causal Reasoning:} Biological interactions in a genome or protein are often bi-directional and spatial, whereas standard LLM reasoning is causal (left-to-right). This mismatch limits the ability of models to natively reason about 3D structural constraints without relying heavily on external, specialized modules \cite{rizvi2025scell, Liu2025ProteinReasoner}.
\end{itemize}

\subsubsection{Nuanced Clinical Judgement and Conflict Resolution}
In the clinical domain, the "reasoning" is often limited to extraction rather than judgment.
\begin{itemize}
    \item \textbf{Evidence Weighting:} While models can identify evidence, they fail to critically evaluate the \textit{strength} or \textit{quality} of that evidence. For instance, distinguishing between a robust functional study and a weak correlation is a task where reasoning models still lag behind human curators. They tend to hallucinate strength where it does not exist or do not downgrade low-quality data \cite{Queen2025CGBENCH}.
    \item \textbf{Conflict Resolution:} When presented with conflicting biological evidence (e.g., one paper claiming pathogenicity and another benign status), current models often struggle to synthesize a coherent resolution or explain the discrepancy, often defaulting to the most recent or most frequent claim rather than reasoning through the methodology \cite{Queen2025CGBENCH, Wang2025BioReasoning}.
\end{itemize}

\subsubsection{Novel Mechanism Generation}
Finally, there is a distinction between "reasoning" which is essentially advanced retrieval of known mechanisms, and true discovery.
\begin{itemize}
    \item \textbf{Novel Mechanism Hypothesis:} Agentic frameworks are currently designed to execute known bioinformatics pipelines. They have not yet demonstrated the ability to hypothesize entirely new biological mechanisms or pathways that are not present in their training corpus. Reliance on "world models" or simulators limits them to the boundaries of current biological knowledge \cite{gao2025scpilot, istrate2025rbio1}.
\end{itemize}

\subsection{Comments on the usage of reasoning model in RAG-like system or agentic framework}

The integration of reasoning models into RAG (Retrieval-Augmented Generation) systems and agentic frameworks has emerged as a dominant strategy to overcome the limitations of static knowledge and context windows in integrative biology.

\subsubsection{Agentic Frameworks and Tool Orchestration}
Recent works demonstrate that reasoning models excel as "orchestrators" or "planners" within agentic systems, capable of breaking down complex biological queries into executable sub-tasks.
\begin{itemize}
    \item \textbf{Omics-Native Reasoning:} The \textit{scPilot} framework exemplifies this by using a "propose-filter-solve" loop. The central reasoning agent (e.g., o1) does not process raw data directly but instead orchestrates a library of bioinformatics tools (like Seurat or Monocle 3) to execute specific operators. This allows the model to perform complex tasks like trajectory inference and GRN prediction by conversing with data via tools, rather than just retrieving text \cite{gao2025scpilot}.
    \item \textbf{Iterative Refinement:} Unlike standard "one-shot" prompting, these agentic frameworks utilize iterative reasoning. If an initial tool execution yields ambiguous results, the reasoning model can refine its hypothesis and request new analyses, mirroring the workflow of a human computational biologist \cite{gao2025scpilot}.
    \item \textbf{Sub-agent Management:} Similarly, other proposals suggest a "Reasoning LLM Orchestrator" that manages specialized sub-agents for data ingestion, clustering, and quality control, automating entire pipelines for tasks like cell type annotation \cite{Wang2025BioReasoning}.
\end{itemize}

\subsubsection{RAG for Contextualization and Grounding}
RAG systems are being adapted to serve as "soft verifiers" or grounding mechanisms for reasoning models, particularly in perturbation analysis.
\begin{itemize}
    \item \textbf{Summarization-based RAG:} The \textit{SUMMER} framework introduces a novel approach where the retrieval component is not just fetching documents but "featurizing" biological entities (genes) into natural language summaries derived from knowledge graphs. The reasoning model then uses these summaries alongside the retrieved experimental outcomes to predict differential expression, effectively translating the graph data into a linguistic format that facilitates logical deduction \cite{wu2025perturbqa}.
    \item \textbf{Soft Verification:} Instead of traditional RAG which retrieves ground truth, the \textit{rbio1} framework uses "world models" (simulators) as external verifiers. The intermediate steps of the reasoning model are evaluated against these simulators, providing a reward signal that guides the reasoning process without the need for expensive experimental wet-lab data \cite{istrate2025rbio1}.
\end{itemize}

\subsection{Number of datasets for training domain-specific reasoning model found}

\begin{itemize}
    \item \textbf{All}: between 1 to 3
    % Found: 3 (C2S-Scale Multimodal Corpus [Rizvi], KEGG/VEP Reasoning Dataset [Fallahpour], SynthPert Synthetic Data [Phillips])

    \item \textbf{Publicly accessible}: between 1 to 3
    % All 3 mentioned above are publicly available via GitHub or Figshare.
\end{itemize}

\subsection{Number of benchmarks/datasets for evaluating domain-specific reasoning model found}

\begin{itemize}
    \item \textbf{All}: between 4 and 6
    % Found: 5 (scQA/C2S [Rizvi], CGBENCH [Queen], PERTURBQA [Wu/Istrate], SCBENCH [Gao], BioReason Benchmarks [Fallahpour])

    \item \textbf{Publicly accessible}: between 4 and 6
    % All 5 benchmarks are publicly accessible.
\end{itemize}

\subsection{Number of domain-specific reasoning models found}

\begin{itemize}
    \item \textbf{All}: between 4 and 6
    % Found: 5 distinct new models/families (C2S-Scale [Rizvi], rbio1 [Istrate], BIOREASON [Fallahpour], ProteinReasoner [Liu], SynthPert-Llama [Phillips]). Note: scPilot and SUMMER are counted as frameworks/methods, not model weights.

    \item \textbf{Publicly accessible}: between 4 and 6
    % 4 models are public (C2S, rbio1, BIOREASON, SynthPert). ProteinReasoner [Liu] source code is marked as "no".
\end{itemize}

\subsection{Number of methods for creating/using domain-specific reasoning models found}

\begin{itemize}
    \item \textbf{All}: many ($>$ 6)
    % Every paper (8 total) introduces a distinct method or framework (e.g., Soft Verification, Omics-Native Reasoning, Cell2Sentence, DNA-LLM Adapter, Evolutionary Profile CoT, etc.).

    \item \textbf{Publicly accessible}: many ($>$ 6)
    % 7 out of 8 papers provide public code/implementations (All except Liu et al.).
\end{itemize}

\section[Literature Review Summary: LS3]{Literature Review Summary: LS3}

\textbf{Cell Biology, Development, Stem Cells and Regeneration.} Structure and function of the cell, cell-cell communication, embryogenesis, tissue differentiation, organogenesis, growth, development, evolution of development, organoids, stem cells, regeneration, therapeutic approaches.

\subsection{Key Findings}

The field has shifted from "computational oracles" performing discrete tasks to models capable of iterative, multi-step logic required for complex biological processes like embryogenesis and tissue regeneration. Several domain-specific reasoning models and frameworks were identified.

BioReason \citep{fallahpour2025bioreason} integrates DNA embeddings with symbolic reasoning to predict disease pathways with 97-98\% accuracy. Cell2Sentence (C2s) \citep{rizvi2025scell} transforms transcriptomic profiles into "cell sentences" where gene names are ordered by their relative expression levels; scaling to 27 billion parameters (C2S-Scale) unlocks emergent reasoning for multicellular contexts. Meanwhile, rbio1 \citep{istrate2025rbio1} is a scientific reasoning model trained using "soft verification" to overcome the lack of immediate laboratory feedback; this paradigm uses biological world models (referred to as Virtual Cell Models, or VCMs) as approximate oracles or surrogate rewards during the training process.

Notable frameworks include: scPilot, an omics-native framework that utilizes reasoning model to converse in natural language while directly inspecting single-cell RNA-seq data and on-demand bioinformatics tools \citep{gao2025scpilot}; PersonaAI, an agentic framework for autonomous hypothesis generation and validation in aging \citep{Cho2026PersonaAI}; MEDEA, an agentic framework for therapeutic discovery that uses a multi-module architecture for research planning, code execution, and reasoning over literature \citep{Sui2026Medea}; DeviceAgent, a multimodal agentic framework that integrates LLM, VLMs, and domain-specific computational tools for bioelectronics research, specifically in developing stretchable mesh electronics for interfacing with human induced pluripotent stem cell-derived cardiomyocytes (hiPSC-CMs) \citep{Lee2025DeviceAgent}.

Current works utilize diverse modalities including raw nucleotides, DNA embeddings, protein sequences, single-cell RNA sequencing (scRNA-seq) data, knowledge graphs, and scientific literature. Models utilize Chain-of-Thought (CoT) prompting and explicit reasoning traces to externalize logical paths. Outputs include mechanistic explanations, "biological traces" of variant effects, and natural language lineage justifications.

\subsection{Identified Gaps}
\begin{itemize}
\item Experimental Bottleneck: Biological predictions are slow and expensive to validate in a laboratory, creating a scarcity of "hard" verification mechanisms. Soft verification via Virtual Cell Models (VCMs) was introduced as a workaround \citep{istrate2025rbio1}. The use of soft verification and virtual cell models allows models to learn the "laws" of biology without being limited by the speed of physical experiments.
\item Current models often struggle to retrieve task-relevant information from their internal parameters unless specifically prompted by an expert to produce verbalized reasoning in that direction \citep{li2026demystifying}.
\item Current models lack a deep understanding of physical and temporal constraints in laboratory work. As a result, while current SOTA models perform well on surface understanding (QA), they struggle significantly with structured generation and deep procedural reasoning involved in complex tissue differentiation protocols.
\item Challenges with Local Open-Weight LLMs: While open-weight LLMs offer model transparency and data control advantages, their performance without further fine-tuning is significantly inferior compared to closed proprietary models in some cases \citep{gao2025scpilot}. The feasibility of fine-tuning is limited by the availability of task-specific datasets.
\end{itemize}

\subsection{Notable domain-specific tasks that have been addressed by reasoning models}
\begin{itemize}
\item Bioelectronics design: The DeviceAgent framework \citep{Lee2025DeviceAgent} was used to develop stretchable mesh electronics for interfacing with human induced pluripotent stem cell-derived cardiomyocytes (hiPSC-CMs).
\item Cell Type Annotation and Analysis: scPilot with GPT o1 was demonstrated to achieve +11\% accuracy over traditional methods in assigning biologically accurate cell-type labels to each cell in an scRNA expression matrix \citep{gao2025scpilot}. Meanwhile, Medea was used in generating therapeutic hypotheses at the level of specific cell types, nominating therapeutic targets in disease-relevant cell types for rheumatoid
arthritis, type 1 diabetes mellitus, Sjogren’s Syndrome, hepatoblastoma, and follicular lymphoma \citep{Sui2026Medea}.
\item Trajectory Inference: scPilot with Gemini-2.5-Pro was able to cut trajectory graph-edit distance by 30\% compared to traditional one-shot prompting in reconstructing lineage trees (cellular developmental progression paths) \citep{gao2025scpilot}.
\item Gene Regulatory Network Prediction: scPilot with GPT o1 was also demonstrated to outperform baseline methods in predicting the existence of a regulatory relationship between a transcription factor (TF) and target gene pair \citep{gao2025scpilot}.
\item Disease Pathway Prediction and Analysis: BioReason achieved very high accuracy (97-98\%) in predicting KEGG pathways from genomic data \citep{fallahpour2025bioreason}, while PersonaAI identified the role of the vascular endothelial niche in the male-specific decline of Adipose Stem and Progenitor Cells \citep{Cho2026PersonaAI}.
\item Therapeutic Discovery: Medea was demonstrated in a variety of tasks such as nominating therapeutic targets in disease-relevant cell types, inferring synthetic lethality in a given cellular context, and predicting personalized response to immunotherapy based on multimodal patient data, including clinical and transcriptomic profiles \citep{Sui2026Medea}.
\end{itemize}

\subsection{Domain-specific tasks that have not been addressed by reasoning models}

\begin{itemize}
\item Robust Design of Multi-Stage Differentiation Protocols: While models can summarize existing protocols, they struggle to design new, robust, and reproducible multi-step differentiation pathways from scratch. Stem cell maturation requires a precise sequence of chemical signals (morphogens) arriving at the exact right moment and concentration; current models lack a deep understanding of the temporal and physical constraints of a laboratory environment. For example, they cannot yet reliably predict how a small variation in timing at Day 3 will impact the final cell population at Day 21, often leading to "immature" or "inconsistent" cell types.
\item Causal Inference in "Non-Linear" Regulatory Networks: While LLMs are excellent at mapping known pathways (e.g., from KEGG), but they struggle with causal discovery in novel, non-linear regulatory networks. Models often default to descriptive bioinformatics, correlating gene expression without understanding the underlying causal mechanics. They fail to predict how a system might evolve under entirely new evolutionary or synthetic pressures.
\item Autonomous "Wet-Lab" Execution and Validation: no current framework has successfully completed a full autonomous research cycle from hypothesis to validated physical results.
\item Predicting multi-scale physiological outcomes from molecular data: Tissue engineering requires integrating structural, functional, and material interactions. LLMs cannot yet reason across scales, from the molecular (Angstroms) to the tissue level (Centimeters), in a way that respects rigorous physiological parameters.
\end{itemize}

\subsection{Comments on the usage of reasoning model in RAG-like system or agentic framework}
Reasoning models are increasingly being used in autonomous agentic frameworks rather than simple chat interfaces. Many of these works focused on multi-modal reasoning models that support not only text but also image. They also rely on tool usage capability to integrate domain-specific computational tools.

\subsection{Number of datasets for training domain-specific reasoning model found}

\begin{itemize}
    \item \textbf{All}: between 4 and 6

    \item \textbf{Publicly accessible}: between 4 and 6
\end{itemize}

\subsection{Number of benchmarks/datasets for evaluating domain-specific reasoning model found}

\begin{itemize}
    \item \textbf{All}: between 4 and 6

    \item \textbf{Publicly accessible}: between 4 and 6
\end{itemize}

\subsection{Number of domain-specific reasoning models found}

\begin{itemize}
    \item \textbf{All}: between 1 to 3

    \item \textbf{Publicly accessible}: between 1 to 3
\end{itemize}

\subsection{Number of methods for creating/using domain-specific reasoning models found}

\begin{itemize}
    \item \textbf{All}: between 4 and 6

    \item \textbf{Publicly accessible}: between 4 and 6
\end{itemize}

\section[Literature Review Summary: LS4]{Literature Review Summary: LS4}

\textbf{Physiology in Health, Disease and Ageing.} Organ and tissue physiology, comparative physiology, physiology of ageing, pathophysiology, interorgan and tissue communication, endocrinology, nutrition, metabolism, interaction with the microbiome, non-communicable diseases including cancer (and except disorders of the nervous system and immunity-related diseases).

\subsection{Key Findings}
\textit{Important datasets, benchmarks, domain-specific reasoning models; what reasoning models are being used; modalities used in existing works (text-only, text-image, or other modalities); how people prompt the model, what kind of outputs they expect (e.g., whether using a defined schema with closed answer set, json formatting, open ended answers etc.), how they use and/or validate the reasoning of models; whether existing works consider aspects like reasoning length/effort as well as cost/efficiency.}

\subsubsection*{Datasets and Benchmarks}
The landscape of domain-specific reasoning benchmarks is rapidly evolving,
with a clear shift from generic medical examinations toward tasks that
require physiological depth, numerical precision, and longitudinal
reasoning.  Key datasets and benchmarks identified in the reviewed
literature include:
 
\begin{itemize}
 
  \item \textbf{MedQA-USMLE.}
        Standard multiple-choice benchmark based on United States Medical
        Licensing Examination questions, used extensively to evaluate
        general medical reasoning.  Reviewed studies including FineMedLM-o1
        and Med-PaLM report substantial performance improvements on this
        dataset~\citep{singhal2025toward, yu2025finemedlmo1enhancingmedicalknowledge,
        lievin2023largelanguagemodelsreason}.
 
  \item \textbf{Diabetes and Endocrinology Board Certification Examination.}
        UK-based subspecialty examination used to probe LLM performance on
        endocrinology-specific knowledge.  All tested general-purpose LLMs
        scored significantly below the human passing threshold, revealing
        substantial gaps in metabolic and hormonal
        reasoning~\citep{poor2025llm}.
 
  \item \textbf{Thyroid Nodule Clinical Dataset.}
        A real-world dataset of 63~pathology-confirmed thyroid nodule cases
        (after exclusion), with ACR-TIRADS and C-TIRADS classifications as
        ground truth.  Used to benchmark LLM-assisted malignancy risk
        stratification; notably, ultrasound images were not provided
        directly---models reasoned from textual
        descriptions only~\citep{dai2025thyroid}.
 
  \item \textbf{Clinical Vignettes (Script Concordance Test).}
        Expert-developed clinical scenarios used in the NEJM AI benchmarking
        study to evaluate how well LLMs update diagnostic and therapeutic
        reasoning when new or conflicting information is
        presented~\citep{mccoy2025assessment}.
 
  \item \textbf{CGMacros.}
        A pilot multimodal dataset coupling continuous glucose monitor (CGM)
        time-series with dietary logs and activity data, designed for
        personalised nutrition modelling and metabolic
        analysis~\citep{gcmacros}.  One of the few physiological datasets
        that combines longitudinal sensor data with macronutrient-level
        dietary records.
 
  \item \textbf{Synthetic Medical Dialogue Dataset.}
        A large-scale synthetically generated corpus used to train
        FineMedLM-o1's long-form reasoning capabilities via supervised
        fine-tuning and direct preference
        optimisation~\citep{yu2025finemedlmo1enhancingmedicalknowledge}.
 
  \item \textbf{NutriBench.}
        A benchmark for carbohydrate estimation from free-text meal
        descriptions, targeting the numerical precision required for
        insulin-dosing and dietary management in
        diabetes~\citep{dhaliwal2025nutribench}.
 
  \item \textbf{CMedCalc-Bench.}
        An emerging Chinese-language benchmark probing precise clinical
        calculations such as drug dosing, fluid management, and
        physiological scoring---an area where current reasoning models
        perform poorly~\citep{khandekar2024medcalcbench}.
 
\end{itemize}
 
Several widely used general benchmarks (MedMCQA, PubMedQA, MMLU-Medical)
are prominent in the broader literature~\citep{lievin2023largelanguagemodelsreason}
but are not domain-specific to physiology or endocrinology; they are
included for completeness but should not be treated as primary evidence of
reasoning capability in this field.

\subsubsection*{Reasoning Models}
% ----------------------------------------------------------------
 
Domain-specific reasoning models have advanced considerably.  Among those
directly relevant to physiology, endocrinology, and metabolism:
 
\begin{itemize}
 
  \item \textbf{FineMedLM-o1}~\citep{yu2025finemedlmo1enhancingmedicalknowledge}:
        A medical LLM trained on synthetic dialogues and reasoning chains
        via supervised fine-tuning and direct preference optimisation.
        Achieves a 23\% average improvement over prior models on major
        medical benchmarks and incorporates test-time training for domain
        adaptation.
 
  \item \textbf{HuatuoGPT-o1}~\citep{huatuogpto12024}:
        Uses verifiable medical problems and a two-stage training pipeline
        combining guided search for complex reasoning trajectories with
        reinforcement learning.  Outperforms medical-specific baselines
        using only 40K training examples.
 
  \item \textbf{Med-PaLM~2}~\citep{singhal2025toward}:
        Instruction-tuned LLM using chain-of-thought and ensemble refinement.
        The first model to exceed the physician passing score on
        USMLE-style questions.
 
  \item \textbf{DeepSeek~R1}~\citep{frontiers2025deepseek}:
        Open-source reasoning model evaluated against medical expert
        reasoning patterns.  Achieves 93\% diagnostic accuracy on MedQA
        cases but exhibits characteristic failure modes relevant to clinical
        deployment, including anchoring bias and overthinking.
 
\end{itemize}
 
Additional models mentioned in the broader literature include
Diabetica-o1~\citep{wei2024adapted} (fine-tuned for diabetes management) and
ESM3~\citep{doi:10.1126/science.ads0018} (protein sequence, structure, and function reasoning for
molecular biology applications). 
 
% ----------------------------------------------------------------
\subsubsection*{Modalities Used in Existing Works}
% ----------------------------------------------------------------
 
The overwhelming majority of reviewed works operate in a
\textbf{text-only} modality.  This is a notable finding given that
endocrinology and metabolic medicine are intrinsically image- and
signal-rich domains:
 
\begin{itemize}
 
  \item Clinical narratives, radiology and pathology reports, and
        structured examination questions are the dominant input format.
 
  \item The thyroid nodule study~\citep{dai2025thyroid} processed ultrasound
        findings as free-text descriptions rather than raw images.  The
        potential value of directly analysing ultrasound images---and the
        performance gap this creates---was not quantified.
 
  \item The oncology clinical reasoning study processed radiology report
        text, not imaging data~\citep{oncology_clinical_reasoning}, with
        models performing guideline-based staging and response classification
        from unstructured narrative.
 
  \item CGMacros~\citep{gcmacros} is an exception: it couples continuous
        time-series physiological data (glucose, activity) with dietary logs,
        though reviewed works that \emph{apply} reasoning models to this
        dataset still operate primarily in a text-converted form of the
        signals.
 
  \item No reviewed study integrated joint text-and-image reasoning for
        endocrinological imaging modalities (thyroid ultrasound, adrenal
        CT, pituitary MRI, DEXA for bone densitometry).  This represents
        a significant and largely unacknowledged gap given that imaging is
        integral to endocrine diagnosis~\citep{sim2025unveiling}.%expression that imaging is central to endocrine diagnosis can be misleading. It is certainly integral, however usualy diagnostic pathway looks  like this: patient's sympthoms and complaints->physical examination/laboratory assesment->if abnormal,then imaging which gives us an anwser about potential source of hormonal excess (e.g autonomusly secreting tumour); therefore I suggest expression 'integral'
 
\end{itemize}
 
%my suggestion: for part of this sentence in brackets (single chain RNA sequencing, DNA transcription regulation,genomic sequennces) 
Multi-omics data (single chain RNA sequencing, DNA transcription regulation, genomic sequences) and continuous physiological
logs (CGM, activity trackers) appear in broader reviews of the
field~\citep{nature2025biomedicine} but are not yet standard inputs to
reasoning models in this domain.
 
% ----------------------------------------------------------------
\subsubsection*{Prompting Approaches}
% ----------------------------------------------------------------
 
Prompting strategies have evolved substantially from simple zero-shot
queries toward structured clinical workflows.  Key approaches identified:
 
\begin{itemize}
 
  \item \textbf{Chain-of-Thought (CoT) Prompting}: The most widely used
        approach, instructing models to produce step-by-step intermediate
        reasoning before giving a final answer.  Used in FineMedLM-o1,
        Med-PaLM~2, and analysed across multiple
        benchmarks~\citep{wei2022chain, singhal2025toward,
        yu2025finemedlmo1enhancingmedicalknowledge}.
 
  \item \textbf{Few-Shot CoT}: Augments prompts with worked examples
        including reasoning chains, as used in Med-PaLM~2 and the CLINICR
        framework~\citep{singhal2025toward, nachane-etal-2024-shot}.
 
  \item \textbf{Self-Consistency Checking}: Multiple reasoning paths are
        sampled independently; the answer supported by the largest subset of
        paths is selected, improving robustness on ambiguous medical
        questions~\citep{wang2022self}.
 
  \item \textbf{Ensemble Refinement}: Med-PaLM~2 combines outputs from
        multiple reasoning passes, further improving accuracy on complex
        multi-step clinical
        questions~\citep{singhal2025toward}.
 
  \item \textbf{Structured Clinical Reasoning Prompts}: Two-stage
        approaches that first translate patient narratives into labelled
        clinical components (history, examination, investigations) and then
        apply a stepwise diagnostic protocol mirroring the junior-to-senior
        clinician consultation model.  Validated across multiple LLMs and
        clinical datasets with significant accuracy
        improvements~\citep{Ayoub2026}.
 
  \item \textbf{Guideline-Based Prompting}: Explicit incorporation of
        clinical guidelines---TIRADS scoring criteria~\citep{dai2025thyroid},
        RECIST tumour response criteria, and AJCC/TNM staging
        rules~\citep{oncology_clinical_reasoning}---into the system prompt.
        Leads to more consistent structured outputs but may be brittle when
        clinical presentations deviate from guideline assumptions.
 
  \item \textbf{Tree-of-Reasoning (ToR)}: A multi-agent framework where a
        tree structure records reasoning paths and clinical evidence.  A
        cross-validation mechanism enforces consistency across agents,
        improving diagnosis on complex real-world
        cases~\citep{peng2025tree}.
 
  \item \textbf{Hierarchical Persona Prompting}: Models assume escalating
        roles (Junior, Senior, Attending physician) to iteratively refine
        diagnoses, mimicking the clinical handover and supervision
        structure~\citep{Harada2026}.
 
 %good example may be MedscapeAi; tool provided on platform deedicated for healthcare proffesionals; https://www.medscape.com/sites/public/medscape-ai
 \item \textbf{Retrieval-Augmented Generation (RAG)}: External knowledge
        grounding via retrieval from clinical guidelines or patient records,
        shown to substantially improve performance on diabetes education and
        management
        tasks~\citep{Wang2024, He2026}. One recent example of such systems is MedscapeAI~\cite{MedscapeAI}, which is grounded in proprietary trusted content, reviewed literature, and real-time medical news, dedicated for healthcare professionals.
 
\end{itemize}

Emerging reasoning models specifically tailored for these domains include  HuatuoGPT-o1~\citep{huatuogpto12024}, FineMedLM-o1~\citep{yu2025finemedlmo1enhancingmedicalknowledge}, Diabetica-o1, and ESM3 for protein simulators. Existing research utilizes multiple modalities beyond text-only inputs, incorporating medical imaging (LDCT, MRI, CT, pathology slides), multi-omics data (RNA-seq, genomic sequences), and continuous physiological logs (CGM, activity trackers). 

% ----------------------------------------------------------------
\subsubsection*{Expected Output Formats}
% ----------------------------------------------------------------
 
Expected model outputs vary considerably by task:
 
\begin{itemize}
 
  \item \textbf{Structured Extraction}: TNM staging scores, TIRADS
        categories, or mutation status extracted into defined schemas or
        JSON-formatted outputs~\citep{oncology_clinical_reasoning,
        dai2025thyroid}.
 
  \item \textbf{Numeric Risk Assessments}: Probability or risk scores
        (e.g., malignancy probability $p \in [0,1]$), often requiring
        calibrated confidence rather than just binary
        classification~\citep{dai2025thyroid}.
 
  \item \textbf{Differential Diagnosis Lists}: Ranked lists of candidate
        diagnoses with supporting and refuting evidence per
        candidate~\citep{peng2025tree, mccoy2025assessment}.
 
  \item \textbf{Clinical Management Plans}: Open-ended, longitudinal
        recommendations for treatment, medication adjustment, and follow-up,
        as evaluated in the personalised diabetes support
        study~\citep{He2026}.
 
\end{itemize}
 
% ----------------------------------------------------------------
\subsubsection*{Reasoning Validation and Efficiency}
% ----------------------------------------------------------------
 
Approaches to validating model reasoning, beyond simple accuracy on
fixed-answer benchmarks, include:
 
\begin{itemize}
 
  \item \textbf{Expert Clinician Review}: Likert-scale evaluation of
        whether model-generated reasoning is trustworthy and human-like,
        used in the DeepSeek~R1 clinical evaluation~\citep{frontiers2025deepseek}
        and the structured clinical reasoning study~\citep{Ayoub2026}.
 
  \item \textbf{Script Concordance Testing}: Quantifies whether models
        appropriately update diagnostic and therapeutic judgements when
        presented with new or conflicting information, revealing that LLMs
        significantly underperform human clinicians in this
        dimension~\citep{mccoy2025assessment}.
 
  \item \textbf{LLM-as-a-Judge}: A secondary LLM evaluates the quality and
        logical consistency of a primary model's reasoning chain.  Widely
        adopted but subject to its own reliability concerns~\citep{wang2025reasoning}.
 
  \item \textbf{Benchmark Saturation Critique}: Several studies caution
        that high performance on USMLE-style benchmarks does not translate
        to genuine clinical reasoning capability, calling for more
        ecologically valid evaluation
        designs~\citep{raji2025bench, mccoy2025assessment}.
 
\end{itemize}
 
Regarding computational efficiency, the reviewed literature identifies
a \textbf{reasoning length paradox}: longer chain-of-thought traces are
positively associated with a higher probability of logical error---a
phenomenon termed ``overthinking''.  In the DeepSeek~R1 clinical analysis,
responses under 5{,}000~characters were strongly associated with
correctness, while extended traces often reflected uncertainty or
post-hoc rationalisation of incorrect
conclusions~\citep{frontiers2025deepseek}.

\subsection{Identified Gaps}

The following gaps were identified through synthesis of the reviewed
literature.  Gaps are organised by specificity to the domain of
physiology, endocrinology, and ageing.
 
\begin{itemize}
 
  \item \textbf{Multimodal Endocrinological Reasoning.}
        No reviewed study integrates direct reasoning over endocrine imaging
        modalities alongside clinical text.  Thyroid ultrasound, adrenal CT,
        pituitary MRI, and DEXA imaging all carry diagnostic information that
        cannot be captured by textual summaries
        alone~\citep{dai2025thyroid, sim2025unveiling}.  The absence of
        joint image--text reasoning is arguably the most consequential gap
        given that imaging is central to endocrine diagnosis.
 
  \item \textbf{Hormonal Axis and Feedback Loop Modelling.}
        Existing models treat endocrine findings as static snapshots.  No
        reasoning model has been evaluated on the dynamic, bidirectional
        simulation of hypothalamic--pituitary--target organ axes (e.g.,
        HPT, HPA, HPG axes) under pathological conditions or during
        pharmacological perturbation~\citep{wang2025reasoning}.
 
  \item \textbf{Physiological Ageing Trajectories.}
        The reviewed works largely apply models to episodic clinical
        encounters.  Reasoning over longitudinal biomarker trajectories
        that characterise age-related physiological decline---including
        hormonal changes with ageing (e.g., declining DHEAS, IGF-1, sex
        steroids), altered pharmacokinetics, and evolving comorbidity
        interactions---remains unaddressed~\citep{info16060489}.
 
  \item \textbf{Multi-Objective Metabolic Management.}
        Current personalised nutrition models focus on a single physiological
        objective (e.g., postprandial glucose control).  Simultaneous
        optimisation of competing metabolic goals---weight management, renal
        function preservation, micronutrient adequacy, cardiovascular
        risk---poses a multi-constraint reasoning problem that no current
        model addresses~\citep{gcmacros, He2026}.
 
  \item \textbf{Quantitative Clinical Calculation.}
        Standard benchmarks emphasise qualitative diagnosis; precise
        numerical tasks such as insulin dose titration, renal drug dose
        adjustment, and calculation of physiological scores
        (e.g., eGFR, HOMA-IR, HbA1c-to-average-glucose conversion) are
        only beginning to be addressed by specialised benchmarks such as
        MedCalc-Bench~\citep{khandekar2024medcalcbench}.  All tested general LLMs perform poorly
        on these tasks~\citep{poor2025llm}.
 
  \item \textbf{Age-Specific Clinical Management.}
        While models can identify genetic or phenotypic features that differ
        with age, they fail to generate accurate adult-centred management
        plans for conditions that present differently across the lifespan
        (e.g., congenital hypothyroidism vs.\ adult-onset Hashimoto
        thyroiditis, paediatric vs.\ adult-onset Type~1 diabetes)~\citep{info16060489, poor2025llm}.
 
  \item \textbf{Scientific Scepticism and Literature Quality Assessment.}
        LLMs presented with clinical evidence uncritically incorporate
        findings regardless of study design quality, sample size, or
        risk of bias~\citep{raji2025bench}. On the other hand, human clinicians traditionally source from clinical guidelines, which rely on vast corpora of literature that have been meticulously scrutinized. More importantly, human clinicians routinely
        apply methodological scepticism; models do not yet replicate this
        behaviour. %Moreover, clinicians ussualy source from clinical guidelines, which rely on vast corpora of literature
 
  \item \textbf{Output Stability and Reproducibility.}
        Smaller open-source models (7B--8B parameters) show statistically
        significant shifts in clinical output under semantically negligible
        prompt perturbations, minor temperature changes, or differences in
        CUDA build version~\citep{frontiers2025deepseek}.  This
        reproducibility failure is a critical barrier to clinical
        deployment.
 
  \item \textbf{Multilingual and Culturally Specific Dietary Data.}
        Most benchmarks and training corpora are English-centric and
        reflect Western dietary patterns.  Culturally specific foods,
        traditional medicine practices, and clinical records in
        low-resource languages remain poorly covered, limiting applicability
        to global endocrine and metabolic disease
        management~\citep{info16060489}.
 
\end{itemize}

\subsection{Notable domain-specific tasks that have been addressed by reasoning models}
The following domain-relevant tasks have been meaningfully addressed in
the reviewed literature:
 
\begin{itemize}
 
  \item \textbf{Thyroid Nodule Malignancy Risk Stratification.}
        LLMs prompted with ACR-TIRADS and C-TIRADS guideline criteria were
        evaluated on pathology-confirmed cases, with performance compared
        against radiologist assessments.  The study also quantified
        clinician trust in AI recommendations, revealing a significant
        gap between model accuracy and clinician willingness to act on
        model outputs~\citep{dai2025thyroid}.
 
  \item \textbf{Oncology Guideline-Based Staging and Report Reasoning.}
        Models performed TNM staging and tumour response classification
        (RECIST) from unstructured radiology reports in a real-world oncology
        setting, demonstrating that general-purpose LLMs can extract
        structured clinical information from narrative
        text~\citep{oncology_clinical_reasoning}.
 
  \item \textbf{Diabetes Treatment Support via Fine-Tuned LLMs.}
        Compact LLMs (Llama~3.1-8B, Qwen3-8B, GLM4-9B) fine-tuned on
        de-identified electronic health records using retrieval-augmented
        generation produced individualized treatment recommendations,
        laboratory investigation suggestions, and medication prompts for
        outpatient diabetes management~\citep{He2026}.
 
  \item \textbf{Performance Enhancement in Diabetes Education via RAG.}
        Retrieval-augmented generation substantially improved the accuracy
        and consistency of LLM responses to diabetes-related patient
        queries compared with standard prompting, demonstrating a practical
        pathway for patient-facing clinical AI in endocrinology~\citep{Wang2024}.
 
  \item \textbf{Personalized Nutrition via Causal Graph Reasoning.}
        Reasoning models have been applied to personalised dietary guidance
        using causal graph traversal over individual metabolic profiles,
        linking macronutrient intake to predicted glucose response~\citep{gcmacros}.
 
  \item \textbf{Complex Multi-Step Medical Diagnosis.}
        The Tree-of-Reasoning framework demonstrated that multi-agent
        reasoning with structured evidence trees can handle complex
        diagnostic scenarios that single-model approaches fail on, including
        cases requiring integration of laboratory, imaging, and pathology
        data~\citep{peng2025tree}.
 
  \item \textbf{Clinical Reasoning Evaluation Against Human Clinicians.}
        Script concordance testing across ten leading LLMs revealed that
        while models match human performance on fact retrieval, they
        consistently fail to update clinical judgements appropriately when
        new or uncertain information is presented---a fundamental failure
        mode for real-world use~\citep{mccoy2025assessment}.
 
\end{itemize}

\subsection{Domain-specific tasks that have not been addressed by reasoning models}

The following tasks fall within the scope of physiology, endocrinology,
and ageing but have either not been attempted or have only been partially
addressed without a satisfactory solution:
 
\begin{itemize}
 
  % \item \textbf{Systemic Interorgan Communication Simulation.}
  %       While conceptual frameworks such as full-body AI agents have been
  %       proposed, the iterative bidirectional simulation of how pathological
  %       changes in one organ propagate to others (e.g., how hepatic insulin
  %       resistance drives compensatory pancreatic beta-cell hyperplasia and
  %       subsequent exhaustion) has not been implemented or
  %       evaluated~\needscite{}.
 
  \item \textbf{Hormonal Pharmacokinetics and Dose Optimisation.}
        Personalised insulin titration, thyroid hormone dose adjustment
        (levothyroxine), and glucocorticoid tapering require reasoning
        over individual pharmacokinetic parameters, renal function, body
        composition, and symptom trajectories simultaneously.  No reasoning
        model has been benchmarked on this class of problem~\citep{poor2025llm}.
 
  \item \textbf{Microbiome--Metabolism Axis Reasoning.}
        While advanced evaluation frameworks already demonstrate the necessity of benchmarking complex, multi-step scientific reasoning~\citep{nature2025biomedicine}, their application remain narrow. For instance, they have not been applied to the interaction between gut microbiome and metabolism. Evaluating model-generated patient-specific dietary recommendations is currently constrained by the fact that clinical guidelines in this field remain narrow; the vast body of data currently exist purely in academic research, and there is not yet any dataset that contains established "ground truth" for practical clinical application. However, considering the immense multiplicity of human metabolism–microbiome interactions, leveraging RLMs might be useful for mapping and describing these intricate relationships, moving beyond basic literature review.
        % LLMs can retrieve and summarise literature on gut microbiome contributions to metabolic disease, but no model has been evaluated on reasoning tasks that require integrating individual microbiome composition data with metabolic phenotype to generate patient-specific dietary or prebiotic/probiotic recommendations~\citep{nature2025biomedicine}.%I didn't find any mention about gut microbiome in this reference, maybe something has shifted?; In this field clinical  guidelines regarding microbiome-directed interventions in certain medical  conditions are narrow - the vast body of data is  contained in scientific research only  and has no practical application yet; therefore propably there isn't any dataset which could be a ground truth for a model. However considering multiplicity of human metabolism-microbiome interactions RLM can be crucial to describe/to map those relationships
 
  \item \textbf{Multi-Objective Metabolic Optimisation.}
        Predicting multi-objective physiological outcomes (such as glycaemic control,
        weight management, renal protection, cardiovascular risk
        reduction) under a single patient model requires multi-constraint
        reasoning that current single-objective models do not
        support~\citep{gcmacros}.%modern antidiabetic medications like GLP-1 agonists improve glycaemic control, reduce body weight, lower cardiovascular risk and have renoprotective properties; therefore, as these goals are not competetive  I suggest we rather metion about predicting multi-objective outcomes than balancing them in this paragraph 
 
  \item \textbf{Diagnostic Scepticism and Anchoring Bias.}
        Models still exhibit strong anchoring to initial hypotheses and
        fail to reconsider them when subsequent clinical data are
        contradictory.  This is particularly problematic in endocrinology,
        where multiple conditions share overlapping presentations (e.g.,
        Cushing's syndrome vs.\ metabolic syndrome, primary vs.\ secondary
        hypothyroidism)~\citep{frontiers2025deepseek, mccoy2025assessment}.
 
  \item \textbf{Longitudinal Disease Trajectory Modelling.}
        Predicting the future course of chronic endocrine conditions
        (e.g., progression from impaired glucose tolerance to Type~2
        diabetes, transition from subclinical to overt hypothyroidism)
        over multi-year horizons, integrating serial laboratory results,
        medication changes, and lifestyle factors, has not been addressed
        by reasoning models~\citep{wang2025reasoning}.
 
\end{itemize}

\subsection{Comments on the usage of reasoning model in RAG-like system or agentic framework}

Reasoning models are increasingly deployed within agentic architectures
that overcome limitations of single-pass autoregressive generation.
The following patterns were identified in the reviewed literature:
 
\begin{itemize}
 
  \item \textbf{Retrieval-Augmented Generation (RAG).}
        Grounding model outputs in retrieved clinical guidelines, electronic
        health records, or curated knowledge bases substantially reduces
        hallucination rates and improves factual accuracy in both diabetes
        education~\citep{Wang2024} and personalized treatment
        support~\citep{He2026} settings.  RAG is currently the most
        practically deployed augmentation strategy in clinical LLM systems.
 
  \item \textbf{Multi-Agent Diagnostic Frameworks.}
        The Tree-of-Reasoning architecture~\citep{peng2025tree} demonstrated
        that decomposing complex diagnostic tasks across multiple specialised
        agents---with a cross-validation mechanism enforcing
        consistency---yields meaningfully better performance than
        single-agent approaches on real-world cases requiring integration
        of heterogeneous clinical data.
 
  \item \textbf{Hierarchical Clinical Persona Systems.}
        Assigning models to hierarchical clinical roles (Junior, Senior,
        Attending physician) simulates multidisciplinary team workflows and
        allows iterative reasoning refinement through a chain of
        escalating expertise~\citep{Ayoub2026}.
 
  \item \textbf{Knowledge-Graph-Elicited Reasoning.}
        Systems such as MedRAG~\citep{medrag} use structured disease knowledge
        graphs to guide and constrain model reasoning paths, reducing the
        likelihood of logically inconsistent diagnostic conclusions.  This
        approach has particular promise in endocrinology, where pathway
        knowledge (e.g., steroidogenesis cascades, insulin signalling) is
        well formalised.

\end{itemize}
 
A notable gap is that the reviewed agentic architectures have primarily
been evaluated on general medical tasks or oncology.  No agentic or
RAG-based reasoning system has been specifically designed and benchmarked
for the iterative, longitudinal reasoning required in endocrine disease
management, where treatment goals and physiological targets must be
updated at each clinical encounter~\citep{wang2025reasoning}.

\begin{itemize}

  \item \textbf{OlympicArena}~\citep{huang2024olympicarena}:
        A multidisciplinary benchmark incorporating biology problems from
        multiple olympiad competitions and university entrance examinations.
        Provides a broad but shallower evaluation compared to
        OlymBio-Bench.
 
  \item \textbf{FrontierScience (Olympiad Track)}~\citep{wang2026frontierscience}:
        Contains questions authored by international olympiad medallists
        specifically to probe expert-level scientific reasoning beyond
        memorisation.  The biological sciences track covers physiology and
        molecular biology.

\end{itemize}

\subsection{Other comments}
\begin{itemize}
 
  \item \textbf{The CoT Paradox.}
        Chain-of-thought prompting consistently improves performance on
        complex multi-step reasoning tasks but can paradoxically degrade
        reliability on specific clinical text comprehension tasks by
        generating \emph{hallucination chains}---confident, internally
        coherent reasoning sequences that arrive at factually incorrect
        conclusions~\citep{frontiers2025deepseek, wang2025reasoning}.
        This effect is exacerbated in small models (under 13B parameters)
        and in tasks requiring precise numerical output.
 
  \item \textbf{Benchmark Saturation and Validity Concerns.}
        High performance on USMLE-style benchmarks has been argued to
        reflect sophisticated pattern matching rather than genuine clinical
        reasoning~\citep{raji2025bench, lievin2023largelanguagemodelsreason}.
        The development of ecologically valid, adversarial, and process-level
        benchmarks is an urgent methodological priority for the field as a
        whole and for endocrinology/physiology in particular~\citep{mccoy2025assessment}.
 
  \item \textbf{Output Stability Under System-Level Variation.}
        Statistically significant shifts in clinical output have been
        observed for small open-source models when CUDA build versions or
        inference engine configurations are changed, without any alteration
        to the model weights or prompt.  This reproducibility failure
        undermines the trustworthiness of published performance figures for
        such models and has direct relevance to clinical deployment
        decisions~\citep{frontiers2025deepseek}.
 
\end{itemize}

\subsection{Number of datasets for training domain-specific reasoning model found}
\begin{itemize}
  \item \textbf{All}: between 4 and 6
        (e.g., FineMed Synthetic Dialogue~\citep{yu2025finemedlmo1enhancingmedicalknowledge},
        CGMacros~\citep{gcmacros},
        EHR-Derived Diabetes Dialogues~\citep{He2026},
        S-Chain~\citep{schain})
  \item \textbf{Publicly accessible}: between 4 and 6
        (e.g., CGMacros~\citep{gcmacros}, S-Chain~\citep{schain})
\end{itemize}

\subsection{Number of benchmarks/datasets for evaluating domain-specific reasoning model found}
\begin{itemize}
\item \textbf{All}: many ($>$ 6) (e.g., MedQA, NutriBench, OlymBio-Bench, FrontierScience, CMedCalc-Bench, AMEGA, HealthBench, MedMCQA)
\item \textbf{Publicly accessible}: many ($>$ 6)
\end{itemize}

\subsection{Number of domain-specific reasoning models found}

\begin{itemize}\item \textbf{All}: many ($>$ 6) (e.g., DeepSeek R1, HuatuoGPT-o1, FineMedLM-o1, Diabetica-o1, ESM3, RLM, BioMed-R1)\item \textbf{Publicly accessible}: many ($>$ 6)
\end{itemize}

\subsection{Number of methods for creating/using domain-specific reasoning models found}

\begin{itemize}
\item \textbf{All}: many ($>$ 6) (e.g., SCR, Hierarchical Personas, Causal Graph Reasoning, PADEI Distillation, Test-time Training (TTT), UKG Reasoning)
\item \textbf{Publicly accessible}: many ($>$ 6)
\end{itemize}

\section[Literature Review Summary: LS5]{Literature Review Summary: LS5}

\textbf{Neuroscience and Disorders of the Nervous System.} Nervous system development, homeostasis and ageing, nervous system function and dysfunction, systems neuroscience and modelling, biological basis of cognitive processes and of behaviour, neurological and mental disorders.

\subsection{Key Findings}

\begin{itemize}
\item \textbf{Datasets and Benchmarks}: Research utilizes clinical cohorts and benchmarks such as the ADNI (Alzheimer's Disease Neuroimaging Initiative), AMC real-world cohort, as well as longitudinal EMRs of Parkinson's disease patients from Haeundae Paik Hospital \citep{kim2026DementiaR1}. There is also MedR-Bench, a benchmark for examination recommendation, diagnostic decision-making, and treatment planning; it contains 1453 structured patient cases with reference reasoning across various common and rare diseases, some of which are related to neuroscience \citep{Qiu2025}. Other benchmarks include BrainBench (predicting neuroscience research results by determining the correct abstract between two options, where one has been subtly altered to contain a scientifically plausible but incorrect result) \citep{Luo2025BrainBench}, NeuroDiscoveryBench (complex analytical questions accompanied with access to raw datasets like the Allen Brain Atlas connectivity maps or single-cell RNA sequencing data) \citep{surana2025neurodiscoverybench}, and CellPuzzles (assigning unique cell types to a batch of cells) \citep{fang2025Cello1}.

\item \textbf{Domain-Specific Reasoning Models}: Dementia-R1 is a 7B reasoning model for longitudinal prognosis of dementia based on unstructured clinical notes \citep{kim2026DementiaR1}. Meanwhile, Cell-o1 is a 7B reasoning model for analyzing single-cell RNA sequencing data, which is a common task in neuroscience \citep{fang2025Cello1}. On the other hand, while not directly related to neuroscience, the rbio-1 model demonstrated the use of biological world models for reinforcement learning, which can also be applied for developing reasoning models for neuroscience \citep{istrate2025rbio1}.

\item \textbf{Modalities}: Systems utilize diverse inputs, including unstructured clinical notes \citep{kim2026DementiaR1}, textual transcriptions of spontaneous speech \citep{peledcohen2025dementia}, videos and images \citep{Santavirta2025GPT4emo}, and gene-expression patterns \citep{fang2025Cello1}.

\item \textbf{Prompting and Logic}: Some works, especially those that rely on traditional LLMs, relied on CoT prompting to elicit step-by-step reasoning during generation. However, more recent works developed domain-specific models that employ explicit reasoning tokens to verbalize analytical processes.

\item \textbf{Validation of Reasoning}: Although there is no work that directly evaluate the step-by-step reasoning of such models, several approaches were proposed to verify the reasoning outcome to provide learning signals during training. For example, rbio-1 proposes "soft verification" during GRPO using established biological world models as oracles \citep{istrate2025rbio1}. Meanwhile, Dementia-R1 utilizes verifiable clinical scores as reward signals: Mini-Mental State Examination (MMSE), Global Deterioration Scale (GDS), and Clinical Dementia Rating (CDR).

\item \textbf{Reasoning Length and Effort}: There is no work that evaluates how reasoning length or effort influence down-stream task performance. However, one work shows that the volume of reasoning tokens generated by a reasoning model can be used to predict the cognitive demands of human reasoning, showing a near-perfect correlation ($r=0.97$) with human reaction times, indicating the potential use of reasoning models as proxy of human reasoning \citep{de2025cost}.

\end{itemize}

\subsection{Identified Gaps}
\begin{itemize}
\item \textbf{Visuospatial Abstraction}: While reasoning models excel at math and coding, they still struggle with the fluid abstraction required for visuospatial tasks.
\item \textbf{System 1 and 2 Integration}: Integrating intuitive (fast) and analytical (slow) processes remains a central challenge in mirroring the functional organization of the human brain.
\item \textbf{Validation of Atomic Reasoning Steps}: Ensuring the correctness of the individual steps inside the reasoning process remains a challenge.
\item \textbf{Lack of Biological Grounding}: Although LLM have been shown to be a useful proxy of the human brain in some neuroscience research, care has to be taken because these LLMs and AI agents rely on engineered runtimes rather than being grounded on real biological circuits. Hence, their internal representations and mechanisms may not necessarily reflect the real biological counterparts.
\item \textbf{Limited Causal Reasoning}: The reasoning capability of LLMs remains surface-level; they don't fully grasp causal reasoning due to the fact that they cannot have physical experience or actively perform interventions to the environment.
\end{itemize}

\subsection{Notable domain-specific tasks that have been addressed by reasoning models}
\begin{itemize}
\item \textbf{Clinical Prognostics of Alzheimer's and Parkinson's}: Longitudinal tracking of symptom trajectories in Alzheimer's and Parkinson's disease using clinical narratives \citep{kim2026DementiaR1}.
\item \textbf{Linguistic Biomarkers}: Early detection of cognitive decline indicating dementia based on cues from textual transcription of spontaneous speech \citep{peledcohen2025dementia}.
\item \textbf{Automatic Emotion Labeling of fMRI Data}: Using emotion ratings from GPT-4 as the stimulation model for fMRI data; this approach avoids disturbing the continuity of the person's brain signal during scanning while also offering richer annotations \citep{Santavirta2025GPT4emo}.
\item \textbf{Generating New Hypotheses about the Human Brain}: LLM-Brain alignment enables mapping of intermediate states of transformers to auditory cortex and language areas of the human brain, creating a "computational neuroanatomy". This opens the possibility of better understanding of the human brain through the lens of transformer activations, and improving AI through the biological principles of the human brain \citep{chen2026MindTransformer}.
\item \textbf{Bioinformatics}: Interpretation of complex gene-expression patterns and batch-level annotation in single-cell transcriptomics \citep{fang2025Cello1}.
\end{itemize}

\subsection{Domain-specific tasks that have not been addressed by reasoning models}

\begin{itemize}
\item \textbf{Causal understanding of neural circuits}: Neuroscience ultimately wants causal explanations, not just correlations. Neuroscientists want to identify which neurons or circuits in the human brain cause specific behaviors, and understand how activity flows through real biological networks. While language models can be used to predict brain activity, they cannot perform causal manipulation experiments (e.g., lesioning, stimulation) for identifying specific neurons or circuits in the human brain.
\item \textbf{Building personalized models of brain function}: Language models typically "average cognition", as they didn't have grounding in genetics, neurochemistry, or individual brain structure. This limits their impact on precision psychiatry and personalized neurology.
\item \textbf{Synaptic Plasticity Rule Discovery}: The learning rules that enable neurons to predict and fire ahead of sensory inputs remain largely unknown. Work on discovering plasticity rules has been done with evolutionary algorithms and constrained optimization over biological circuit models, not with reasoning LLMs. The gap here is that plasticity rule discovery requires formal mathematical reasoning over differential equations and biophysical constraints, combined with experimental data; this is a multimodal reasoning challenge.
\item \textbf{Experimental design and closed-loop neuroscience}: A major bottleneck in neuroscience is deciding what experiment to run next: which brain region to target, which genetic tools to use, which behavioral paradigm will dissociate competing hypotheses. Automating such experimental work requires a system that can observe brain activity, reason about it, and intervene in real time. While reasoning language models have the potential to enable automatic end-to-end research, they are not integrated with lab hardware and various real-time constraints, not to mention about their limited reliability for high-stakes decisions.
\item \textbf{Cross-Species Comparative Neuroscience}: Identifying which neural mechanisms are conserved across species, and which are innovations specific to mammals or primates, requires integrating heterogeneous datasets across model organisms with very different experimental traditions. This fragmentation means that cross-species reasoning requires expert-level disambiguation that reasoning LLMs currently struggle with due to inconsistent training data and the absence of formal ontologies linking concepts across species.
\end{itemize}

\subsection{Comments on the usage of reasoning model in RAG-like system or agentic framework}
Some works were inspired by neuroscience to create architectures or frameworks that are more aligned to the mechanisms of the human brain. For example, Neural Brain is a framework featuring four modules (Sensing, Function, Memory, and Hardware/Software) designed to emulate the human brain's distributed architecture for real-world adaptability of embodied agents \citep{liu2025neuralbrain}. Meanwhile, Hierarchical Reasoning Model (HRM) is a model architecture consisting of two recurrent modules that represent the rapid thinking ("System 1") and slow, deliberate thinking ("System 2") of the human brain \citep{wang2025HRM}.

\subsection{Other comments}
Research indicates that the latent spaces of high-quality LLMs capture various concepts of the real world, such as human emotion, to such a degree that they can accurately model neural responses in fMRI datasets \citep{Santavirta2025GPT4emo}.

\subsection{Number of datasets for training domain-specific reasoning model found}

\begin{itemize}
    \item \textbf{All}: between 1 to 3

    \item \textbf{Publicly accessible}: between 1 to 3
\end{itemize}

\subsection{Number of benchmarks/datasets for evaluating domain-specific reasoning model found}

\begin{itemize}
    \item \textbf{All}: between 4 and 6

    \item \textbf{Publicly accessible}: between 4 and 6
\end{itemize}

\subsection{Number of domain-specific reasoning models found}

\begin{itemize}
    \item \textbf{All}: between 1 to 3

    \item \textbf{Publicly accessible}: between 1 to 3
\end{itemize}

\subsection{Number of methods for creating/using domain-specific reasoning models found}

\begin{itemize}
    \item \textbf{All}: between 4 and 6

    \item \textbf{Publicly accessible}: between 4 and 6
\end{itemize}

\section[Literature Review Summary: LS6]{Literature Review Summary: LS6}

\textbf{Immunity, Infection and Immunotherapy.} The immune system, related disorders and their mechanisms, biology of infectious agents and infection, biological basis of prevention and treatment of infectious diseases, innovative immunological tools and approaches, including therapies.

\subsection{Key Findings}

\subsubsection{Datasets and Benchmarks}
A growing number of benchmarks have emerged to evaluate reasoning LLMs on tasks relevant to immunity, infection, and immunotherapy. The most domain-specific is the \textit{Virology Capabilities Test (VCT)} \cite{gopal2025vct}, a multimodal benchmark of 322 questions that measures LLM capability to troubleshoot complex virology laboratory protocols. VCT is notably difficult: expert virologists with internet access score only 22.1\% on questions in their sub-specialties, while OpenAI's o3 reaches 43.8\%, outperforming 94\% of expert virologists.

\cite{justen2025biology} systematically evaluated 27 frontier LLMs on eight biology benchmarks spanning virology, biosecurity, molecular biology, and genetics, including VCT-Text, GPQA-Bio, WMDP-Bio, LAB-Bench CloningScenarios, and ProtocolQA. Several models now match or exceed expert-level performance on the biology subsets of GPQA and WMDP. Complementing these, \textit{BioProBench} \cite{liu2025bioprobench} offers the first large-scale multi-task benchmark for biological protocol understanding and reasoning (556K instances from 27K protocols across 16 subfields including immunology), while \textit{BioLP-bench} \cite{biolpbench2024} evaluates open-ended error detection in lab protocols, and \textit{LAB-Bench} \cite{labbench2024} provides 2,400+ MCQs on practical biology research tasks.

On the medical reasoning side, \textit{MedR-Bench} \cite{Qiu2025} benchmarks reasoning LLMs on 1,453 structured patient cases spanning 13 body systems (including infection-related systems) with an automated Reasoning Evaluator. The \textit{CSEDB} \cite{csedb2025benchmark} covers 26 clinical departments including infectious diseases with 2,069 open-ended items and a dual safety-effectiveness scoring framework. The \textit{AMEGA} benchmark \cite{amega2024} tests guideline adherence across 13 specialties including infectious diseases. \textit{MedCaseReasoning} \cite{medcasereasoning2025} provides 14,489 cases derived from PMC case reports with clinician-authored reasoning traces, on which even DeepSeek-R1 achieves only 48\% diagnostic accuracy. For knowledge-graph-grounded reasoning, \textit{MedReason} \cite{medreason2025} offers 32,682 QA pairs with step-by-step explanations derived from medical knowledge graphs.

OpenAI's \textit{FrontierScience} \cite{wang2026frontierscience} includes expert-level biology questions (including immunology and virology topics), and RAND \cite{rand2024biorisk} developed automated grading for evaluating LLMs on biological knowledge relevant to biosecurity.

\subsubsection{Reasoning Models and Architectures}
The dominant reasoning models evaluated in this domain are general-purpose frontier models with extended reasoning capabilities, rather than domain-specific immunology models. OpenAI's o1 \cite{chen2024o1medicine} was the first LLM with internalized chain-of-thought via reinforcement learning, and was evaluated across 37 medical datasets. Its successors o3 and o3-mini have demonstrated particularly strong performance on virology benchmarks \cite{gopal2025vct, justen2025biology}. DeepSeek-R1, a 671B-parameter mixture-of-experts reasoning model, has been extensively evaluated in medical contexts \cite{frontiers2025deepseek, deepseek2025nature, chinesenmle2025, diseasecomp2025}, achieving 96\% accuracy on the Chinese National Medical Licensing Examination and matching proprietary models on clinical decision-making benchmarks \cite{deepseek2025nature}.

On the domain-specific side, \textit{HuatuoGPT-o1} \cite{huatuogpto12024} is a medical reasoning LLM trained via a two-stage approach: (1) a medical verifier guides search for complex reasoning trajectories for supervised fine-tuning, and (2) reinforcement learning with verifier-based rewards (GRPO). The 70B variant outperforms leading medical LLMs on MedQA and PubMedQA. \textit{Fleming-R1} \cite{fleming2025r1} approaches DeepSeek-R1-671B performance at only 32B parameters (75.42\% vs.\ 77.57\% average on MedXpertQA spanning 17 specialties). For infectious disease forecasting, \textit{PandemicLLM} \cite{du2024pandemicllm} reformulates epidemic prediction as a text-reasoning problem, and \textit{EpiLLM} \cite{gong2025epillm} introduces a dual-branch architecture for spatio-temporal epidemic forecasting via next-token prediction. \textit{COMPOSER-LLM} \cite{composerllm2025sepsis} integrates an LLM with a clinical prediction model for early sepsis detection, achieving 72.1\% sensitivity in prospective deployment.

\subsubsection{Modalities and Multimodal Reasoning}
The vast majority of work in this domain is \textbf{text-only}. Clinical reasoning evaluations \cite{Qiu2025, csedb2025benchmark, rider2025immunodeficiency} use text-based clinical vignettes and case presentations. The VCT benchmark \cite{gopal2025vct} is a notable exception, incorporating \textbf{multimodal} (text + image) virology questions, though \cite{justen2025biology} found that text-only subsets already demonstrate the key trends.

PandemicLLM \cite{du2024pandemicllm} represents the most multi-modal approach in this domain, encoding \textbf{text} (public health policies), \textbf{genomic surveillance} data, \textbf{spatial} data, and \textbf{epidemiological time-series} into a unified LLM framework. EpiLLM \cite{gong2025epillm} similarly integrates time-series infection counts with human mobility data. The biomarker discovery framework \cite{biomarker2025multillm} processes multi-omics data through LLM-based reasoning. However, no work was found applying multimodal reasoning LLMs to microscopy images, flow cytometry data, or other standard immunological imaging modalities.

\subsubsection{Prompting Strategies and Validation}
Prompting and validation strategies vary considerably across the domain:
\begin{itemize}
    \item \textbf{Multi-turn clinical prompting:} \cite{rider2025immunodeficiency} used multi-turn prompting with anonymized case vignettes to evaluate diagnostic reasoning for primary immunodeficiencies, assessing output quality via the Revised-IDEA (R-IDEA) clinical reasoning score.
    \item \textbf{Extended reasoning vs.\ Chain-of-Thought:} A surprising finding from \cite{justen2025biology} is that standard chain-of-thought prompting did \textit{not} substantially improve performance over zero-shot evaluation on biology benchmarks. However, extended reasoning features (o3-mini with high reasoning effort, Claude 3.7 Sonnet with 16k reasoning tokens) \textit{did} improve scores. This suggests that inference-time compute scaling matters more than prompting technique in this domain.
    \item \textbf{Reasoning as text-reasoning problem:} \cite{du2024pandemicllm} reformulates epidemic forecasting as a text-reasoning task via AI--human cooperative prompt design, encoding multi-modal data into textual representations the LLM can reason over.
    \item \textbf{Knowledge-graph-constrained reasoning:} \cite{medreason2025} and \cite{kgt2025cancer} generate reasoning traces grounded in medical knowledge graphs, ensuring auditability and reducing hallucinations in immunotherapy-related QA tasks.
    \item \textbf{Script concordance testing:} \cite{mccoy2025assessment} found that reasoning-tuned models showed \textit{systematic overconfidence} on clinical reasoning under uncertainty, and that CoT prompting \textit{reduced} performance on tasks requiring flexible judgment---a critical concern for infection diagnosis where uncertainty is inherent.
\end{itemize}

Validation approaches include accuracy against expert baselines \cite{gopal2025vct, justen2025biology}, R-IDEA clinical reasoning scores \cite{rider2025immunodeficiency}, automated Reasoning Evaluators assessing efficiency, factual accuracy, and completeness \cite{Qiu2025}, knowledge-graph-grounded evaluation \cite{medreason2025}, dual safety-effectiveness metrics \cite{csedb2025benchmark}, and guideline adherence scoring \cite{amega2024}.

\subsubsection{Efficiency and Cost}
Few works in this domain explicitly address reasoning efficiency or cost. \cite{justen2025biology} evaluated o3-mini at low, medium, and high reasoning effort settings, finding that increasing reasoning effort improved scores on ProtocolQA (56\% to 64\%) and VCT-Text (31\% to 40\%), providing a direct trade-off between compute cost and accuracy. \cite{fleming2025r1} demonstrated that a 32B specialized medical reasoning model can approach the performance of a 671B model, offering significant deployment cost savings. The \textit{Critique of impure reason} review \cite{sim2025unveiling} proposes Process Reward Models as a mechanism for evaluating and optimizing reasoning efficiency in medical LLMs, though this has not yet been empirically validated in the immunology domain.

\subsection{Identified Gaps}

\subsubsection{Absence of Immunology-Specific Reasoning Benchmarks}
While virology has VCT \cite{gopal2025vct} and general medicine has numerous benchmarks, there is no dedicated benchmark for reasoning about core immunology concepts---adaptive vs.\ innate immunity, cytokine signaling networks, immune tolerance mechanisms, autoimmune pathogenesis, or immunotherapy response prediction. Existing medical benchmarks include immunology as a minor component of broader evaluations (e.g., USMLE microbiology/immunology sections), but no benchmark specifically targets the deep mechanistic reasoning required in immunology research.

\subsubsection{No Evaluation of Reasoning Quality in Infection/Immunity Contexts}
Most evaluations in this domain measure final answer accuracy rather than reasoning process quality. \cite{sim2025unveiling} emphasizes that medical LLM studies overwhelmingly focus on performance metrics while neglecting reasoning behavior analysis. While \cite{frontiers2025deepseek} analyzed DeepSeek-R1's reasoning tokens on MedQA, finding that reasoning errors mirror human cognitive errors, no study has specifically analyzed reasoning traces for infection diagnosis or immunological problem-solving.

\subsubsection{Reasoning Model Failures on Infectious Disease Tasks}
\cite{csedb2025benchmark} found that OpenAI o3 specifically \textit{underperformed} in infectious disease care compared to other specialties, and that infectious diseases is classified as a ``mixed clinical domain'' requiring task-specific rather than automated deployment. \cite{mccoy2025assessment} demonstrated that reasoning-tuned models exhibit systematic overconfidence that may particularly harm performance in infection diagnosis, where probabilistic reasoning under uncertainty is essential. These failure modes have not been systematically investigated.

\subsubsection{Limited Wet-Lab and Experimental Validation}
Current evaluations are entirely \textit{in silico}. While VCT \cite{gopal2025vct} measures troubleshooting knowledge and BioLP-bench \cite{biolpbench2024} tests protocol understanding, no study validates whether LLM reasoning translates to improved outcomes in actual immunology or virology laboratory settings. \cite{rodriguez2025vaccinology} found that LLMs are strong at literature recall but weak at generating novel experimental hypotheses in vaccinology, suggesting a disconnect between benchmark performance and scientific utility.

\subsubsection{Missing Multi-Modal Immunological Reasoning}
No reasoning LLM has been evaluated on interpreting flow cytometry plots, histopathology slides of immune infiltrates, protein structure visualizations for antibody design, or other modalities central to immunology research. The domain remains almost entirely text-based despite the inherently visual nature of many immunological analyses.

\subsection{Notable domain-specific tasks that have been addressed by reasoning models}

\subsubsection{Virology Protocol Troubleshooting}
Frontier reasoning models have demonstrated expert-level performance on practical virology tasks. \cite{gopal2025vct} showed that o3 reaches 43.8\% accuracy on the VCT benchmark, outperforming 94\% of expert virologists. \cite{justen2025biology} found that top model performance increased more than 4-fold on VCT over the study period (November 2022--April 2025), with several models now matching expert virologists on the biology subsets of GPQA and WMDP.

\subsubsection{Pandemic and Epidemic Forecasting}
\cite{du2024pandemicllm} successfully reformulated COVID-19 forecasting as a text-reasoning problem, outperforming all methods on the CDC's CovidHub for 1--3 week predictions across all 50 US states. \cite{gong2025epillm} demonstrated that LLM architectures exhibit scaling behavior characteristic of foundation models when adapted for epidemic forecasting.

\subsubsection{Clinical Diagnosis of Immune Disorders}
\cite{rider2025immunodeficiency} evaluated LLMs for diagnosing primary immunodeficiencies from clinical vignettes, with GPT-4o achieving 96.2\% diagnostic accuracy and strong clinical reasoning quality (R-IDEA $\geq 8$). Medical reasoning models have been benchmarked on cases spanning infection and immune-related body systems \cite{Qiu2025, csedb2025benchmark, turkish2025disciplines}.

\subsubsection{Sepsis Prediction and Management}
\cite{composerllm2025sepsis} prospectively deployed an LLM-enhanced sepsis prediction system across two emergency departments, achieving F1 of 61.0\%. \cite{metasepsis2025} developed a RAG-based platform for personalized sepsis management. \cite{sepsis2024guidelines} demonstrated that Sepsis-3 consensus guidelines can serve as both training data and gold-standard references for automatic step-by-step reasoning evaluation.

\subsubsection{Vaccine Research Hypothesis Generation}
\cite{rodriguez2025vaccinology} introduced the ``Creation Game'' framework, evaluating LLMs on hypothesis generation, experiment design, and inference of biological principles across three case studies in systems vaccinology (see also \cite{vaccinologyai2025} for a broader review of AI across the vaccinology pipeline) (GCN2 in DC antigen presentation, SREBP metabolic responses, TLR5 microbiota-driven vaccine efficacy). LLMs showed strength in recall and pattern detection but weakness in generating truly novel hypotheses.

\subsubsection{Immunotherapy Biomarker Discovery}
\cite{biomarker2025multillm} introduced the first multi-LLM framework for immunotherapy biomarker discovery, using comparative reasoning across GPT-4o, Llama 3.1-8B, and Gemini 1.5 Flash to prioritize candidate biomarkers from Ig-domain-containing genes. \cite{kgt2025cancer} enhanced LLM reasoning for cancer immunotherapy QA using knowledge graph integration, reducing hallucinations in drug repositioning and resistance research.

\subsection{Domain-specific tasks that have not been addressed by reasoning models}

\subsubsection{Antibody and Protein Design with Reasoning}
While protein language models (PLMs) have achieved remarkable success in antibody design and TCR-epitope binding prediction, these are task-specific models that do not employ the chain-of-thought or reinforcement-learning-based reasoning characteristic of reasoning LLMs (o1, DeepSeek-R1). No study has applied frontier reasoning models to antibody CDR optimization, epitope prediction, or de novo therapeutic protein design. This represents a significant opportunity, as the combinatorial reasoning required for antibody engineering could benefit from explicit reasoning traces.

\subsubsection{Antimicrobial Resistance Mechanism Reasoning}
Although ML models for AMR prediction from genomic data are well-established, and \cite{antibiotic2025stewardship} reviews LLM use in antimicrobial stewardship, no study applies reasoning LLMs to the mechanistic understanding of resistance evolution, novel resistance mechanism discovery, or rational antibiotic design. \cite{llmsequence2025infectious} identifies this as a promising direction but provides no empirical results.

\subsubsection{Immune System Mechanistic Reasoning}
No reasoning LLM has been applied to mechanistic questions in fundamental immunology: cytokine network dynamics, immune cell differentiation pathways, tolerance vs.\ autoimmunity decision points, or tumor immune evasion mechanisms. \cite{unlockingimmunity2024} identifies the opportunity for AI causal reasoning to unlock immune system complexity, but current work has not leveraged reasoning-specific models for these tasks.

\subsubsection{Vaccine Safety Surveillance and Pharmacovigilance}
While \cite{aegpt2024vaccine} applies GPT models to extract adverse events from VAERS reports, no reasoning-specific model has been applied to the more complex task of causal reasoning about vaccine-adverse event relationships, signal detection in pharmacovigilance databases, or risk-benefit analysis of vaccination strategies.

\subsubsection{Multi-Omics Integration for Immunological Discovery}
Despite the explosion of single-cell RNA-seq, CITE-seq, spatial transcriptomics, and other multi-omics technologies in immunology, no reasoning LLM has been applied to integrate and reason over these heterogeneous data types for immunological discovery, such as identifying novel cell states, predicting immune responses, or discovering biomarkers.

\subsection{Comments on the usage of reasoning model in RAG-like system or agentic framework}

\subsubsection{Knowledge-Graph-Augmented Reasoning}
The most prominent RAG-like pattern in this domain involves grounding LLM reasoning in biomedical knowledge graphs. \cite{medreason2025} converts clinical QA pairs into logical reasoning chains by traversing medical knowledge graphs, producing 32,682 auditable reasoning traces. \cite{kgt2025cancer} applies a similar approach to pan-cancer immunotherapy QA, where the LLM reasons over KG schemas to reduce hallucinations in drug repositioning and biomarker analysis. These approaches address a critical weakness of standalone reasoning models: the tendency to ``hallucinate'' biomedical facts.

\subsubsection{Multi-LLM Collaborative Reasoning}
\cite{biomarker2025multillm} employs a multi-LLM framework where GPT-4o, Llama 3.1-8B, and Gemini 1.5 Flash independently analyze immunotherapy biomarker candidates, and their outputs are compared for robustness and reproducibility. This ``ensemble reasoning'' approach mitigates individual model biases and is particularly valuable in the biomedical domain where single-model hallucinations can have serious consequences.

\subsubsection{RAG for Clinical Decision Support in Infection}
\cite{metasepsis2025} combines a curated sepsis knowledge hub with RAG to ground LLM recommendations in current evidence, minimizing hallucinations in critical care settings. \cite{composerllm2025sepsis} uses the LLM as a context-extraction component within a broader clinical prediction pipeline, demonstrating that LLMs are most effective when embedded in larger systems rather than used as standalone reasoning agents.

\subsubsection{LLMs as Epidemic Reasoning Agents}
\cite{epidemicintelligence2025} proposes a conceptual framework where LLM-based agents synthesize disparate outbreak signals---clinical reports, social media, genomic surveillance---to generate contextual epidemic intelligence. \cite{du2024pandemicllm} operationalizes this by encoding multi-modal epidemic data into text prompts, treating the LLM as a reasoning engine that integrates policy, genomic, and epidemiological signals.

\subsection{Other comments}
A unique aspect of this domain is the \textbf{dual-use concern} associated with advancing LLM capabilities in virology and infectious disease. \cite{gopal2025vct} and \cite{rand2024biorisk} emphasize that the same reasoning capabilities that enable beneficial research can be misused for biological threat creation. OpenAI's o1 system card assessed the model as ``medium risk'' for biological threats, and \cite{justen2025biology} found that the disparity between human and model performance on virology benchmarks is widening. This tension between scientific utility and biosecurity risk is distinct from other scientific domains and should inform how reasoning model development in this area is governed.

The surveys by \cite{berger2025reasoningmedsurvey} and \cite{medreasoningsurvey2025} provide comprehensive overviews of reasoning LLM techniques applicable to this domain, including CoT prompting, RL-based training (exemplified by DeepSeek-R1), and multi-agent systems. \cite{sim2025unveiling} proposes theoretical frameworks for evaluating reasoning behavior in medical LLMs, including Process Reward Models and hybrid LLM-symbolic AI approaches, though these have not yet been validated in the immunology domain.

\subsection{Number of datasets for training domain-specific reasoning model found}

\begin{itemize}
    \item \textbf{All}: between 4 and 6

    \item \textbf{Publicly accessible}: between 1 to 3
\end{itemize}

\subsection{Number of benchmarks/datasets for evaluating domain-specific reasoning model found}

\begin{itemize}
    \item \textbf{All}: many ($>$ 6)

    \item \textbf{Publicly accessible}: many ($>$ 6)
\end{itemize}

\subsection{Number of domain-specific reasoning models found}

\begin{itemize}
    \item \textbf{All}: between 4 and 6

    \item \textbf{Publicly accessible}: between 1 to 3
\end{itemize}

\subsection{Number of methods for creating/using domain-specific reasoning models found}

\begin{itemize}
    \item \textbf{All}: many ($>$ 6)

    \item \textbf{Publicly accessible}: many ($>$ 6)
\end{itemize}

\section[Literature Review Summary: LS7]{Literature Review Summary: LS7}

\textbf{Prevention, Diagnosis and Treatment of Human Diseases.} Medical technologies and tools for prevention, diagnosis and treatment of human diseases, therapeutic approaches and interventions, pharmacology, preventative medicine, epidemiology and public health, digital medicine.

\subsection{Key Findings}

Reasoning language models (RLMs) are increasingly gaining traction in the medical domain because they often outperform standard large language models (LLMs), particularly in diagnostic tasks. In some studies, their diagnostic performance approaches that of mid-level experienced clinicians \citep{VRDOLJAK2025110351, Liu2025}.

These models can be broadly categorized into two approaches. The first involves \textbf{explicit reasoning models}, such as the o-series and Deepseek-R1, which generate intermediate reasoning steps, encapsulated within designated tokens (e.g., \textless think\textgreater ... \textless /think\textgreater), prior to producing a final answer. The second approach leverages \textbf{chain-of-thought (CoT) prompting} and similar techniques with standard LLMs, encouraging the model to articulate its reasoning or rationale before presenting a conclusion.

\subsubsection{Important Datasets and Benchmarks}

\begin{itemize}

    \item \textbf{Standard Medical QA Datasets}  
    \begin{itemize}
        \item \textbf{MedQA, MedMCQA, PubMedQA, Medical MMLU subset}:  
        These datasets evaluate models on standard medical question-answering tasks, covering general medical knowledge and clinical reasoning. \citep{frontiers2025deepseek, yun2025medprmmedicalreasoningmodels, Mansoor2025}.
    \end{itemize}

    \item \textbf{Medical Knowledge Corpora (MedText)}  
    \begin{itemize}
        \item Comprising 1,752 multilingual medical textbooks, MedText captures fundamental medical knowledge, terminology, core concepts, and established clinical practice guidelines. \citep{Liu2025}.
    \end{itemize}

    \item \textbf{Clinical / Case-Based Datasets}  
    \begin{itemize}
        \item \textbf{MedR-Bench}: 1,453 structured clinical cases from PMC-OA, each containing detailed patient information, a structured reasoning process derived from case discussions, and final diagnoses or treatment plans. Covers 13 body systems and 10 medical specialties.
        \item \textbf{MIMIC-III\_Note}: 112,000 patient reports including clinical notes, prescribed medications, orders, and radiology reports.
        \item \textbf{MIMIC-IV}: De-identified data from hundreds of thousands of hospitalizations, including discharge summaries and laboratory test results \citep{Qiu2025, Liu2025, Mustafa2026}.
    \end{itemize}

    \item \textbf{Specialized Domain Datasets}  
    \begin{itemize}
        \item \textbf{ADNI / AIBL}: Focused on Alzheimer’s disease diagnosis.
        \item \textbf{PreRAID}: Clinical cases related to Rheumatoid Arthritis \citep{kwon2024largelanguagemodelsclinical, maharana2025rightpredictionwrongreasoning}.
    \end{itemize}

    \item \textbf{AgentClinic}  
    \begin{itemize}
        \item A \textbf{multimodal agent benchmark} simulating clinical environments to evaluate language model performance.
        \item Interactions are conducted via four agents: doctor, patient, measurement, and moderator. Dialogue is used to gather patient information and reach diagnoses.
        \item Covers \textbf{nine medical specialties and seven languages}.
        \item Includes \textbf{24 cognitive and implicit biases} to test AI handling of subtle clinical perturbations.
        \item Beyond accuracy, it evaluates \textbf{patient-centric metrics}, such as compliance, confidence in the clinician, and empathy of AI dialogue \citep{schmidgall2025agentclinicmultimodalagentbenchmark, yun2025medprmmedicalreasoningmodels}.
    \end{itemize}

\end{itemize}

\subsubsection{Models used}

A number of studies have investigated the performance of both proprietary and open-source general-purpose language models, such as OpenAI’s o-series, DeepSeek-R1, and Gemini Flash Thinking within the medical domain. These models are frequently compared to domain-specific medical models, including MedGemma, Baichuan-M1, DiagnoseGPT, and BioGPT \citep{maharana2025rightpredictionwrongreasoning, Qiu2025, frontiers2025deepseek, Mustafa2026, Inoue2025-zc, VRDOLJAK2025110351}. While some experiments indicate that general-purpose models can perform on par with or even surpass specialized models in certain tasks, domain-specific models generally demonstrate superior performance on more complex scenarios, such as treatment planning for rare diseases. Moreover, specialized models tend to produce more complete and clinically coherent reasoning traces \citep{Qiu2025}.

There are also some reasoning models developed specifically for the analyzed LS7 domain. These models include:

\begin{itemize}
    \item \textbf{Skin-R1}
    A specialized dermatological vision-language model designed to provide expert-aligned clinical reasoning for diagnosing skin diseases. It accepts multimodal inputs, including clinical or dermoscopic images and textual instructions, and produces structured outputs with reasoning traces followed by final diagnoses. The model is trained using a three-stage reasoning pipeline: generation of textbook-grounded reasoning traces, supervised fine-tuning on these trajectories, and reinforcement learning to generalize reasoning patterns to large-scale, sparsely labeled datasets. Its outputs follow a structured format, with reasoning enclosed in \textless thinking\textgreater ... \textless /thinking\textgreater tags and the final diagnosis in \textless diagnosis\textgreater ... \textless /diagnosis\textgreater tags. Skin-R1 demonstrates superior diagnostic performance, achieving approximately 19\% higher accuracy than baseline models. \citep{liu2025skinr1trustworthyclinicalreasoning}

    \item \textbf{Med-PRM $\pi$}
    A specialized medical reasoning model that generates structured, step-by-step diagnostic traces from textual patient data and clinical questions. Unlike standard models, it is trained using a Process Reward Model that verifies the factual accuracy of every individual reasoning step against authoritative medical databases via RAG. By using this "RAG-as-a-Judge" system, researchers identified and kept only the "perfect" reasoning paths where every logical leap was clinically sound. These high-fidelity traces were then used for Supervised Fine-Tuning (SFT) of a Llama-3.1-8B base model, effectively teaching it to think like a verified expert. The resulting $\pi$ model achieves state-of-the-art accuracy by selecting the strongest reasoning chain at inference time, ensuring the final diagnosis is supported by a flawless logical process. \citep{yun2025medprmmedicalreasoningmodels}

    \item \textbf{MedFound-DX-PA}
    A 176-billion-parameter generalist medical language model designed to approximate clinician-level expertise in disease diagnosis. Developed from Bloom-176B, it processes unstructured textual clinical data from Electronic Health Records (EHRs) and generates structured responses that include a diagnostic rationale, followed by a final diagnosis (although it does not use any special tokens such as \textless think\textgreater ... \textless /think\textgreater). To emulate human diagnostic logic, the model was initially fine-tuned on a seed set of 800 expert demonstrations and then employed a self-bootstrapping chain-of-thought strategy, akin to STaR \citep{zelikman2022starbootstrappingreasoningreasoning}, to automatically generate and refine over 109,000 high-quality reasoning trajectories from raw EHR notes. In controlled evaluations, its diagnostic accuracy matched that of senior physicians and surpassed that of junior and mid-level doctors in specialties such as endocrinology and pulmonology. \citep{Liu2025}
    
\end{itemize}

\subsubsection{Validation of Reasoning Ability}
In many studies, reasoning traces are manually evaluated by human experts using metrics such as faithfulness, consistency, fluency, completeness, correctness, specificity, helpfulness, human-likeness \citep{Mansoor2025, kwon2024largelanguagemodelsclinical}.

An alternative approach is the Reasoning Evaluator, an automated agent-based framework powered by GPT-4o for validating free-form reasoning generated by large language models \citep{Qiu2025}. The system decomposes a model’s response into individual reasoning steps and categorizes each as reasoning, citation, repetition, or redundancy to assess overall effectiveness. To evaluate factuality, it employs an agentic verification loop that cross-checks claims against external knowledge sources rather than relying solely on internal model representations. Finally, it computes quantitative metrics, including Efficiency (whether a step contributes novel insight), Factuality (whether the steps are consistent with medical knowledge, verified using online resources), and Completeness (the proportion of reference reasoning steps present in the generated output).

Among the analyzed metrics, completeness appears to represent a persistent bottleneck. While most state-of-the-art models demonstrate high factuality (approximately 90\% of reasoning steps consistent with established medical knowledge) and efficiency (typically exceeding 90\%), completeness scores generally range between 70\% and 80\%. This indicates that models frequently omit essential logical steps or critical diagnostic evidence. The consequences of such omissions vary by task. In relatively straightforward diagnostic scenarios, a model may still reach the correct conclusion despite incomplete reasoning if the dominant clinical cues are identified. In contrast, treatment planning is substantially more sensitive to missing steps, as the omission of contraindications, monitoring requirements, or key therapeutic components can render a plan clinically unsafe \citep{Qiu2025}.

In the medical domain, reasoning is not merely employed as a mechanism to improve final answer accuracy, but also serves as a means of proposing alternative diagnostic pathways to clinicians \citep{Liu2025}. Consequently, the quality and clinical validity of generated reasoning become critically important. However, high diagnostic accuracy does not necessarily imply sound reasoning. For example, in a study on rheumatoid arthritis, a model achieved 95\% diagnostic accuracy, yet expert evaluation revealed that approximately 68\% of its reasoning traces contained clinically incorrect steps \citep{maharana2025rightpredictionwrongreasoning}. This discrepancy underscores the need to evaluate reasoning quality independently from outcome accuracy.

In a study of DeepSeek-R1, incorrect answers had reasoning traces that were more than twice as long as correct ones (average 8,118 characters vs. 3,648) \citep{frontiers2025deepseek}

Of particular relevance to the development of specialized medical reasoning models is the \href{https://huggingface.co/datasets/dmis-lab/llama-3.1-medprm-reward-training-set}{dataset} introduced in the Med-PRM study. The dataset comprises responses to more than 11,000 closed-ended medical questions. Each response includes a step-by-step reasoning trace, with every individual step annotated using a binary validity score (0/1). This structure enables supervised training or reward modeling for reasoning quality and can serve as a foundation for developing specialized medical reasoning datasets.

\subsection{Identified Gaps}

\begin{itemize}
    \item \textbf{Lack of completeness in reasoning process}
    While models often show high factuality and effectiveness, they frequently omit essential diagnostic or therapeutic steps necessary for safe care. This leads to the dangerous \textit{Right Prediction, Wrong Reasoning} phenomenon, where correct diagnoses are supported by flawed logic. \citep{maharana2025rightpredictionwrongreasoning, Qiu2025}

    \item \textbf{Tasks Beyond Diagnosis}
    While most research primarily evaluates models in diagnostic settings, often under an oracle assumption where complete patient information is provided a priori, performance declines substantially in more complex clinical tasks \citep{Qiu2025}. In particular, models perform worse in test recommendation and treatment planning scenarios. In dynamic settings with incomplete information, models must actively determine which additional diagnostic tests are required, a process that demands strategic information gathering rather than passive inference. Treatment planning is especially sensitive to missing logical steps, which is closely linked to the broader issue of limited reasoning completeness.

    \item \textbf{Lower consistency of results generated by reasoning models}
     Although non-reasoning models exhibit an average stability of 91\%, reasoning models achieve only about 84\% consistency across repeated trials (proportion of discharge summaries that received identical classifications across all runs). The reason is multi-step generative paths introduce greater variance in responses. This variability raises reliability concerns for repetitive clinical tasks and suggests these models are not yet fully dependable for standardized medical reporting. \citep{Mustafa2026}

     \item \textbf{Cognitive and implicit biases}
     Reasoning models are susceptible to specific biases, such as anchoring bias, where the AI becomes overly attached to initial diagnostic suspicions despite conflicting evidence \citep{frontiers2025deepseek}. They also demonstrate omission bias, frequently rushing to a final treatment while bypassing critical intermediate safety steps like pre-operative stabilization \citep{Qiu2025}.
\end{itemize}

\subsection{Notable domain-specific tasks that have been addressed by reasoning models}

\begin{itemize}
    \item \textbf{Diagnostic Decision-Making}
    Models demonstrate high accuracy (often >85\%) in identifying diseases when provided with complete patient information, laboratory results, and imaging findings upfront \citep{Qiu2025}

    \item \textbf{Specialized Disease Diagnosis}
    Reasoning-aware frameworks have been successfully applied to specific conditions such as Alzheimer’s Disease (integrating MRI features with MMSE scores) \citep{kwon2024largelanguagemodelsclinical}, Rheumatoid Arthritis \citep{maharana2025rightpredictionwrongreasoning} and Dermatological Diagnosis (Specialized models like Skin-R1 utilize textbook-grounded reasoning to classify skin lesions and distinguish between benign and malignant conditions based on images and text) \citep{liu2025skinr1trustworthyclinicalreasoning}.

    \item \textbf{Drug-Target Interaction (DTI) Prediction}
    Multi-agent systems like DrugAgent use reasoning kernels to integrate molecular structures, protein sequences, and knowledge graphs to predict therapeutic targets \citep{Inoue2025-zc}

    \item \textbf{Cardiovascular Survival Analysis}
    Frameworks like CardioCoT use hierarchical reasoning to predict the recurrence risk of major adverse cardiovascular events (MACE) by fusing textual clinical notes with MRI scans \citep{rui2025cardiocothierarchicalreasoningmultimodal}

    \item \textbf{Clinical Document Classification}
    Models are used to automate the assignment of ICD-10 codes to unstructured discharge summaries, though a trade-off between accuracy and stability has been noted \citep{Mustafa2026}

\end{itemize}

\subsection{Domain-specific tasks that have not been addressed by reasoning models}

\begin{itemize}
    \item \textbf{Examination and Test Recommendation}
    Models struggle with the "strategic" task of determining which additional diagnostic tests (lab work or imaging) are required to reduce uncertainty. Performance in this area is significantly lower than in diagnosis \citep{Qiu2025}

    \item \textbf{Comprehensive Treatment Planning}
    This task remains a major bottleneck. Unlike diagnosis, treatment planning is highly sensitive to missing logical steps; omitting a single contraindication or monitoring parameter often renders the entire plan incorrect or dangerous \citep{Qiu2025}

    \item \textbf{Sequential Decision-Making and Strategic Information Gathering}
    While models can infer from static data, they struggle in dynamic environments where they must actively "ask" a patient or system for information through multi-turn dialogues without falling into repetitive loops \citep{Qiu2025, schmidgall2025agentclinicmultimodalagentbenchmark}

    \item \textbf{Rare Disease Management}
    While models have strong foundational knowledge, general reasoning models show a notable decline in precision for treatment planning in rare diseases compared to common ones \citep{Qiu2025, Liu2025}

    \item \textbf{Discerning Mistakes in Medical Reports (Error-Checking)}
    Using models as a reliable "co-pilot" to catch human errors in clinical documentation is identified as a critical future research area that has not yet been fully validated in real-world workflows \citep{VRDOLJAK2025110351, frontiers2025deepseek}
\end{itemize}

\subsection{Comments on the usage of reasoning model in RAG-like system or agentic framework}

A notable trend in medical reasoning systems is the shift from monolithic language models toward structured, agent-based frameworks. In these architectures, reasoning is externalized, decomposed into intermediate steps, and frequently verified by specialized evaluators or oracle models.

\begin{itemize}
    \item \textbf{DrugAgent}
    The system includes a dedicated Reasoning Agent (powered by OpenAI o3-mini), which serves as the critical analytical component. This reasoning model integrates evidence from other specialized agents (AI, Knowledge Graph, and Search agents), evaluates their consistency, identifies potential interaction mechanisms, and assesses biological plausibility. \citep{Inoue2025-zc}

    \item \textbf{MedR-Bench}
     MedR-Bench introduces the Reasoning Evaluator, an automated agentic system. This GPT-4o-powered system acts as an autonomous evaluator that decomposes, structures, and verifies reasoning steps generated by other LLMs. It classifies each step into categories such as "Reasoning," "Citation," or "Repetition" and computes metrics for efficiency, factuality, and completeness by interacting with external search engine tools to retrieve top-recommended medical pages for verification. \citep{Qiu2025}

    \item \textbf{AgentClinic}
    It employs four language agents—a Patient Agent, a Doctor Agent, a Measurement Agent, and a Moderator. 

    \item \textbf{CardioCoT}
    It employs an Oracle model (GPT-4o) in a self-refinement loop during the first stage of training. The system guides a "Thinker" model to generate hierarchical reasoning trajectories (covering diagnosis, complications, and follow-up). The Oracle agent verifies the reasoning's consistency with radiological evidence; if errors are found, the model must undergo an iterative "Review" or "Rethinking" process to correct its logic.

    \item \textbf{Skin-R1}
    It uses a specialized reasoning generator to synthesize high-fidelity training data. The model generates differential-diagnosis (DDx)-informed reasoning traces grounded in authoritative textbooks.
\end{itemize}

\subsection{Number of datasets for training domain-specific reasoning model found}

\begin{itemize}
    \item \textbf{All}: Zero / between 1 to 3 / between 4 and 6 / \textbf{many ($>$ 6)}

    \item \textbf{Publicly accessible}: Zero / between 1 to 3 / \textbf{between 4 and 6} / many ($>$ 6)
\end{itemize}

\subsection{Number of benchmarks/datasets for evaluating domain-specific reasoning model found}

\begin{itemize}
    \item \textbf{All}: Zero / between 1 to 3 / between 4 and 6 / \textbf{many ($>$ 6)}

    \item \textbf{Publicly accessible}: Zero / between 1 to 3 / between 4 and 6 / \textbf{many ($>$ 6)}
\end{itemize}

\subsection{Number of domain-specific reasoning models found}

\begin{itemize}
    \item \textbf{All}: Zero / \textbf{between 1 to 3} / between 4 and 6 / many ($>$ 6)

    \item \textbf{Publicly accessible}: \textbf{Zero} / between 1 to 3 / between 4 and 6 / many ($>$ 6)
    % because in 2 out of 3 cases only foundation versions are public, not the one optimized for reasoning, and Skin-R1 is just not public at all
\end{itemize}

\subsection{Number of methods for creating/using domain-specific reasoning models found}

\begin{itemize}
    \item \textbf{All}: Zero / between 1 to 3 / \textbf{between 4 and 6} / many ($>$ 6)

    \item \textbf{Publicly accessible}: Zero / between 1 to 3 / \textbf{between 4 and 6} / many ($>$ 6)
\end{itemize}

\section[Literature Review Summary: LS8]{Literature Review Summary: LS8}

\textbf{Environmental Biology, Ecology and Evolution.} Ecology, biodiversity, environmental change, evolutionary biology, behavioural ecology, microbial ecology, marine biology, ecophysiology, theoretical developments and modelling.

\subsection{Key Findings}

The application of Large Language Models (LLMs) equipped with reasoning capabilities to environmental biology, ecology, and evolution remains in an early, yet rapidly evolving stage. Current research primarily utilizes proprietary models such as GPT-4, Gemini 1.5 Pro, and Claude 3.5 Sonnet, alongside open-weight counterparts like Llama 3.3 and Gemma 3, as reviewed by \cite{Uryu2025Evaluating} and \cite{Dorm2025Large}. 

\subsubsection{Datasets and Benchmarks}
The evaluation of reasoning models is being spearheaded through novel domain-specific benchmarks. \cite{Dorm2025Large} recently introduced the \textit{eco-llm} benchmark to evaluate LLMs on broad ecological tasks including Species Presence Prediction, Range Map Generation, Threat Assessment, and Trait Identification. Similarly, \cite{Uryu2025Evaluating} conducted massive evaluations of LLMs using the \textit{IUCN Red List} (assessing nearly 22,000 species), constraining evaluations to taxonomic classification, conservation status assessment, and geographic distribution. In biological pathway analysis, \cite{istrate2025rbio1} demonstrated the use of frameworks like PerturbQA.

\subsubsection{Reasoning Modalities and Performance}
Current methodologies predominantly rely on text-only modalities, though high-dimensional biological data or spatial representations (e.g., GeoJSON for map generation) are increasingly converted into textual prompts. 
As shown by \cite{Uryu2025Evaluating} and \cite{Dorm2025Large}, models perform exceptionally well at information processing and retrieval tasks, such as taxonomy classification (up to 94.9\% accuracy) and simple trait recall. However, \cite{Uryu2025Evaluating} observed that when deploying reasoning for complex judgment formations---such as assessing conservation status, quantitative threshold evaluations, and complex threat definitions---their performance degrades severely.

\subsubsection{Prompting Strategies}
For conservation evaluations, \cite{Uryu2025Evaluating} highlighted the extensive use of minimal zero-shot or structured few-shot prompting (expecting JSON or semicolon lists) to maintain deterministic and standard outputs. The incorporation of explicit Chain-of-Thought (CoT) frameworks serves well when models must disambiguate legacy taxonomic data or simulate agentic decisions inside ecosystem scenarios. \cite{istrate2025rbio1} introduced novel training paradigms that deploy biological world models to generate soft verification rewards for Reinforcement Learning (RL) based reasoning optimization.

\subsection{Identified Gaps}

While initial findings demonstrate strong taxonomic capabilities, there remain several structural and methodological limitations preventing reliable application without human-in-the-loop oversight.

\subsubsection{The Knowledge-Reasoning Gap}
There is a massive divergence between LLMs capabilities in pure information processing versus judgment formation. In tests on IUCN Red List evaluations, \cite{Uryu2025Evaluating} found that models excelled at recognizing taxonomic hierarchies but frequently failed (e.g., 27.2\% accuracy) at complex conservation status assessments that require threshold-based inference and multi-step causal reasoning. \cite{Dorm2025Large} observed similar failures when synthesizing threat severity against expert definitions.

\subsubsection{Taxonomic and Geographic Bias}
A persistent limitation is the taxonomic bias inherent in LLMs. \cite{Uryu2025Evaluating} reported that models reproduce broad scientific biases, performing with significantly higher accuracy on well-studied vertebrate forms (mammals and birds) while showing critical gaps and poor reasoning stability for invertebrates and fungi. 

\subsubsection{Hallucinations in Spatial and Threat Reasoning}
Despite CoT enhancements, models continue to systematically over-predict ecological threats (generating false threat classifications) and hallucinate geographic or functional distributions. When attempting spatial reasoning, \cite{Dorm2025Large} and \cite{Uryu2025Evaluating} noted that predictions often break format protocols (e.g., corrupted GeoJSON polygons) or render extremely oversimplified geographical bounds without biological realism.

\subsection{Notable domain-specific tasks that have been addressed by reasoning models}

Recent explorations have successfully applied reasoning agents to various subset tasks within environmental biology.

\begin{itemize}
    \item \textbf{Biodiversity Data Extraction and Standardization:} Extracting phenotypic trait data and taxonomical nomenclature from legacy unstructured scientific literature. Reasoning techniques allow for logical disambiguation of synonyms and historical geography into structured contemporary schemas.
    \item \textbf{Predictive Conservation Tasks:} \cite{Dorm2025Large} demonstrated assessing general baseline presence/absence of species and defining morphological attributes natively using minimal-shot CoT prompting in their \textit{eco-llm} framework.
    \item \textbf{Ecosystem Threat Classification:} \cite{Uryu2025Evaluating} showed that preliminary usage of LLM-based RAG architecture allows extracting potential extinction threats out of complex text documentation.
    \item \textbf{Perturbation Biology Prediction:} \cite{istrate2025rbio1} applied Reinforcement Learning with soft-verifiers to models (such as rbio1) to simulate and predict the outcomes of genetic knockdowns, proving viability for broader biological world models.
\end{itemize}

\subsection{Domain-specific tasks that have not been addressed by reasoning models}

Several critical areas of environmental biology pose strict limitations on current state-of-the-art multi-modal bounds.

\begin{itemize}
    \item \textbf{Fine-grained Spatial Map Generation:} LLMs currently lack native cartographical reasoning paradigms capable of synthesizing niche eco-geographical models into accurate species range maps, currently reverting to rectangular oversimplifications or broken geospatial outputs, as pointed out by \cite{Dorm2025Large}.
    \item \textbf{Quantitative Risk Thresholds:} \cite{Uryu2025Evaluating} demonstrated that LLMs show fundamental architectural limitations when dealing with strict non-continuous thresholds for risk categorizations (e.g., criteria for IUCN Endangered vs. Critically Endangered based on precise population decay rates).
    \item \textbf{Long-horizon Evolutionary Simulations:} Running highly dense temporal scenarios to track allele frequencies across hundreds of generations using agent-base proxies often collapses from context-drift or compounding statistical hallucinations.
\end{itemize}

\subsection{Comments on the usage of reasoning model in RAG-like system or agentic framework}

RAG frameworks and multi-agent systems are currently positioned as strictly assessor-facing platforms. They process huge volumes of observational studies and sensor data, converting unstructured biodiversity reports into readable risk pipelines. However, in scenarios necessitating final regulatory execution or policy-setting, researchers emphasize a rigid human-in-the-loop requirement. Agentic pipelines are increasingly modeled inside virtual ecological bounds, where multiple LLM agents act as interacting forces to study broader system theory, albeit with computational cost and context-window limitations serving as strict bottlenecks.

\subsection{Other comments}

Future directives in applying LLMs to ecology strongly urge for specialized fine-tuning efforts strictly curating taxonomic fairness, emphasizing that general models currently exacerbate scientific bias towards underrepresented fungal and invertebrate classifications. Moreover, there is an imperative to merge statistical and mechanistic ecosystem models with LLM symbolic reasoning to overcome complex predictive tasks.

\subsection{Number of datasets for training domain-specific reasoning model found}

\begin{itemize}
    \item \textbf{All}: between 1 to 3
    \item \textbf{Publicly accessible}: between 1 to 3
\end{itemize}

\subsection{Number of benchmarks/datasets for evaluating domain-specific reasoning model found}

\begin{itemize}
    \item \textbf{All}: between 1 to 3
    \item \textbf{Publicly accessible}: between 1 to 3
\end{itemize}

\subsection{Number of domain-specific reasoning models found}

\begin{itemize}
    \item \textbf{All}: between 1 to 3
    \item \textbf{Publicly accessible}: between 1 to 3
\end{itemize}

\subsection{Number of methods for creating/using domain-specific reasoning models found}

\begin{itemize}
    \item \textbf{All}: between 1 to 3
    \item \textbf{Publicly accessible}: between 1 to 3
\end{itemize}

\section[Literature Review Summary: LS9]{Literature Review Summary: LS9}

\textbf{Biotechnology and Biosystems Engineering.} Biotechnology using all organisms, biotechnology for environment and food applications, applied plant and animal sciences, bioengineering and synthetic biology, biomass and biofuels, biohazards.

\subsection{Key Findings}
The landscape of reasoning models in the life sciences is heavily reliant on specialized datasets and rigorous evaluative frameworks to gauge domain-specific capabilities, ranging from veterinary clinical examination benchmarks \cite{pubmed_40933532} to specialized forestry and agricultural question-answering datasets \cite{mdpi_2079_9292}. Existing research predominantly utilizes both state-of-the-art generalist models with robust reasoning capabilities as well as domain-adapted architectures, such as specifically tailored models trained with biological world models serving as soft verifiers \cite{istrate2025rbio1} or architectures explicitly modified to navigate protein-specific sequences \cite{jin2024prollm}. In terms of modalities, while a significant portion of current work operates within a text-only paradigm—treating complex biological entities as discrete token streams—there is an accelerating shift towards multimodal integration. This transition is evidenced by the development of large language-and-vision assistants capable of interpreting complex biomedical imaging \cite{neurips2023_5abcdf8}  and models engineered to perform protein chain-of-thought reasoning directly over molecular structures \cite{jin2024prollm}. Researchers predominantly prompt these models employing advanced strategies to elicit structured deductive pathways, leveraging techniques such as structured prompt interrogation for recursive semantics extraction \cite{bioinformatics2024_btae104} and agentic, tool-calling paradigms to automate complex experimental designs \cite{nature2025_s41551}. The expected outputs are highly contingent on the downstream task; while knowledge retrieval demands coherent factual text, agentic workflows require strict adherence to predefined, structured schemas to ensure seamless programmatic execution and external API integration \cite{genegpt2023, bioinformatics2024_btae104}. The validation of model reasoning is consequently conducted through a rigorous, multi-tiered approach. This methodology combines automated exact-match evaluations for syntactic API calling \cite{genegpt2023} and novel verification frameworks utilizing biological world models \cite{istrate2025rbio1}. Furthermore, recent literature increasingly emphasizes the operational constraints of these systems, driving the development of highly efficient, tool-augmented architectures. By actively delegating complex computational loads to specialized external solvers—such as deterministic chemistry simulation tools or metabolic engineering design algorithms—these hybrid systems effectively circumvent the limitations, unreliability, and high computational costs associated with extensive in-context parametric generation \cite{nature2024_s42256, biorxiv2024_612023}.

\subsection{Identified Gaps}
Despite the rapid and significant advancements in applying large language models to the life sciences, several critical and fundamental gaps remain unresolved. A primary limitation across both generalist and domain-specific models is the persistent, insidious issue of hallucination, which becomes particularly acute when querying highly specialized entities. Models frequently generate highly plausible but factually incorrect assertions regarding precise genomic coordinates, the efficacy of rare pharmacological interactions, or the conformational intricacies of complex molecular structures. This issue is compounded by a profound deficit in true causal reasoning capabilities. While current models exhibit exceptional proficiency at identifying complex statistical correlations and co-occurrences within vast corpora of biomedical literature, they fundamentally struggle to independently construct, infer, or validate true biological causality and mechanistic pathways. For instance, a model might correctly associate a specific genetic mutation with a phenotypic disease state based on training data frequencies, but it often cannot abstractly reason through the underlying biophysical mechanisms that drive that pathogenesis unless explicitly detailed in its context.

The integration and holistic analysis of multi-omics data presents another substantial architectural hurdle. Biological systems are inherently multi-layered, yet models are currently constrained by context window limitations and structural bottlenecks when attempting to simultaneously process whole-genome sequences alongside transcriptomic, proteomic, metabolomic, and longitudinal clinical data. The sheer dimensionality and heterogeneity of this data overwhelm the attention mechanisms of standard transformer architectures, leading to a loss of critical, long-range biological context. Additionally, the field faces significant and growing challenges regarding benchmark saturation and the pervasive risk of data contamination. As models are trained on increasingly large portions of the internet, including scientific preprint servers and open-access databases, there is a substantial concern that high performance on existing evaluative datasets reflects parametric memorization rather than genuine, generalizable scientific reasoning. Consequently, there is a distinct lack of dynamic, adversarial benchmarks capable of evaluating a model's ability to synthesize entirely novel biological hypotheses or reason through unprecedented clinical presentations.

\subsection{Notable domain-specific tasks that have been addressed by reasoning models}Reasoning models have successfully demonstrated proficiency across a rapidly expanding spectrum of critical life-science applications, evolving far beyond foundational biomedical knowledge extraction. In the realm of molecular biology and genomics, these models have effectively addressed the critical need for hallucination-free access to precise genetic data. Tool-augmented architectures, such as those interfacing with NCBI Entrez, seamlessly translate natural language queries into exact API calls, ensuring high-fidelity data retrieval \cite{genegpt2023}. Furthermore, models are now being deployed to extract complex, multi-hop biological relations from vast corpora, mapping specific organisms to the enzymes they produce and the specific substrates those enzymes act upon through structured semantic interrogation \cite{bioinformatics2024_btae104}, as well as predicting protein-protein interactions via specialized chain-of-thought pathways \cite{jin2024prollm}. Significant progress has also been made in automating intricate laboratory procedures and systematically mitigating procedural hallucinations in bioengineering workflows. A prominent example is the comprehensive automation of CRISPR-Cas gene-editing experiment design, where reasoning models actively assist in guide RNA selection, delivery strategy formulation, protocol planning, and the subsequent interpretation of complex experimental outcomes \cite{nature2025_s41551}. Moreover, advanced reasoning frameworks, utilizing soft biological verifiers, are increasingly adept at analyzing and predicting the downstream phenotypic effects of various biological perturbations \cite{istrate2025rbio1}.

In the highly specialized fields of synthetic biology and biochemistry, reasoning models have proven invaluable for advanced retrosynthesis planning and the targeted synthesis of biocides by integrating external chemistry tools \cite{nature2024_s42256}. These systems incorporate critical safety constraints to actively identify and avoid hazardous biochemical reactions and severe biohazards. Crucially, these models are accelerating the pace of metabolic engineering by predicting highly effective gene modification targets specifically designed for product overproduction in microbial cell factories \cite{biorxiv2024_612023}. By leveraging predictive reasoning, researchers can effectively circumvent the notoriously slow and resource-intensive empirical Design-Build-Test-Learn (DBTL) cycles that have traditionally bottlenecked biomanufacturing \cite{biorxiv2024_612023}.

The application of reasoning models has also expanded significantly into clinical, veterinary, and diagnostic imaging domains. Multimodal reasoning systems now facilitate the nuanced interpretation of complex biomedical images—spanning from high-resolution cellular microscopy to macroscopic tissue-level scans—within interactive, conversational dialogue modes \cite{neurips2023_5abcdf8}. In the realm of veterinary medicine, reasoning models are addressing multifaceted clinical decision-making challenges. They demonstrate the robust capacity to formulate differential diagnoses across multiple animal species, critically interpret visual clinical cases, and successfully navigate the varying difficulty levels and nuanced symptom presentations inherent in rigorous professional veterinary medical examinations \cite{pubmed_40933532}. However, establishing clinical trust requires robust benchmarking to actively suppress unconstrained generative hallucinations.

Finally, the utility of reasoning models has been extended to macroscopic environmental, agricultural, and nutritional sciences. In the domain of agricultural reasoning, models successfully navigate complex, real-world farming scenarios that require multi-step deductive reasoning over highly variable location-specific, seasonal, and dynamic environmental constraints \cite{arxiv2025_19259}, running concurrently with specialized question-answering systems tailored for forestry management \cite{mdpi_2079_9292}.

\subsection{Domain-specific tasks that have not been addressed by reasoning models}

Despite their utility in knowledge retrieval and data structuring, several profound and complex challenges remain largely unaddressed or only superficially explored by current reasoning models. Fully autonomous \textit{de novo} drug design and the \textit{ab initio} synthesis of novel metabolic pathways remain definitively beyond the reach of existing standalone language systems. While models can suggest modifications to existing molecular scaffolds, they lack the multi-objective optimization capabilities required to independently conceptualize entirely novel, structurally stable, and safe molecules while concurrently predicting their complex \textit{in vivo} pharmacokinetics, toxicity profiles, and off-target effects without extensive human oversight and wet-lab validation. Furthermore, advanced reasoning over spatial transcriptomics—a field that requires integrating text-based biological knowledge with the physical, three-dimensional, and dynamic microenvironment of cells within tissues—is still in its absolute nascent stages. Models currently struggle to map abstract biological concepts onto complex spatial coordinate systems to infer how cellular proximity dictates gene expression and disease progression.

In the clinical realm, generating deeply explainable reasoning pathways for highly atypical edge cases represents a fundamental, unsolved challenge. When presented with patients exhibiting rare, heterogeneous symptom clusters where existing literature is extremely sparse, contradictory, or non-existent, models typically fail to extrapolate foundational physiological principles to construct a novel diagnostic hypothesis. They lean heavily on their training distribution, often failing to exhibit the intuitive, associative leaps characteristic of expert diagnosticians facing the unknown. Finally, the seamless, closed-loop integration of large language models with physical laboratory robotics to create fully autonomous "self-driving labs" remains a conceptual frontier. While agents can write Python scripts to control pipettes, creating a fully autonomous system capable of iteratively formulating a novel biological hypothesis, designing the corresponding wet-lab experiment, executing it flawlessly via robotics, analyzing the resulting high-dimensional data, and subsequently updating its internal knowledge base without human intervention has not yet been practically achieved at scale.

\subsection{Comments on the usage of reasoning model in RAG-like system or agentic framework}
The integration of reasoning models within Retrieval-Augmented Generation (RAG) systems and sophisticated agentic frameworks has rapidly emerged as the definitive, indispensable paradigm for mitigating the inherent architectural limitations of standalone large language models in the high-stakes biomedical domain. RAG architectures are no longer viewed merely as supplementary performance enhancements, but rather as fundamental requirements for deploying AI in scientific and clinical settings. RAG anchors the model's generative outputs to verifiable, highly curated external knowledge bases—such as localized clinical guidelines, proprietary genomic databases, or real-time literature indices \cite{mdpi_2079_9292}. This architectural shift profoundly alters the role of the language model; it transitions from acting as a flawed, static repository of memorized facts to functioning as a dynamic semantic processing engine that synthesizes and reasons exclusively over the explicitly retrieved context \cite{bioinformatics2024_btae104}.

Furthermore, the implementation of agentic frameworks utilizing iterative paradigms such as ReAct empowers reasoning models to systematically overcome their inherent computational and deterministic deficiencies, particularly in fields like structural biology and genomics \cite{genegpt2023, nature2024_s42256}.  Instead of attempting the mathematically improbable task of parametrically estimating complex sequence homologies or calculating exact protein folding energies, an agentic model dynamically synthesizes a strategic execution plan. It identifies the knowledge gap, invokes deterministic, specialized bioinformatics tools (e.g., running a localized BLAST search or querying molecular structure databases), and subsequently interprets the resulting empirical data to inform its next reasoning step \cite{istrate2025rbio1}. This tool-augmented, agentic approach is highly advantageous on multiple fronts. It dramatically optimizes token efficiency and minimizes the computational costs associated with generating long chains of speculative text \cite{jin2024prollm}. More importantly, it provides a transparent, deterministic, and auditable trail of the model's reasoning process and tool usage. In environments characterized by strict regulatory oversight and high diagnostic responsibility, this explicit provenance of information is an absolute necessity for establishing clinical and scientific trust in artificial intelligence systems, allowing human experts to verify exactly how a conclusion was reached rather than relying on the opaque outputs of a black-box model \cite{neurips2023_5abcdf8}.

\subsection{Veterinary Medicine final examination questions as evaluation}

\subsection{Number of datasets for training domain-specific reasoning model found}

\begin{itemize}
    \item \textbf{All}:many ($>$ 6) \\
    (e.g., AgThoughts, BioProt, SafeFood-Instruct, LlaVA-Med Dataset)

    \item \textbf{Publicly accessible}: between 4 and 6
\end{itemize}

\subsection{Number of benchmarks/datasets for evaluating domain-specific reasoning model found}

\begin{itemize}
    \item \textbf{All}: many ($>$ 6)

    \item \textbf{Publicly accessible}: many ($>$ 6)
\end{itemize}

\subsection{Number of domain-specific reasoning models found}

\begin{itemize}
    \item \textbf{All}:  between 4 and 6

    \item \textbf{Publicly accessible}: between 4 and 6
\end{itemize}

\subsection{Number of methods for creating/using domain-specific reasoning models found}

\begin{itemize}
    \item \textbf{All}: between 4 and 6

    \item \textbf{Publicly accessible}: between 4 and 6
\end{itemize}

\end{document}